\documentclass{article}

\usepackage{arxiv}

\usepackage[utf8]{inputenc}
\usepackage[T1]{fontenc}
\usepackage{amsmath}
\usepackage{amssymb}
\usepackage{amsfonts}
\usepackage{graphicx}
\graphicspath{{./}}
\usepackage{fix-cm}
\usepackage{newtxmath}
\usepackage{latexsym}
\usepackage[colorlinks,bookmarksopen,bookmarksnumbered,citecolor=blue,linkcolor=blue,urlcolor=blue]{hyperref}
\usepackage{url}
\usepackage{microtype}
\usepackage{algorithm}
\usepackage{algpseudocode}
\usepackage{enumitem}
\usepackage{booktabs}
\usepackage{multirow}
\usepackage{makecell}
\usepackage[table]{xcolor}
\usepackage{rotating}
\usepackage{tabularx}
\usepackage{array}
\usepackage{calc}
\usepackage{ulem}
\usepackage[numbers]{natbib}
\usepackage{cleveref}
\hypersetup{hypertexnames=false}

\Crefname{figure}{Figure}{Figures}
\crefname{figure}{Figure}{Figures}
\Crefname{table}{Table}{Tables}
\crefname{table}{Table}{Tables}
\allowdisplaybreaks

\title{SAT-RTS: A systematic framework for tactical knowledge extraction and visualization-based analysis in real-time strategy games}

\author{
\normalfont
\begin{tabular}{c}
Chunhui Bai$^{1,3,4}$, Changhe Li$^{2,5,*}$, Yuqiang Li$^{2,5}$, Lei Liu$^{2,5}$, Shoufei Han$^{2,5}$\\
\texttt{baichunhui@cug.edu.cn}, \texttt{changhe.lw@gmail.com}, \texttt{yuqiangli@aust.edu.cn}\\
\texttt{liulei970507@163.com}, \texttt{hanshoufei@gmail.com}
\end{tabular}\\[0.4em]
\begin{minipage}{0.96\textwidth}
\small
$^{1}$ School of Artificial Intelligence and Automation, China University of Geosciences, Wuhan 430074, China\\
$^{2}$ State Key Laboratory of Digital Intelligent Technology for Unmanned Coal Mining, Anhui University of Science \& Technology, Huainan 232001, China\\
$^{3}$ Hubei Key Laboratory of Advanced Control and Intelligent Automation for Complex Systems, Wuhan 430074, China\\
$^{4}$ Engineering Research Center of Intelligent Technology for Geo-Exploration, Ministry of Education, Wuhan 430074, China\\
$^{5}$ School of Artificial Intelligence, Anhui University of Science \& Technology, Hefei 231131, China\\
$^{*}$ Corresponding author.
\end{minipage}
}

\begin{document}
\maketitle

\begin{abstract}
Efficient tactical knowledge extraction and analysis in real-time strategy (RTS) games micromanagement are constrained by the high-dimensional coupled state-action sequential data and the black-box decision-making process. Current research rarely provides a hierarchical visualization-based attribution analysis from the perspective of data decoupling and abstraction. To facilitate interpretable tactical knowledge extraction and visualization-based analysis in RTS games, a systematic framework named state-action-tactic analysis pipeline (SAT-RTS) is proposed. To decipher the deep-seated drivers of critical decisions in RTS learning systems, this work integrates interpretable visualization with the automated extraction of latent tactical patterns from high-dimensional sequence data. By adapting a cluster-centric BK-tree algorithm and incorporating specialized distance metrics designed to quantify multi-aspect similarities, the proposed framework facilitates robust state-stream abstraction. Furthermore, a rule-based multi-label extraction method is developed to transform unstructured state-action sequences into discrete and interpretable tactical labels, effectively bridging the gap between raw behavioral data and high-level tactical insights. By holistically integrating these computational methods into a hierarchical visualization-based pipeline, the proposed framework effectively addresses the challenges of processing massive real-time data streams while providing fitness landscape visualizations and analytical insights to decipher deep-seated tactical drivers. Comprehensive experiments demonstrate that the proposed SAT-RTS significantly enhances the interpretability and efficiency of tactical analysis in complex RTS environments.
\end{abstract}


\section{Introduction}
\label{sec:Introduction}

Real-time strategy (RTS) games, characterized by real-time confrontation among multiple agents and at hierarchical strategic levels, have been regarded as a major challenge for artificial intelligence (AI) algorithms \citep{ontanon_Survey_2013}. Difficulties become particularly complex under the coupling effect of large-scale state-action space and black-box spatial-temporal reasoning. Deep Reinforcement learning (DRL) methods have been widely applied to RTS micromanagement as an ideal testbed \citep{vinyals_StarCraft_2017}. However, the massive state space and reliance on sparse rewards often result in highly fragmented and unstructured policies, making the underlying tactical logic difficult to interpret \citep{wang_Deep_2024a, qi_Distributed_2026a}. Compounding this issue, even well-trained learning systems optimized for reward maximization lack the transparency necessary to reveal the underlying decision-making mechanisms \citep{puiutta_Explainable_2020}.

\begin{figure}[ht]
    \centering
    \includegraphics[width=1\linewidth]{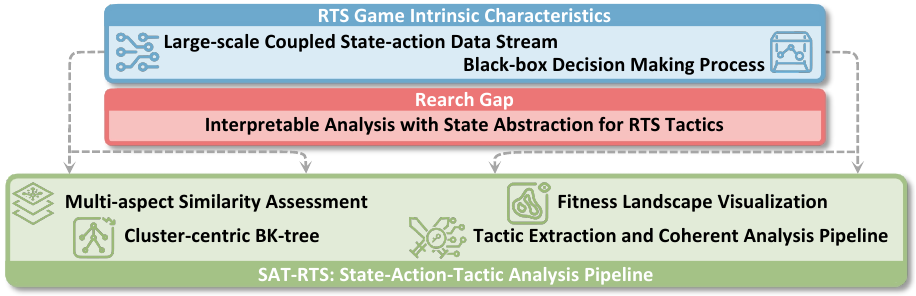}
    \vspace{-10pt}
    \caption{Key challenges in tactical extraction and analysis for RTS micromanagement and the motivation of the proposed SAT-RTS pipeline}
    \label{fig:motivation}
\end{figure}

Large-scale and black-box nature of RTS games arises from multiple coupled factors. The multi-level tactical requirements necessitate that numerous units execute parameterized actions across continuous maps, resulting in game states that predominantly manifest as large-scale and high-dimensional data \citep{cao_Reducing_2025}. Stochasticity and decision-observation disparities obscure the complex coordination and causal logic among units, hindering AI applications in RTS and even real-world scenarios. Consequently, addressing large-scale state-action space processing and interpretable analysis in RTS games has emerged as pivotal research directions \citep{xu_Applicable_2022}. To better navigate these complexities, Figure~\ref{fig:motivation} illustrates the key challenges in RTS tactic analysis and the conceptual motivation of the proposed framework.

Existing research has advanced state abstraction \citep{perkins_Terrain_2010,jo_FoX_2024} and interpretable analysis \citep{kuan_Visualizing_2017,haley_Cluster_2020,kozik_Mimicking_2021} for RTS games to address the challenges of large-scale state spaces and black-box decision-making. However, research remains limited in exposing the underlying tactical rationales that drive critical decisions within a coherent framework. In particular, the massive entanglement of state-action data increases the difficulty of distilling tactical intent from complex game streams, while the lack of fitness landscape and visualization techniques further limits the interpretable analysis of reward-driven Reinforcement learning (RL) policies \citep{heuillet_Explainability_2021}.

To address the existing limitations, this work proposes the interpretable state-action-tactic analysis pipeline for RTS micromanagement (SAT-RTS). The primary contributions of this work are summarized as follows:
\begin{itemize}[label=$\bullet$]
    \item It proposes a multi-aspect similarity assessment framework to mitigate the analytical challenges posed by massive coupled state-action data. By quantifying similarities across states, state-transition sequences and action patterns, the method establishes hierarchical normative criteria that facilitate subsequent high-level analysis.
    \item It designs an efficient cluster-centric BK-tree algorithm to cluster complex state data streams into representative nodes. Crucially, the fitness landscape visualization and action sequence pattern mining for RTS micromanagement are developed as a bridge between black-box behaviors and intuitive tactics, constituting a multi-level visual toolset for tactical analysis.
    \item It implements a rule-based multi-label tactic extraction method and comprehensive attribution analyses to transform subjective tactic perceptions into explicit semantic knowledge. Furthermore, the decoupling of state-action data empowers the quantitative cross-analysis of states and tactics, providing in-depth tactical insights and elucidating the underlying drivers of critical decision-making.
\end{itemize}

The remainder of the paper is organized as follows. Section~\ref{sec:Related Work} provides an overview of the previous related work. Section~\ref{sec:Methodologies} describes the method proposed in this work in detail. Section~\ref{sec:Experimental Results and Analysis} shows the comparative experimental results, the fitness landscape visualization for RTS game micromanagement and the comprehensive tactic analysis. Section~\ref{sec:Conclusion} concludes the work and outlines future works.

\section{Related work}
\label{sec:Related Work}
This section briefly reviews the related works for large-scale state space abstraction and interpretable analysis in RTS games. The former focuses on reducing high-dimensional observation complexity, while the latter discusses the underlying decision-making logic.

\subsection{Large-scale state abstraction}

Large-scale state space presents an inherent challenge in RTS games \citep{vinyals_Grandmaster_2019}. Various state abstraction techniques are widely employed in learning and optimization, primarily comprising compression and aggregation. Compression preserves the most essential state features to reduce data dimensionality \citep{westphal_InformationTheoretic_2024,wang_Building_2024}, whereas aggregation consolidates numerous similar states into the same formation semantics, contributing to interpretability in model training or attribution analysis \citep{zeng_Hierarchical_2023,nayyar_Learning_2024}. Open problems including the trade-off between abstraction accuracy and training time in large-scale environments are particularly pronounced in RTS games \citep{xu_Applicable_2022}.

\subsubsection{State compression}

State compression can effectively reduce high-dimensional state data \citep{dockhorn_State_2023}, yet in RTS games, balancing representational fidelity, semantic interpretability and scalability remains challenging. Discretization of continuous features is a common approach, yet suffers from information loss and degradation in decision accuracy \citep{micic_Developing_2011}. Similarly, probability-based army clustering focuses solely on unit type composition, lacking spatial distribution and HP information critical to combat effectiveness \citep{synnaeve_Dataset_2012}. Neural networks encode high-dimensional states within low-dimensional latent spaces as implicit dimensionality reduction \citep{kim_Surrogateassisted_2024}. Although lightweight semantic layers improve upon this by preserving decision quality \citep{li_Accelerating_2020}, they remain focused on computational efficiency rather than explicit tactical interpretation. Domain-specific predicate sets and DSLs can achieve semantic representations through rule-based formulations \citep{lee_State_2008, marino_Evolving_2022}, yet their applicability is constrained by problem complexity and rule set scalability \citep{dockhorn_State_2023}.

Spatial abstraction through geometric representation, notably the theoretical foundation of topological nodes centered on choke points \citep{perkins_Terrain_2010}, has been extensively implemented in game-tree search in RTS games \citep{uriarte_Gametree_2014, uriarte_HighLevel_2014, uriarte_Combat_2018}. Furthermore, influence map provides an effective spatial abstraction that encodes terrain and unit distribution, preserving firepower distribution patterns via numerical matrix \citep{micic_Developing_2011,park_MCTS_2015}. Nevertheless, spatial abstraction methods face limitations in state similarity assessment and applicability to advanced techniques, exemplified by the challenge of performing similarity-based consolidation and efficient storage-retrieval of influence map representations in RL.

\subsubsection{State aggregation}
While aggregating large-scale states through tactical formation similarity can mitigate the curse of dimensionality, direct clustering based on Euclidean distance remains difficult due to complex unit distributions. To address this, recent studies have shifted toward either identifying formations as learnable tactical elements, or aggregating states with similar outcomes.

Regarding the former, formation awareness can be incorporated into online learning to tackle the challenges of large-scale and sparse-reward problems. The semantic clustering module (SCM) leverages feature dimensionality reduction (FDR) to achieve automated online semantic segmentation of state spaces synchronously during training \citep{zhang_Enhancing_2025}. Nevertheless, this method is deeply integrated into the online training process, and its effectiveness in RTS environments remains unverified. Formation-aware exploration (FoX) partitions the joint state space into local observations and defines formations based on observation differences and various index sets \citep{jo_FoX_2024}. However, this formation definition is designed to serve efficient diversity exploration in state spaces and intrinsic formation maintenance. The applicability of min-max agent interaction relationships exhibits potential limitations in complex formations, and formation interpretability is constrained by feature dimensionality reduction and encoding.

Alternatively, from a result-oriented perspective, similar states should exhibit similar rewards and transitions. Elastic Monte Carlo Tree Search (MCTS) \citep{xu_Elastic_2023} and size-constrained state abstraction \citep{xu_Strategy_2024} achieve automatic state aggregation in RTS games by approximate Markov decision process (MDP) homomorphism. However, these approaches face limitations including reliance on sampling estimation, lack of semantic interpretability, and delayed processing.

Existing large-scale state abstraction methods often struggle with oversimplification or an excessive reliance on online learning. These limitations prevent them from offering the efficient and lightweight similarity metrics necessary for the rapid clustering of real-time state streams, particularly in complex environments of RTS micromanagement. Accordingly, it remains difficult for these methods to implement systematic raw data organization for tactic analysis.

\subsection{Interpretable analysis}
The decision-making process in RTS games is complex and multidimensional, characterized by coupled state-action interactions and long-horizon tactic dependencies \citep{wallner_Brief_2019}. This complexity has driven researchers to seek interpretability through both the discovery of tactical semantics and the development of visual analytics \citep{metoyer_Explaining_2010}. While tactical semantics discovery focuses on extracting behavioral patterns and causal relationships from trajectories, visual analytics transforms multidimensional insights into accessible graphical representations.

\subsubsection{Tactical semantics discovery}
The discovery of tactical semantics focuses on the algorithmic extraction of high-level logical patterns and causal relationships from entangled game trajectories. Specific data structures facilitate human-interpretable tactical representations, including derivation trees for automatic strategy script synthesis \citep{marino_Evolving_2022}, behavior trees for representing gameplay styles \citep{kozik_Mimicking_2021}, and network graphs for mapping step-by-step problem-solving trajectories \citep{kleinman_Understanding_2022}. However, these tree-based or network-based methods often incur high computational costs due to extensive simulations.

Beyond explicit structural modeling, another research trajectory explores the implicit discovery of tactics through high-dimensional feature encoding. Deep embeddings can map raw data into low-dimensional latent spaces. Building on this, clustering analysis on technology trees \citep{haley_Cluster_2020} and natural language episode tags \citep{mathes_CODEX_2023} can respectively facilitate game strategy identification and behavior summarization. Similarly, interpretable play-style descriptions can be generated through deep unsupervised clustering of player trajectories \citep{ingram_Generating_2023}. However, static clusters struggle to capture temporal dynamics, and the tactical semantics of clusters rely heavily on domain knowledge. Furthermore, latent representation clustering may struggle to capture the granular micro-management analysis required in RTS games.

\subsubsection{Visual analytics}

Visualizations involving only statistical data in RTS games cannot adequately present fine-grained strategic information and are not discussed in this paper. For a process-oriented presentation, visualization can be applied to low-level combat elements, including maps, states and action sequences. For instance, heatmaps overlaid on game maps visualize predictions of enemy targets under incomplete information \citep{izumigawa_Building_2020}. By using curved arrows for movements and attack lines between units, the system visualizes offensive and defensive micro-tactics, allowing for a granular analysis of formation effectiveness and unit-level interactions \citep{kuan_Visualizing_2017}. However, these visualizations focus on battlefield details, overlooking the identification and highlighting of similar tactical patterns.

Furthermore, visualization can also be macroscopic, such as fitness landscapes and spatio-temporal dynamics. Fitness landscape analysis in abstract RTS games can explicitly reveal the underlying problem structure \citep{keaveney_Analysing_2008}, yet the generalizability of these findings to extremely complex environments remains to be verified. Storyline visualization to summarize the movement of heroes has been introduced in League of Legends, effectively condensing complex spatial data into a temporal narrative of team cohesion \citep{wallner_Visualizing_2023}. VisuaLeague employs density-based spatial clustering of applications with noise (DBSCAN) to aggregate spatio-temporal events, enabling the identification of long-term strategic patterns and the analysis of player tendencies through interactive heatmaps \citep{afonso_VisuaLeague_2021}. Similarly, space-time cubes (STC) achieve 3D visualization for complex spatiotemporal tasks like critical game events and player trajectories \citep{sufliarsky_Space_2023}.

Although current interpretable game analysis covers low-level game elements and high-level spatio-temporal correlations, a unified architecture that bridges state, action, and tactical levels for cohesive analysis and lucid visualization has yet to be realized. This fragmentation limits the ability to discern how discrete low-level actions manifest as intricate high-level tactics.

\section{Proposed method}
\label{sec:Methodologies}

\begin{figure*}[th]
    \centering
    \includegraphics[width=\linewidth]{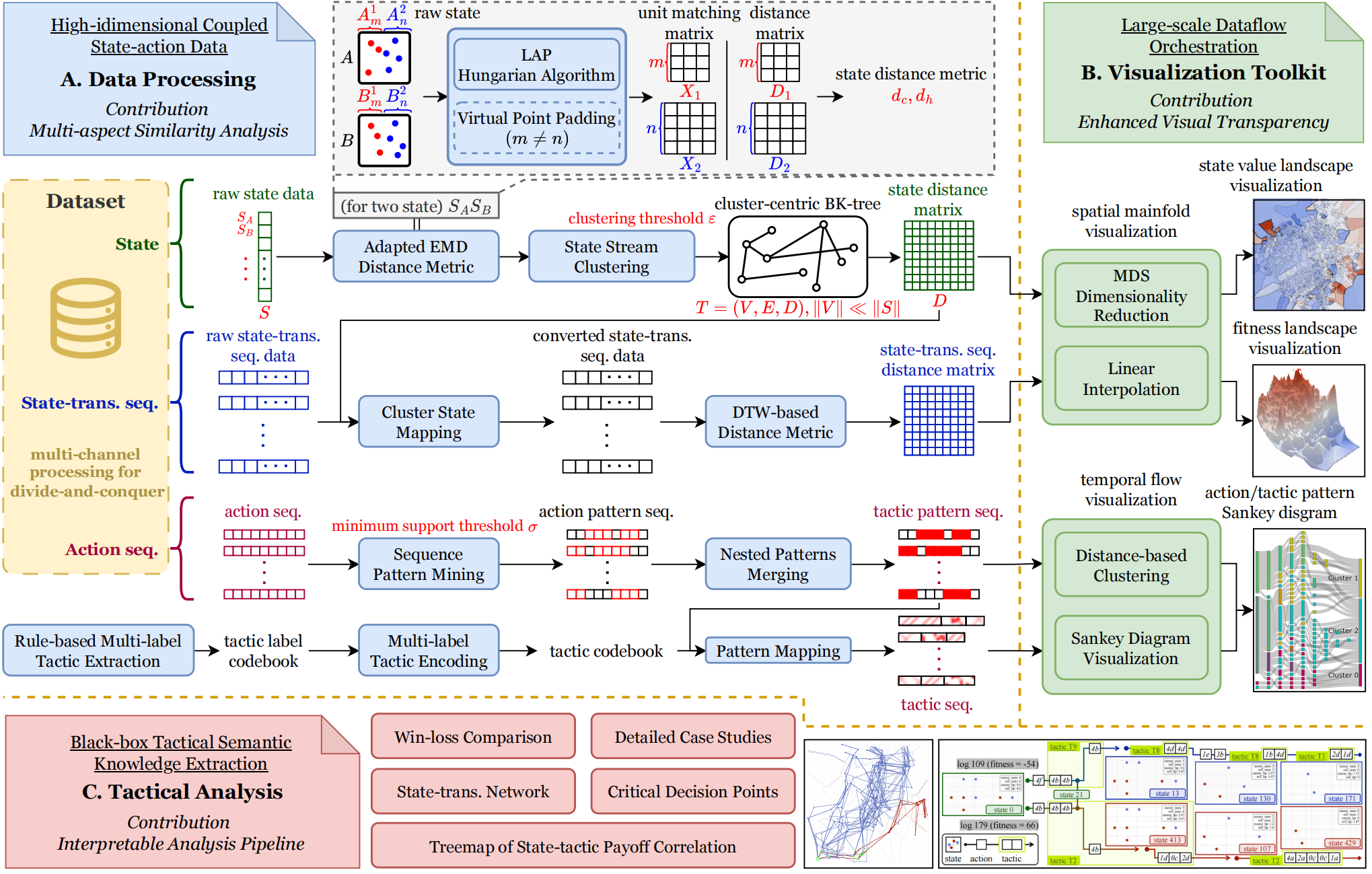}
    \vspace{-20pt}
    \caption{The overall architecture of the SAT-RTS pipeline}
    \label{fig:SAT-RTS framework}
\end{figure*}

To achieve an accessible multi-level tactical analysis in large-scale state space, this work proposes the multi-aspect analysis pipeline. This framework comprises battlefield situation similarity assessment based on the Hungarian algorithm \citep{kuhn_Hungarian_1955} and cluster-centric BK-trees, alongside fitness landscape visualization integrating DTW-based distance and dimensionality reduction. Additionally, the proposed framework incorporates action sequence pattern mining and tactic extraction with analysis. The overall architecture of the SAT-RTS pipeline is illustrated in Figure~\ref{fig:SAT-RTS framework}.

\subsection{Battlefield situation similarity assessment}

\subsubsection{Adapted EMD distance metric based on the Hungarian algorithm}
\label{sec:Adapted EMD Distance Metric based on the Hungarian Algorithm}

Within the StarCraft~II environment, the battlefield situation is represented as a state vector through the observer, which typically comprises the three-dimensional coordinates and current HP of each unit from both players. However, due to limitations in the underlying engine, only global observation information can be obtained, and unit matching cannot be guaranteed. Consequently, the inability to determine the corresponding units from the state vectors poses a challenge for state distance measurement, specifically the combat unit matching problem. Wasserstein distance, also known as Earth Mover's Distance (EMD), utilizes optimal transport theory to measure the difference between two probability distributions. This concept can be extended to optimal unit matching in point sets to achieve distance measurement between unit sets.

This work proposes the adapted EMD for game state distance metric, where the unit matching problem can be transformed into a linear assignment problem (LAP) by constructing a cost matrix where entries represent the Euclidean distance between units in different states. Given two states $S_A$ and $S_B$, let point set 
\begin{equation}
\label{eq:state_A}
    A = A^1_m+A^2_n = \{a^1_1, a^1_2, ..., a^1_m\} + \{a^2_1, a^2_2, ..., a^2_n\}
\end{equation}
represents the positions of $m$ units from player 1 and $n$ units from player 2 in state $S_A$, and point set 
\begin{equation}
\label{eq:state_B}
    B = B^1_m+B^2_n = \{b^1_1, b^1_2, ..., b^1_m\} + \{b^2_1, b^2_2, ..., b^2_n\}
\end{equation}
represents the positions of $m$ units from player 1 and $n$ units from player 2 in state $S_B$. For the distance matrices, let $D^1$ denote the $m \times m$ distance matrix and $D^2$ denote the $n \times n$ distance matrix, where $D^1_{ij}$ represents the Euclidean distance between unit $i$ in $S_A$ and unit $j$ in $S_B$ for player 1, and $D^2_{ij}$ represents the Euclidean distance between unit $i$ in $S_A$ and unit $j$ in $S_B$ for player 2. The objective is to determine binary assignment matrices $X^1$ and $X^2$ that minimize the total sum of distances between all matched units
\begin{equation}
\label{eq:objective}
    \min (\sum_{i=1}^{m} \sum_{j=1}^{m} D^1_{ij} X^1_{ij} + \sum_{i=1}^{n} \sum_{j=1}^{n} D^2_{ij} X^2_{ij}),
\end{equation}
subject to:
\begin{align}
\sum_{j=1}^{m} X^1_{ij} &= 1 \quad \forall i \in \{1, 2, \ldots, m\},\label{eq:constraint1} \\ 
\sum_{i=1}^{m} X^1_{ij} &= 1 \quad \forall j \in \{1, 2, \ldots, m\},\label{eq:constraint2} \\
\sum_{j=1}^{n} X^2_{ij} &= 1 \quad \forall i \in \{1, 2, \ldots, n\},\label{eq:constraint3} \\
\sum_{i=1}^{n} X^2_{ij} &= 1 \quad \forall j \in \{1, 2, \ldots, n\},\label{eq:constraint4} \\
X^1_{ij} &\in \{0, 1\} \quad \forall i, j \in \{1, 2, \ldots, m\},\label{eq:constraint5} \\
X^2_{ij} &\in \{0, 1\} \quad \forall i, j \in \{1, 2, \ldots, n\},\label{eq:constraint6}
\end{align}
where the constraints ensure that each unit for both players is matched exactly once between the two states.

To address scenarios with inconsistent unit quantities, the adapted EMD metric extends the Hungarian-based assignment by introducing virtual point padding. Given two point sets $A = A^1_m+A^2_n$ and $B = B^1_{m'}+B^2_{n'}$ with potentially different cardinalities, this work first extends the smaller set to match the larger cardinality by adding virtual points
\begin{align}
&|A^1_{m^*}|=|B^1_{m^*}|=\max(m,m'), \\ 
&|A^2_{n^*}|=|B^2_{n^*}|=\max(n,n'),
\end{align}
where virtual points are positioned at distance $D_v$ from all real points, with $D_v \gg \max_{i,j} d(a_i,b_j)$ to explicitly enhance the distance difference between the states of unit scale mismatch.

For comprehensive state distance calculation in RTS scenarios, this work incorporates both coordinate distribution and HP differences. After optimal matching is achieved, the unit coordinate distribution distance $d_c$ and unit HP difference $d_h$ can be expressed as
\begin{equation}
\label{eq:8}
     d_c = \sum_{i=1}^{m} \sum_{j=1}^{m} D^1_{ij} X^1_{ij} + \sum_{i=1}^{n} \sum_{j=1}^{n} D^2_{ij} X^2_{ij}
\end{equation}
and
\begin{equation}
\label{eq:9}
d_h =  \sum_{i=1}^{m} \sum_{j=1}^{m} \left| H^{1}_{Ai} - H^{1}_{Bj} \right| X^1_{ij} + \sum_{i=1}^{n} \sum_{j=1}^{n} \left| H^{2}_{Ai} - H^{2}_{Bj} \right| X^2_{ij},
\end{equation}
where $H^{1}_{Ai}$ and $H^{1}_{Bj}$ represent the HPs of units for player 1 in states $S_A$ and $S_B$, respectively, and $H^{2}_{Ai}$ and $H^{2}_{Bj}$ represent the HPs for player 2.

\begin{figure*}[th]
	\centering
		\centering
		\includegraphics[width=\linewidth]{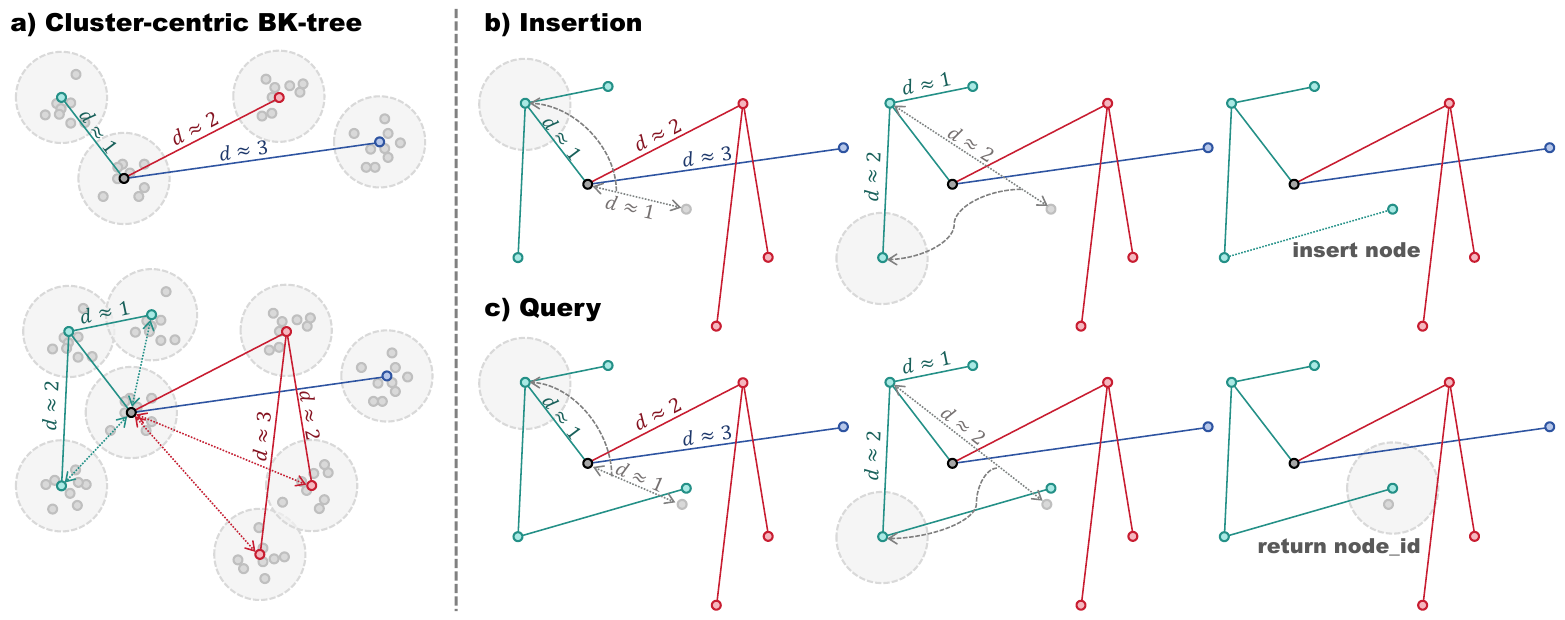}
    \vspace{-20pt}
    \caption{Schematic diagram of the cluster-centric BK-tree with insertion and query}
    \label{fig:bktree}
\end{figure*}

The Hungarian algorithm provides $O(n^3)$ time complexity for the optimal assignment problem, significantly improving upon the $O(n^3 \log n)$ complexity of traditional linear programming approaches for EMD computation. From a practical RTS game perspective, the construction of virtual points in cases of inconsistent unit quantities results in larger state distance values, which conforms to the tactical intuition that mismatched unit counts represent fundamentally different battlefield situations. This characteristic enhances the applicability of threshold-based clustering methods while maintaining theoretical foundation in optimal transport theory.

\subsubsection{Stream clustering based on the cluster-centric BK-tree}

The extensive simulation in RTS games generates a vast influx of raw states, which requires efficient battlefield situation grouping for management. To address this requirement, a stream clustering algorithm termed the cluster-centric BK-tree is proposed for RTS game state. Unlike conventional methods that store exhaustive state information, the cluster-centric BK-tree retains only representative cluster centers based on the state distance metric and clustering threshold. This selective maintenance significantly reduces the memory overhead and accelerates the insertion and query of nodes, thereby facilitating real-time grouping of large-scale game states.

The BK-tree is a metric tree traditionally optimized for spell checking and fuzzy searching via triangle inequality-based pruning \citep{burkhardwa_Approaches_1973}. In this work, the BK-tree is adapted for the stream clustering of RTS game states by redefining nodes as cluster centers rather than individual data points. Furthermore, the conventional edit distance is replaced by a specialized state distance metric, as elaborated in Eqs. (\ref{eq:8}) and (\ref{eq:9}), to accommodate the high-dimensional nature of battlefield situations. This adaptation leverages the efficient recursive partitioning of the BK-tree to achieve rapid insertion and query.

As shown in Figure \ref{fig:bktree}, the cluster-centric BK-tree algorithm maintains a metric tree and introduces a clustering threshold $\epsilon$ to store only the cluster center nodes. The tree can be defined as $T=(V, E, D)$, where $V$ is a finite non-empty set representing the set of nodes, $E$ is the set of pairs of nodes in $V$, representing the set of edges and $E \subseteq \{(u, v, d) \mid u, v \in V, u \neq v, d=D(u, v)\}$, $D$ is a custom distance metric representing the distance associated with each edge.

Figure \ref{fig:bktree} b) and Figure \ref{fig:bktree} c) illustrate the insertion and query processes of the cluster-centric BK-tree, respectively. The algorithm process is described as follows:
\begin{enumerate}[leftmargin=33pt,label={Step \arabic*:},itemindent=0pt,parsep=0pt]
    \item Initialize the root node $V_0$ with the initial game state.
    \item For the new state $V_i$, compute the distance $d=D(V_0, V_i)$ and round it to $d'=\lfloor d+0.5\rfloor$.
        \begin{itemize}[label=$\bullet$]
            \item If $d' \leq \epsilon$, return $V_0$.
            \item Otherwise, proceed to Step 3.
        \end{itemize}
    \item For node $V_0$, check if there exists node $V_j$ and $(V_0, V_j, \allowbreak D(V_0, V_j)) \in E_{\text{out}}(V_0)$ such that $d' = D(V_0, V_j)$, expressed by the following statement:
    \begin{equation*}
        \begin{aligned}
        &\exists V_j \in V_{\text{out}}(V_0) \mid (V_0, V_j, D(V_0, V_j)) \in E_{\text{out}}(V_0), \quad \\
        &\text{s.t.} \quad d' = D(V_0, V_j).
        \end{aligned}
    \end{equation*}
        \begin{itemize}[label=$\bullet$]
            \item If such a node exists, set $V_j$ as the new root and repeat Step 2.
            \item Otherwise, proceed to Step 4.
        \end{itemize}
    \item Insert $V_i$ as a child of $V_0$ and create the edge $(V_0, V_i, d(V_0, \allowbreak V_i))$. Return $V_i$.
\end{enumerate}

The cluster-centric BK-tree algorithm handles large-scale data by retaining only the cluster center data through a clustering threshold. It achieves rapid query of state grouping through efficient pruning. The structure of the metric tree enables it to add nodes instantaneously without the need for frequent adjustment. It exhibits low computational complexity and superior real-time performance, thereby fulfilling the computational requirements for stream clustering of game states in RTS games and realizing the similarity assessment of battlefield situation. In terms of clustering accuracy, although the clustering threshold $\epsilon$ is introduced, the construction of extremely distant virtual points in the designed state distance metric makes this clustering method still applicable to the scenarios discussed in this work.

Furthermore, Eqs. (\ref{eq:8}) and (\ref{eq:9}) provide two distance metrics for game states, representing the distance of unit coordinate distribution and the difference in unit HPs, respectively. In order to clearly represent the similarity between game states, this work employs a hierarchical construction based on the cluster-centric BK-tree. Specifically, the primary clustering is conducted according to the $d_c$ from Eq. (\ref{eq:8}) to group game states based on unit distribution. Subsequently, the secondary clustering is carried out according to the $d_h$ from Eq. (\ref{eq:9}) to further group game states based on the differences in HPs. By implementing the hierarchical construction of the cluster-centric BK-tree, both coarse-grained and fine-grained classifications of game states can be obtained.

\subsection{State-transition sequence similarity assessment}
\label{sec:State-transition sequence similarity assessment}

\subsubsection{State-transition sequence distance metric}
The design of the state distance metrics and the state clustering algorithm efficiently addresses the issue of assessing the similarity of battlefield situations and achieve a substantial reduction in the size of state space. Benefiting from this, the similarity between state-transition sequences, which correspond to the unfolding of combat events in RTS game micromanagement, can be analyzed.

Dynamic Time Warping (DTW) is an effective technique for non-linear sequence alignment, traditionally utilized in trajectory matching \citep{sakoe_Dynamic_1978}. This work extends the application of DTW to the domain of RTS game state-transition sequences. By integrating a specialized state metric, the similarity of state-transition sequence can be effectively assessed.

To implement DTW, a cost matrix $D$ is constructed where $D[i][j]$ represents the accumulated cost between the first $i$ states of the first sequence and the first $j$ states of the second sequence. The matrix is initialized with $D[0][0]=0$ and remaining elements in the first row and column set to infinity. Each element is computed using the state distance metric $d$ and the minimum accumulated cost from preceding states. The optimal matching path is found by backtracking from $D[m][n]$ to $D[0][0]$, where the final matrix value represents the overall sequence distance, with smaller values indicating higher similarity.

\subsubsection{Fitness landscape visualization}
\label{sec:Fitness landscape visualization}

RTS game micromanagement represents a heavily constrained optimization problem where exhaustive inspection of the full search space is computationally infeasible. To circumvent this challenge, the analysis is anchored on the set of solutions sampled by trained RL algorithms. Given that these algorithms effectively concentrate on high-fitness regions, the distributional modality of optimal solutions can be faithfully reflected within these focused areas. This targeted approach ensures that the most relevant problem characteristics can be captured.

\begin{figure*}[th]
	\centering
	\begin{minipage}{0.41\linewidth}
		\centering
		\includegraphics[width=\linewidth]{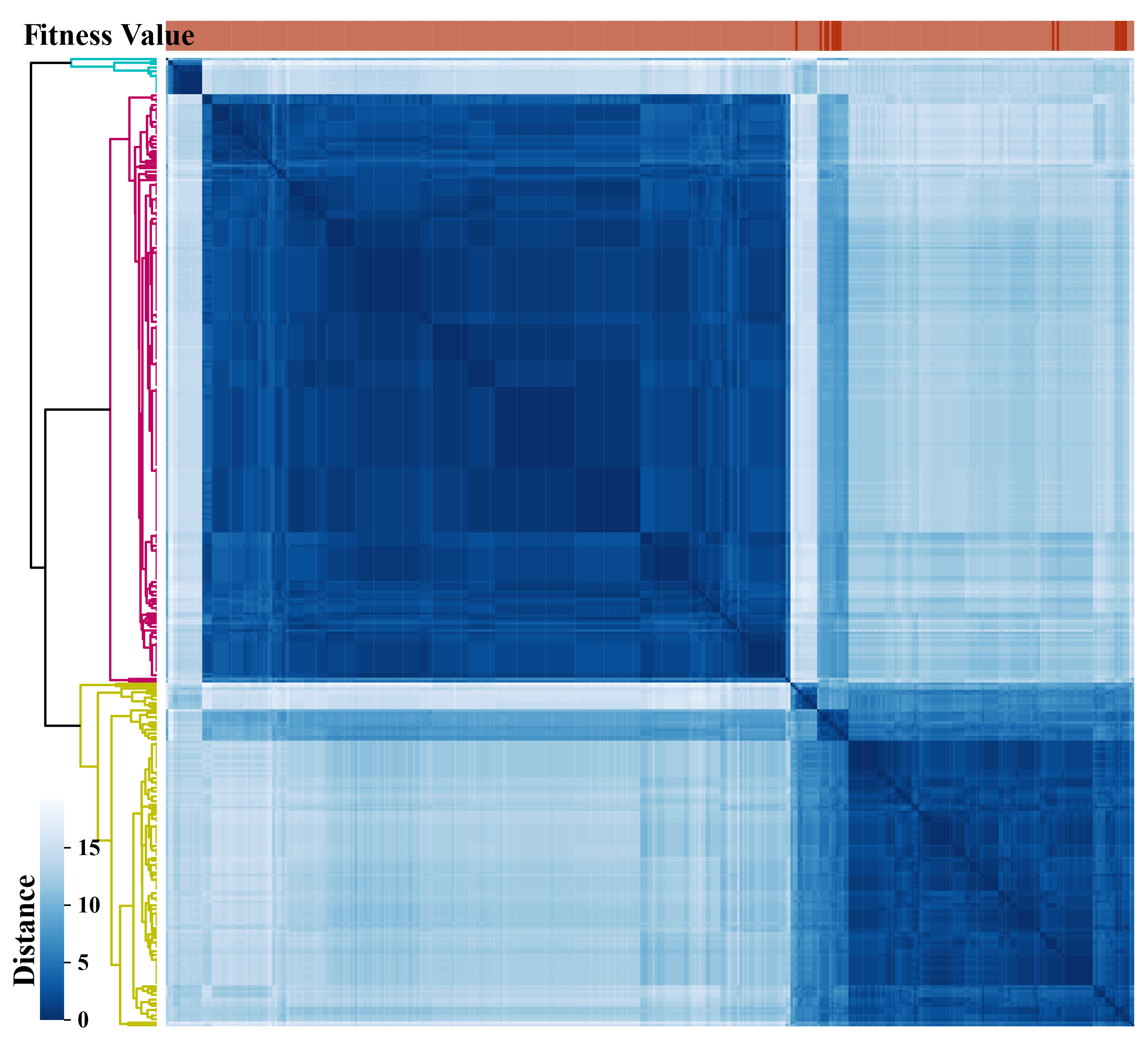}
		a) Heatmap of distance of state-transition sequences with clustering
		\label{fig:heatmap}
	\end{minipage}
	\hfill
	\begin{minipage}{0.46\linewidth}
		\centering
		\includegraphics[width=\linewidth]{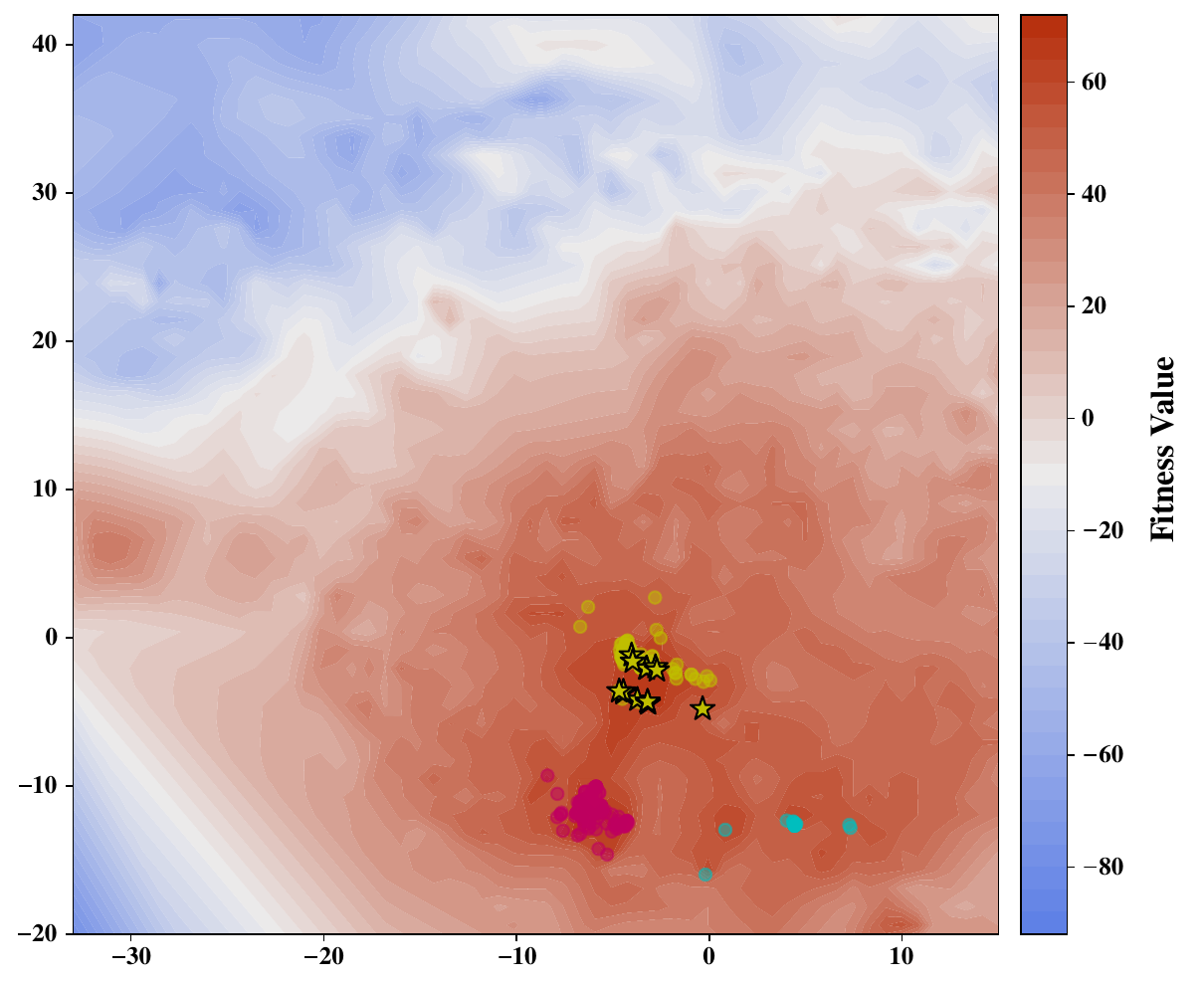}
		b) Fitness landscape with global (sub-)optima with clustering
		\label{fig:fitness_landscape_2d}
	\end{minipage}
	\hfill
	\begin{minipage}{0.11\linewidth}
		\centering
		\includegraphics[width=\linewidth]{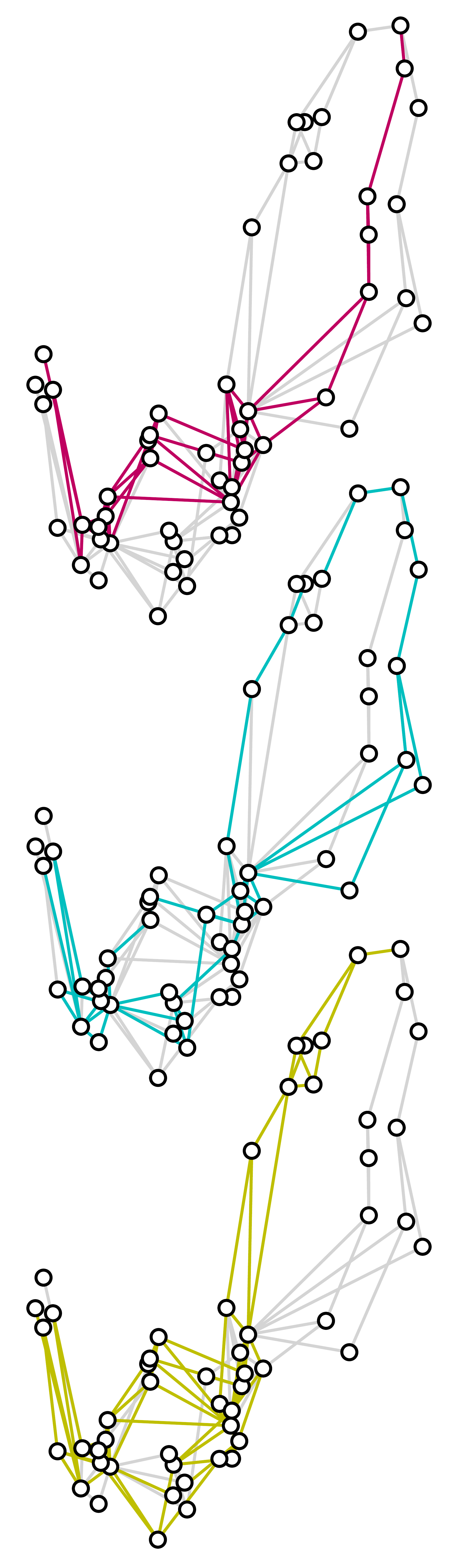}
		c) Actual sequences
		\label{fig:all_clusters}
	\end{minipage}
	\caption{State-transition sequence distance metric and fitness landscape visualization}
	\label{fig:fitness landscape visualization}
    \vspace{-10pt}
\end{figure*}

This work implements fitness landscape visualization based on the distance metric defined between state-transition sequences. By integrating this metric with multidimensional scaling (MDS), the sampled solution distribution is embedded into a low-dimensional manifold, with linear interpolation subsequently employed to approximate the fitness landscape within the convex hull of the points. Notably, global (sub-)optima are presented as illustrative examples to enhance the interpretability of the proposed method. As shown in Figure \ref{fig:fitness landscape visualization}, the visualization encompasses three key elements:
\begin{enumerate}
    \item[a)] A distance heatmap of exemplar solutions accompanied by its derived hierarchical clustering.
    \item[b)] The fitness landscape generated via MDS and linear interpolation, marking the positions of global (sub-)optima.
    \item[c)] The actual state-transition sequences, where nodes are spaced by inter-state distances and colored edges represent different clusters.
\end{enumerate}

The proposed visualization method is analogously applicable to the visualization of state value landscapes. In this context, the analytical focus shifts from the search space of sequential solutions to the state space of combat scenarios. Specifically, state values are calculated based on the differential in average total HP between opposing sides, providing a quantitative metric of snapshot advantage. By applying the same dimensionality reduction and interpolation pipeline to the distance matrix of clustered states, the state value landscape can be constructed to capture the essential characteristics of the state space.

\subsection{Tactic similarity assessment}

\subsubsection{Action sequence pattern mining}

Beyond analyzing state-transition trajectories, mining action sequence patterns offers significant insights into the decision-making space by capturing the underlying tactical logic of agents. In RTS micromanagement, tactics are often implicitly represented as specific combinations of actions. Although different tactical inclinations may yield identical objectives, they manifest as distinct action sequence patterns. By capturing these patterns, particularly the coupling between individual actions and collective interests, similar tactics can be effectively categorized and compared.

Given the strict adjacency constraints in tactical execution, this work employs an exhaustive search with a minimum support threshold ($\sigma$) to mine sequential action patterns. Unlike candidate-generation (e.g., Apriori \citep{agrawal_Fast_1994}) or projection-based paradigms (e.g., PrefixSpan \citep{jianpei_PrefixSpan_2001}), this approach ensures a complete enumeration of patterns for subsequent analysis. To mark the original sequences, nested patterns are merged based on a bi-objective optimization criterion that aims to maximize coverage length while minimizing the total number of patterns. 

In alignment with the discussion of fitness landscape visualization, the global (sub-)optima are again employed here as illustrative examples to maintain consistency and enhance interpretability. As shown in Table \ref{tab:Examples of original and marked action sequences with sequential action patterns}, the action sequences of these exemplar solutions are presented alongside the extracted sequential action patterns. These patterns, identified as frequent occurrences across the entire sampled solution set, are used to mark the original sequences. To ensure the most representative tactical marking, nested patterns are merged based on a bi-objective optimization criterion that maximizes coverage length while minimizing the total number of patterns.

\subsubsection{Tactic extraction and analysis}

While tactic execution is a forward process, extracting implicit tactics from action sequences remains challenging. This work refers to tactic extraction as the identification of actionable knowledge from high-fitness solutions. By integrating mined action sequence patterns with state-transition similarity and fitness landscape visualizations, this work analyzes the tactical consistency across similar solutions to facilitate a comprehensive understanding of decision-making logic.

\begin{table*}[thbp]
\centering
\renewcommand{\arraystretch}{0.9}
\caption{Examples of original and marked action sequences with sequential action patterns}
\label{tab:Examples of original and marked action sequences with sequential action patterns}
\begin{tabular}{>{\centering\arraybackslash}m{0.025\textwidth}>{\centering\arraybackslash}m{0.44\textwidth}p{0.46\textwidth}}
\toprule
\multicolumn{1}{c}{ID} & \multicolumn{1}{c}{Original/Marked action sequences} & \multicolumn{1}{c}{Sequential action patterns} \\
\midrule

\multirow{2}{*}[-10pt]{A1} & 
\multicolumn{1}{>{\vspace*{1.0mm}}m{0.44\textwidth}}{$4b4b4b4j1j0c1j2j4d2e2d3d1d1d1d1d0c$} & \multirow{2}{0.46\textwidth}{$\langle4b4b4b4j1j0c\rangle$, $\langle4b4b4b4j1j\rangle$, $\langle4b4b4j1j0c\rangle$,\\ $\langle2d3d1d1d1d1d\rangle$, $\langle2d3d1d1d1d\rangle$, $\langle3d1d1d1d\rangle$,\\ $\langle1j2j\rangle$, $...$} \\[4pt]
\cmidrule(lr){2-2}
& \multicolumn{1}{>{\vspace*{1.0mm}}m{0.44\textwidth}}{
  $[4b4b4b4j1j0c]$$[1j2j]$$4d2e$$[2d3d1d1d1d1d]$$0c$} & \\[4pt]
\midrule

\multirow{2}{*}[-10pt]{A2} & 
\multicolumn{1}{>{\vspace*{1.0mm}}m{0.44\textwidth}}{$4b4b4b4j1j0c1j2j4d2d2d3d1d1d2d1d3d$} & \multirow{2}{0.46\textwidth}{$\langle4b4b4b4j1j0c\rangle$, $\langle4b4b4b4j1j\rangle$, $\langle4b4b4j1j0c\rangle$, $\langle2d2d3d1d1d\rangle$, $\langle2d3d1d1d\rangle$, $\langle2d2d3d1d\rangle$, \\ $\langle2d1d3d\rangle$, $\langle2d1d\rangle$, $\langle1d3d\rangle$, $\langle1j2j\rangle$, $\langle0c3a\rangle$, $...$} \\[4pt]
\cmidrule(lr){2-2}
& \multicolumn{1}{>{\vspace*{1.0mm}}m{0.44\textwidth}}{
  $[4b4b4b4j1j0c]$$[1j2j]$$4d$$[2d2d3d1d1d]$$[2d1d3d]$} & \\[4pt]
\midrule

\multirow{2}{*}[-10pt]{B1} & 
\multicolumn{1}{>{\vspace*{1.0mm}}m{0.44\textwidth}}{$4b1b1b4b4b0h0e2d1d4a2a0c1a1a1a0c$} & \multirow{2}{0.46\textwidth}{$\langle4b1b1b4b4b0h0e2d\rangle$, $\langle4b1b1b4b4b0h0e\rangle$,\\$\langle1b4b4b0h\rangle$, $\langle4a2a0c\rangle$, $\langle4a2a\rangle$, $\langle2a0c\rangle$, \\$\langle1a1a1a\rangle$, $\langle1a1a\rangle$, $...$} \\[4pt]
\cmidrule(lr){2-2}
& \multicolumn{1}{>{\vspace*{1.0mm}}m{0.44\textwidth}}{
  $[4b1b1b4b4b0h0e2d]$$1d$$[4a2a0c]$$[1a1a1a]$$0c$} & \\[4pt]
\midrule

\multirow{2}{*}[-10pt]{B2} & 
\multicolumn{1}{>{\vspace*{1.0mm}}m{0.44\textwidth}}{$4b1b1b4b4b0h0e2d4a4a2a0c2a2a3a1a$} & \multirow{2}{0.46\textwidth}{$\langle4b1b1b4b4b0h0e2d\rangle$, $\langle4b1b1b4b4b0h0e\rangle$, \\$\langle1b4b4b0h\rangle$, $\langle4a4a2a0c\rangle$, $\langle4a4a2a\rangle$, $\langle4a2a0c\rangle$,\\ $\langle2a2a3a1a\rangle$, $\langle2a2a3a\rangle$, $\langle2a3a1a\rangle$, $...$} \\[4pt]
\cmidrule(lr){2-2}
& \multicolumn{1}{>{\vspace*{1.0mm}}m{0.44\textwidth}}{
  $[4b1b1b4b4b0h0e2d]$$[4a4a2a0c]$$[2a2a3a1a]$} & \\[4pt]

\bottomrule
\end{tabular}
\end{table*}

\begin{figure}[th]
	\centering
		\centering
		\includegraphics[width=0.92\linewidth]{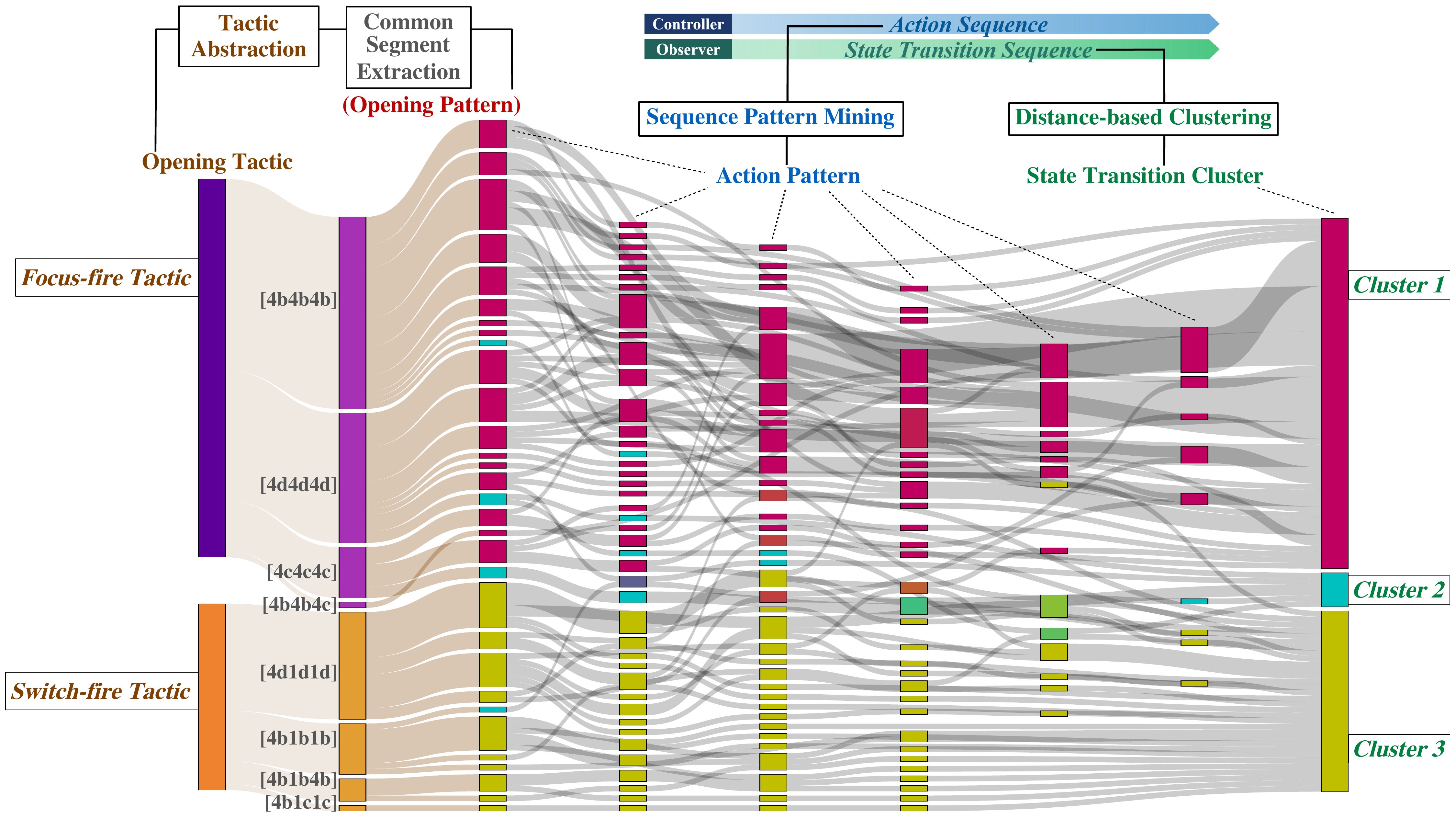}
    \vspace{-10pt}
    \caption{Sankey diagram of action sequence patterns and tactic extraction}
    \label{fig:action_pattern}
    \vspace{-10pt}
\end{figure}

Consistent with previous examples, the previously identified global (sub-)optima are further examined to facilitate tactic extraction. Table \ref{tab:Examples of original and marked action sequences with sequential action patterns} and Figure~\ref{fig:action_pattern} characterize these solutions by depicting the temporal flow of mined patterns. Specifically, the first two columns link manually identified opening tactics to their action segments (e.g., focus-fire for [4b4b4b] and switch-fire for [4b1b1b]). The strong correlation between these patterns and the cluster labels in Figure~\ref{fig:fitness landscape visualization} reveals a causal link between action sequences and the resulting distributional modality, highlighting the decisive impact of opening tactics on combat outcomes.

Sankey diagrams provide an organized visualization for analyzing combat elements. In each diagram, the first column on the left represents tactics extracted from common segments of opening action patterns, which are themselves displayed in the second column. The rightmost column denotes the cluster labels of samples, while the intermediate columns represent all mined frequent action sequence patterns. The flow lines between these rectangular blocks delineate specific causal and temporal relationships, including the generative dependency between tactics and segments, the temporal progression of action sequences, and the attribution of samples to cluster labels. The color of blocks reflects their distribution across clusters, where color consistency with cluster labels indicates strong correlation.

In addition, this work implements a rule-based tactic extraction approach to ensure scalability and reproducibility, which comprises three main steps:
\begin{enumerate}[leftmargin=33pt,label={Step \arabic*:},itemindent=0pt, itemsep=0pt]
    \item Establish a human-readable rule set where each rule generates a tactic label based on action sequence characteristics. The tactic label codebook is presented in \text{Table~S3} in Section S4 of the Supplementary Material.
    \item Map each pattern to interpretable tactical concepts via the rule set, yielding a multi-label encoding.
    \item Combine activated labels into a unified tactical signature for large-scale analysis.
\end{enumerate}
This formalization facilitates robust tactic comparison across diverse solutions without manual annotation.

\section{Experimental results and analysis}
\label{sec:Experimental Results and Analysis}

This section presents comprehensive experimental results and analysis, mainly including the effectiveness and efficiency of stream clustering with the cluster-centric BK-tree under different distance metrics, visualization of state-value landscapes and fitness landscapes, analysis of action sequence and tactic pattern flows, and interpretable analysis of tactic extraction.

\subsection{Experimental design and parameter sensitivity analysis}

Both synthetic data and actual collected data have been utilized for comprehensive validation of the proposed method. Spe-cifically, synthetic data with true cluster labels enables targeted assessment of clustering accuracy under different distance metrics. For empirical validation, 8,100 complete small-scale combat simulations were conducted using the StarCraft Multi-Agent Challenge (SMAC) platform\footnote{\url{https://github.com/oxwhirl/smac}.}. These simulations spanned four distinct StarCraft~II scenarios and were implemented through specific RL algorithms.

\begin{figure}[ht]
	\centering
	\begin{minipage}{0.24\linewidth}
		\centering
		\includegraphics[width=\linewidth]{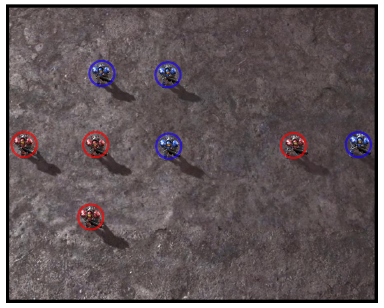}
		a) sce1
	\end{minipage}
	\begin{minipage}{0.24\linewidth}
		\centering
		\includegraphics[width=\linewidth]{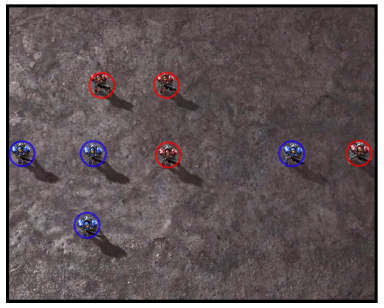}
		b) sce1m
	\end{minipage}
    \begin{minipage}{0.24\linewidth}
		\centering
		\includegraphics[width=\linewidth]{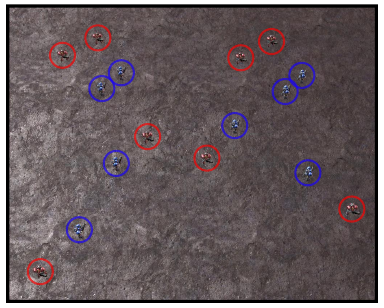}
		c) sce2
	\end{minipage}
	\begin{minipage}{0.24\linewidth}
		\centering
		\includegraphics[width=\linewidth]{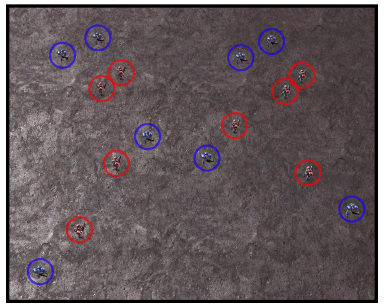}
		d) sce2m
	\end{minipage}
    \caption{Designed experimental scenarios with unit distributions of both sides swapped in mirror maps}
    \label{fig:scenarios}
    \vspace{-10pt}
\end{figure}

All experiments and data processing tasks were performed on a high-performance workstation equipped with an AMD Ryzen 9 7900X 12-Core Processor and 64.0 GB of RAM. No GPU acceleration was employed during the data organization and clustering phases. During these simulations, all raw states and state-transition sequences were recorded. The combat scale and the number of sampled raw game states are summarized in Table~\ref{tab:Combat scale and number of sampled raw states across scenarios}. The designed experimental scenarios, in which the unit distribution of both sides is swapped in mirror maps, are illustrated in Figure~\ref{fig:scenarios}.


\begin{table}[htbp]
\vspace{-10pt}
\centering
\caption{Combat scale and number of sampled raw game states across scenarios}
\label{tab:Combat scale and number of sampled raw states across scenarios}
\renewcommand{\arraystretch}{0.9}
\setlength{\tabcolsep}{4pt}
\footnotesize
\begin{tabular}{@{}lcccc@{}}
\toprule
& \textit{sce1} & \textit{sce1m} & \textit{sce2} & \textit{sce2m} \\
\cmidrule{2-5}
Combat scale & M4 vs M4 & M4 vs M4 & M8 vs M8 & M8 vs M8 \\
Number of states & 80175 & 84788 & 127299 & 111745 \\
\bottomrule
\end{tabular}
\vspace{-10pt}
\end{table}

\begin{figure}[ht]
    \centering
    \begin{minipage}{0.24\linewidth}
        \centering
        \includegraphics[width=\linewidth]{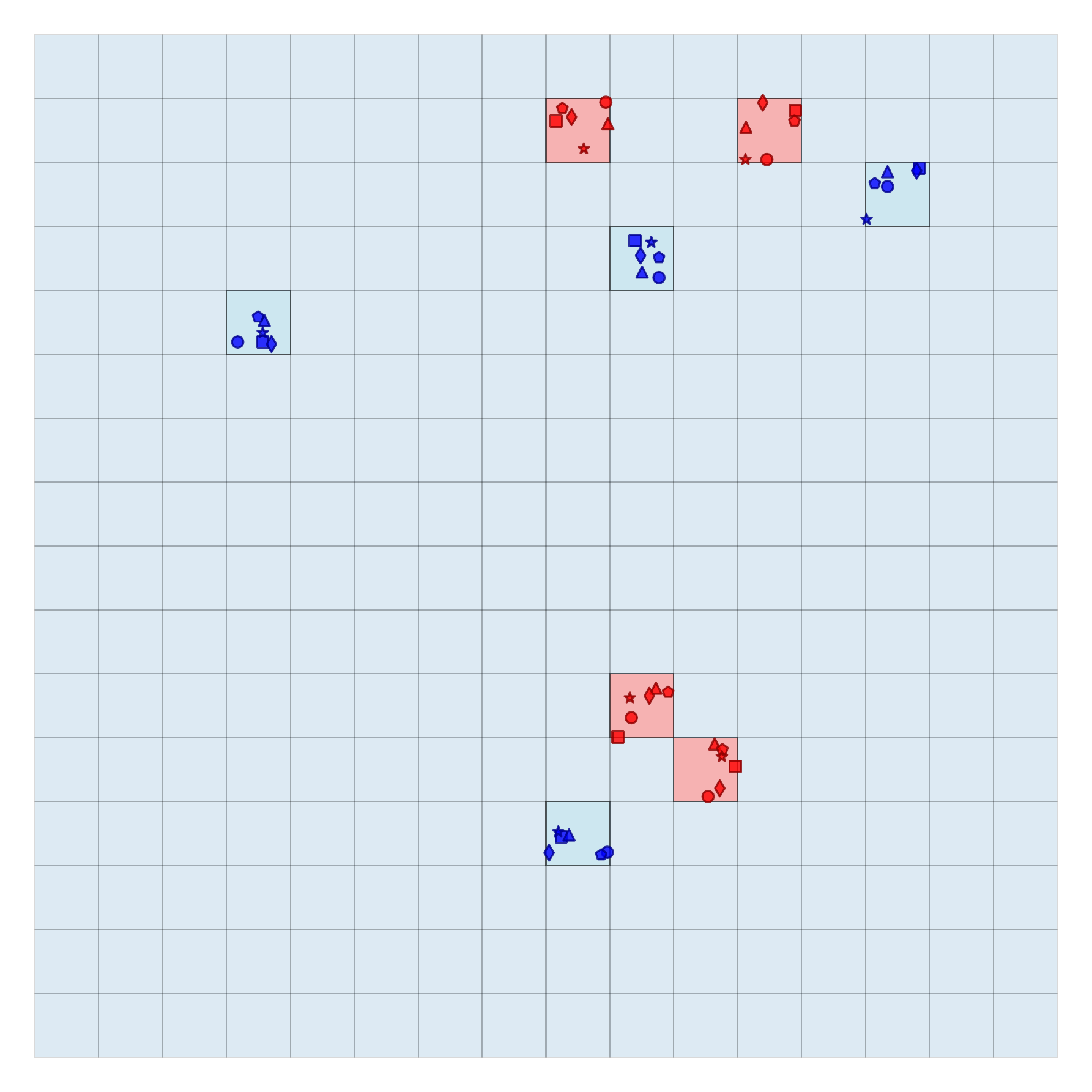}
    \end{minipage}
    \begin{minipage}{0.24\linewidth}
        \centering
        \includegraphics[width=\linewidth]{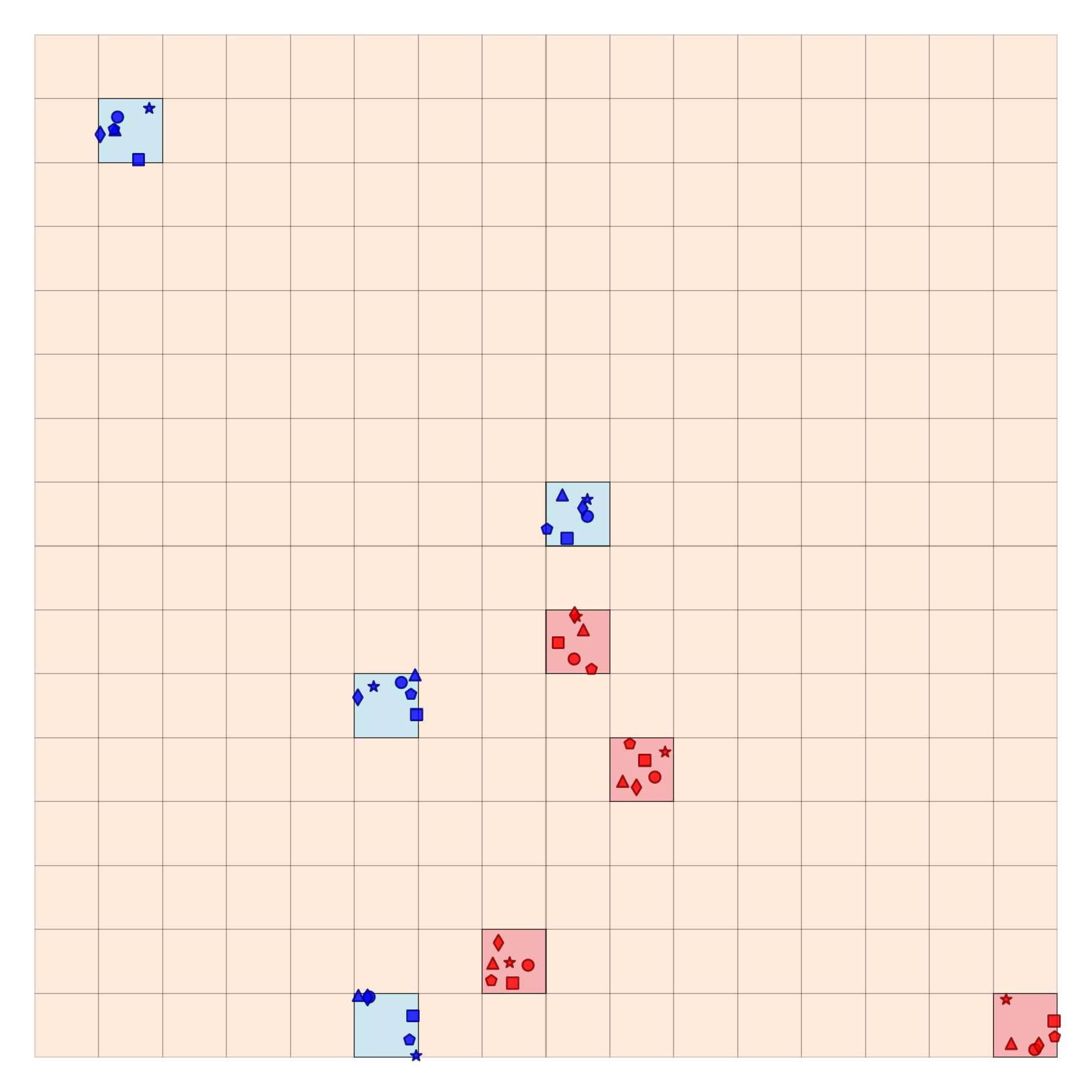}
    \end{minipage}
    \begin{minipage}{0.24\linewidth}
        \centering
        \includegraphics[width=\linewidth]{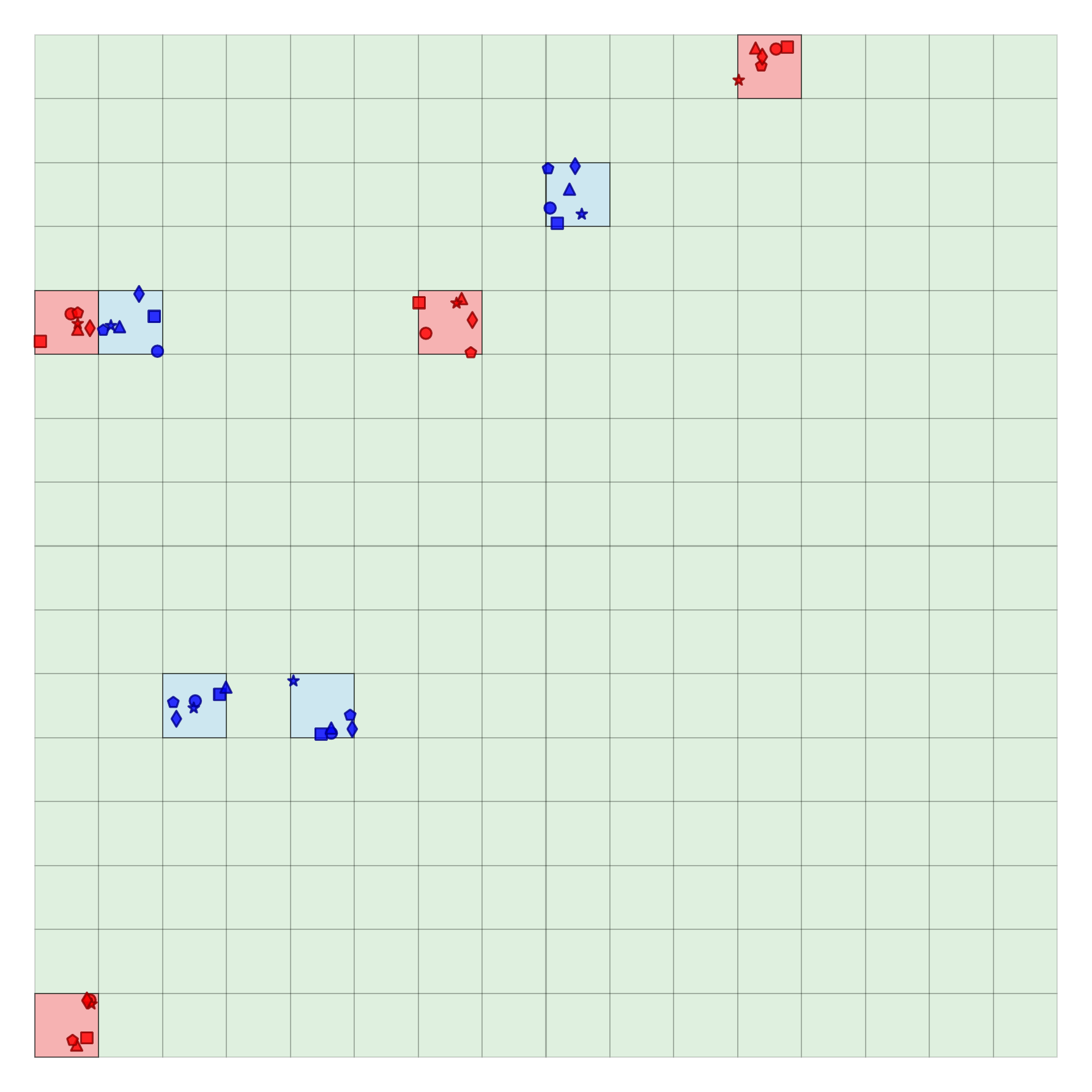}
    \end{minipage}
    \begin{minipage}{0.24\linewidth}
        \centering
        \includegraphics[width=\linewidth]{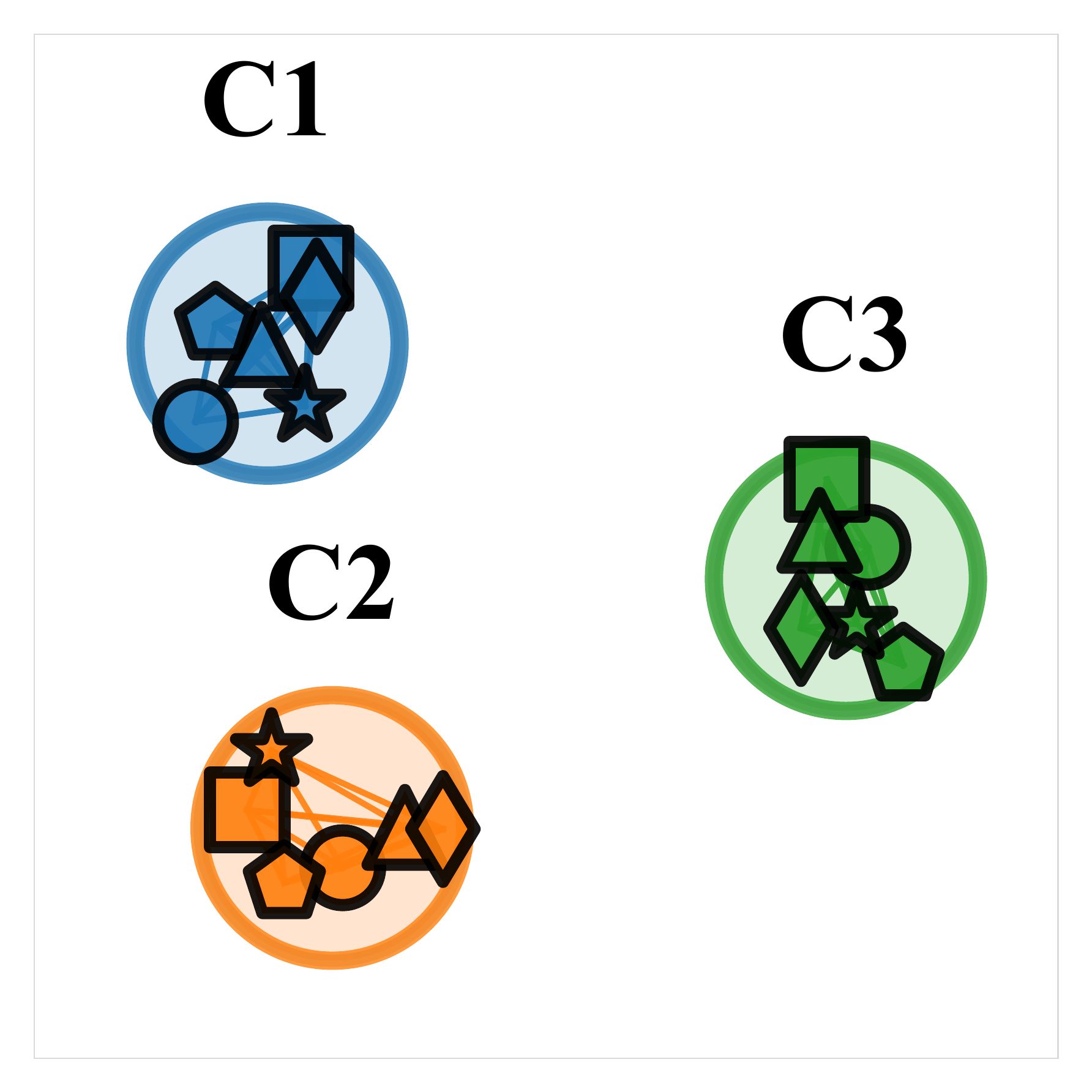}
    \end{minipage}
    \caption{Illustration of synthetic data and clustering results for three true cluster labels with six samples per cluster}
    \label{fig:classification_example}
\end{figure}

Experiments on synthetic datasets with true clustering labels are performed to comprehensively evaluate the effectiveness of the proposed methods. The representation of RTS game state can be formalized as the spatial unit distribution. Batch data is generated through grid partitioning combined with random point selection, where samples with the same selected grids share the same true cluster labels. Figure~\ref{fig:classification_example} shows the data generation and clustering results for three clusters. Each cluster containing six samples, utilizing red markers for player 1 units and blue markers for player 2 units. Specifically, combat unit scales across different clusters are allowed to be inconsistent, facilitating the synthetic data better reflecting actual situations for validating clustering effectiveness.

\begin{figure}[ht]
	\centering
		\centering
		\includegraphics[width=\linewidth]{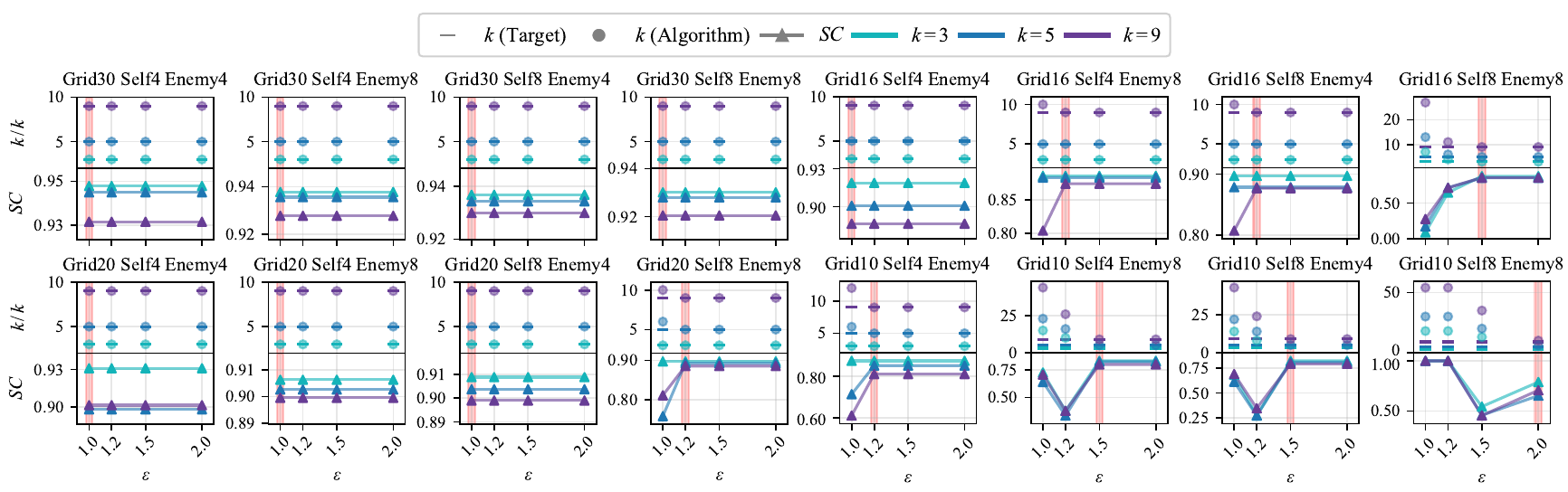}
    \vspace{-20pt}
    \caption{Threshold sensitivity analysis of clustering results under different grid granularities and combat unit scales}
    \label{fig:threshold_sensitivity}
    \vspace{-10pt}
\end{figure}

Considering the impact of the clustering threshold $\epsilon$ on clustering results, the tolerance for thresholds varies across different grid partitioning granularities and combat unit scales. Figure~\ref{fig:threshold_sensitivity} shows the sensitivity of clustering results to $\epsilon$ variations under different experimental configurations. Subfigures on the same row represent the same grid partitioning granularities, where higher granularity results in smaller differences between samples within the same cluster and lower sensitivity to $\epsilon$ changes. Subfigures on the same column represent the same combat unit scales, where more complex scenarios exhibit higher sensitivity to $\epsilon$ variations.

For each subfigure, colors represent the true number of clusters in the synthetic data, the horizontal axis represents different thresholds, the upper half of the vertical axis represents the difference between the target number of clusters $k$ and the algorithm output $\hat{k}$, while the lower half represents the silhouette coefficient. The red bar indicates the optimal $\epsilon$ for the experimental configuration, evaluated based on maintaining minimal difference between $k$ and $\hat{k}$ while achieving the highest silhouette coefficient. In the subsequent experiments, the clustering threshold $\epsilon$ is configured to the optimal value following the same manner.

\subsection{Stream clustering with the cluster-centric BK-tree under different distance metrics}

The cluster-centric BK-tree is a distance-based algorithm for stream clustering, offering the advantage of adaptability to various distance metrics. Different distance metrics have been utilized for comparison. Table \ref{tab:clustering_results_with_different_metrics} shows the comparison of clustering performance with the cluster-centric BK-tree under different distance metrics and true numbers of clusters. For scenarios with inconsistent combat unit scales, accuracy based on ground truth is additionally presented to measure performance. Additionally, comparisons of the t-SNE visualization of clustering results are presented in \text{Figures~S1} and \text{S2} in Section S1 of the Supplementary Material.

\begin{table}[ht]
\renewcommand{\arraystretch}{1}
\centering
\caption{Comparison of clustering performance with the cluster-centric BK-tree under different distance metrics and true numbers of clusters}
\label{tab:clustering_results_with_different_metrics}
\setlength{\tabcolsep}{2pt}
\footnotesize

\begin{tabularx}{\columnwidth}{@{}Xcccccc@{}}
\toprule
\textbf{Consistent Combat Unit Scale} & \multicolumn{2}{c}{$k=9$} & \multicolumn{2}{c}{$k=50$} & \multicolumn{2}{c}{$k=100$} \\
\cmidrule(lr){1-1}\cmidrule(lr){2-3}\cmidrule(lr){4-5}\cmidrule(lr){6-7}
Distance Metric & SC & Time/s & SC & Time/s & SC & Time/s \\
\midrule
Chamfer Distance & 0.8902 & 0.01 & 0.8747 & 0.34 & 0.8717 & 1.27 \\
Hausdorff Distance & 0.8961 & 0.02 & 0.8829 & 0.43 & 0.8795 & 1.63 \\
2-Wasserstein & 0.9223 & 0.03 & 0.9128 & 0.90 & 0.9101 & 3.19 \\
1-Wasserstein (EMD) & 0.9205 & 0.03 & 0.9096 & 0.79 & 0.9066 & 3.21 \\
Adapted EMD (ours) & 0.9205 & 0.02 & 0.9096 & 0.28 & 0.9066 & 1.28 \\
\midrule
\midrule
\textbf{Inconsistent Combat Unit Scale} & \multicolumn{3}{c}{$k=25$} & \multicolumn{3}{c}{$k=81$}\\
\cmidrule(lr){1-1}\cmidrule(lr){2-4}\cmidrule(lr){5-7}
Distance Metric & Accuracy & SC & Time/s & Accuracy & SC & Time/s \\
\midrule
Chamfer & 0.8800 & 0.6074 & 0.02 & 0.5021 & 0.3521 & 0.11 \\
Hausdorff & 0.9067 & 0.7647 & 0.03 & 0.3786 & 0.4970 & 0.17 \\
2-Wasserstein & 1.0000 & 0.8936 & 0.10 & 0.9198 & 0.7442 & 1.20 \\
1-Wasserstein (EMD) & 1.0000 & 0.8453 & 0.09 & 0.5617 & 0.4826 & 0.72 \\
Adapted EMD (ours) & 1.0000 & 0.8759 & 0.01 & 1.0000 & 0.7844 & 0.04 \\
\bottomrule
\end{tabularx}
\end{table}

The results indicate a clear trade-off between computational overhead and distribution sensitivity. Baseline geometric metrics, such as Chamfer and Hausdorff distances, exhibit high efficiency but struggle to capture global distribution features, leading to a significant decline in silhouette coefficient and accuracy as $k$ increases or when unit scales become inconsistent. Conversely, while Optimal Transport (OT)-based metrics (e.g., 1-W and 2-W) provide robust distribution awareness, their polynomial computational complexity poses challenges for real-time RTS stream clustering.

The comparative analysis demonstrates that the adapted EMD integrated with the cluster-centric BK-tree achieves superior performance across multiple dimensions. It provides enhanced distribution awareness via optimal matching, handling of unit count mismatch through extremely distant virtual points padding, and improved computational efficiency through Hungarian algorithm. These advantages are particularly evident in RTS game state stream clustering where unit counts vary significantly and real-time performance requirements are stringent. In the subsequent experiments, the adapted EMD will be uniformly utilized as the distance metric for the cluster-centric BK-tree.

\subsection{Performance analysis and comparative study of state stream clustering}

\begin{figure*}[th]
    \centering
    \includegraphics[width=\linewidth]{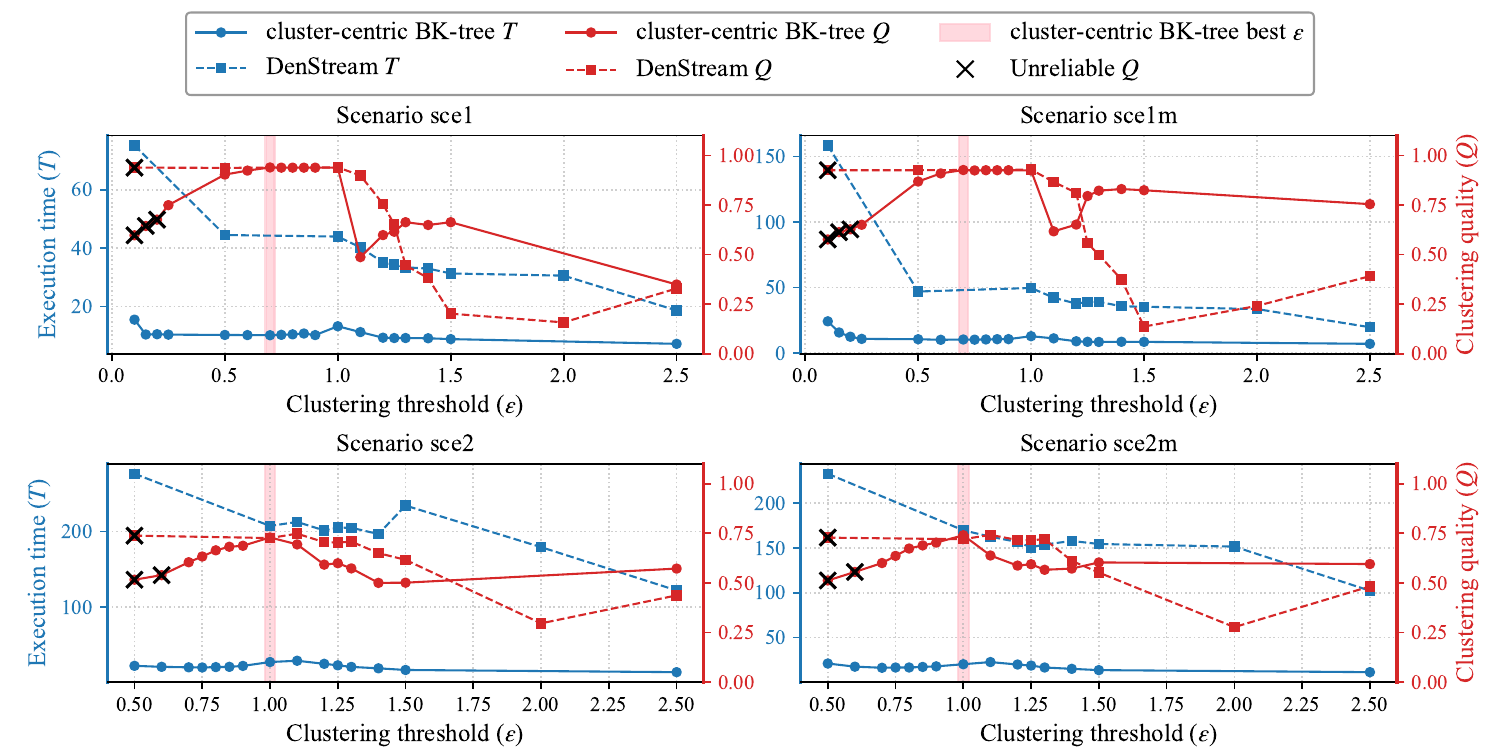}
    \vspace{-20pt}
    \caption{Comparison of clustering performance metrics of the cluster-centric BK-tree and DenStream with various $\epsilon$ across different scenarios}
    \label{fig:clustering_performance}
    \vspace{-2pt}
\end{figure*}

\begin{figure*}[th]
    \centering
    \begin{minipage}{0.22\linewidth}
        \centering
        \includegraphics[width=\linewidth]{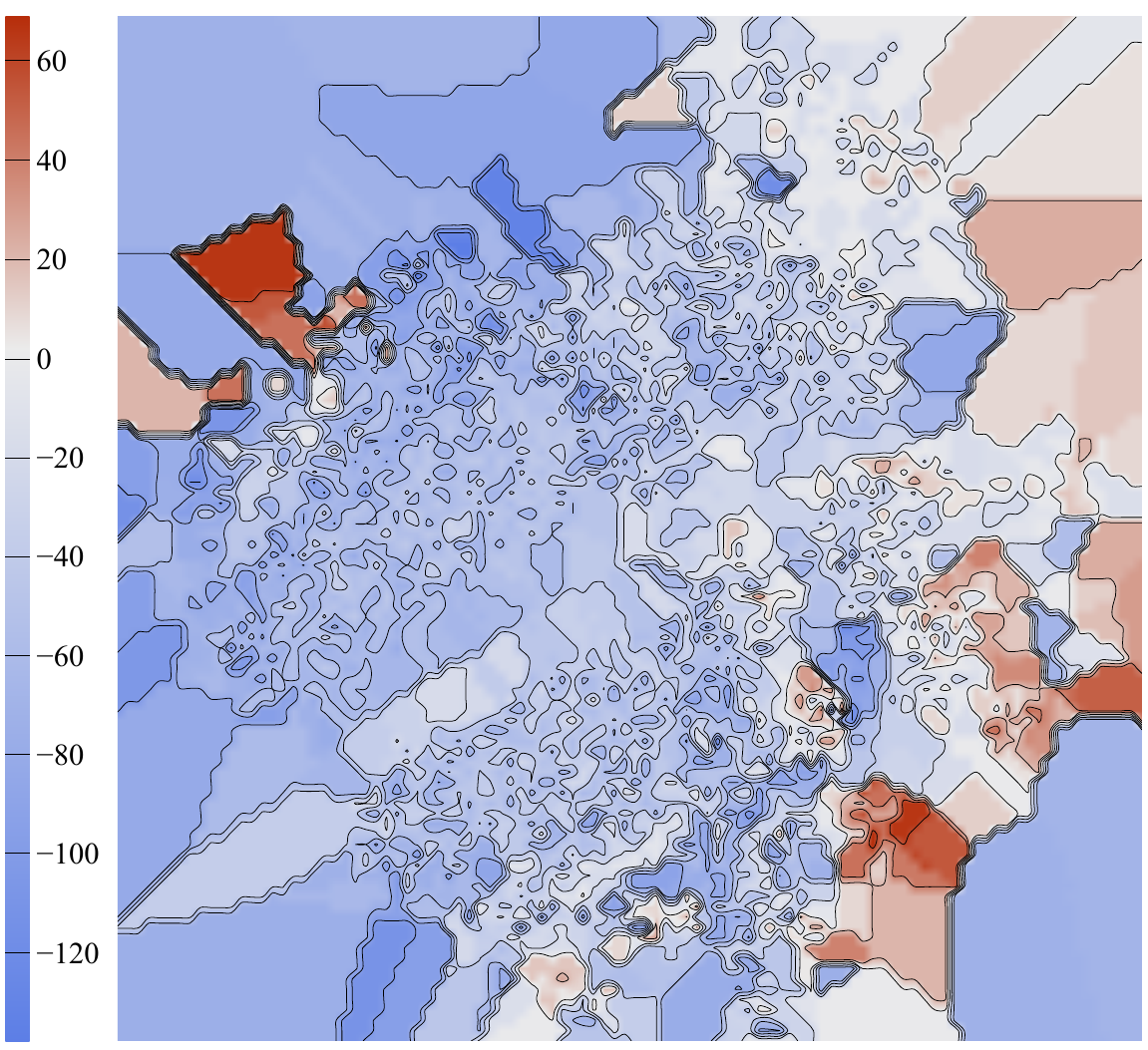}
        a) \textit{sce1} state landscape
    \end{minipage}
    \hfill
    \begin{minipage}{0.22\linewidth}
        \centering
        \includegraphics[width=\linewidth]{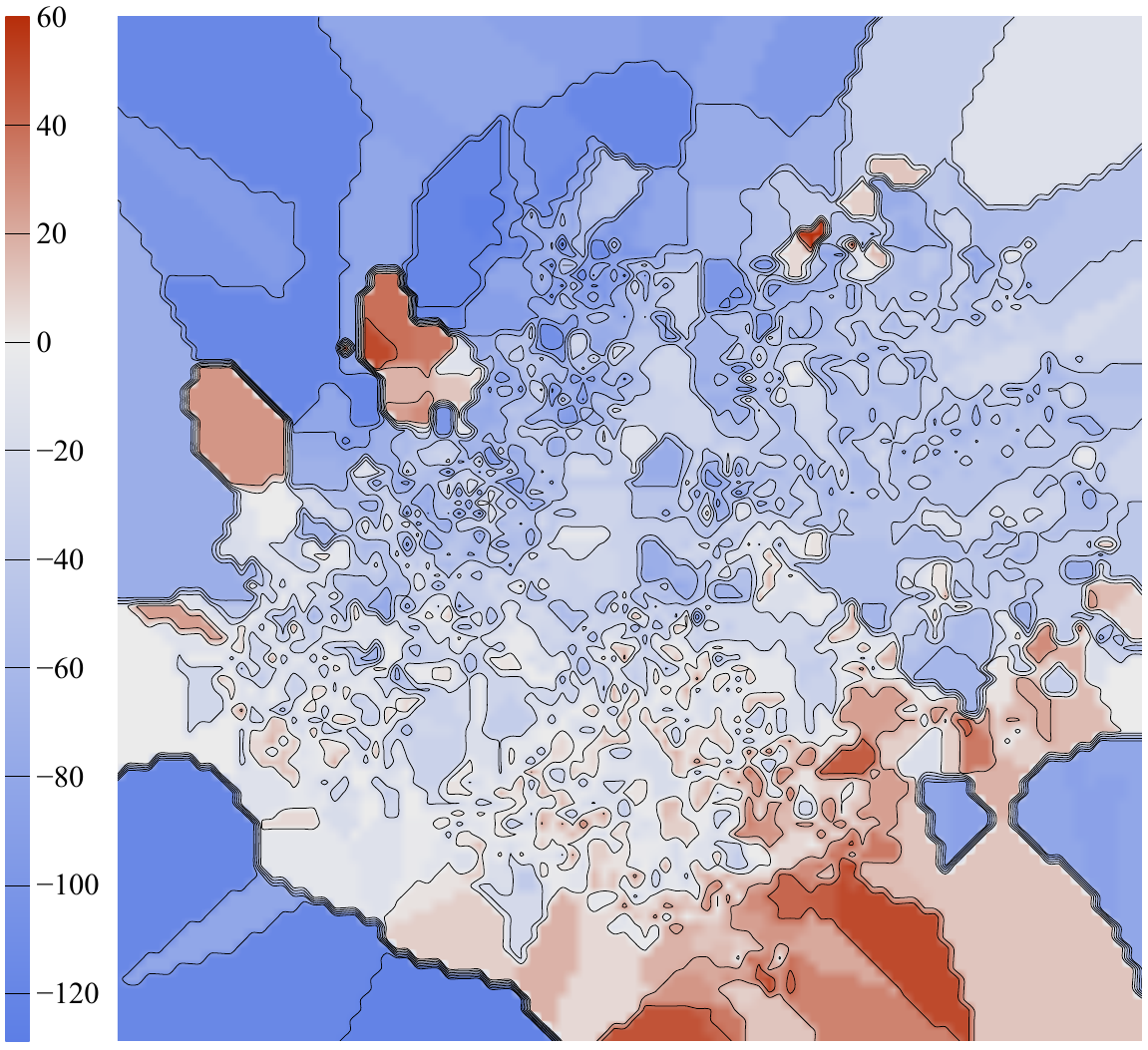}
        b) \textit{sce1m} state landscape
    \end{minipage}
    \hfill
    \begin{minipage}{0.22\linewidth}
        \centering
        \includegraphics[width=\linewidth]{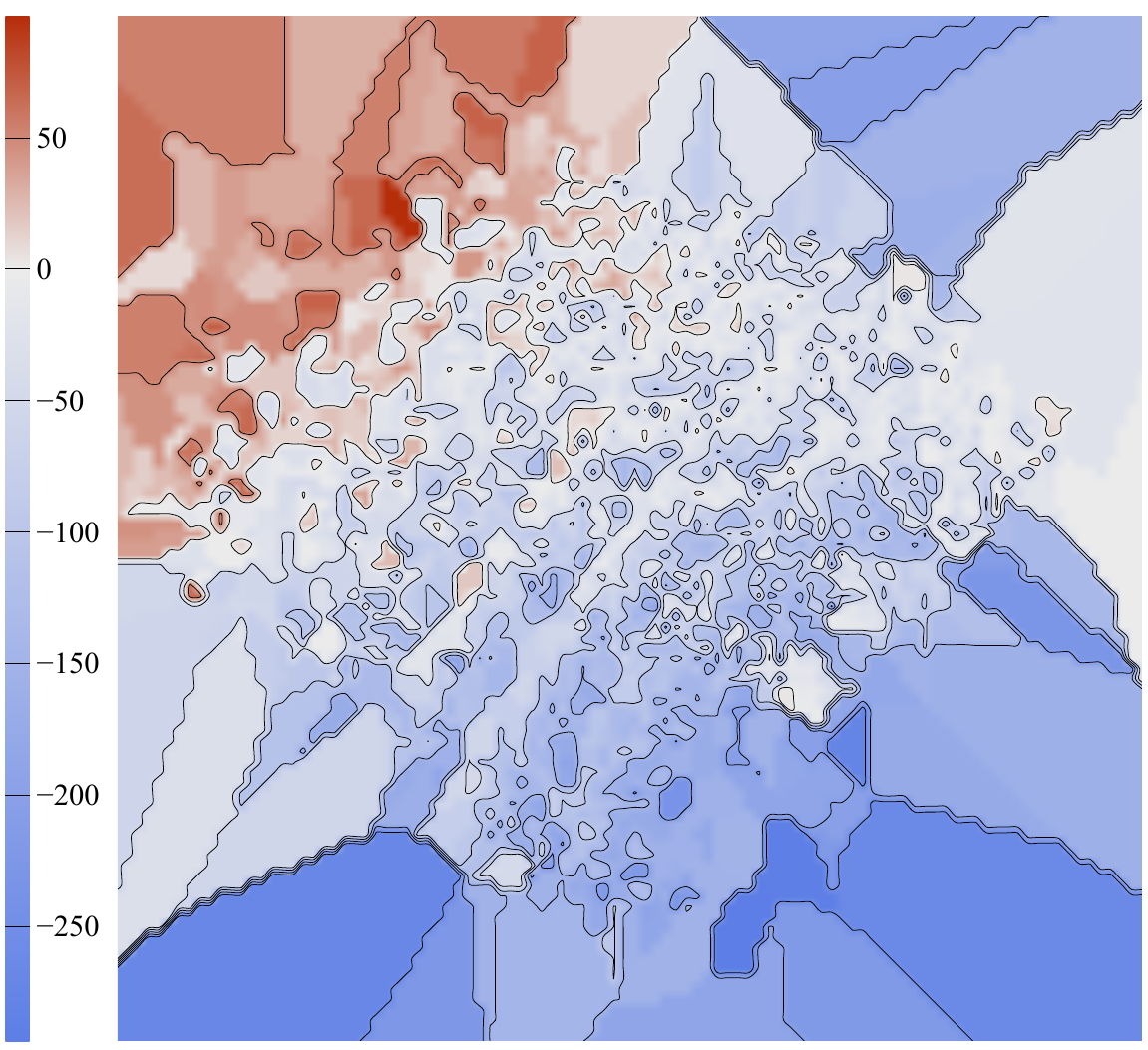}
        c) \textit{sce2} state landscape
    \end{minipage}
    \hfill
    \begin{minipage}{0.22\linewidth}
        \centering
        \includegraphics[width=\linewidth]{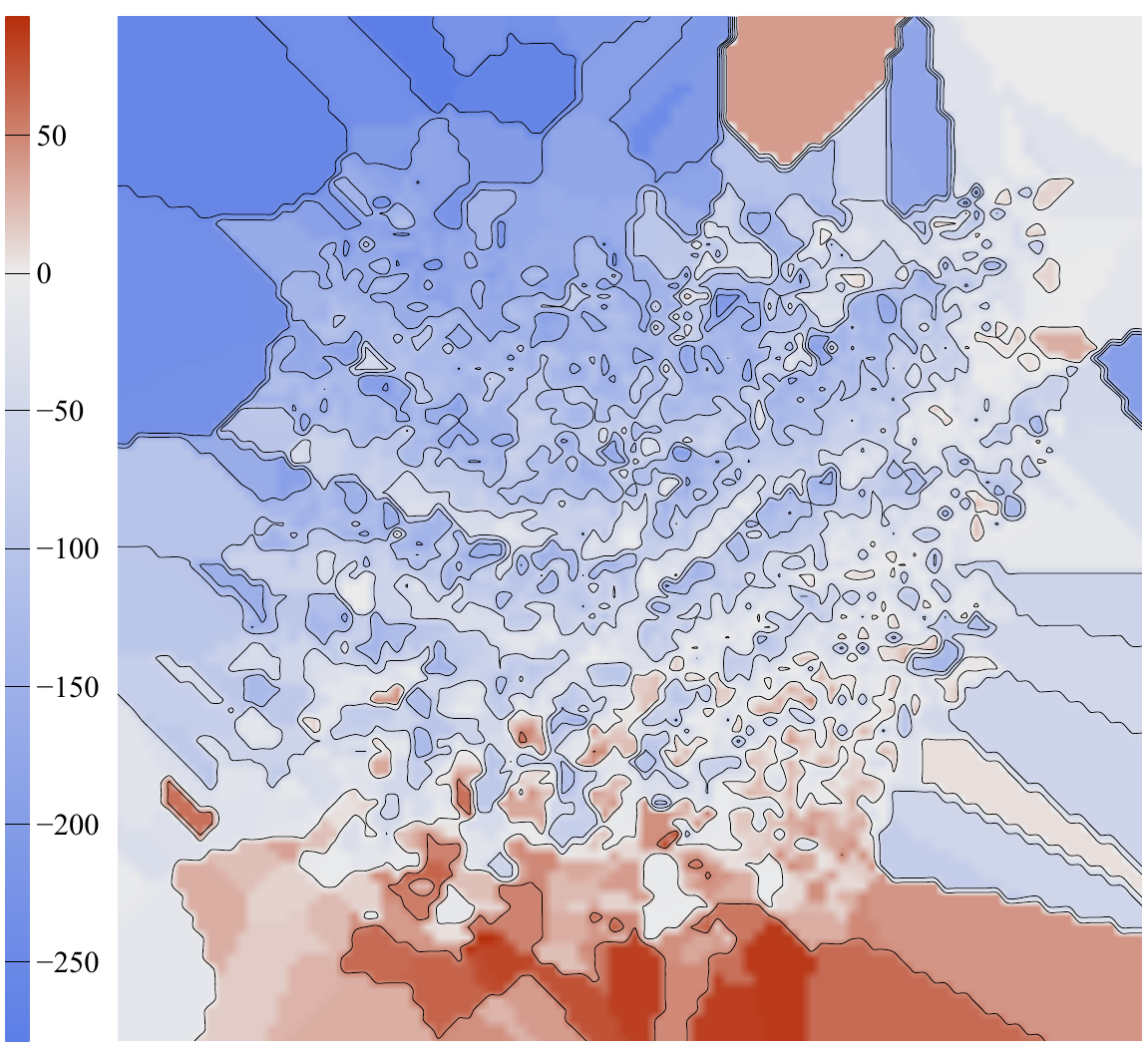}
        d) \textit{sce2m} state landscape
    \end{minipage}
    \caption{State value landscape visualization in different scenarios}
    \label{fig:state_value_landscape}
\end{figure*}

\begin{figure*}[th]
    \centering
    \begin{minipage}{0.22\linewidth}
        \centering
        \includegraphics[width=\linewidth]{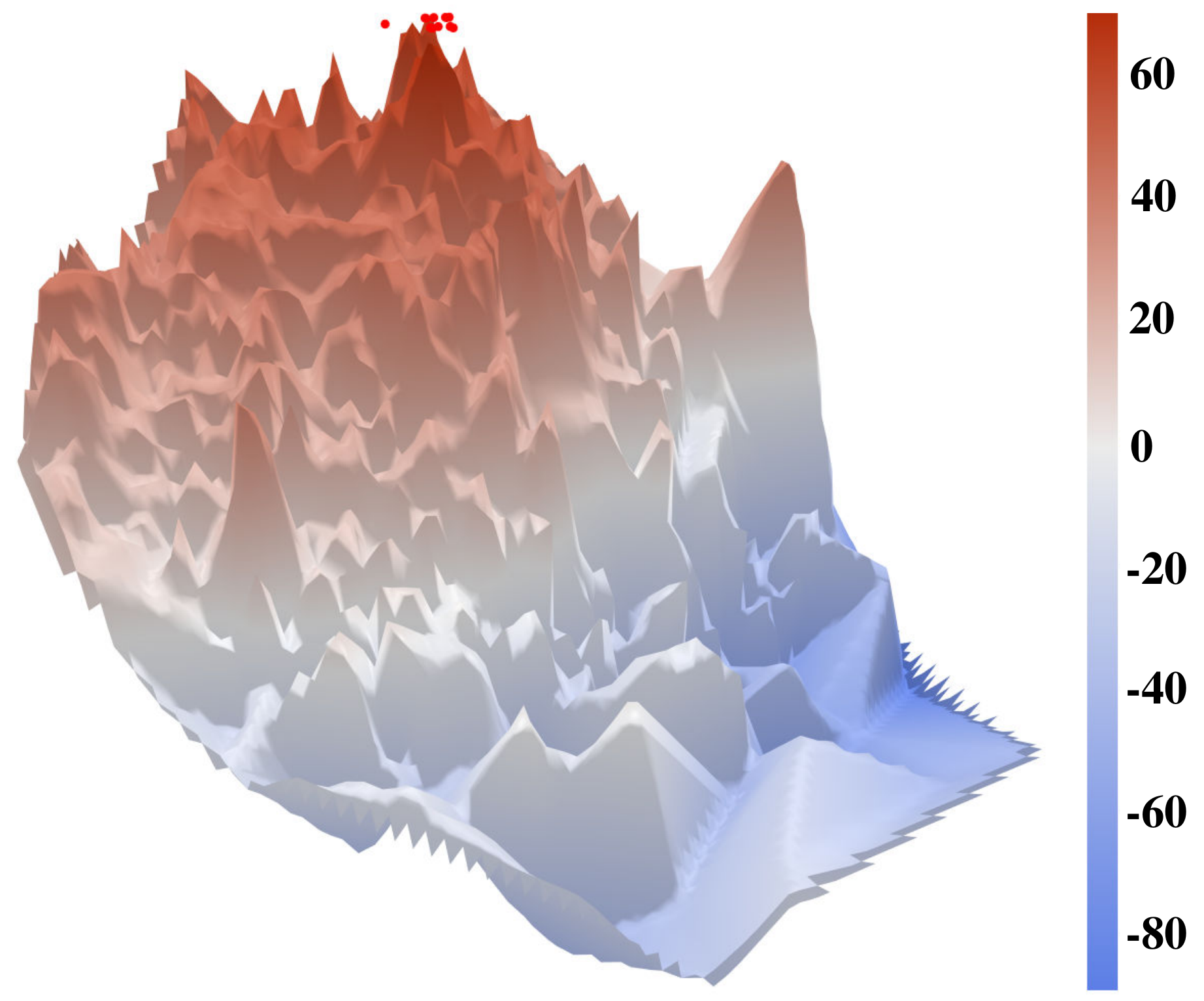}
        a) \textit{sce1} with optima
    \end{minipage}
    \hfill
    \begin{minipage}{0.22\linewidth}
        \centering
        \includegraphics[width=\linewidth]{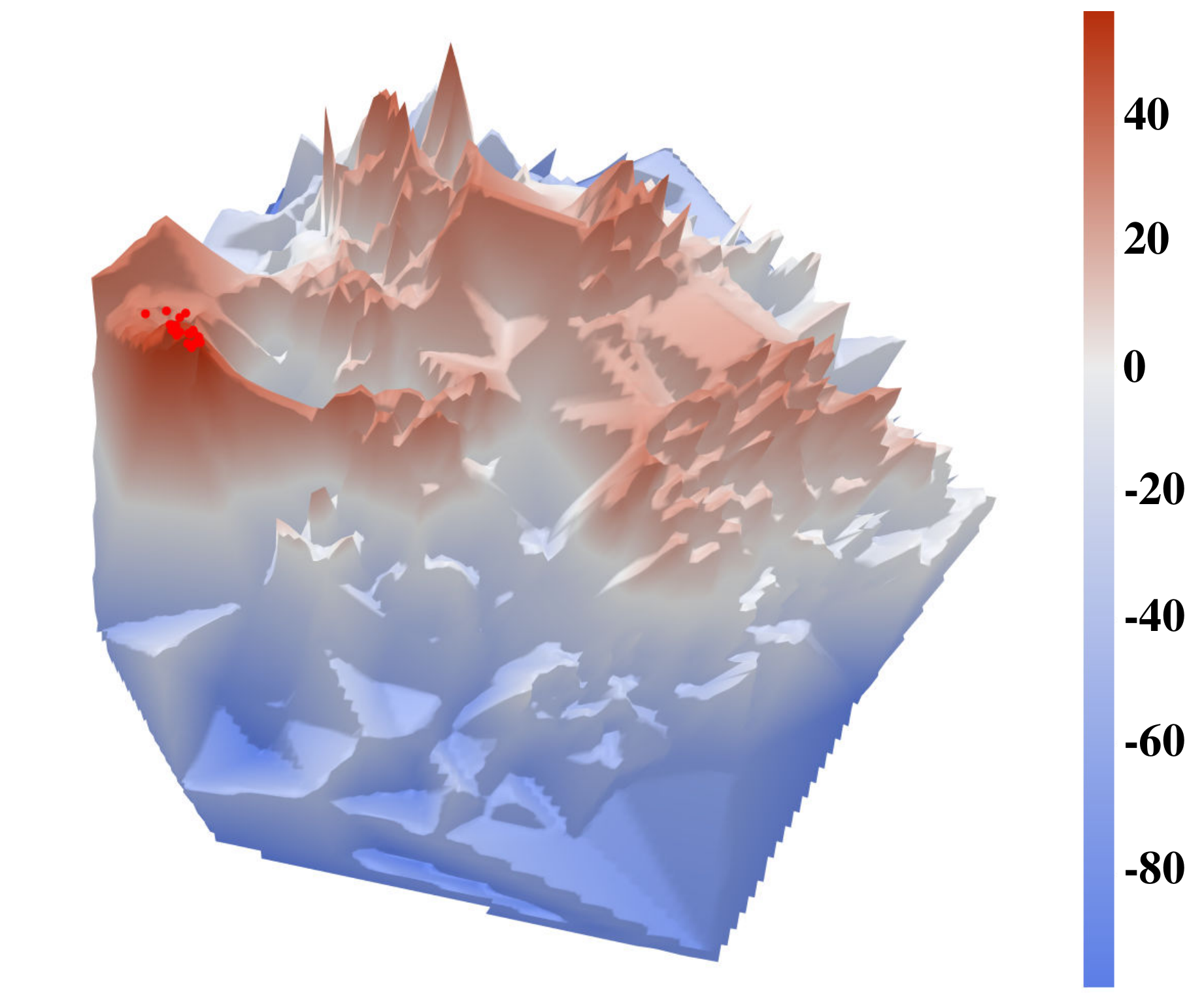}
        b) \textit{sce1m} with optima
    \end{minipage}
    \hfill
    \begin{minipage}{0.22\linewidth}
        \centering
        \includegraphics[width=\linewidth]{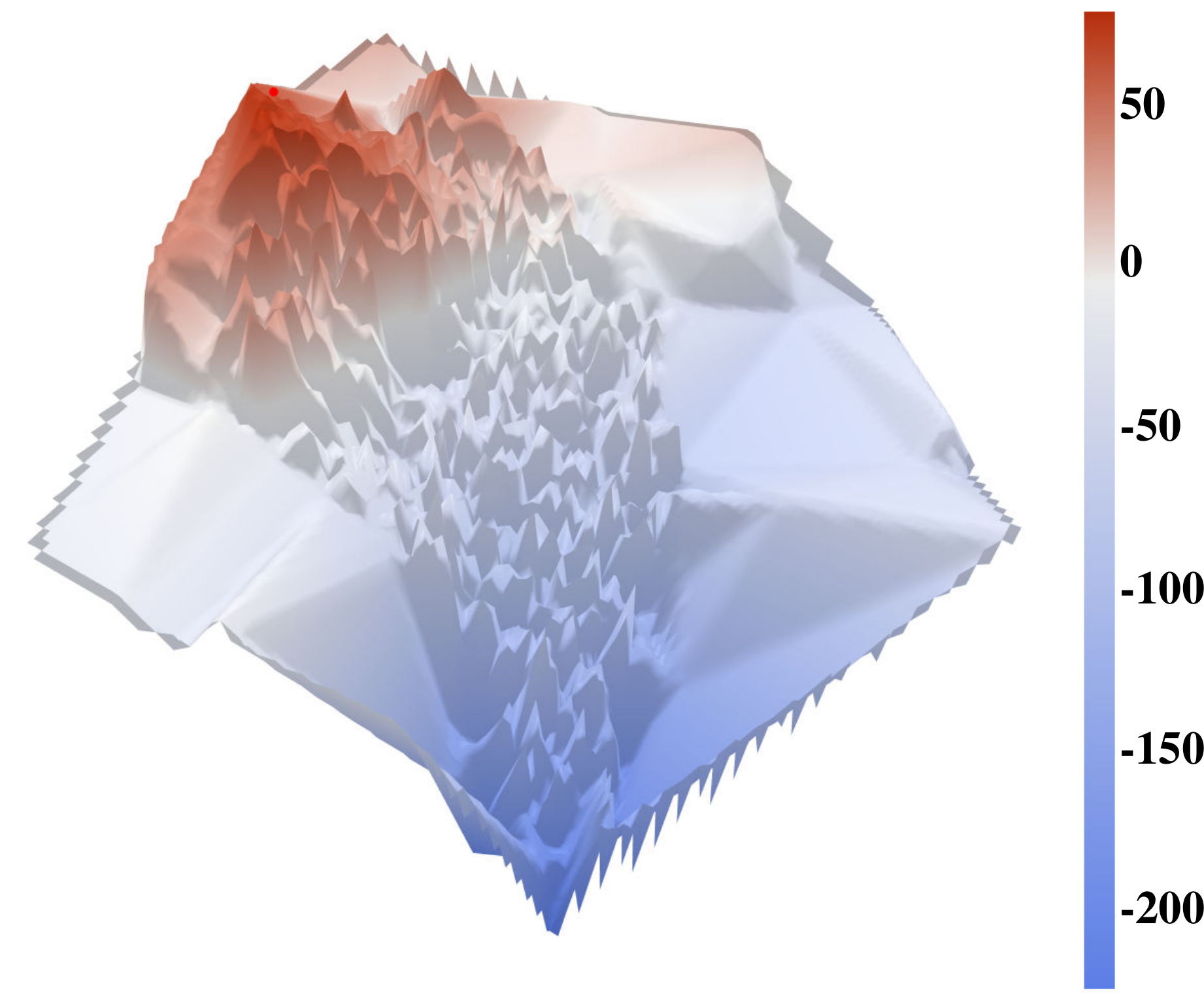}
        c) \textit{sce2} with optima
    \end{minipage}
    \hfill
    \begin{minipage}{0.22\linewidth}
        \centering
        \includegraphics[width=\linewidth]{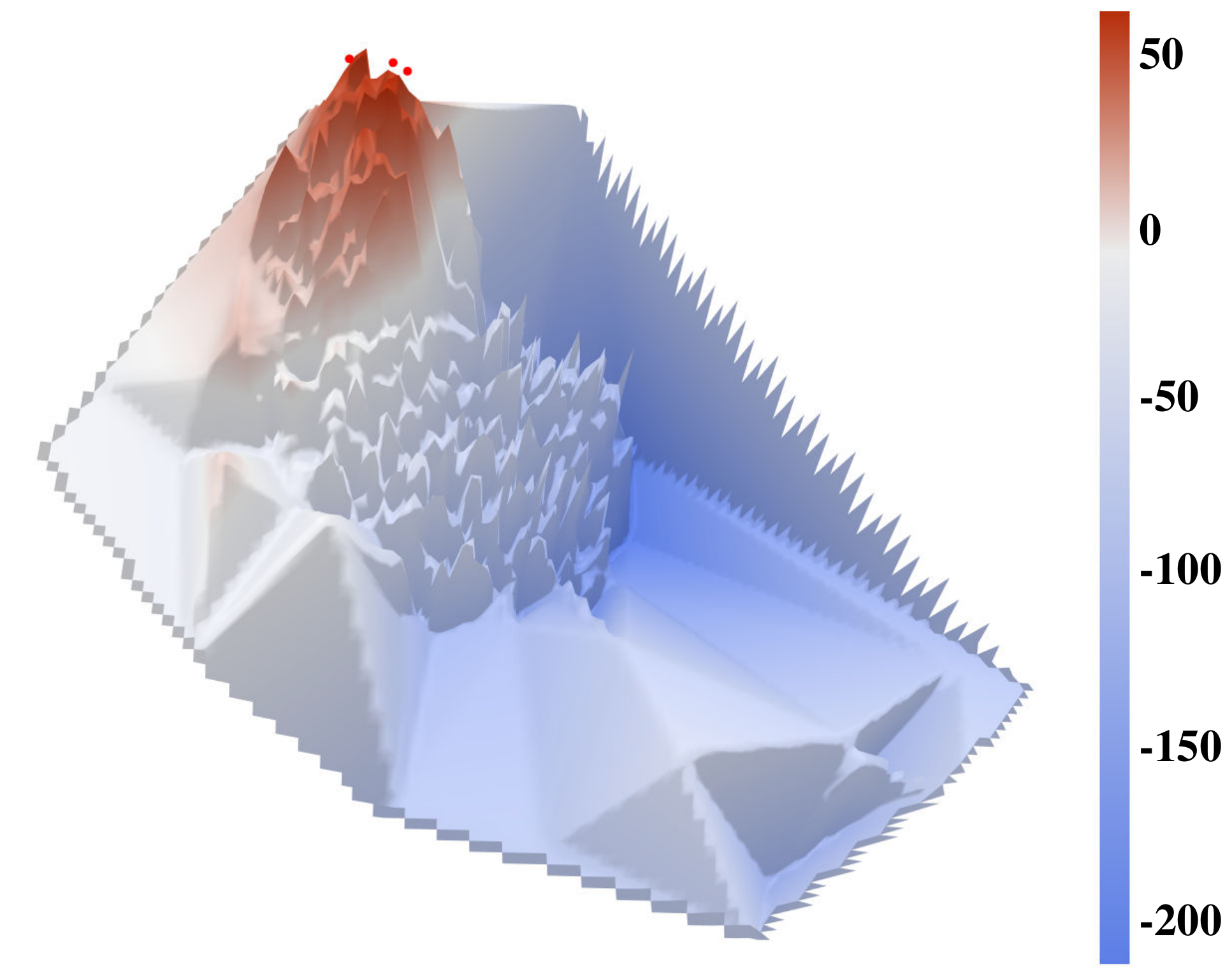}
        d) \textit{sce2m} with optima
    \end{minipage}
    \\
    \begin{minipage}{0.22\linewidth}
        \centering
        \includegraphics[width=\linewidth]{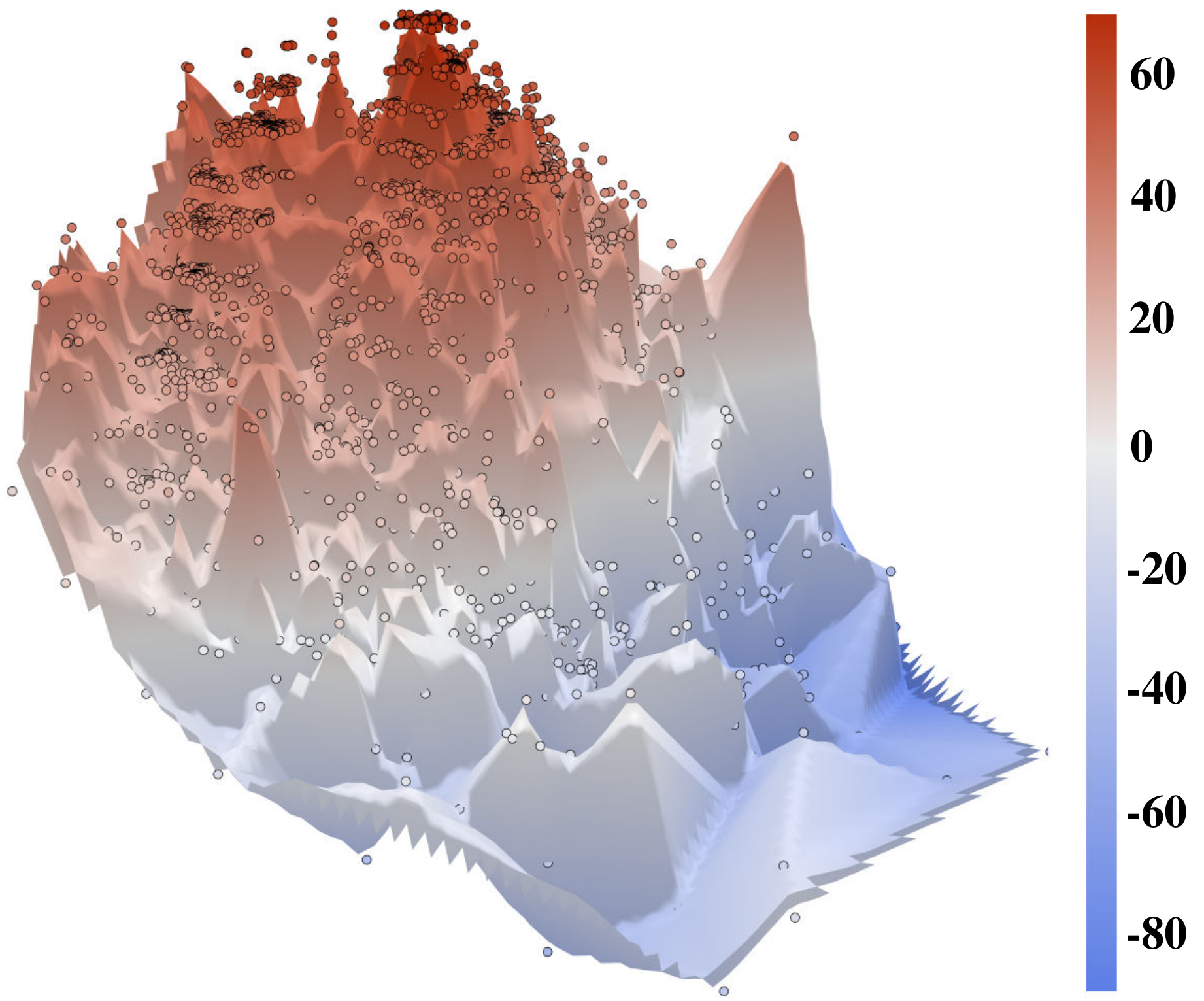}
        e) \textit{sce1} with samples
    \end{minipage}
    \hfill
    \begin{minipage}{0.22\linewidth}
        \centering
        \includegraphics[width=\linewidth]{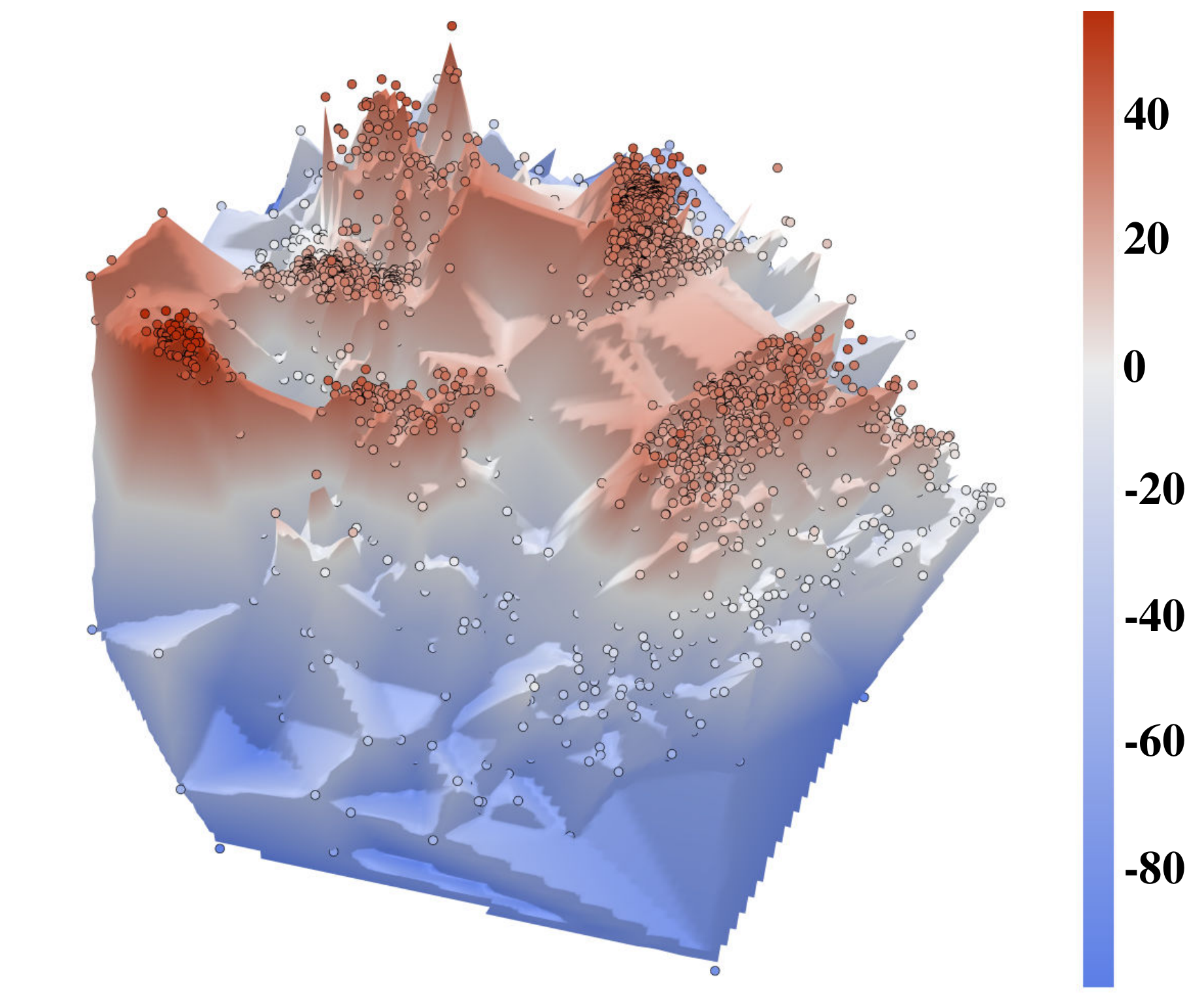}
        f) \textit{sce1m} with samples
    \end{minipage}
    \hfill
    \begin{minipage}{0.22\linewidth}
        \centering
        \includegraphics[width=\linewidth]{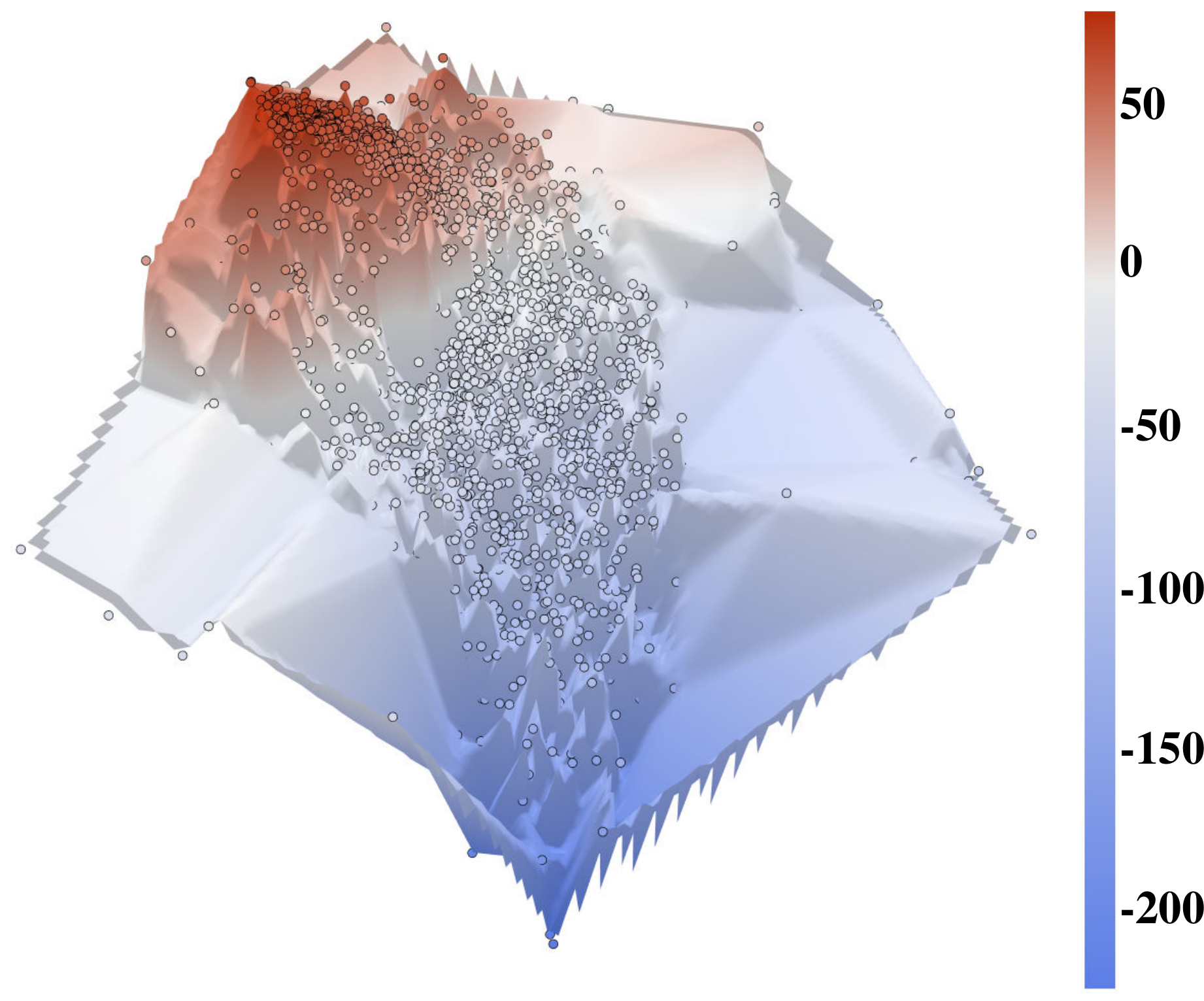}
        g) \textit{sce2} with samples
    \end{minipage}
    \hfill
    \begin{minipage}{0.22\linewidth}
        \centering
        \includegraphics[width=\linewidth]{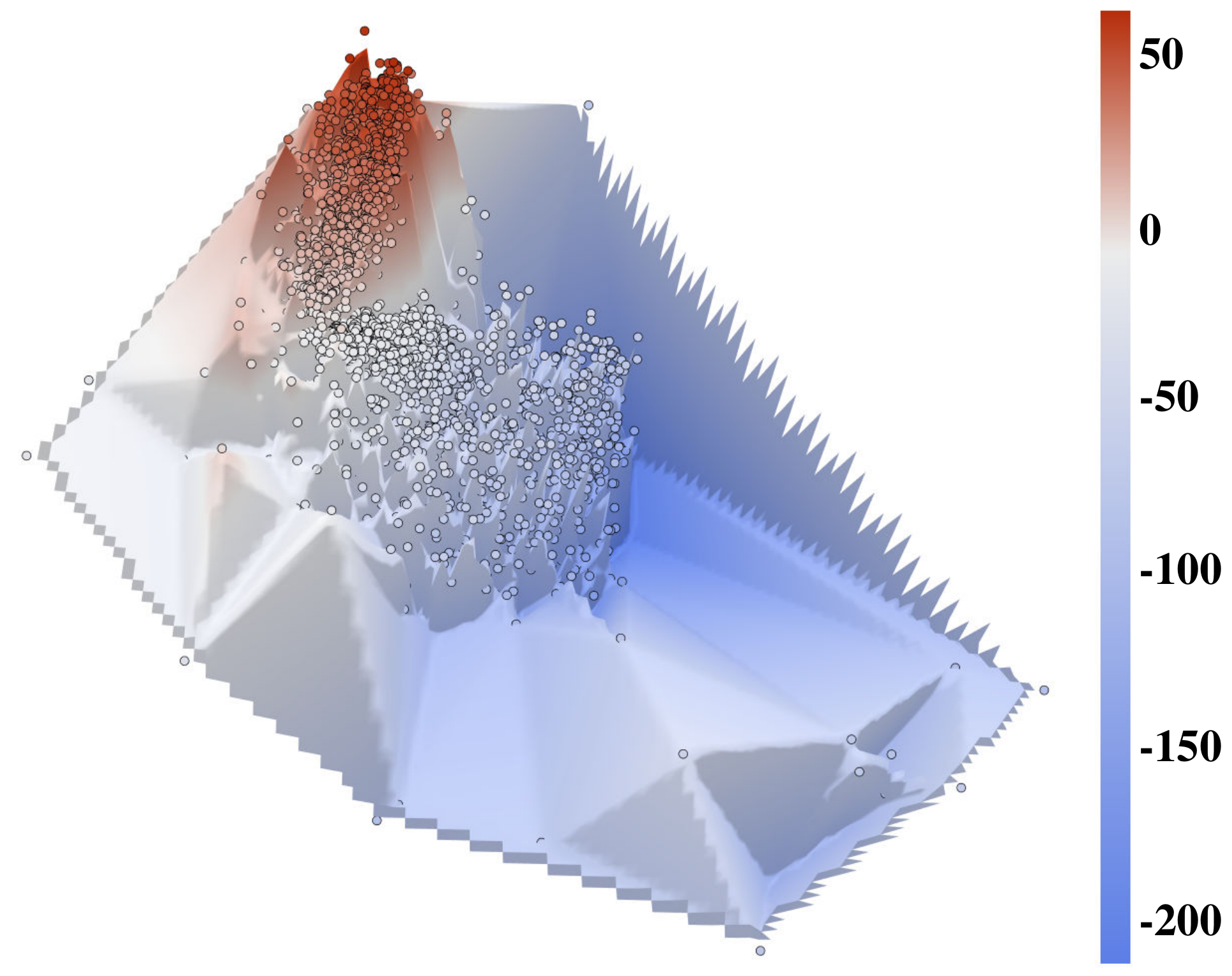}
        h) \textit{sce2m} with samples
    \end{minipage}
    \caption{Fitness landscape visualization in different scenarios}
    \label{fig:fitness_landscape}
\end{figure*}

To validate the performance of the proposed cluster-centric BK-tree algorithm for game state stream clustering, numerical experiments were conducted across various scenarios and thresholds $\epsilon$. The results are compared against DenStream \citep{fengcao_Densitybased_2006}, which shares conceptual similarities in maintaining density-based micro-clusters. Figure~\ref{fig:clustering_performance} presents the comparative results for execution time $T$ and clustering quality $Q$ (measured by the silhouette coefficient). Notably, the data points marked with a cross symbol on the $Q$ curves indicate instances where the number of clusters $N$ is excessively large, rendering the corresponding silhouette coefficient unreliable for meaningful evaluation. Detailed numerical data for both algorithms are provided in \text{Table~S1} and \text{Table~S2} in Section S3 of the Supplementary Material, while the parameter sensitivity analysis of DenStream is shown in \text{Figure~S3} in Section S2 of the Supplementary Material.

As illustrated in Figure~\ref{fig:clustering_performance}, the cluster-centric BK-tree achieves satisfactory clustering outcomes in StarCraft~II datasets when an appropriate threshold $\epsilon$ is selected. Specifically, in scenarios \textit{sce1} and \textit{sce1m}, the clustering quality $Q$ exceeds $0.9$ for $\epsilon \in [0.6, 1.0]$. While $Q$ is sensitive to $\epsilon$, both algorithms achieve comparable optimal clustering quality across all scenarios given an appropriately tuned $\epsilon$, substantiating the adaptability of the proposed state distance metric to RTS data streams.

Under general circumstances, a smaller clustering threshold $\epsilon$ leads to a greater number of clusters $N$. However, the execution time $T$ exhibits an interesting non-monotonic phenomenon: $T$ does not significantly increase with the reduction of $\epsilon$, but generally presents two peaks within a relatively small range. This is attributed to the synergistic effect of the insertion/query pruning mechanisms of the cluster-centric BK-tree and the greedy cluster assignment strategy. A detailed case study of this internal efficiency logic in the scenario \textit{sce1} is provided in Section S3 of the Supplementary Material.

Comparative analysis with the DenStream algorithm shows that while both algorithms achieve comparable optimal clustering qualities, the cluster-centric BK-tree demonstrates a significant advantage in computational efficiency. In complex scenarios involving over 100,000 states, the proposed method completes the task within approximately 20 seconds. This substantiates that the proposed state distance metric and the cluster-centric BK-tree algorithm are highly adaptable for high-dimensional RTS game data streams.

\subsection{Fitness landscape visualization and problem characterization}

Fitness landscape visualization offers insights into the structural characterization of problem complexity, mapping the distribution of solutions from a high-dimensional search space into an interpretable analytical framework. Analogously, the visualization of the state value landscape serves to capture the key structural features of the state space, projecting instantaneous advantages into a spatial representation that delineates the underlying complexity.

As illustrated in Figure \ref{fig:state_value_landscape}, the state value landscapes across various scenarios reveal the structural distribution of tactical advantages within the combat environment. The state values are represented by a bipolar color gradient, where deeper red tones signify a decisive advantage for Player 1, and those in blue for Player 2. This spatial representation not only captures the intrinsic evaluative logic of the state space but also establishes a transparent tool for subsequent tactical analysis.

As illustrated in Figure~\ref{fig:fitness_landscape}, the fitness landscapes pertaining to the StarCraft~II micromanagement are characterized by the following notable attributes:

\begin{enumerate}[label={$($\arabic*$)$}]
    \item \textit{Multi-modality} denotes the coexistence of multiple global optima within the landscape. In StarCraft~II micromanagement, diverse tactical combinations (e.g. focus-fire, kiting, or decoy maneuvers) can yield equivalent high-fitness outcomes. This inherent diversity in optimal decision sequences, spanning various objectives and timescales, manifests as a multi-modal distribution where distinct tactical peaks represent equally viable strategies for victory.
    \item \textit{Ruggedness} is defined by the high density and erratic distribution of local optima, which significantly amplifies search complexity. Fitness landscapes demonstrate exceptional sensitivity to tiny perturbations in the high-dimensional action space, driven by the intricately interrelated nature of critical decisions such as unit positioning, target selection, and tactical switching.
    \item \textit{Funnel-like topography} characterizes the global fitness landscape as a vast basin of attraction where a pronounced gradient clearly separates high-fitness solutions from poor ones. This structure reflects the snowball effect prevalent in RTS combat, where a significant advantage or disadvantage, once established, becomes increasingly difficult to reverse.
\end{enumerate}

The interplay between these structural features creates a highly intricate optimization environment. From a global perspective, the funnel-like topography supplies a directional signal that guides algorithms toward high-fitness regions, yet the reliability of this gradient is constantly challenged by local ruggedness, which increases search costs through a profusion of sub-optimal peaks. Furthermore, the presence of multi-modality ensures that these high-fitness regions are not singular but distributed across diverse tactical clusters. The feasibility of an efficient search therefore hinges on an algorithm's capacity to concurrently navigate global directionality, escape local optima, and maintain tactical diversity across the multi-modal landscape.

\subsection{Action sequence pattern mining and tactic extraction}
\label{sec:Action sequence pattern mining and tactic extraction}

The above analysis is framed from the perspective of states and state-transition sequences, representing the outcomes of decisions. In RTS games, however, action sequences are just as crucial, which reveal the decision-making process itself. To trace the deeper causes from outcomes, action sequence patterns and higher-level tactic extraction are distilled and rendered as Sankey diagrams.

\begin{figure*}[th]
  \centering

  \begin{minipage}{0.285\linewidth}
    \centering
    \includegraphics[width=\linewidth]{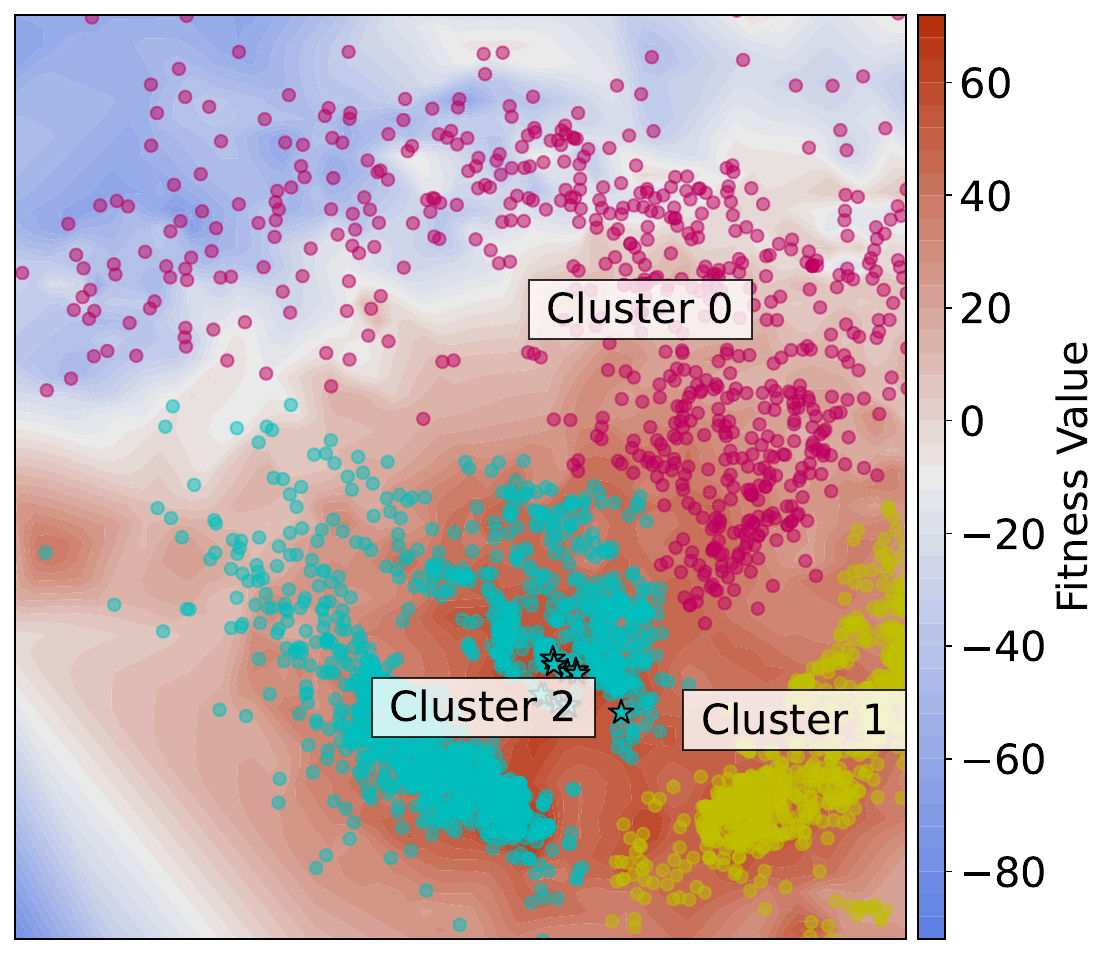}\\
    {\small a) fitness landscape}
  \end{minipage}%
  \hfill
  \begin{minipage}{0.34\linewidth}
    \centering
    \begin{minipage}{0.48\linewidth}
      \centering
      \includegraphics[width=\linewidth]{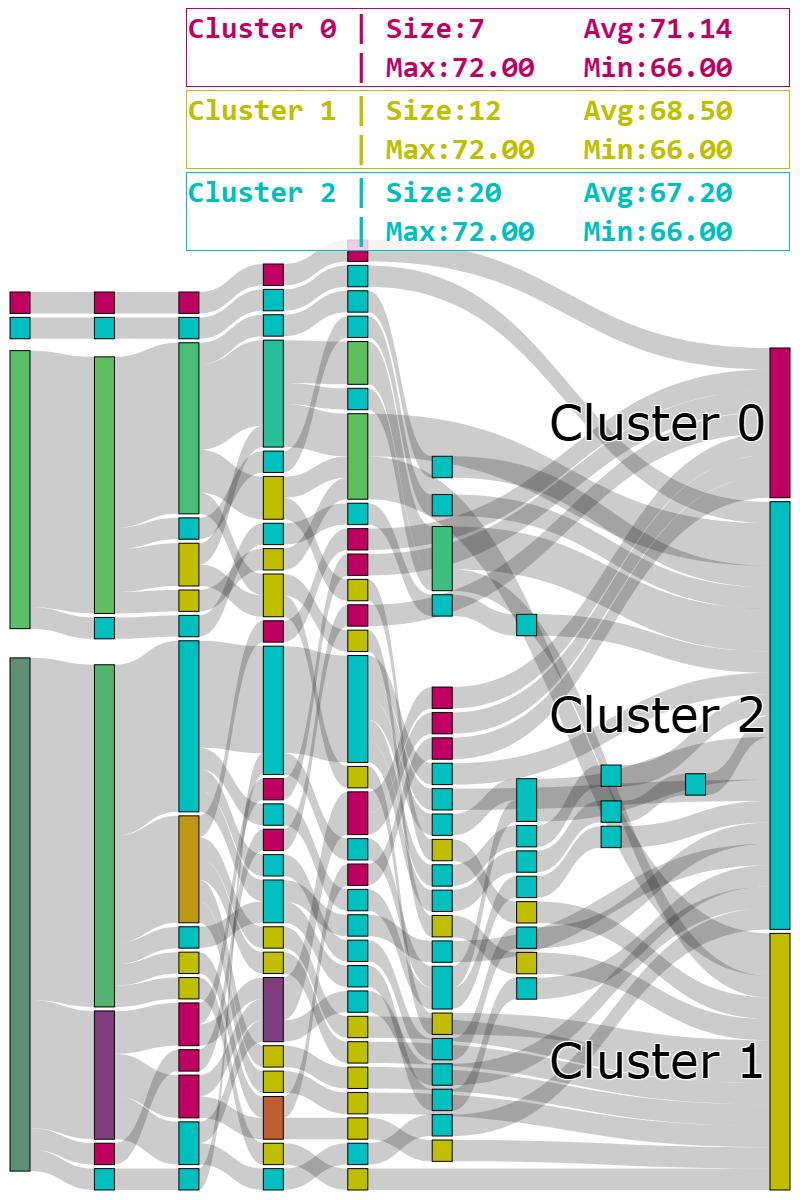}
    \end{minipage}\hfill
    \begin{minipage}{0.48\linewidth}
      \centering
      \includegraphics[width=\linewidth]{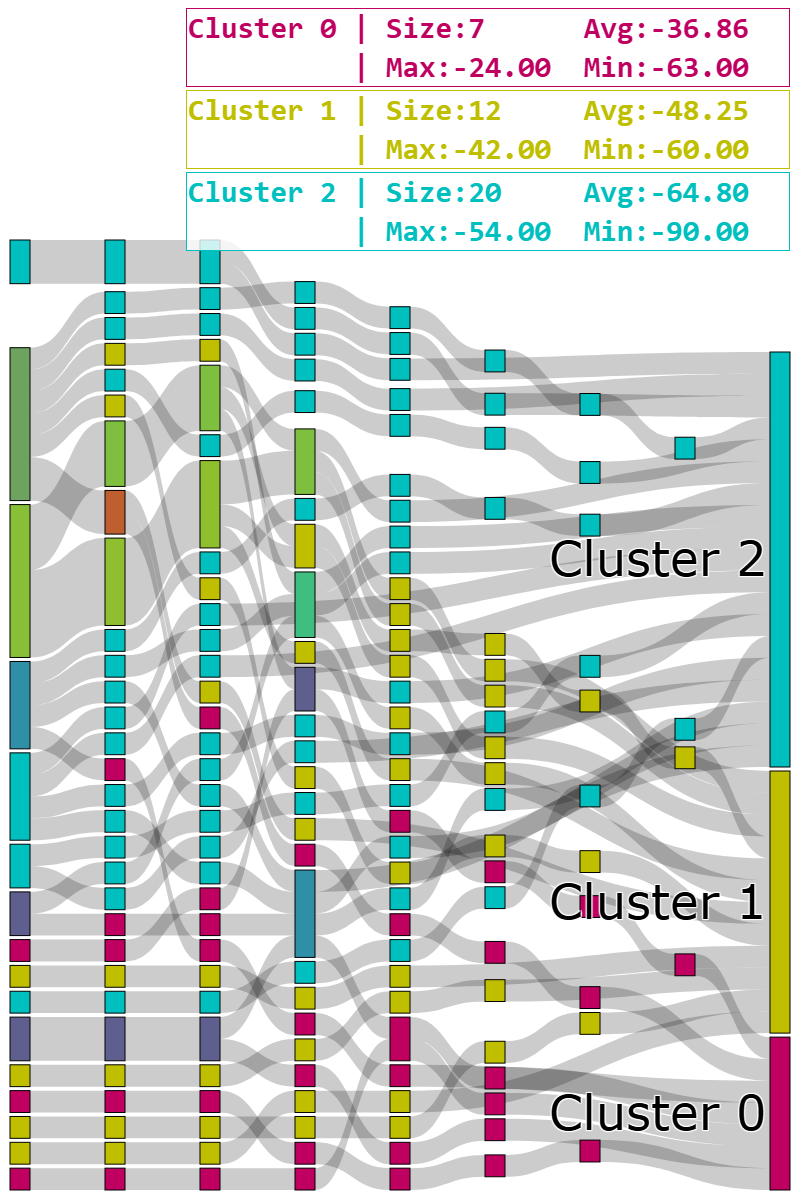}
    \end{minipage}\\[2pt]
    {\small b) action Sankey (left: win, right: loss)}
  \end{minipage}%
  \hfill
  \begin{minipage}{0.34\linewidth}
    \centering
    \begin{minipage}{0.48\linewidth}
      \centering
      \includegraphics[width=\linewidth]{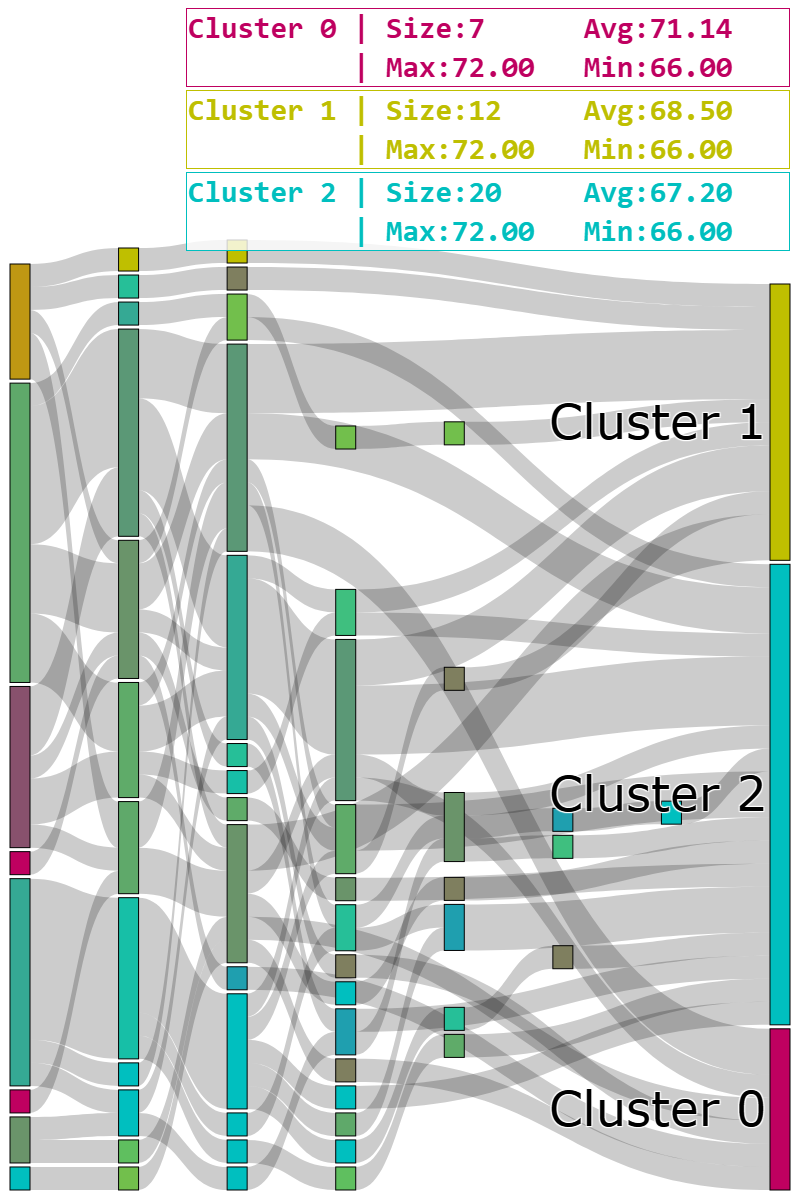}
    \end{minipage}\hfill
    \begin{minipage}{0.48\linewidth}
      \centering
      \includegraphics[width=\linewidth]{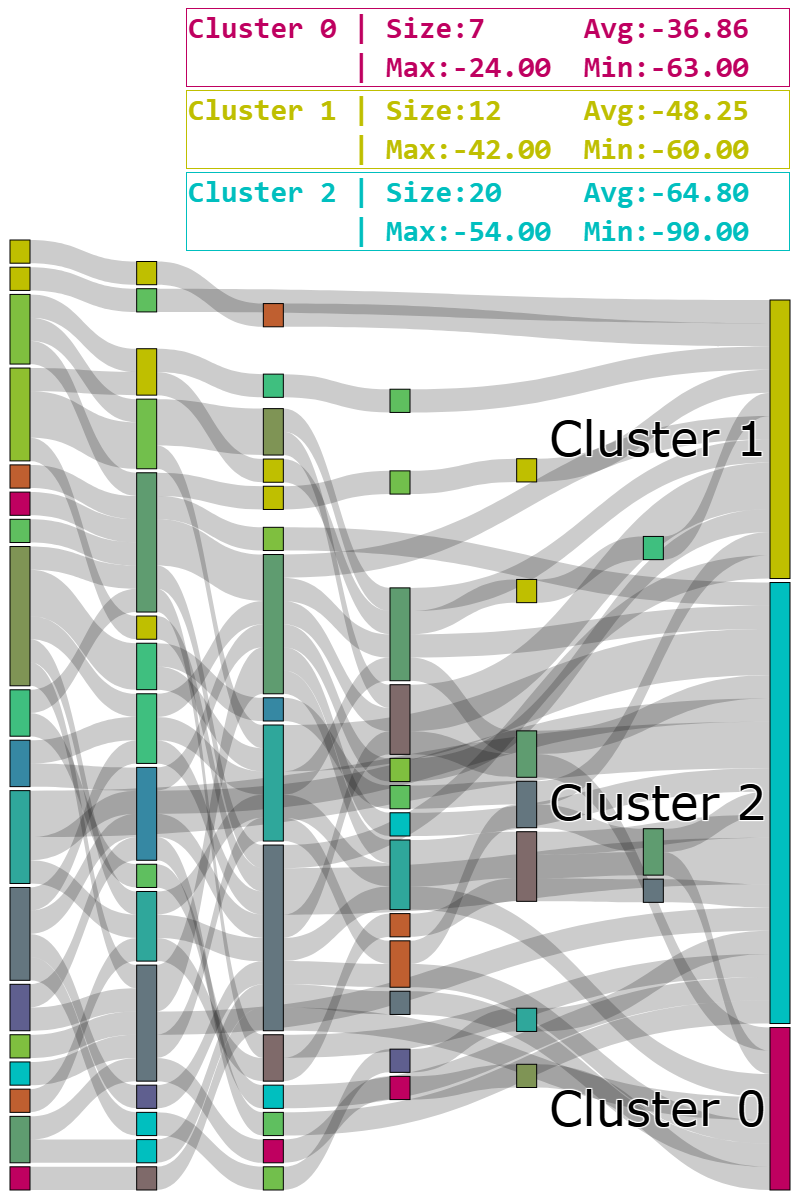}
    \end{minipage}\\[2pt]
    {\small c) tactic Sankey (left: win, right: loss)}
  \end{minipage}

  \caption{Comprehensive Sankey diagram of action sequence and tactic patterns with Agglomerative algorithm for clustering}
  \label{fig:comprehensive_sankey_diagram_Agglomerative}
\end{figure*}

As a representative example, the detailed results of Ward's hierarchical clustering method \citep{wardjr._Hierarchical_1963} with $k=3$ are presented in Figure~\ref{fig:comprehensive_sankey_diagram_Agglomerative}, where Figure~\ref{fig:comprehensive_sankey_diagram_Agglomerative} a) represents the fitness landscape and clustering results, Figure~\ref{fig:comprehensive_sankey_diagram_Agglomerative} b) and c) characterize the process elements in victorious and defeated solutions, from perspectives of action and tactic patterns, respectively. Comprehensive results for different clustering algorithms and cluster numbers are presented in \text{Figures~S4} to \text{Figures~S11} in Section S6 of the Supplementary Material.

\begin{figure*}[th]
    \centering
    \begin{minipage}{0.24\linewidth}
        \centering
        \includegraphics[width=\linewidth]{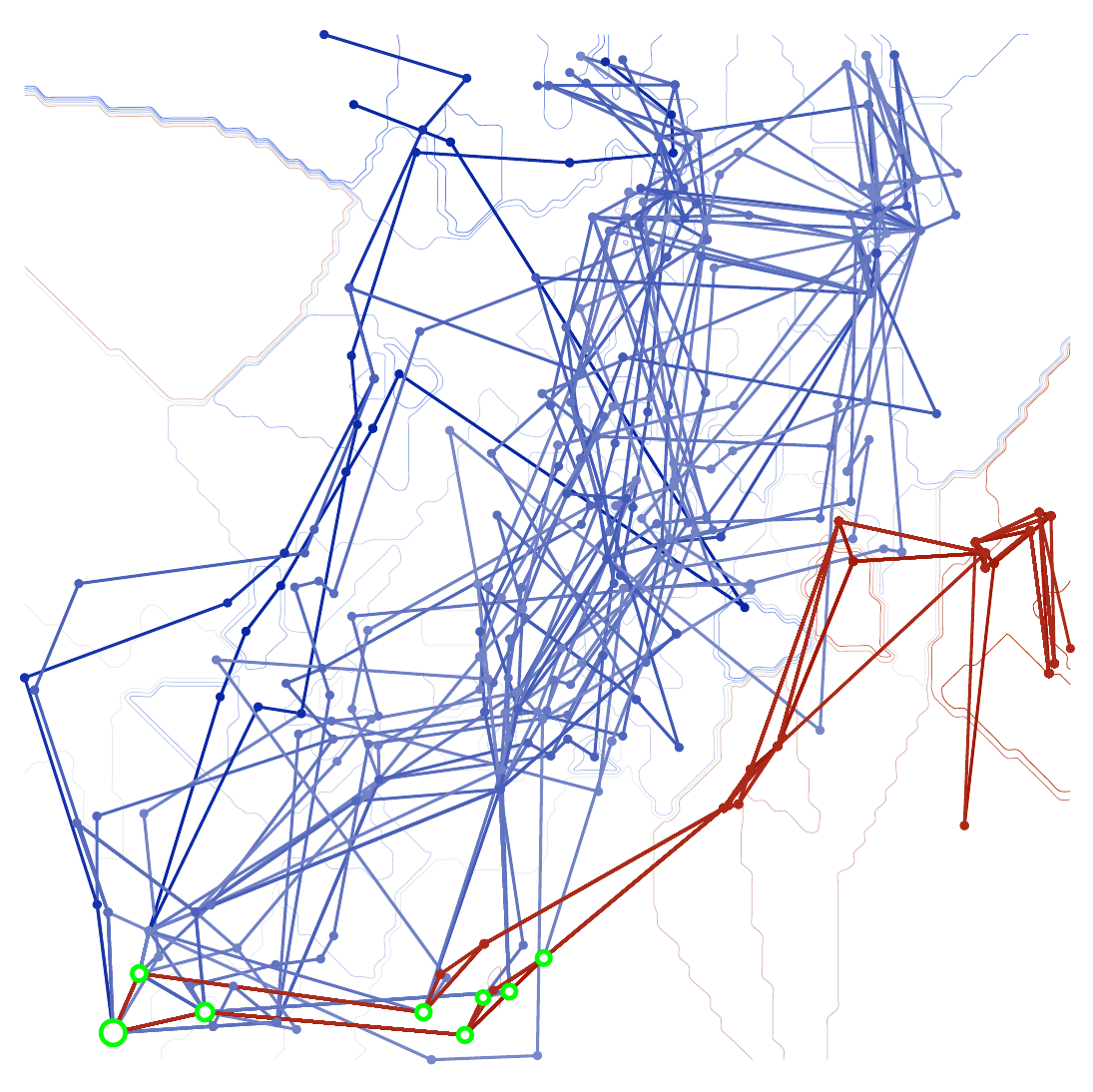}
        a) \textit{sce1} state landscape
    \end{minipage}
    \begin{minipage}{0.24\linewidth}
        \centering
        \includegraphics[width=\linewidth]{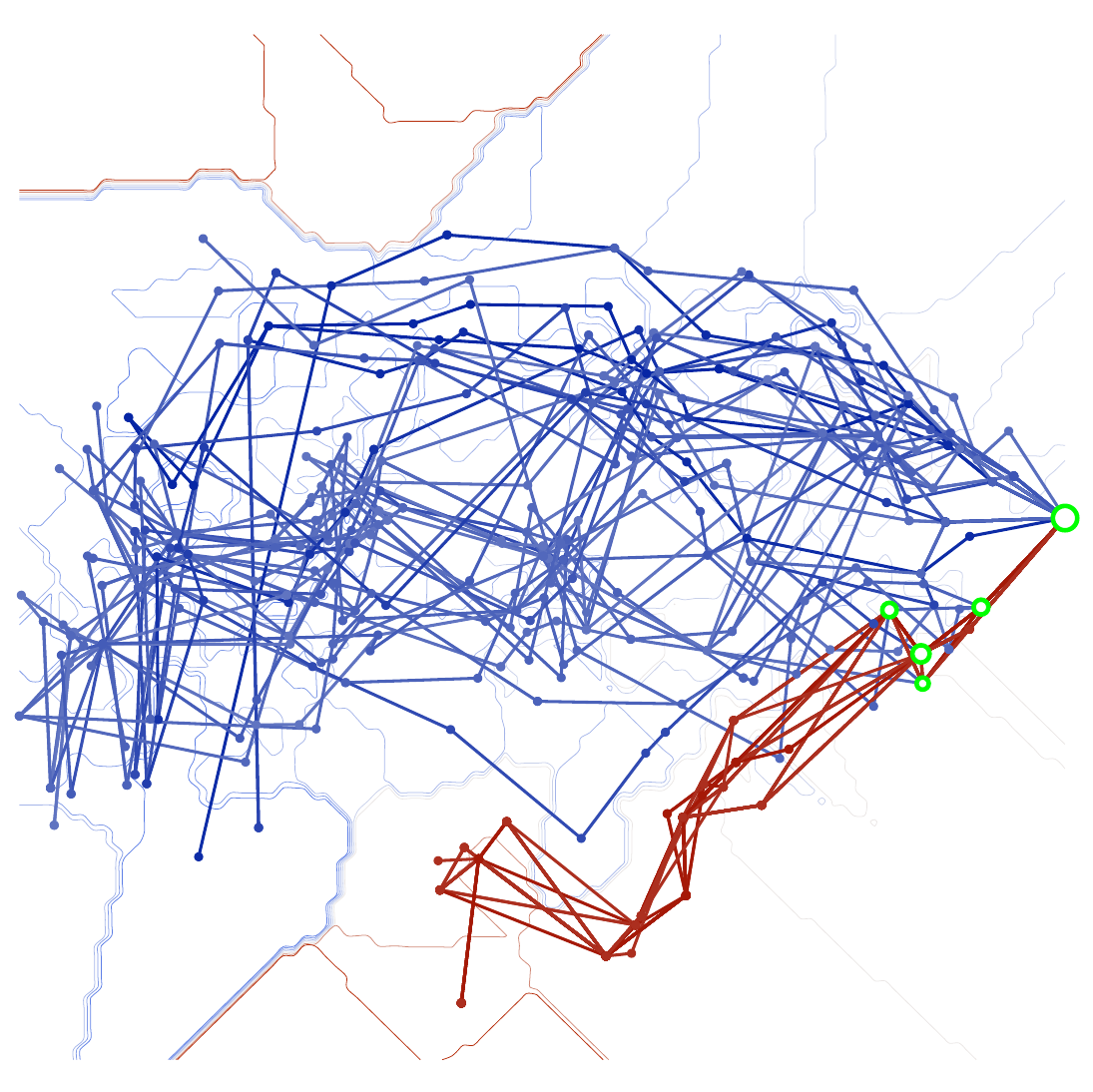}
        b) \textit{sce1m} state landscape
    \end{minipage}
    \begin{minipage}{0.24\linewidth}
        \centering
        \includegraphics[width=\linewidth]{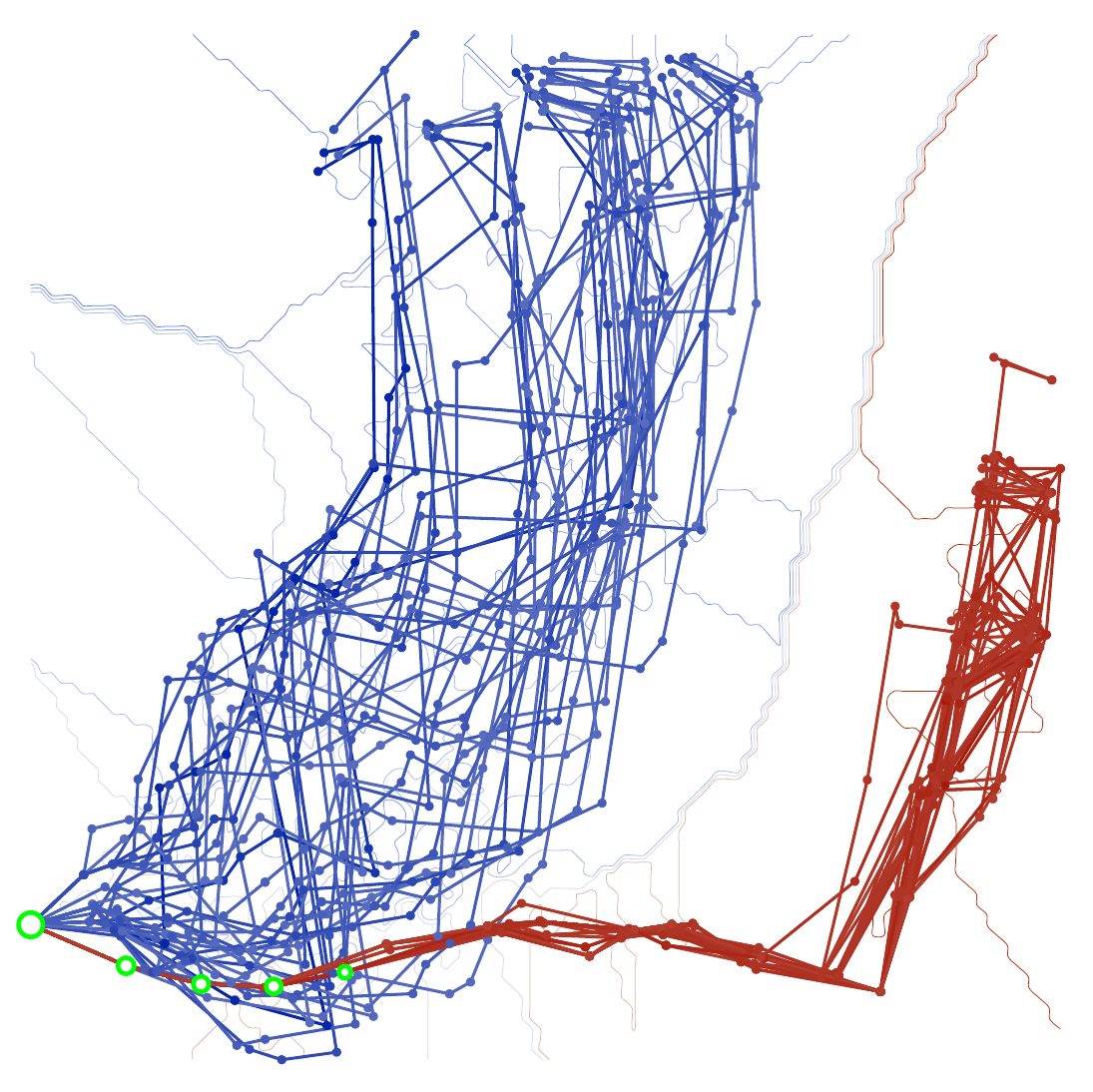}
        c) \textit{sce2} state landscape
    \end{minipage}
    \begin{minipage}{0.24\linewidth}
        \centering
        \includegraphics[width=\linewidth]{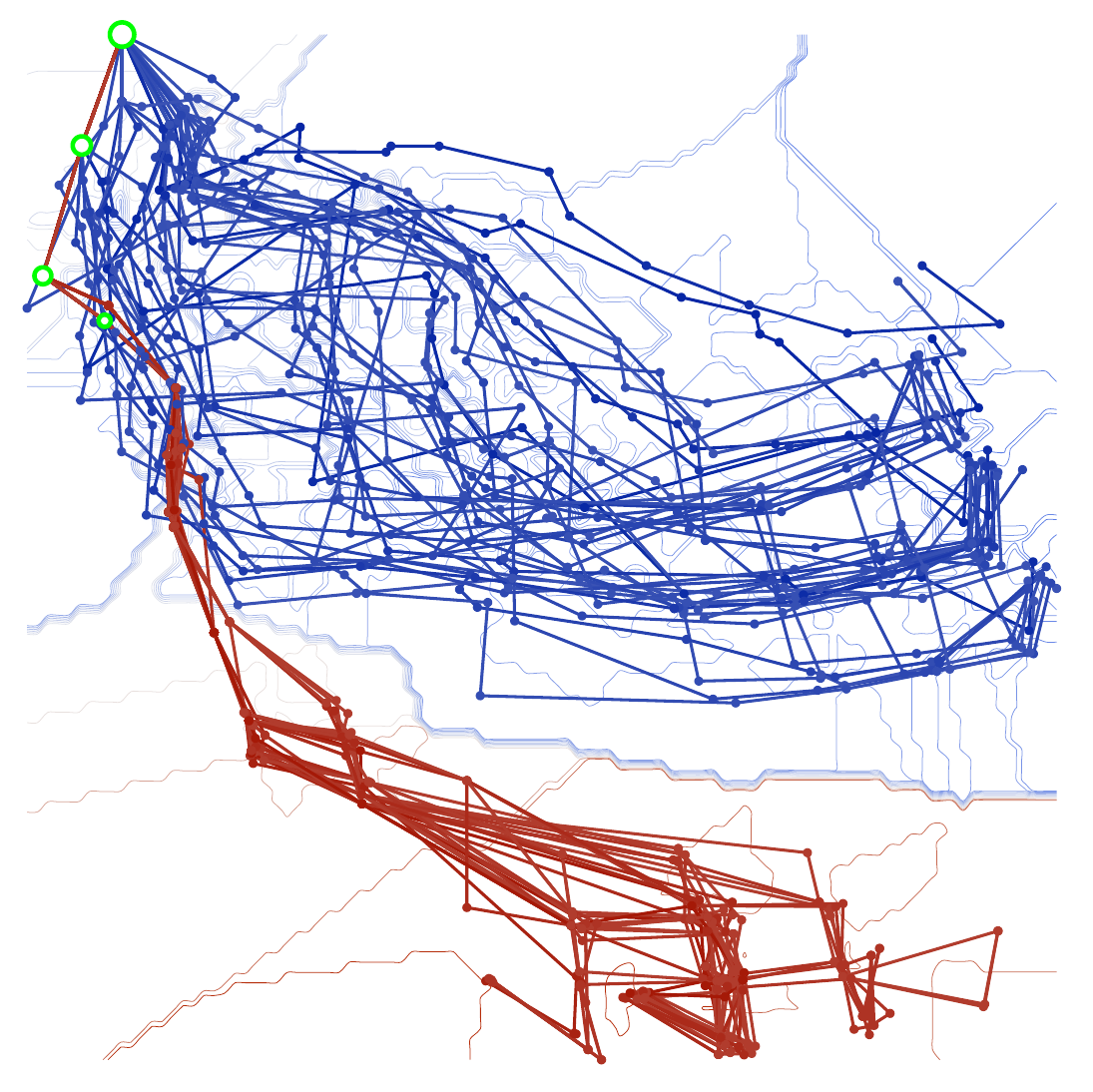}
        d) \textit{sce2m} state landscape
    \end{minipage}
    \\
    \begin{minipage}{0.24\linewidth}
        \centering
        \includegraphics[width=\linewidth]{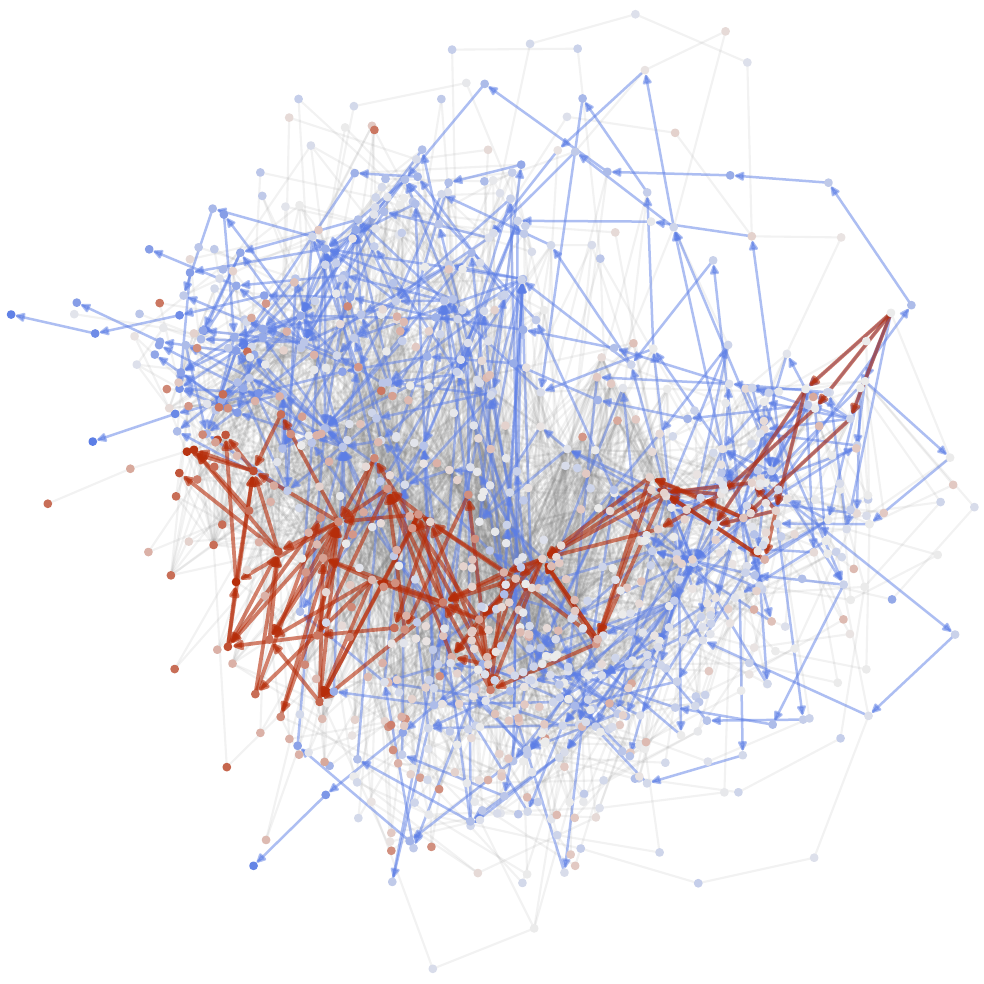}
        e) \textit{sce1} state graph
    \end{minipage}
    \begin{minipage}{0.24\linewidth}
        \centering
        \includegraphics[width=\linewidth]{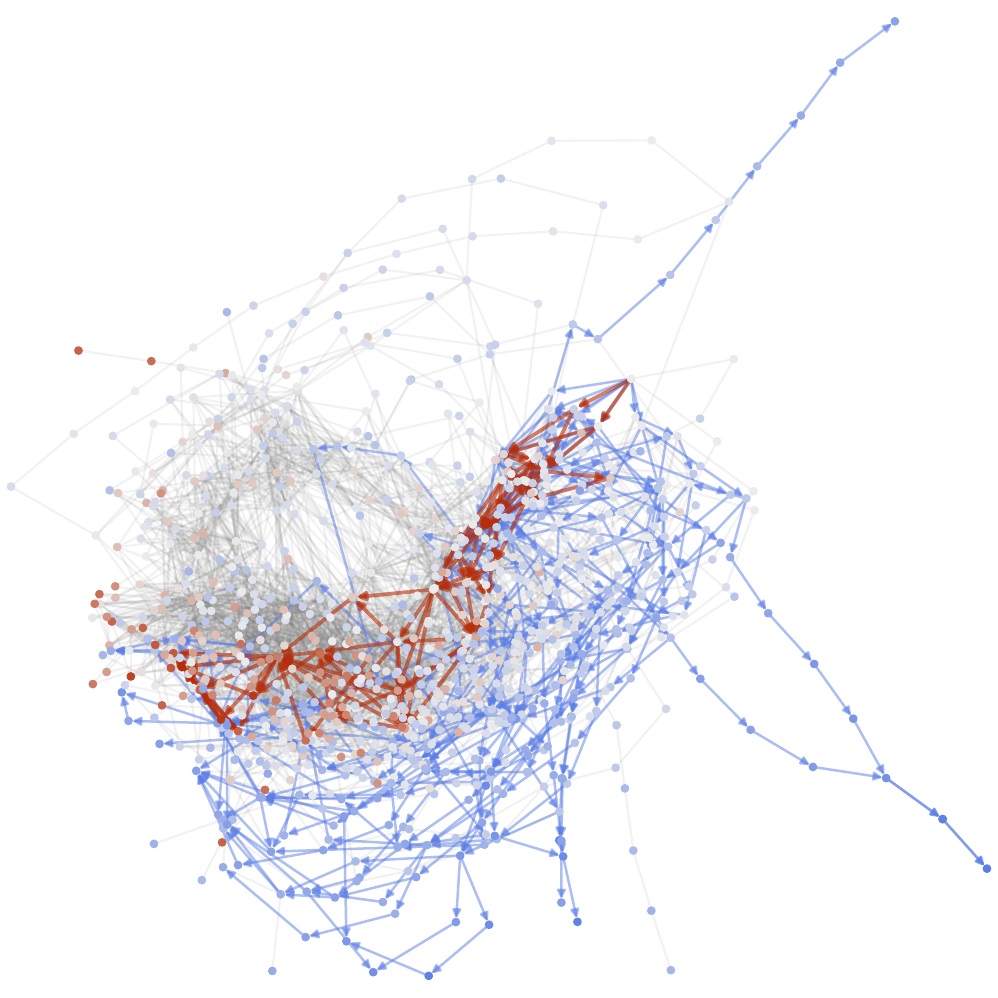}
        f) \textit{sce1m} state graph
    \end{minipage}
    \begin{minipage}{0.24\linewidth}
        \centering
        \includegraphics[width=\linewidth]{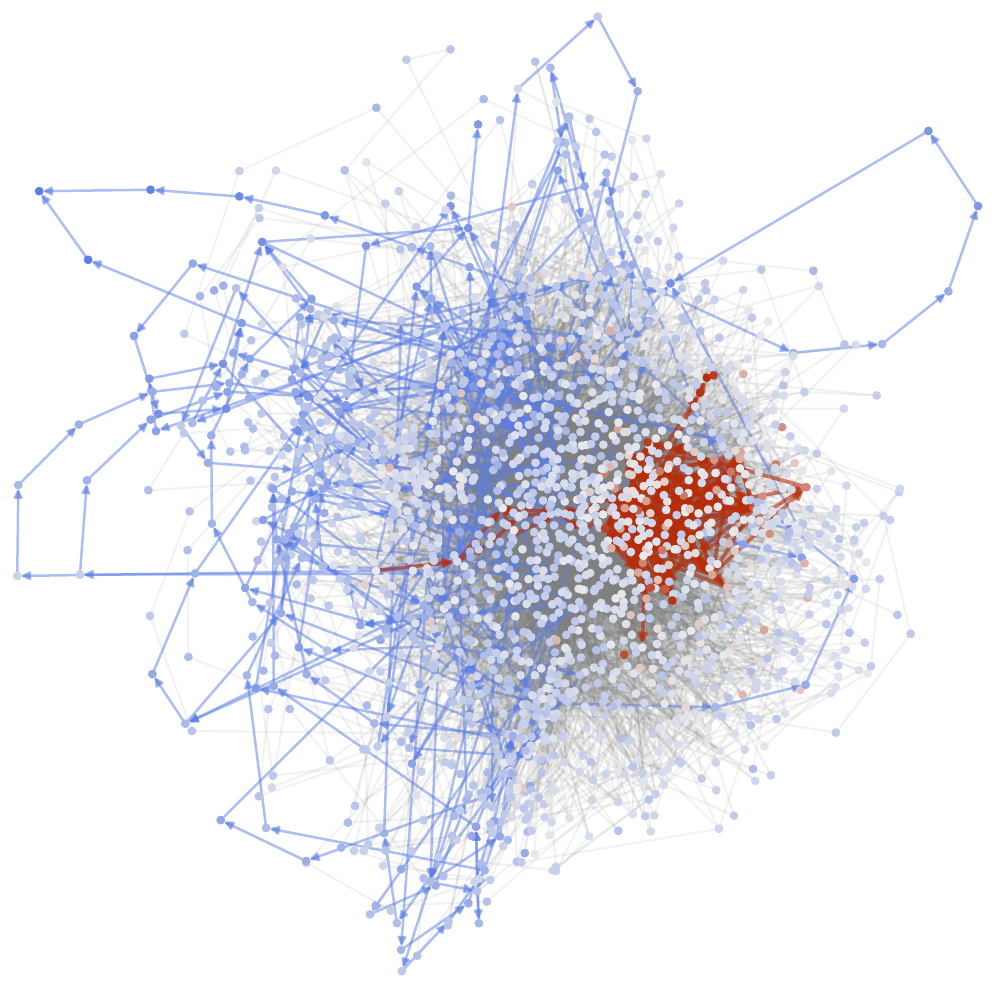}
        g) \textit{sce2} state graph
    \end{minipage}
    \begin{minipage}{0.24\linewidth}
        \centering
        \includegraphics[width=\linewidth]{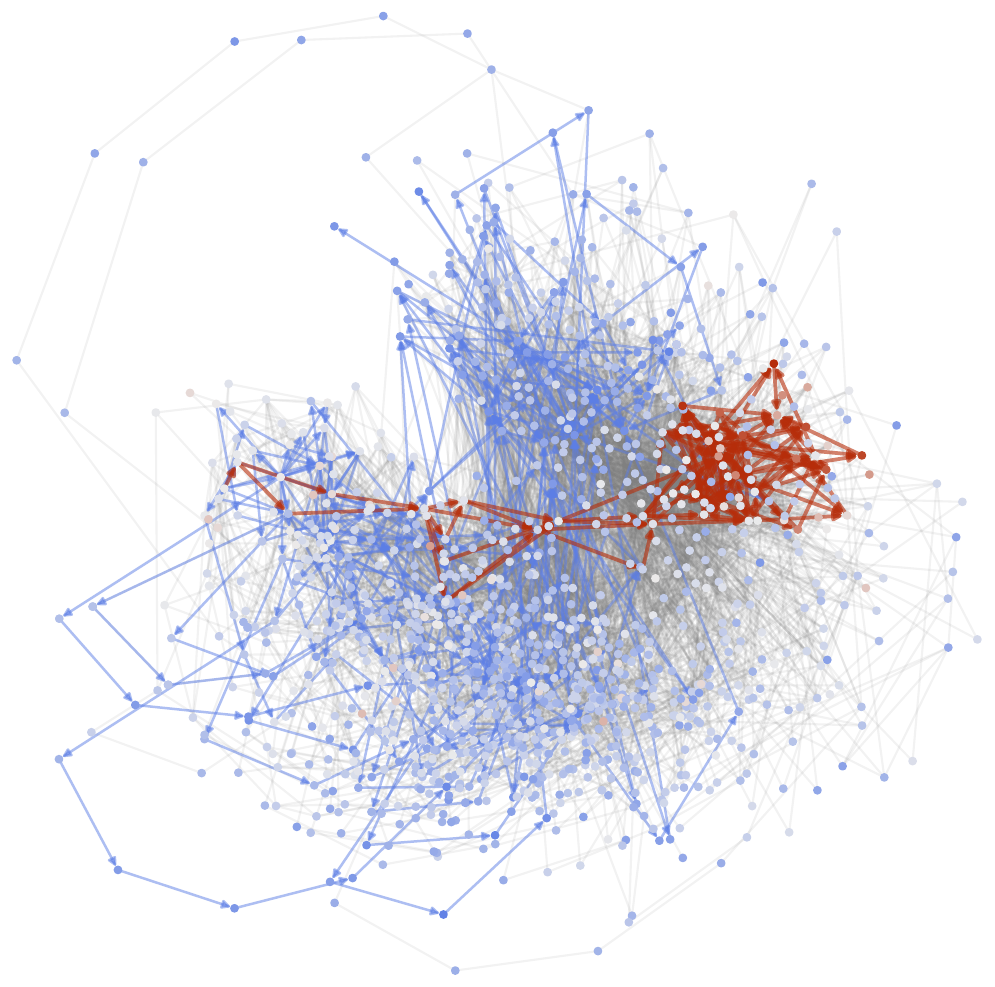}
        h) \textit{sce2m} state graph
    \end{minipage}
    \\
    \begin{minipage}{0.24\linewidth}
        \centering
        \includegraphics[width=\linewidth]{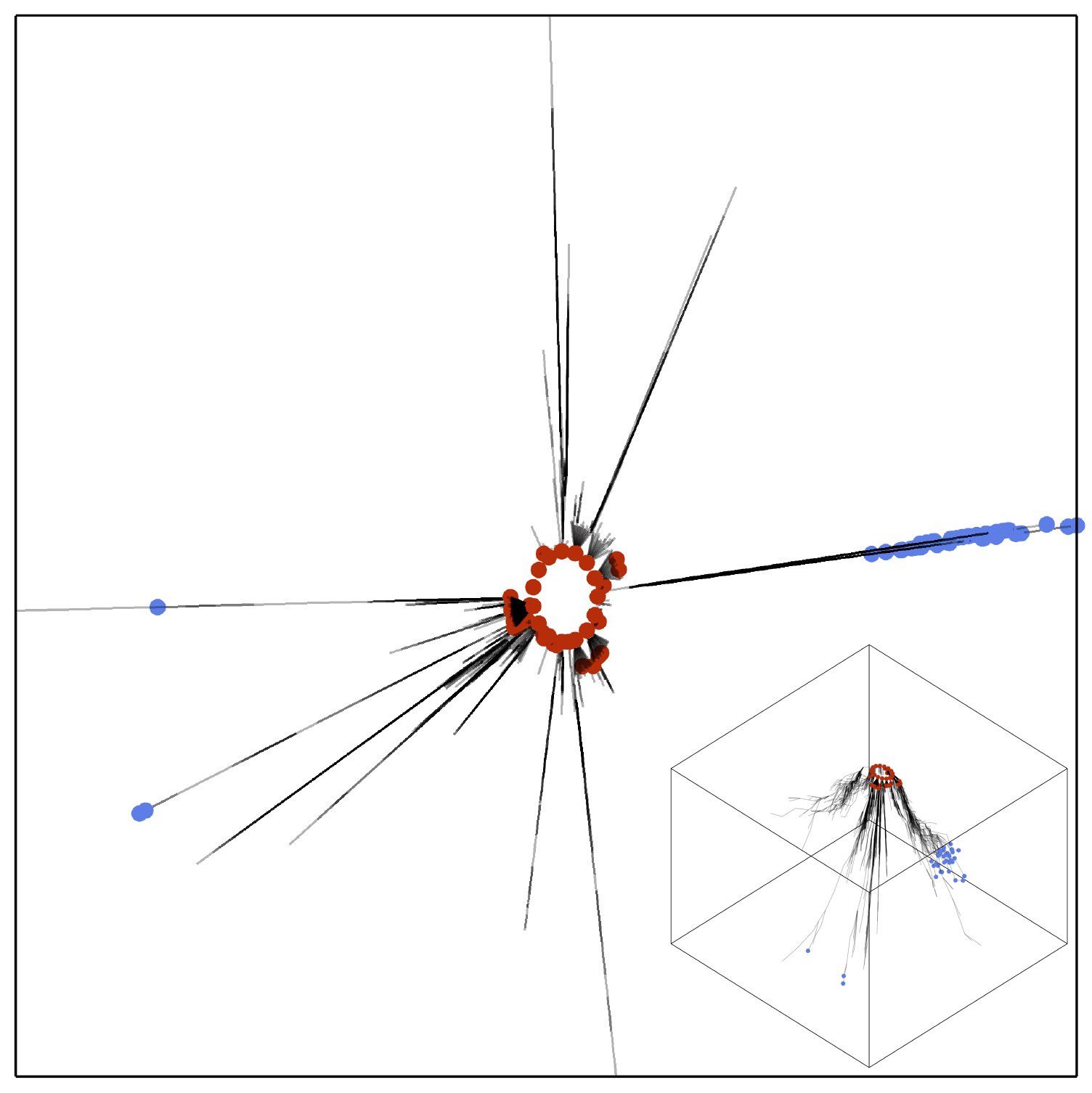}
        i) \textit{sce1} NBN
    \end{minipage}
    \begin{minipage}{0.24\linewidth}
        \centering
        \includegraphics[width=\linewidth]{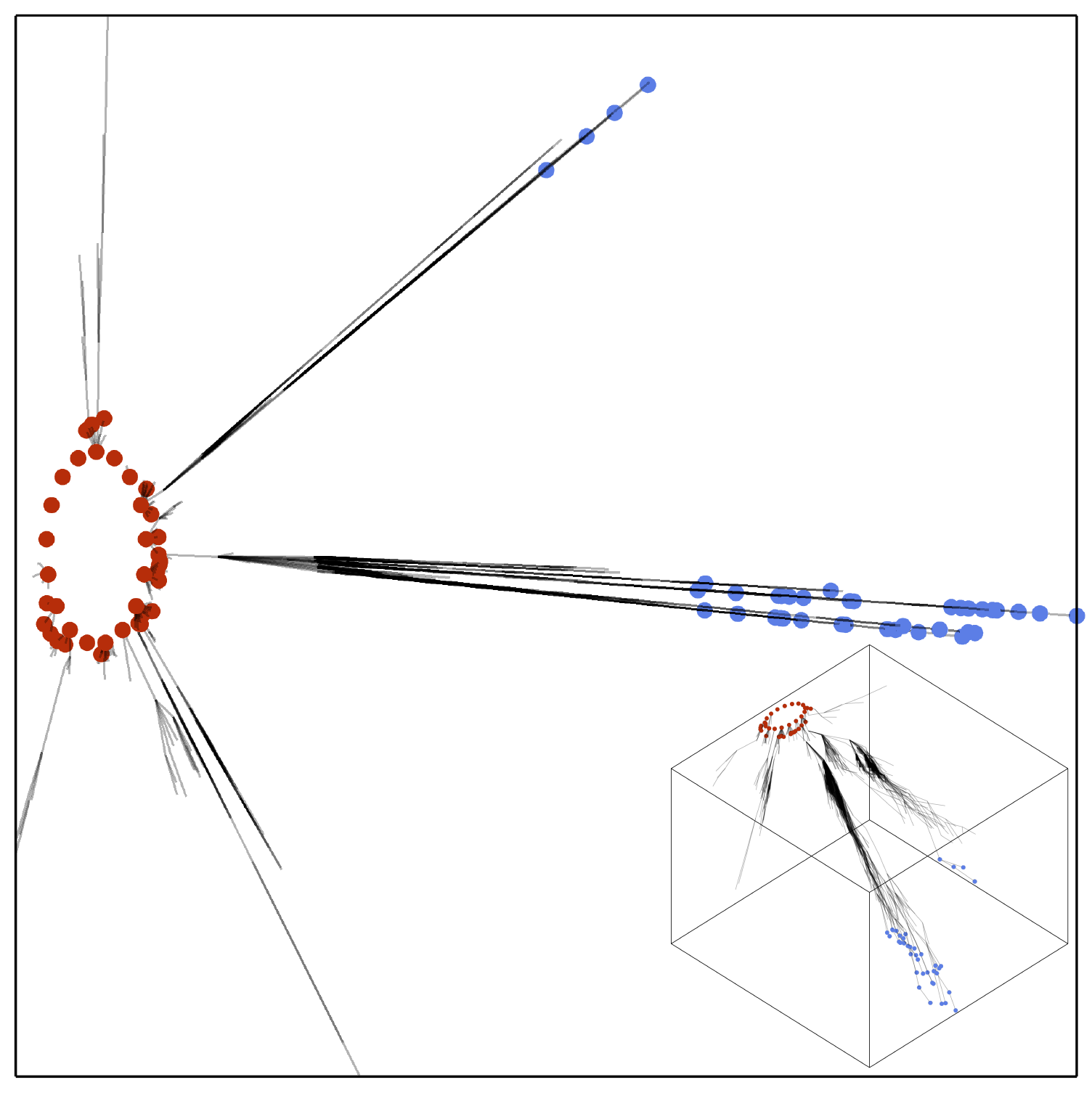}
        j) \textit{sce1m} NBN
    \end{minipage}
    \begin{minipage}{0.24\linewidth}
        \centering
        \includegraphics[width=\linewidth]{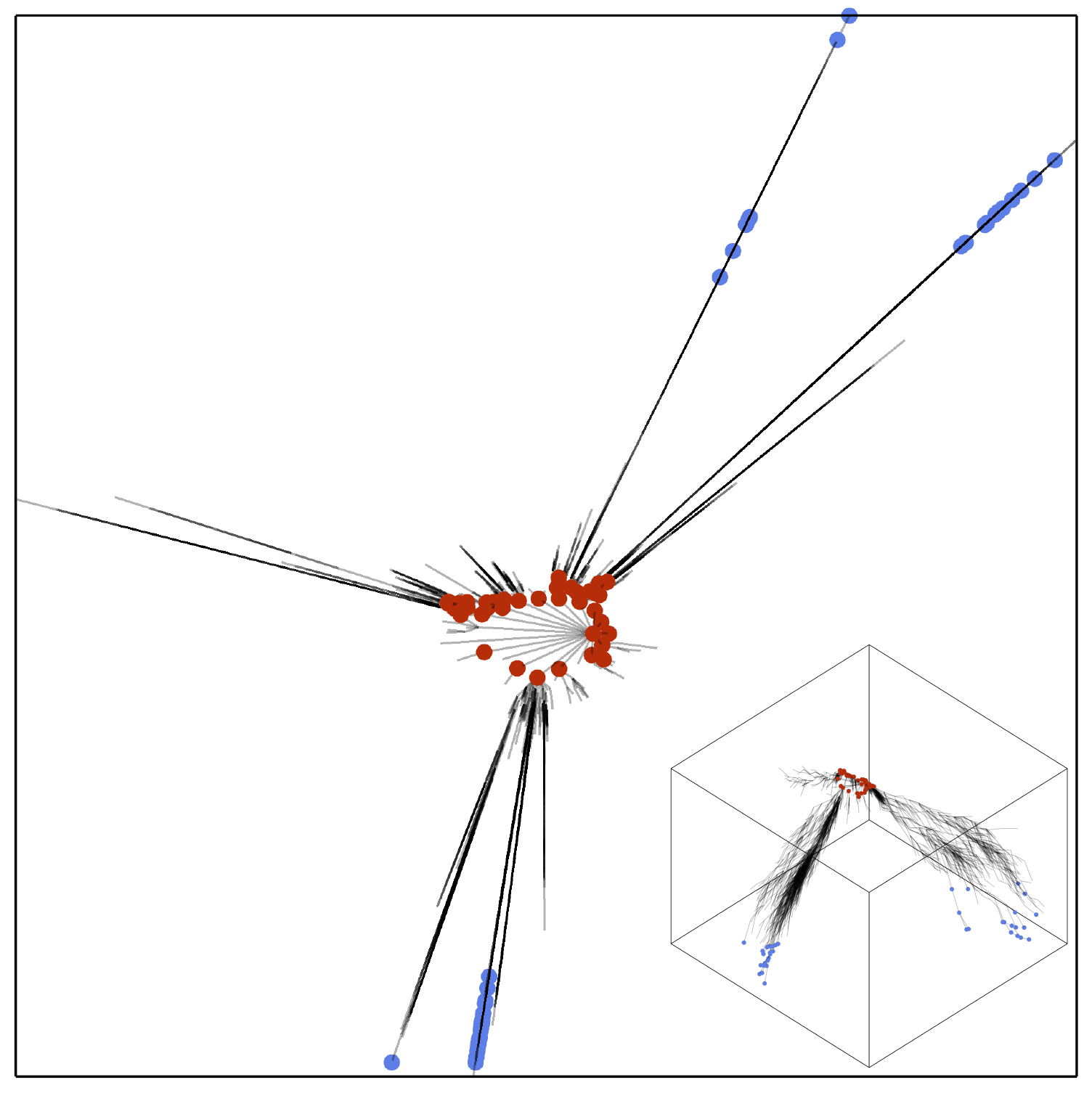}
        k) \textit{sce2} NBN
    \end{minipage}
    \begin{minipage}{0.24\linewidth}
        \centering
        \includegraphics[width=\linewidth]{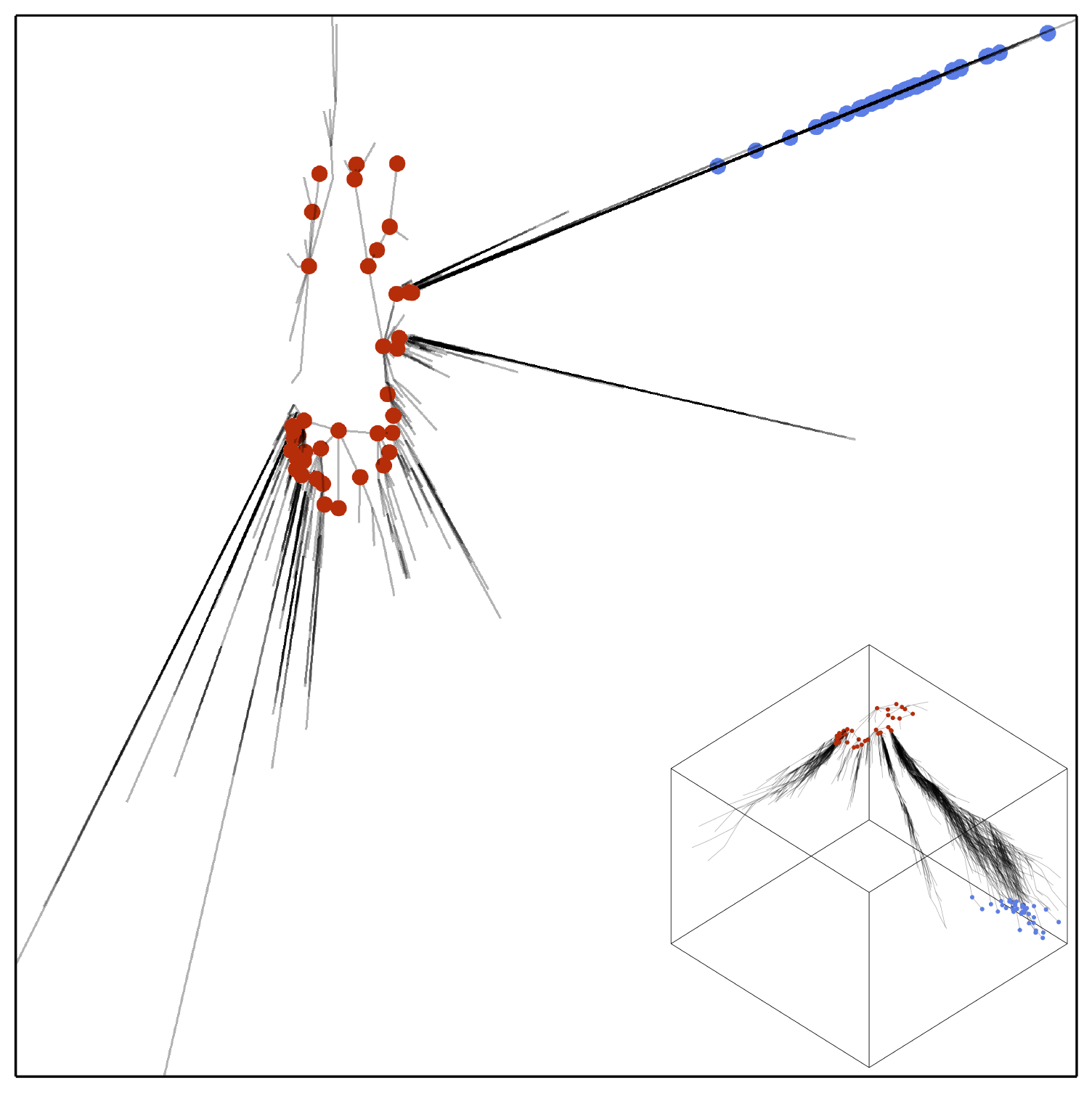}
        l) \textit{sce2m} NBN
    \end{minipage}
    \caption{Comparison of different analytical forms}
    \label{fig:analysis_pipeline_comparison}
\end{figure*}

The color analysis of pattern blocks reveals three main phenomena. First, many action pattern blocks show colors consistent with their corresponding cluster labels, indicating strong pattern-cluster correlations. Second, blocks within the same flow exhibit largely consistent colors, suggesting coherent action sequence development. Third, tactics demonstrate varying degrees of correlation with specific cluster colors, reflecting their differential effectiveness across different solution quality levels. These color-based observations collectively indicate clear causal relationships between tactics, action sequences, and clustering outcomes.

Comparative analysis across solution quality sets reveals significant differences in tactical behavior. High-fitness victorious solutions demonstrate greater tactical consistency and stronger causal relationships with clustering results, while low-fitness defeated solutions contain more numerous and dispersed tactics. This pattern highlights the critical role of opening tactics in achieving victorious solutions, with only a few high-fitness tactics successfully attaining higher fitness values, whereas numerous inefficient tactics lead to poor results. The pronounced gap in tactical scale between high and low-fitness solutions further confirms that specific, well-chosen tactics are key determinants of solution quality.

Tactical Sankey diagrams demonstrate effective pattern consolidation, as evidenced by the reduced number of blocks compared to action sequence diagrams. The enhanced contrast between tactic patterns in victorious versus defeated solutions becomes more pronounced in this simplified representation, further strengthening the conclusion that specific tactics are crucial factors in generating high-fitness solutions.

\subsection{Comparison of different analytical forms}
\label{sec:Comparison of different analytical forms}

Figure \ref{fig:analysis_pipeline_comparison} compares the SAT-RTS analysis pipeline (a--d) with state graph (e--h) and Nearest Better Network (NBN) \citep{diao_NearestBetter_2025} (i--l). To ensure a consistent baseline, all methods utilize the same data sampling, with red and blue markers denoting high-fitness victorious and low-fitness defeated solutions, respectively.

Graph networks serve as a versatile and efficient data structure for modeling the intricate relational dependencies within large-scale sequential data. A primary limitation of graph-based network structures lies in their inability to clearly represent local proximity relationships between states. While edge weights may, to some extent, reflect inter-nodal distances, such representations lack clarity in large-scale graphs and are inherently constrained by the underlying graph layout algorithms.

As an advanced topological abstraction for fitness landscape analysis, NBN offers a uniquely structured, macro-level perspective for characterizing complex problem features from a search perspective. However, while it effectively maps the global correspondence between objective and decision spaces, NBN is limited in its ability to delineate the fine-grained coupling relationships inherent in temporal sequential solutions.

The SAT-RTS pipeline achieves a synergistic coupling of spatial geometric structures and dynamic evolution logic. By projecting high-dimensional state-transition sequences onto a continuous state value landscape, this approach preserves the local proximity between states while explicitly visualizing the tactical gradients that drive transition trajectories. Unlike discrete graph layouts, the state landscape provides a globally consistent reference frame where the convergence of victorious solutions and the divergence of defeated ones are intuitively localized. This spatial grounding facilitates the precise identification of critical decision points, where divergent tactical selections from identical states yield polarized outcomes, effectively bridging the gap between static structural features and underlying decision-making causalities.

\subsection{Comprehensive interpretable tactical analysis}
\label{sec:Comprehensive interpretable tactical analysis}

\begin{figure*}[th]
    \centering
    \begin{minipage}{0.32\linewidth}
        \centering
        \includegraphics[width=\linewidth]{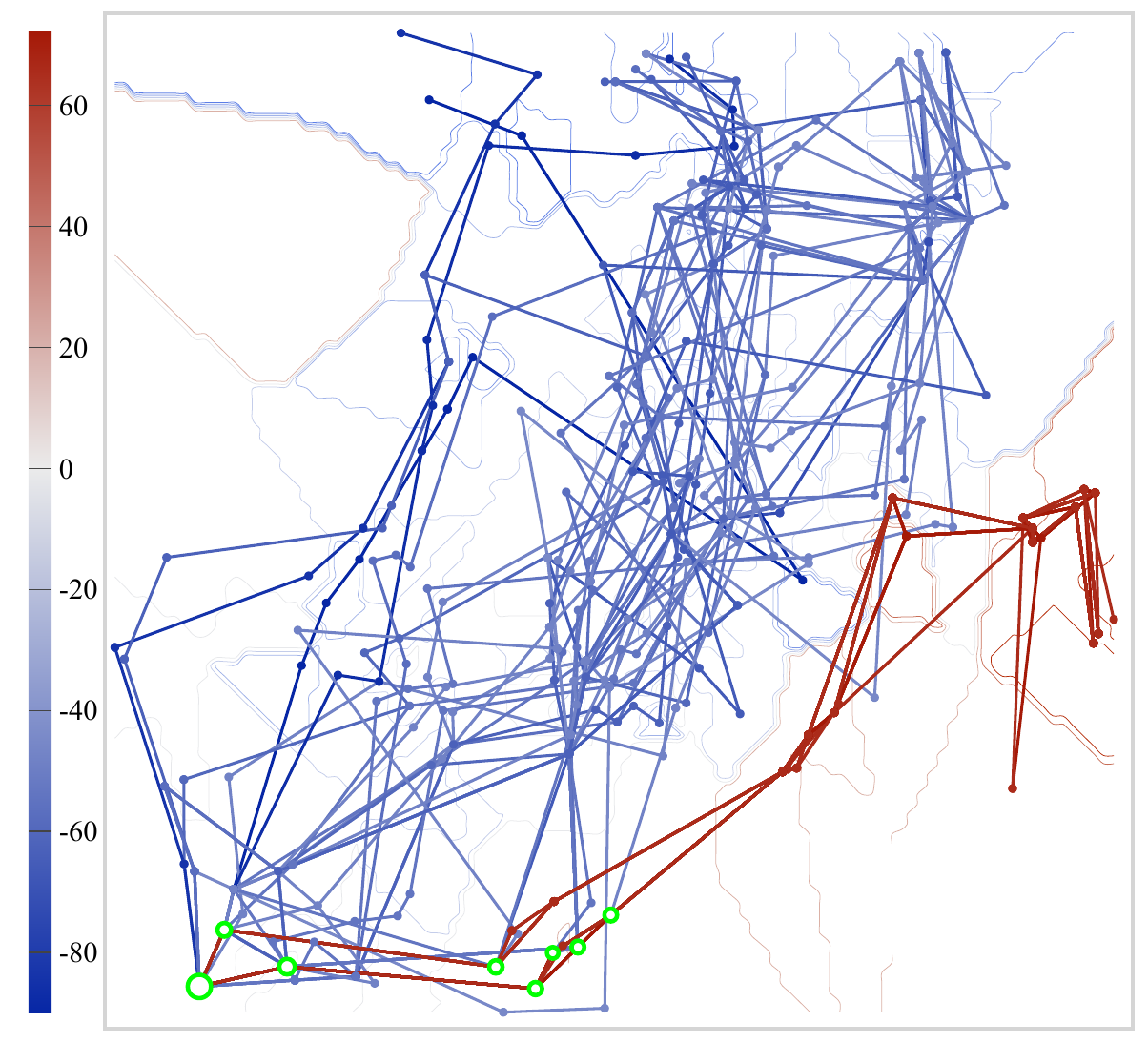}
        a) \textit{sce1}
    \end{minipage}
    \begin{minipage}{0.32\linewidth}
        \centering
        \includegraphics[width=\linewidth]{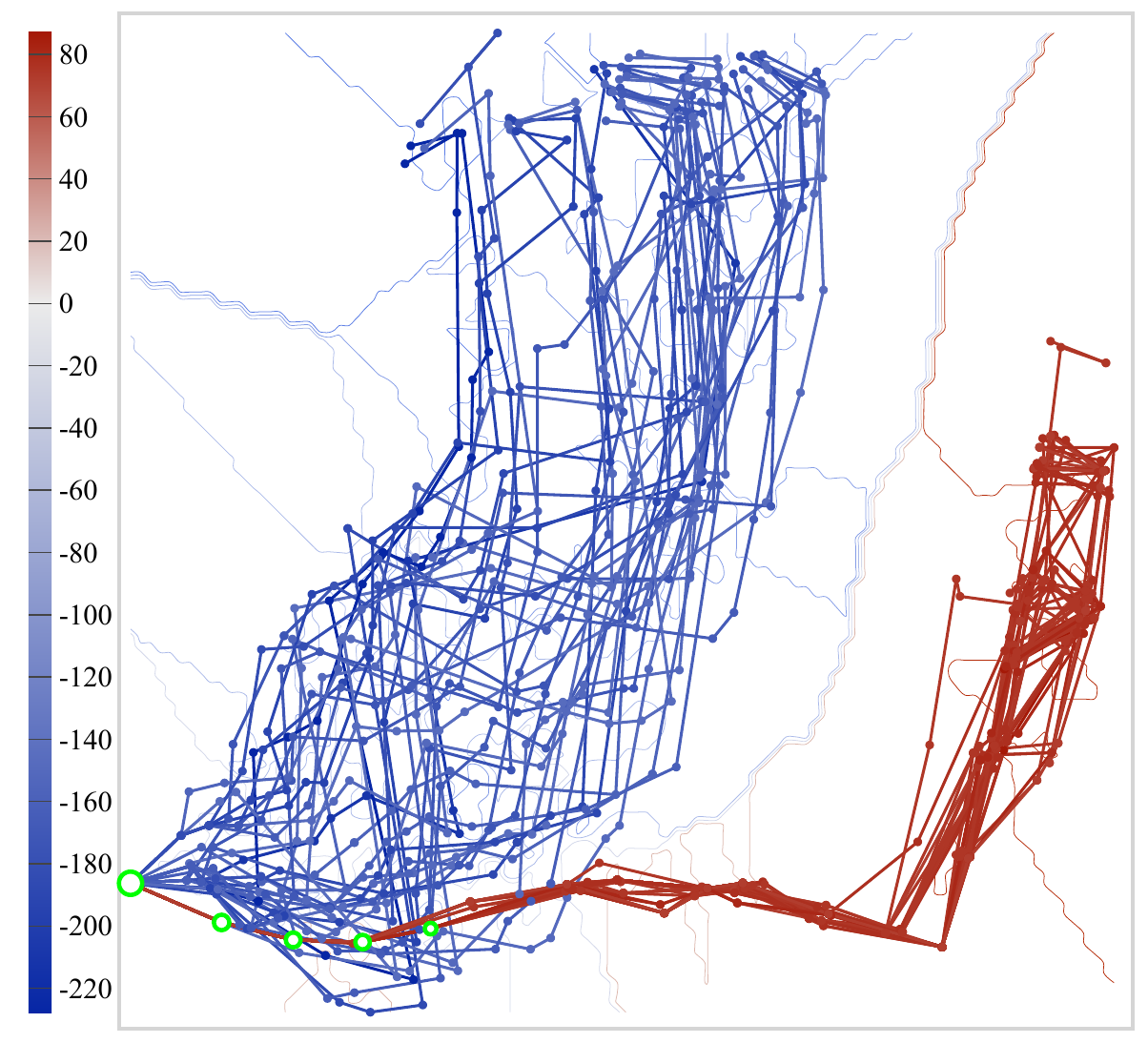}
        b) \textit{sce2}
    \end{minipage}
    \begin{minipage}{0.32\linewidth}
        \centering
        \includegraphics[width=\linewidth]{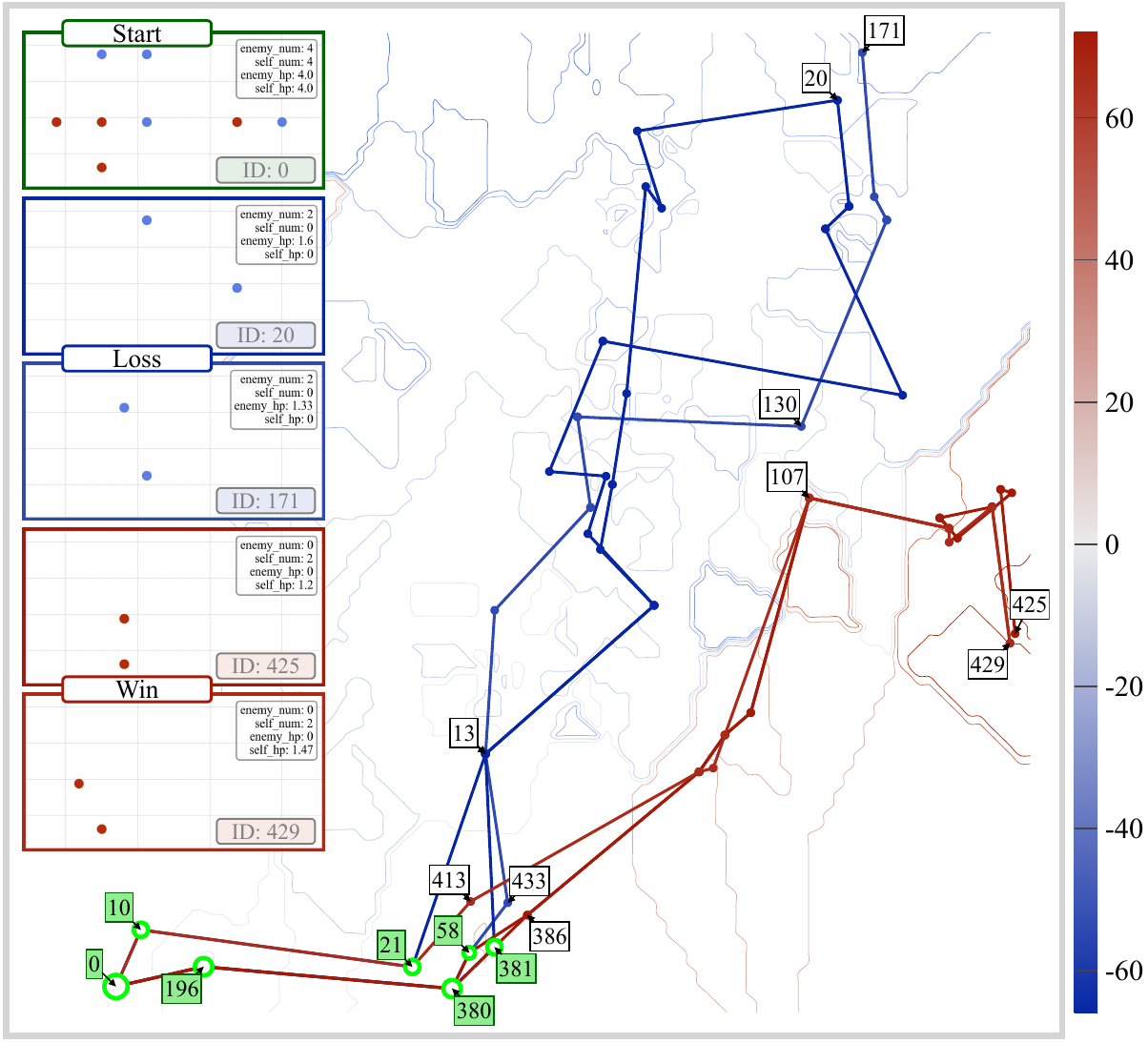}
        c) representative solutions in \textit{sce1}
    \end{minipage}
    \caption{State-transition networks in state value landscapes}
    \label{fig:state_landscape}
\end{figure*}

State clustering and visualization provide highly interpretable analysis for action sequence pattern mining and tactic extraction. Figure~\ref{fig:state_landscape}~a) and Figure~\ref{fig:state_landscape}~b) illustrate state-transition networks of sampled solutions in the state value landscape (following the same methodology as Sec.~\ref{sec:Action sequence pattern mining and tactic extraction}, where high-fitness victorious and low-fitness defeated solutions are sampled based on fitness values and distinguished by a color bar). In the figure, contour lines represent the state value landscape (simplified from Figure~\ref{fig:state_value_landscape}), nodes and edges represent states and their transitions in sampled solutions, and green nodes represent states that appear in both victorious and defeated solutions (such as the initial state, the node with no in-degree in the bottom-left corner).

\begin{figure*}[th]
    \centering
    \includegraphics[width=\linewidth]{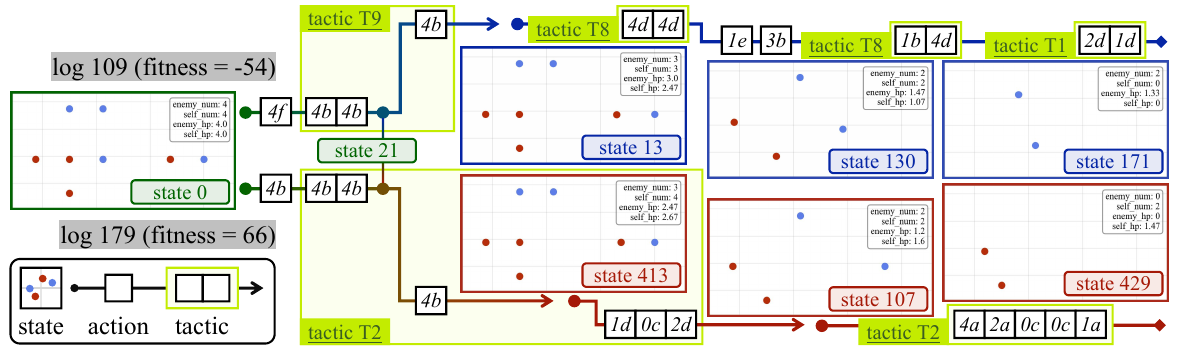}
    \vspace{-20pt}
    \caption{Detailed process of the two sample solutions (solution 109 and 179)}
    \label{fig:interpretable_analysis}
\end{figure*}

Two main conclusions can be summarized from state-transition networks. First, state-transition networks of low-fitness solutions are more disordered and extensive, while high-fitness solutions are more concentrated, indicating that only a few specific state space exploration strategies are high-quality and can lead to final victory, whereas aimless exploration typically yields low returns. Second, there exists a clear distributional separation between high-fitness and low-fitness solutions. This conforms to the heuristic expectation that recovering from a disadvantageous position is notably difficult, which is further accentuated in the more complex \textit{sce2} scenario.

Considering the win-loss differences caused by different tactics applied to the same node, the points represented by green nodes can be regarded as critical decision points. Figure~\ref{fig:state_landscape}~c) illustrates two victorious and two defeated solutions as examples, with representative state IDs and their distributions annotated in the figure. Figure~\ref{fig:interpretable_analysis} details the processes of the two sample solutions. The primary elements comprise states, script-based actions, and the corresponding tactics, the coding scheme and descriptions of which are detailed in \text{Table~S4} in Section S5 of the Supplementary Material.

Taking solutions 109 and 179, which both pass through state 21, as examples, the transition [state 21-state 13] leads to defeat in solution 109, while transition [state 21-state 413] leads to victory in solution 179. After this point, the difference between the two example solutions becomes increasingly apparent. The tactics executed around the critical decision point 21 warrant special attention. Figure~\ref{fig:interpretable_analysis} reveals that the main reason for solution 109's failure is the execution of action 4f (a defensive script) in the opening phase, which wasted firepower attack opportunities. Although subsequent actions were identical to those in the victorious solution 179, significant differences in unit health values had already emerged at states 13 and 413. From state 13 onward, the disadvantageous situation gradually accumulated, ultimately resulting in failure. The main reason for solution 179's success is the execution of T2 tactic (greedy attack with collaborative mode switching) in the opening phase, specifically embodied in the action sequence [4b4b4b] representing all red units collectively concentrating fire on the nearest unit (the blue unit at the center). After eliminating this unit, the subsequent [1d0c2d] represents red units performing local collaborative attacks on nearby weakest units according to distribution, thereby establishing a significant advantage.

\begin{figure*}[th]
    \centering
    \begin{minipage}{0.48\linewidth}
        \centering
        \includegraphics[width=\linewidth]{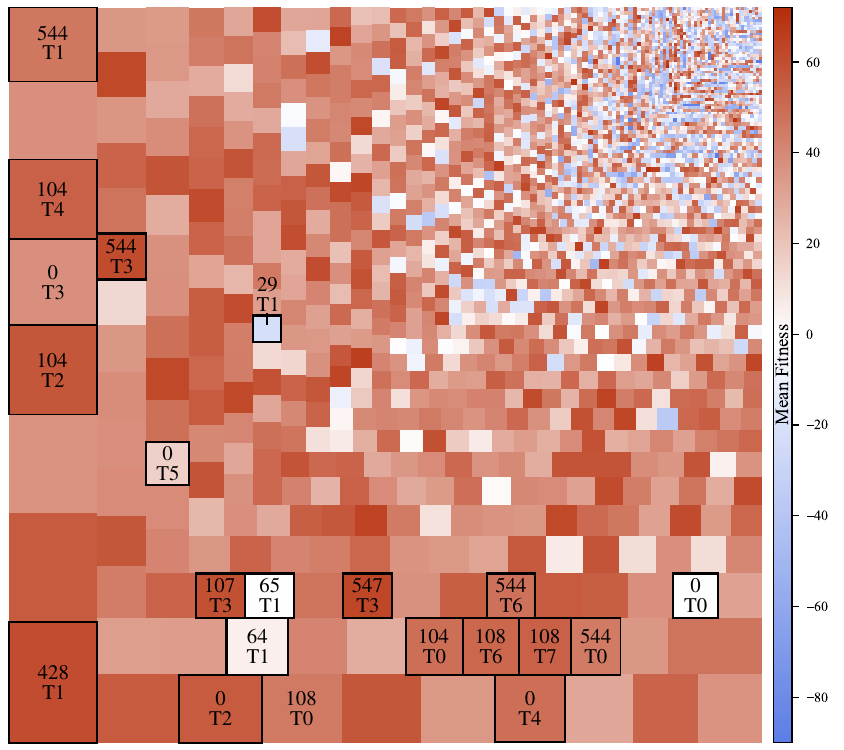}
        a) \textit{sce1}
    \end{minipage}
    \begin{minipage}{0.48\linewidth}
        \centering
        \includegraphics[width=\linewidth]{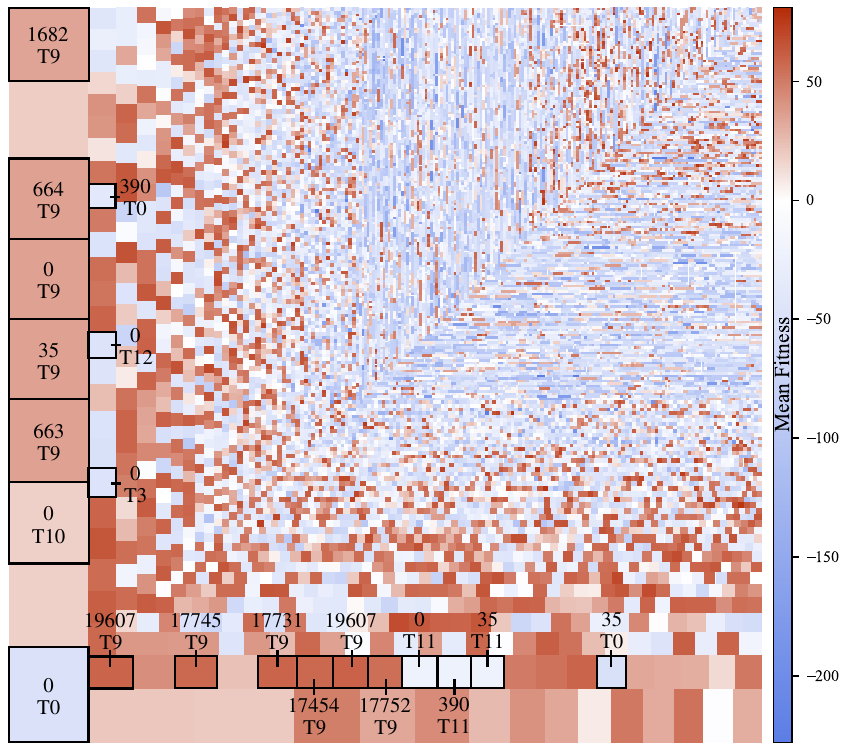}
        b) \textit{sce2}
    \end{minipage}
    \caption{Treemap of state-tactic payoff correlation}
    \label{fig:state_tactic_correlation}
\end{figure*}

Tactics themselves are not inherently good or bad, but different tactics should be adapted to different states. To further analyze the relationship between states and tactics, cross-statistical analysis has been performed. By counting all (state, tactic, fitness value) triplets in the sample, Figure~\ref{fig:state_tactic_correlation} shows a treemap of state-tactic pairs. Each rectangle represents a specific tactic executed in a particular state, with its area representing frequency and its color representing average outcome. Specifically, larger rectangles indicate higher frequency of executing that tactic in that state, while colors closer to deep red represent higher fitness values when executing that tactic in that state, and vice versa.

The treemap visualization and detailed analysis in \text{Table~S5} and \text{Table~S6} in Section S7 of the Supplementary Material reveal several key insights into state-tactic relationships. First, the effectiveness of tactics is highly state-dependent, with specific tactics showing optimal performance in particular state configurations. For instance, independent greedy attacks excel in states with clear numerical advantages, while coordinated tactics prove essential in scenarios with dispersed targets. Second, opening tactics play a crucial role in determining final outcomes, as evidenced by the significant performance differences in state 0 across various tactical approaches. Third, the broad applicability of certain tactics like T9 in \textit{sce2} demonstrates that effective tactics can adapt to diverse state conditions while maintaining high performance. These findings underscore the importance of tactical adaptability and state-aware decision-making in achieving optimal outcomes. Additionally, the characteristics presented in treemaps differ across varying difficulty scenarios. In the challenging \textit{sce2} scenario, high-yield state-tactic pairs are significantly fewer, manifestly demonstrating the differences in state exploration difficulty between the two scenarios.

\section{Conclusion}
\label{sec:Conclusion}
This work presents the SAT-RTS method, an interpretable state-action-tactic analysis pipeline designed to address the analytical challenges of high-dimensional coupled data in RTS micromanagement. By implementing multi-aspect similarity assessments, the framework successfully quantifies relationships across states, state-transitions, and action patterns, establishing lightweight normative criteria necessary for high-level analysis. The integration of the cluster-centric BK-tree algorithm with fitness landscape visualization enables the efficient simplification of massive state data streams, thereby enhancing the structural clarity of the underlying problem and creating a multi-level visual toolset. Furthermore, the rule-based multi-label tactic extraction method, alongside comprehensive attribution analyses, effectively elucidates the decision-making logic inherent in otherwise opaque black-box behaviors. Ultimately, the proposed pipeline achieves a cohesive and interpretable tactical analysis that bridges the gap between low-level actions and high-level tactics.

Future works will focus on integrating fitness landscape visualization into knowledge-guided learning models, aiming to establish a functional bridge between RL and intelligent optimization. Furthermore, efforts will be directed toward mining the temporal causal relationships of action sequence patterns. By exploring more robust and universally applicable tactical knowledge mining methods, subsequent studies seek to gain a deeper understanding of the complex, multi-objective characteristics inherent in RTS micromanagement.

\section*{Acknowledgements}
This work was supported in part by the National Natural Science Foundation of China under Grants 62476006, 62506004, and 62306279, in part by the Hubei Provincial Natural Science Foundation of China under Grant 2023AFA049, and in part by the Fundamental Research Funds of the AUST under Grant 2024JBZD0007.

\section*{Acknowledgements}
This work was supported in part by the National Natural Science Foundation of China under Grants 62476006, 62506004, and 62306279, in part by the Hubei Provincial Natural Science Foundation of China under Grant 2023AFA049, and in part by the Fundamental Research Funds of the AUST under Grant 2024JBZD0007.

\bibliographystyle{elsarticle-num}
\bibliography{reference/ref-abbr}

\clearpage
\appendix
\section{Supplementary Material for SAT-RTS}
This document provides supplementary technical evidence for the paper "SAT-RTS: A systematic framework for tactical knowledge extraction and visualization-based analysis in real-time strategy games", including clustering visualization, parameter sensitivity studies, tactical codebook definitions, and extended experimental metrics. All figures and tables herein are prefixed with "S" (e.g., Figure S1, Table S1).

\renewcommand{\thesubsection}{S\arabic{subsection}}

\setcounter{figure}{0}
\setcounter{table}{0}
\renewcommand{\thefigure}{S\arabic{figure}}
\renewcommand{\thetable}{S\arabic{table}}
\renewcommand{\theHfigure}{S\arabic{figure}}
\renewcommand{\theHtable}{S\arabic{table}}

\makeatletter
\renewcommand{\fnum@figure}{Figure~\thefigure}
\renewcommand{\fnum@table}{Table~\thetable}
\makeatother

\subsection{Comparison of t-SNE Visualization of Clustering Results}

\begin{figure*}[ht]
    \centering
    \begin{minipage}{0.19\linewidth}
        \centering
        \includegraphics[width=\linewidth]{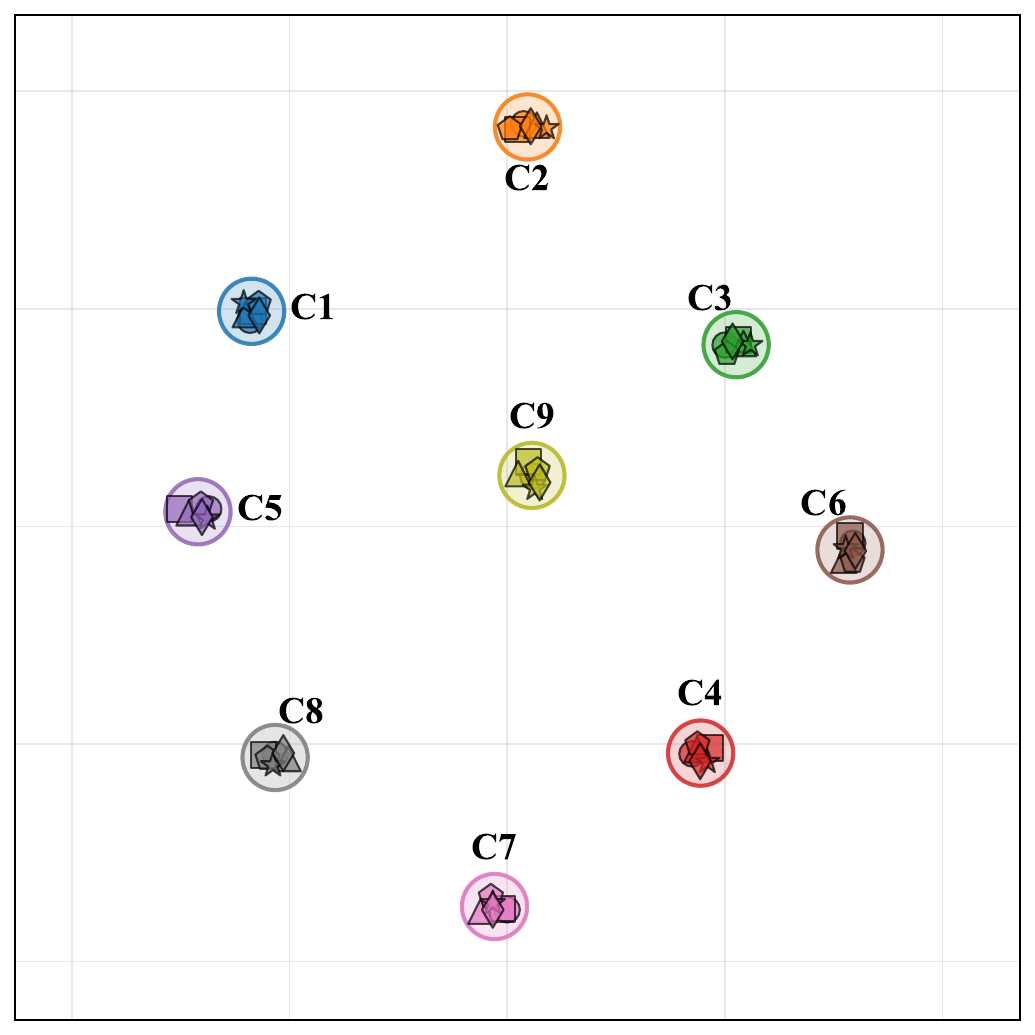}
        Chamfer ($k=9$)
    \end{minipage}
    \begin{minipage}{0.19\linewidth}
        \centering
        \includegraphics[width=\linewidth]{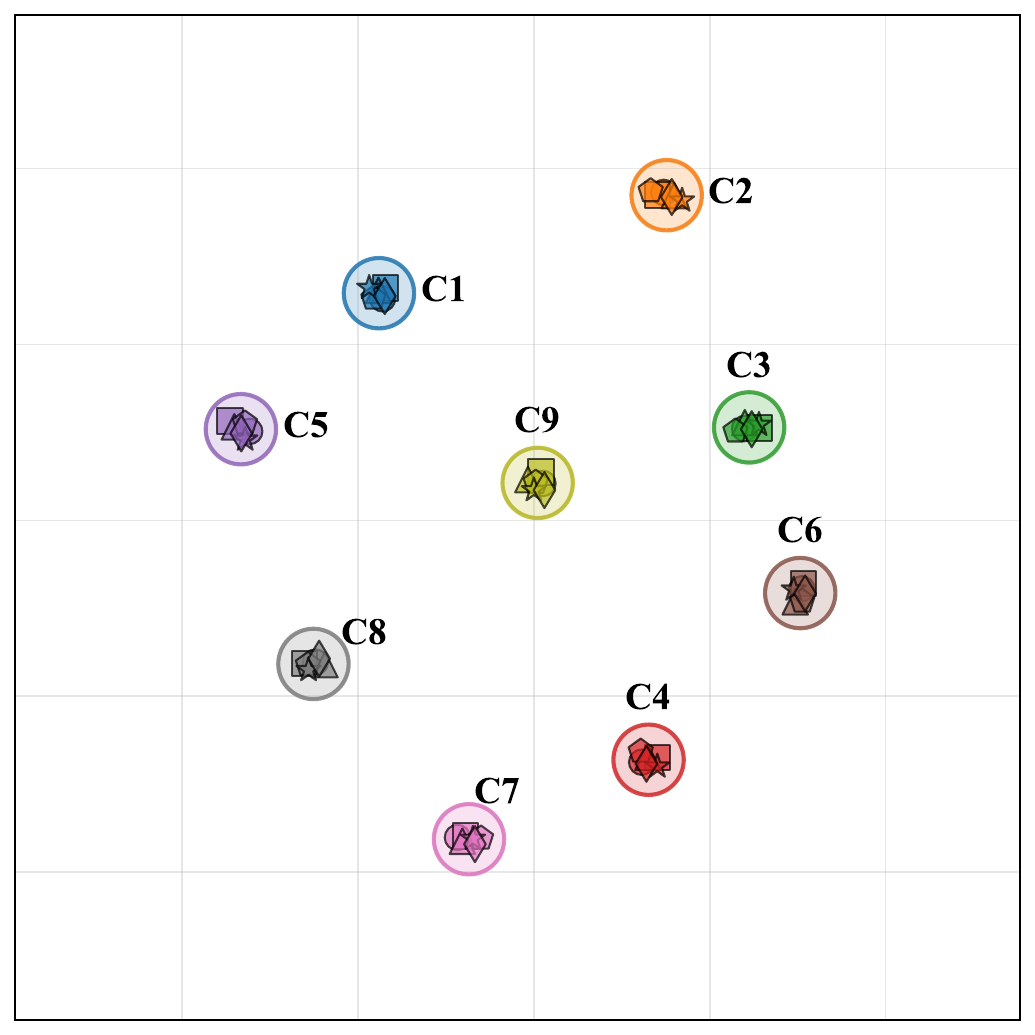}
        Hausdorff ($k=9$)
    \end{minipage}
    \begin{minipage}{0.19\linewidth}
        \centering
        \includegraphics[width=\linewidth]{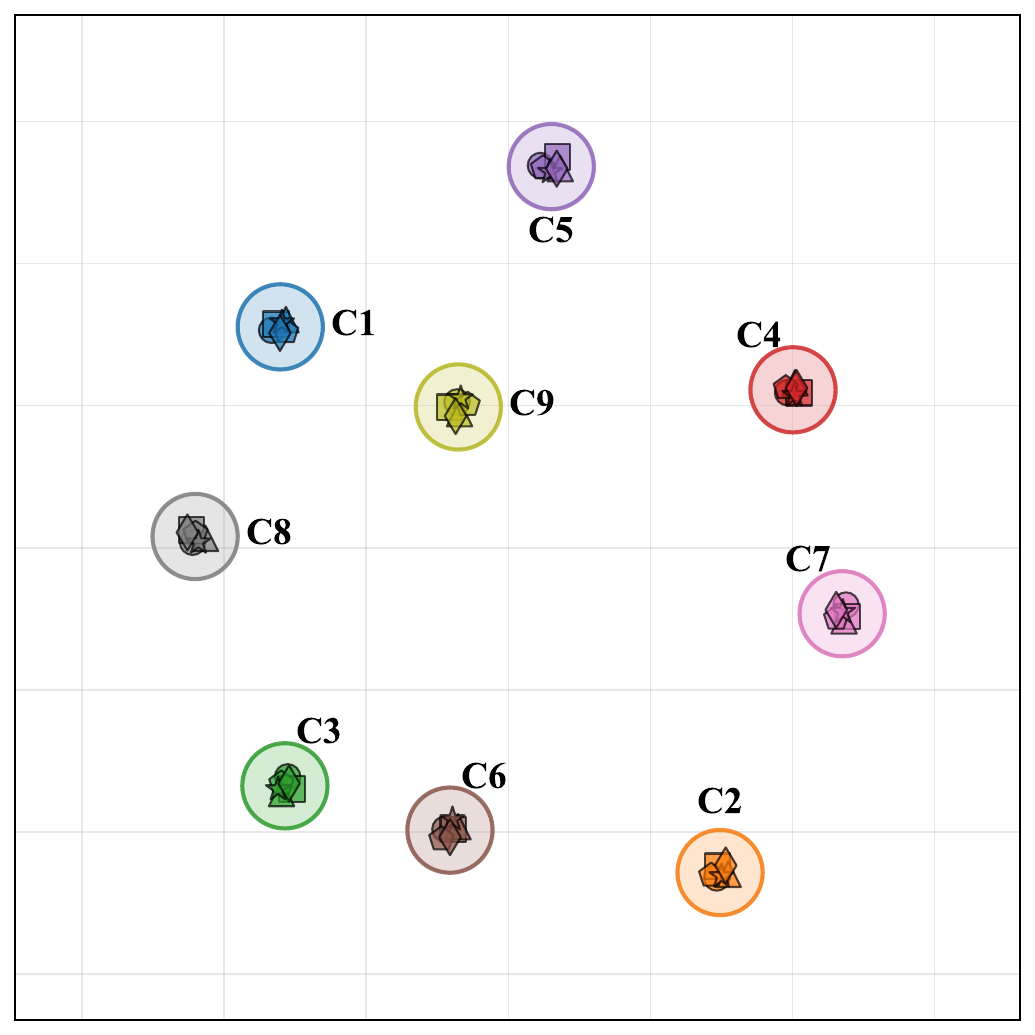}
        2-Wass. ($k=9$)
    \end{minipage}
    \begin{minipage}{0.19\linewidth}
        \centering
        \includegraphics[width=\linewidth]{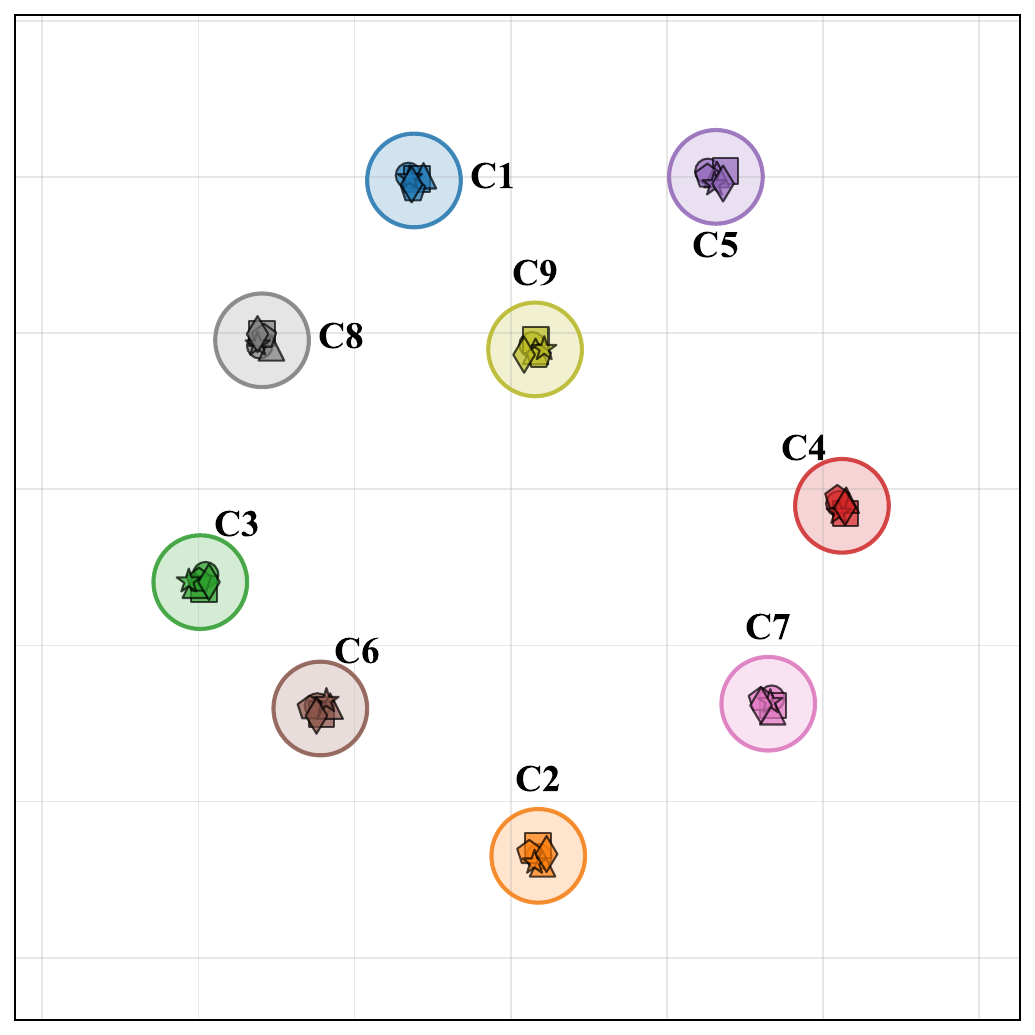}
        EMD ($k=9$)
    \end{minipage}
    \begin{minipage}{0.19\linewidth}
        \centering
        \includegraphics[width=\linewidth]{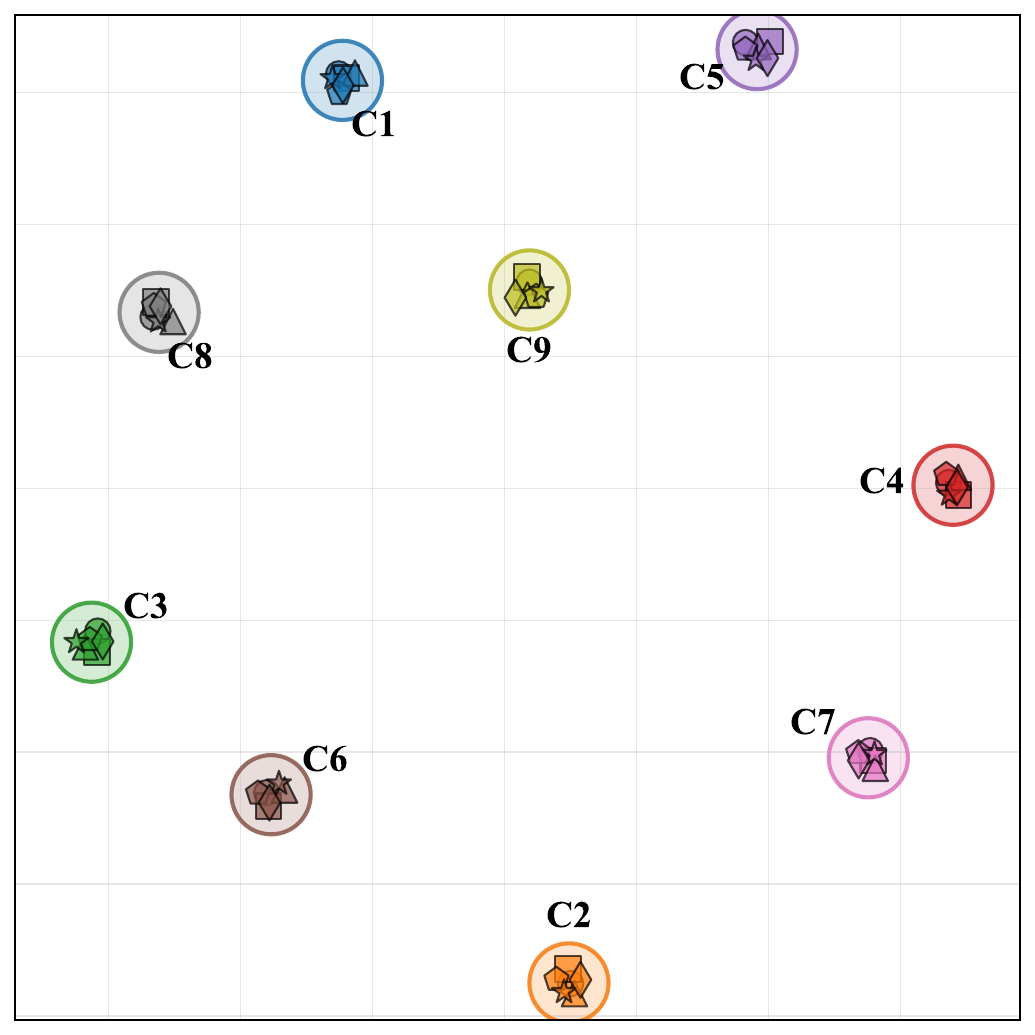}
        Adp. EMD ($k=9$)
    \end{minipage}
    \\
    \begin{minipage}{0.19\linewidth}
        \centering
        \includegraphics[width=\linewidth]{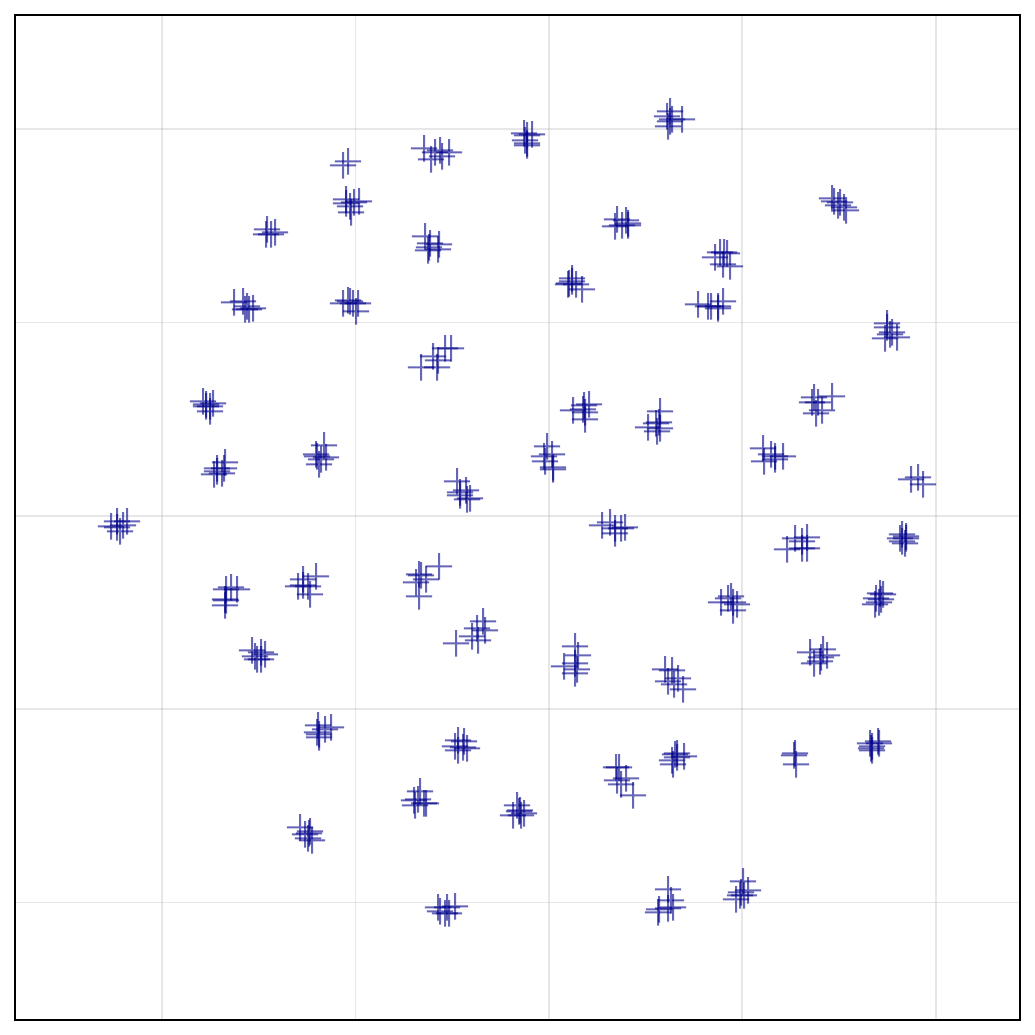}
        Chamfer ($k=50$)
    \end{minipage}
    \begin{minipage}{0.19\linewidth}
        \centering
        \includegraphics[width=\linewidth]{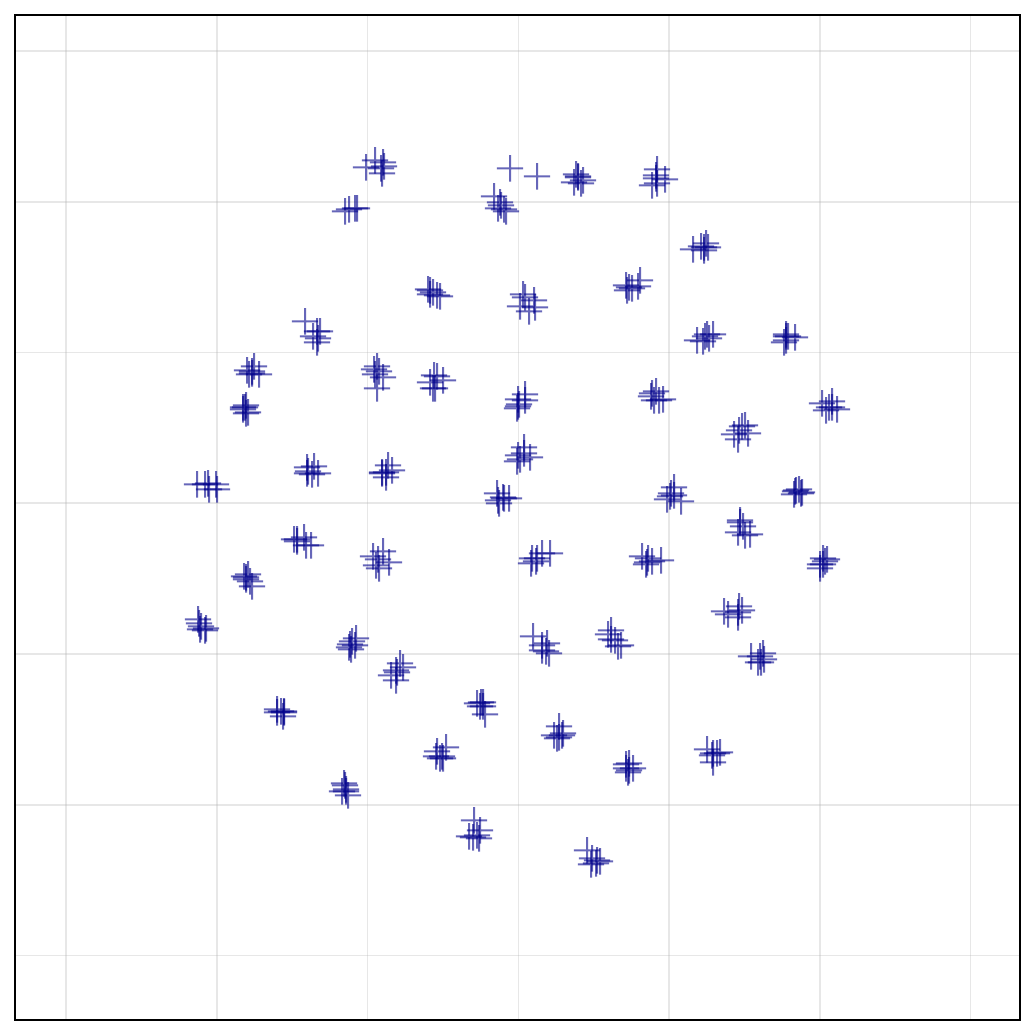}
        Hausdorff ($k=50$)
    \end{minipage}
    \begin{minipage}{0.19\linewidth}
        \centering
        \includegraphics[width=\linewidth]{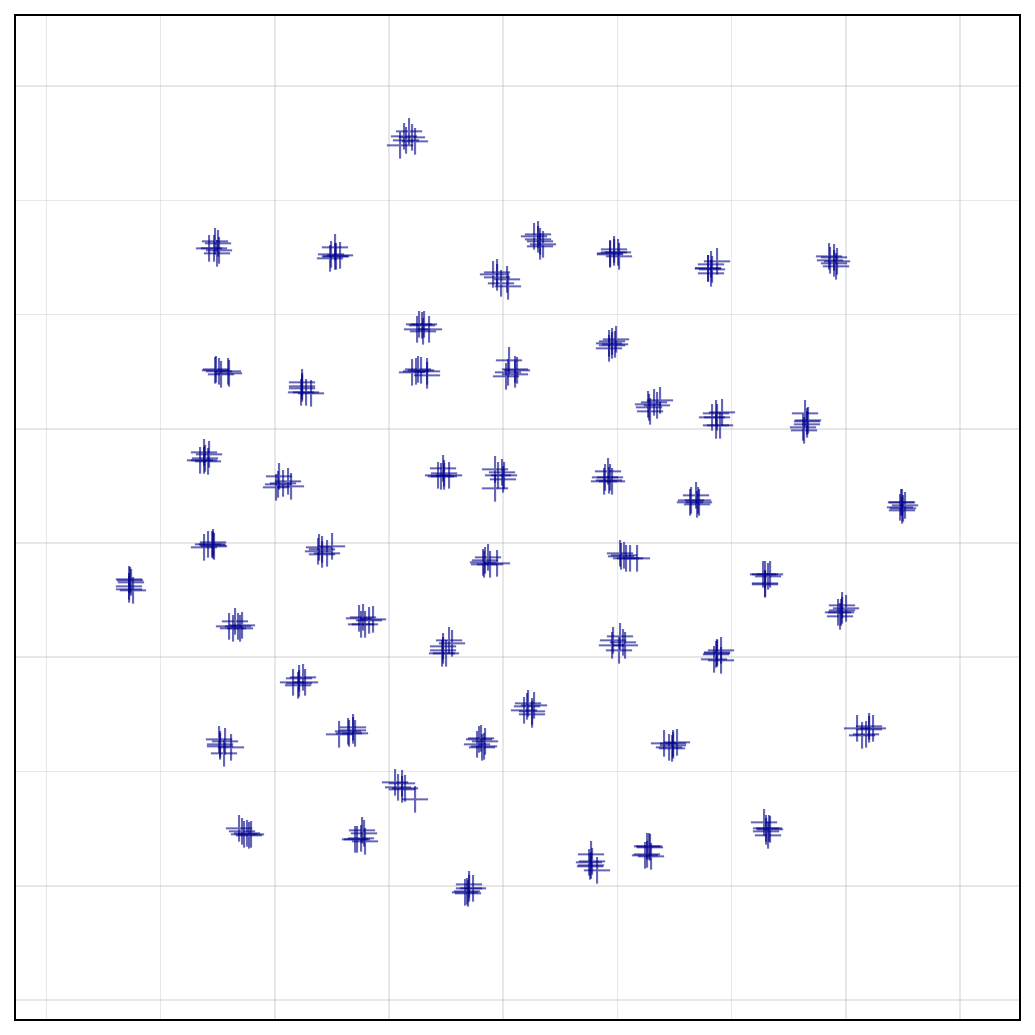}
        2-Wass. ($k=50$)
    \end{minipage}
    \begin{minipage}{0.19\linewidth}
        \centering
        \includegraphics[width=\linewidth]{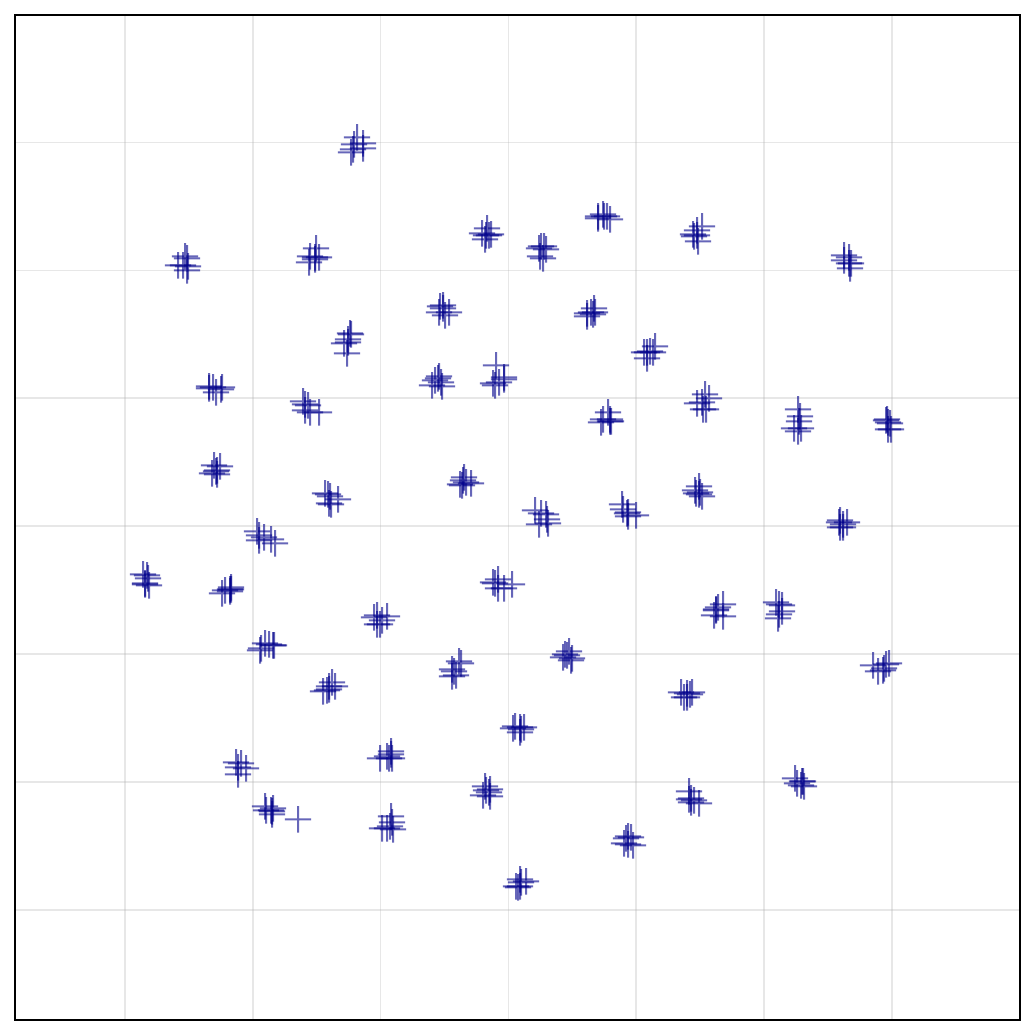}
        EMD ($k=50$)
    \end{minipage}
    \begin{minipage}{0.19\linewidth}
        \centering
        \includegraphics[width=\linewidth]{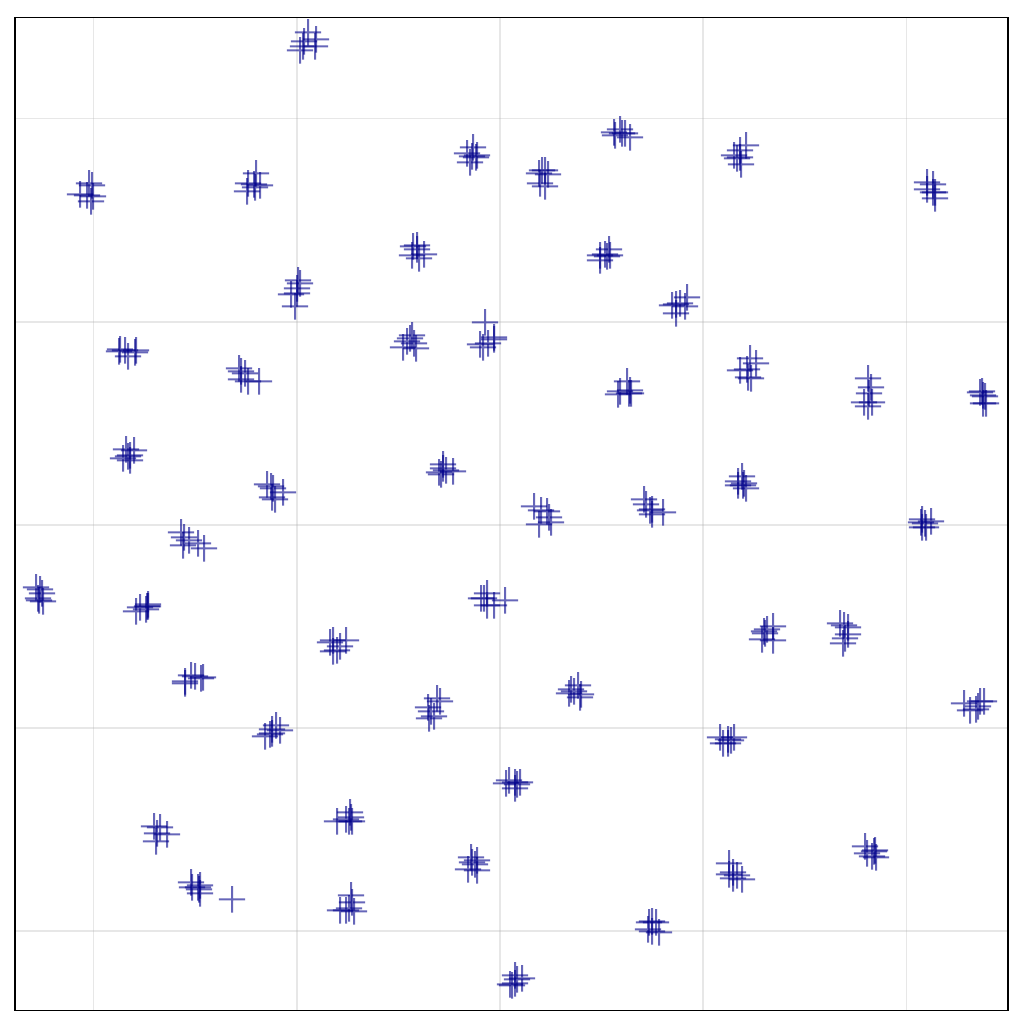}
        Adp. EMD ($k=50$)
    \end{minipage}
    \\
    \begin{minipage}{0.19\linewidth}
        \centering
        \includegraphics[width=\linewidth]{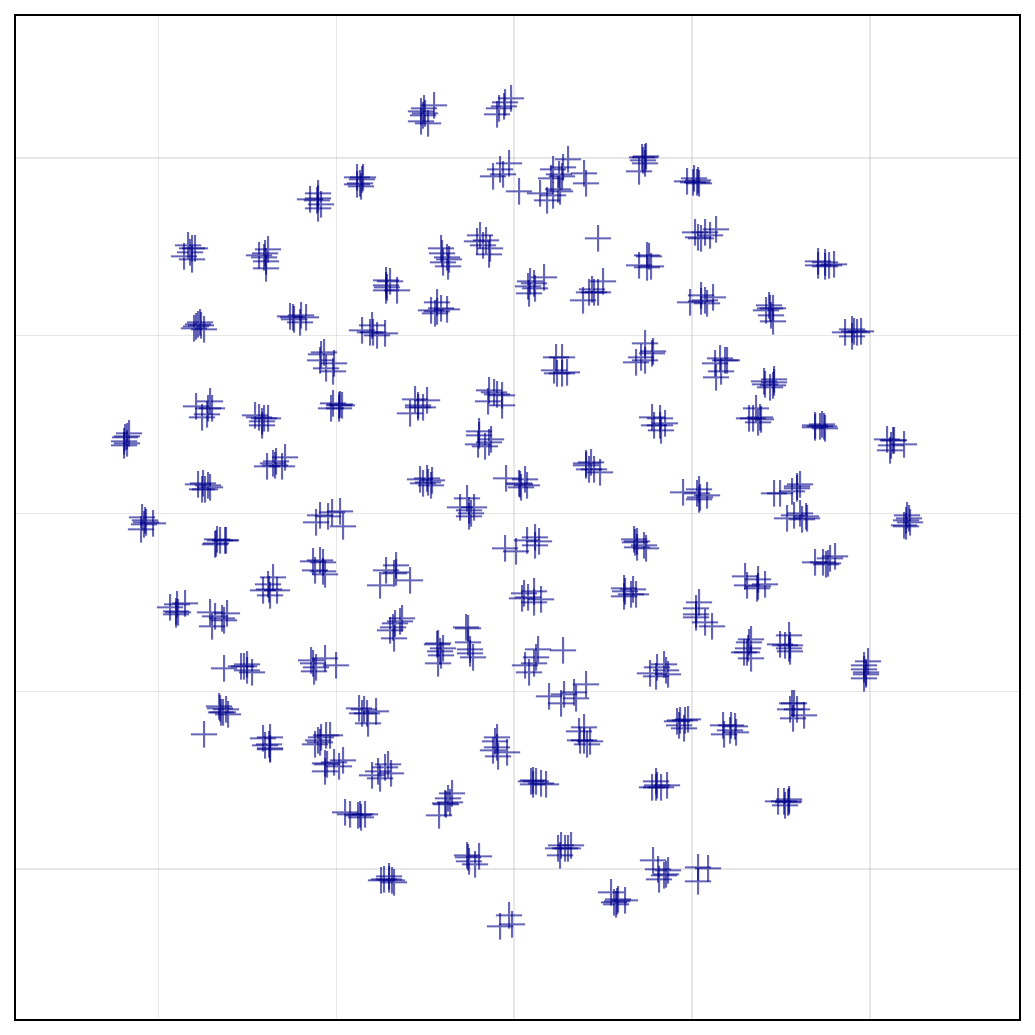}
        Chamfer ($k=100$)
    \end{minipage}
    \begin{minipage}{0.19\linewidth}
        \centering
        \includegraphics[width=\linewidth]{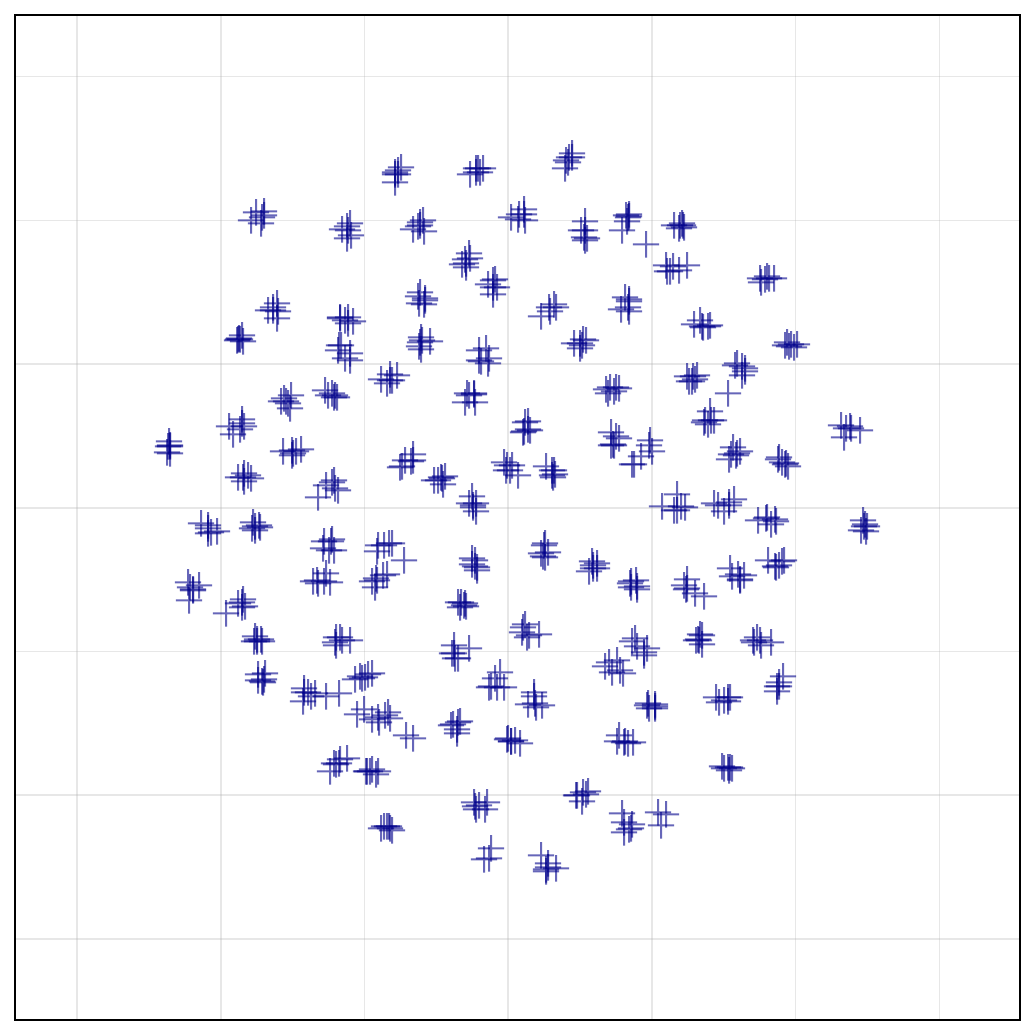}
        Hausdorff ($k=100$)
    \end{minipage}
    \begin{minipage}{0.19\linewidth}
        \centering
        \includegraphics[width=\linewidth]{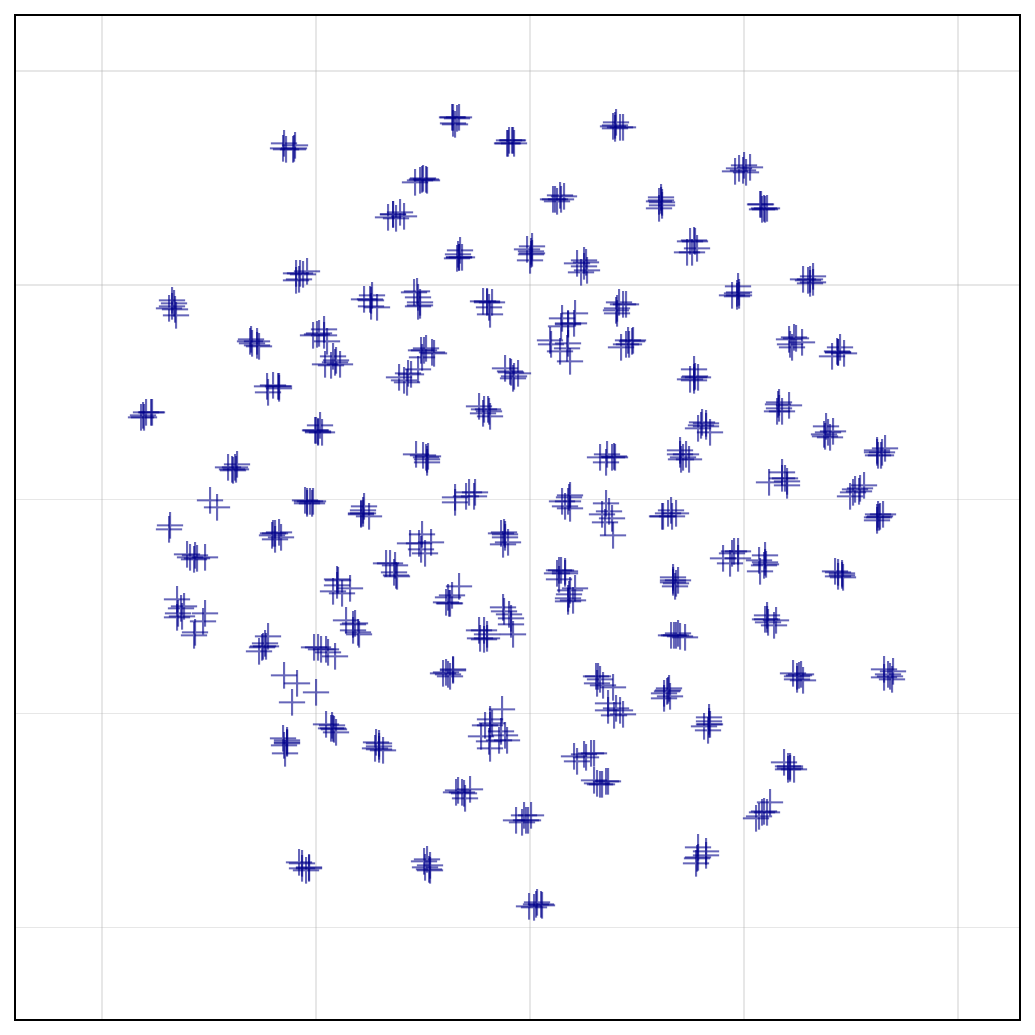}
        2-Wass. ($k=100$)
    \end{minipage}
    \begin{minipage}{0.19\linewidth}
        \centering
        \includegraphics[width=\linewidth]{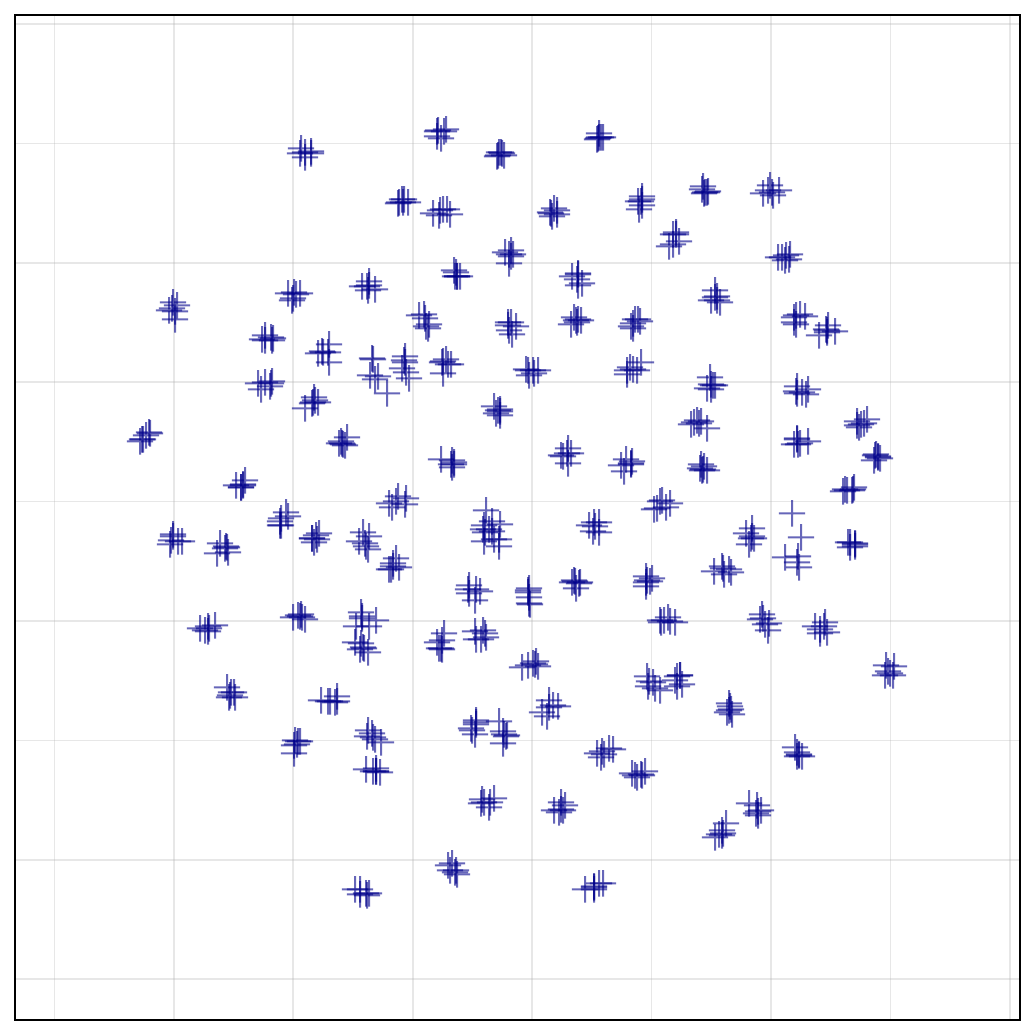}
        EMD ($k=100$)
    \end{minipage}
    \begin{minipage}{0.19\linewidth}
        \centering
        \includegraphics[width=\linewidth]{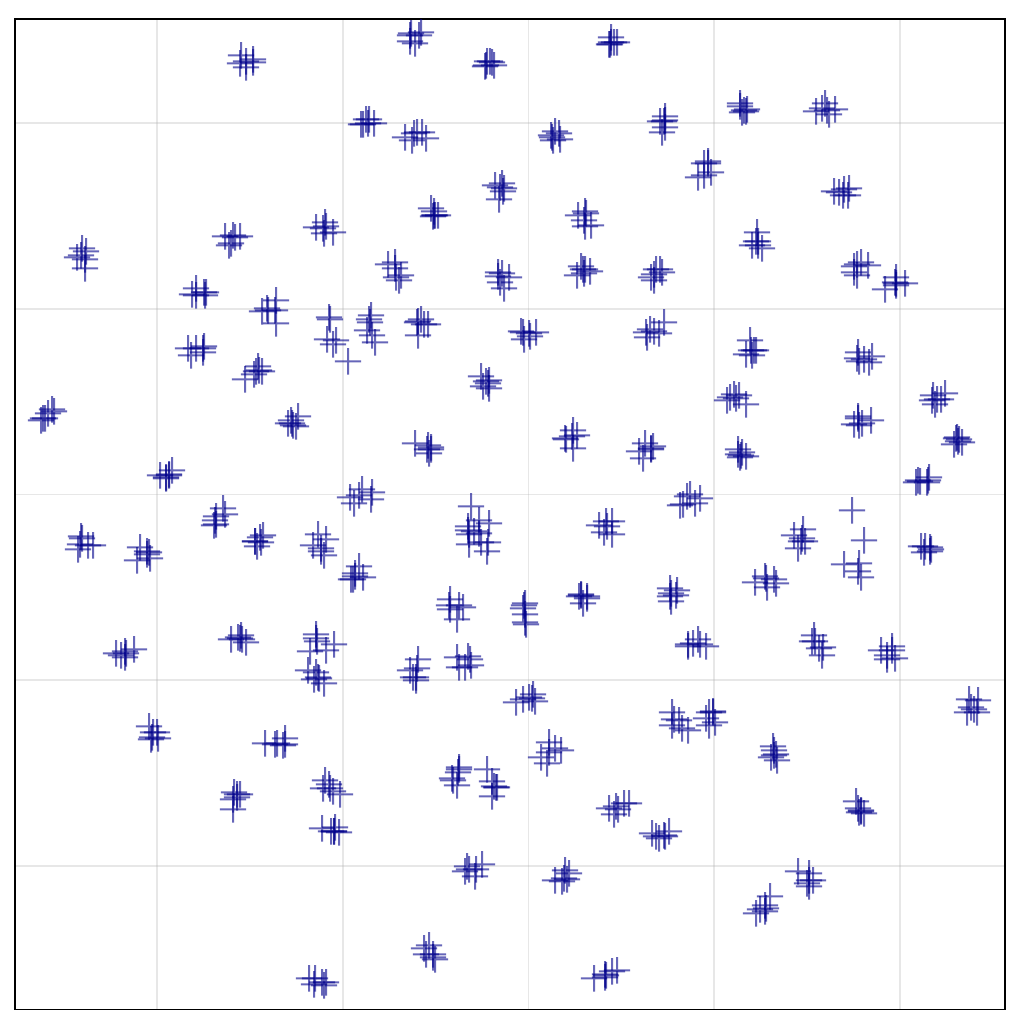}
        Adp. EMD ($k=100$)
    \end{minipage}
    \\
    \caption{Comparison of t-SNE visualization of clustering results with cluster-centric BK-tree under different distance metrics and true numbers of clusters}
    \label{fig:classification_results}
\end{figure*}

Figure~\ref{fig:classification_results} shows the comparison of t-SNE visualizations of clustering results with cluster-centric BK-tree under different distance metrics and true numbers of clusters for scenarios with consistent combat unit scale. Clustering results under adapted EMD and Wasserstein family distance metrics are highly consistent, while Chamfer and Hausdorff show significant differences due to their neglect of unit distribution information. 

\begin{figure*}[ht]
    \centering
    \begin{minipage}{0.19\linewidth}
        \centering
        \includegraphics[width=\linewidth]{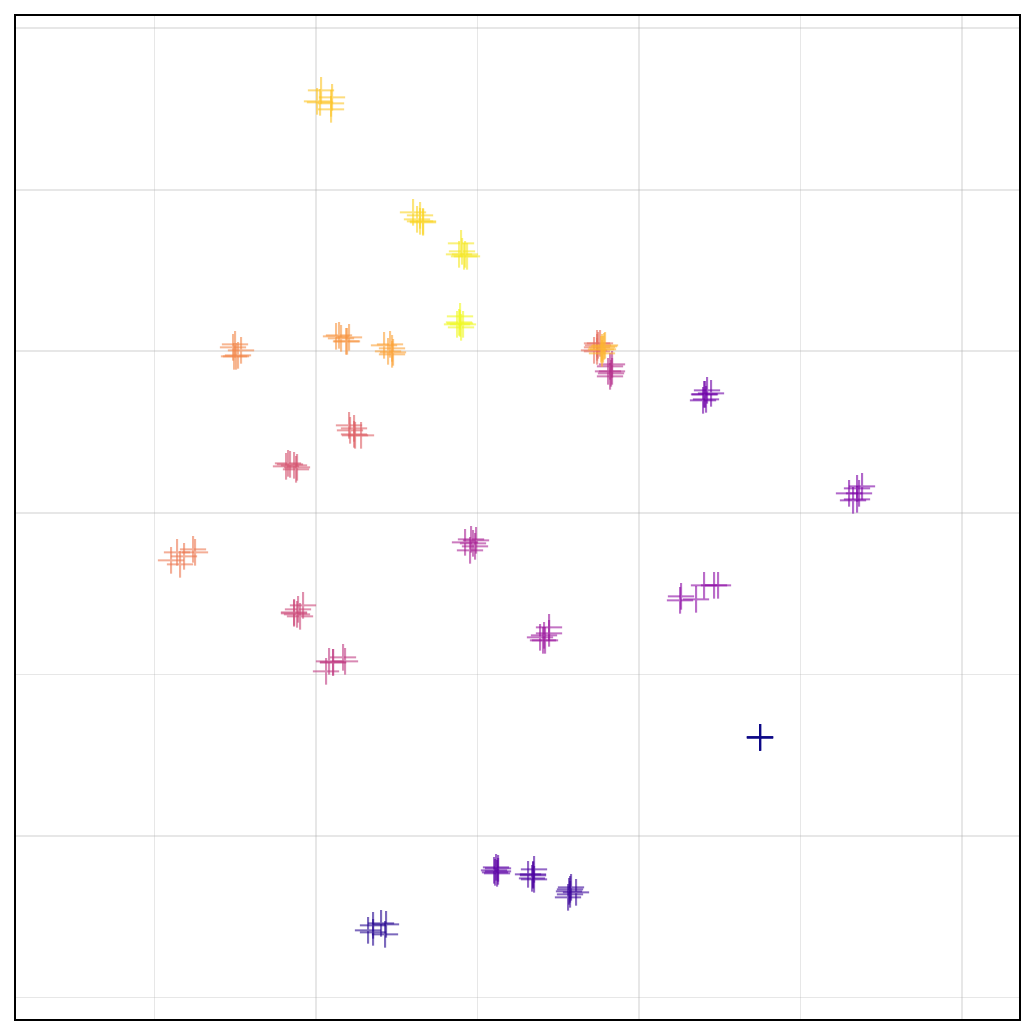}
        Chamfer ($k=25$)
    \end{minipage}
    \begin{minipage}{0.19\linewidth}
        \centering
        \includegraphics[width=\linewidth]{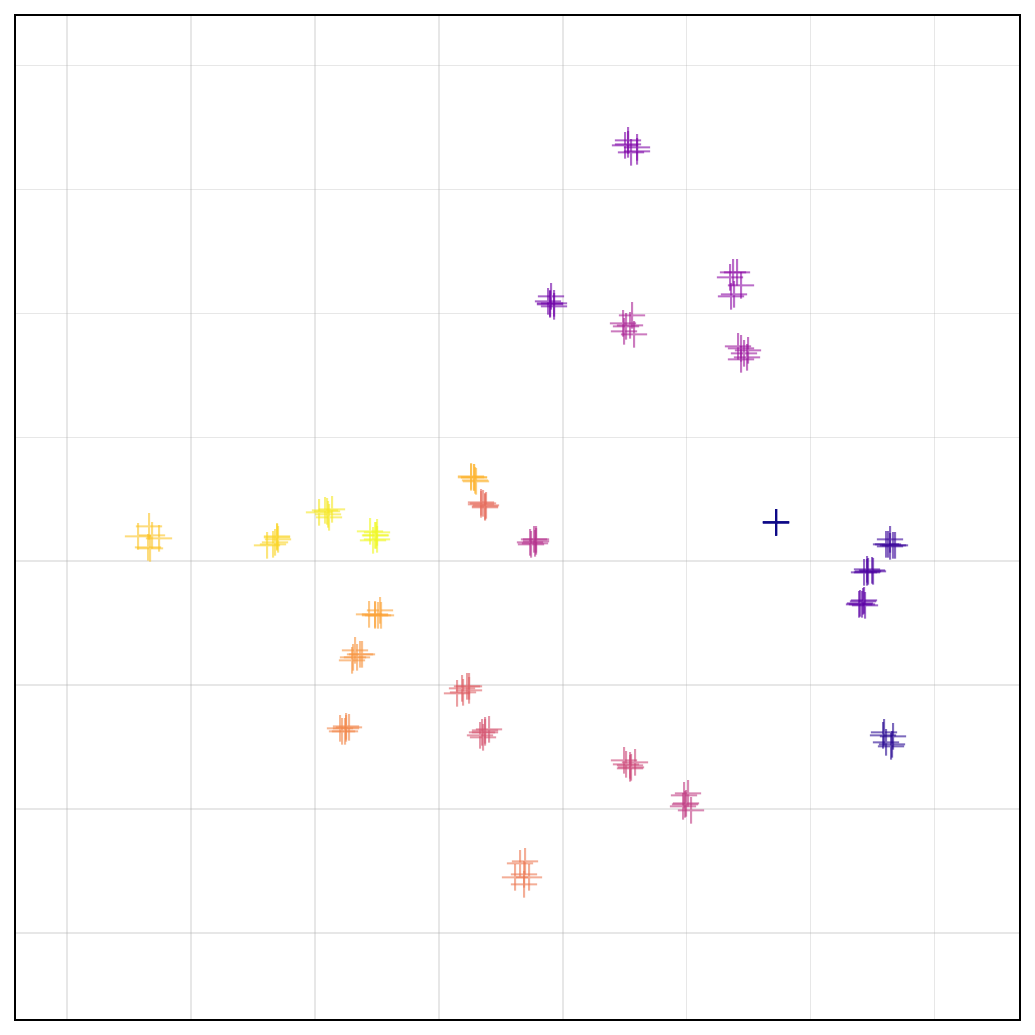}
        Hausdorff ($k=25$)
    \end{minipage}
    \begin{minipage}{0.19\linewidth}
        \centering
        \includegraphics[width=\linewidth]{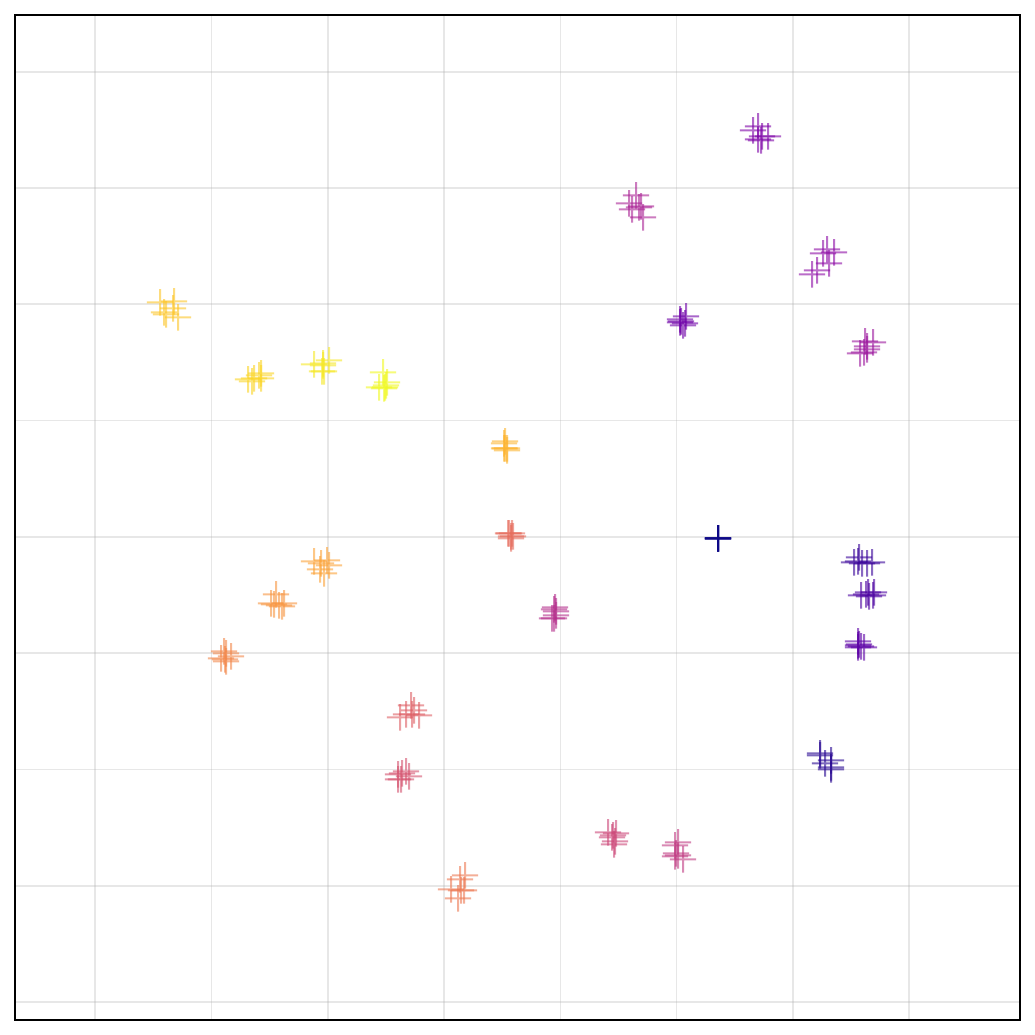}
        2-Wass. ($k=25$)
    \end{minipage}
    \begin{minipage}{0.19\linewidth}
        \centering
        \includegraphics[width=\linewidth]{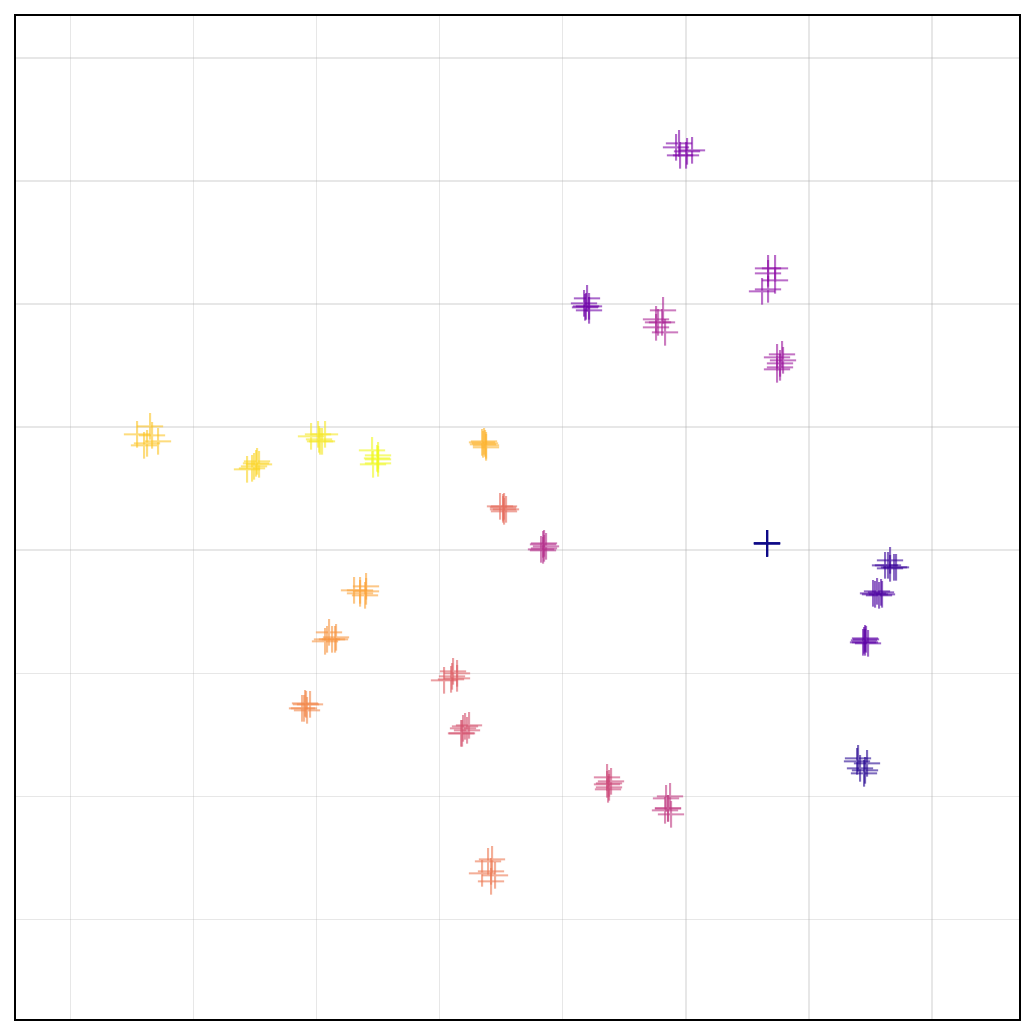}
        EMD ($k=25$)
    \end{minipage}
    \begin{minipage}{0.19\linewidth}
        \centering
        \includegraphics[width=\linewidth]{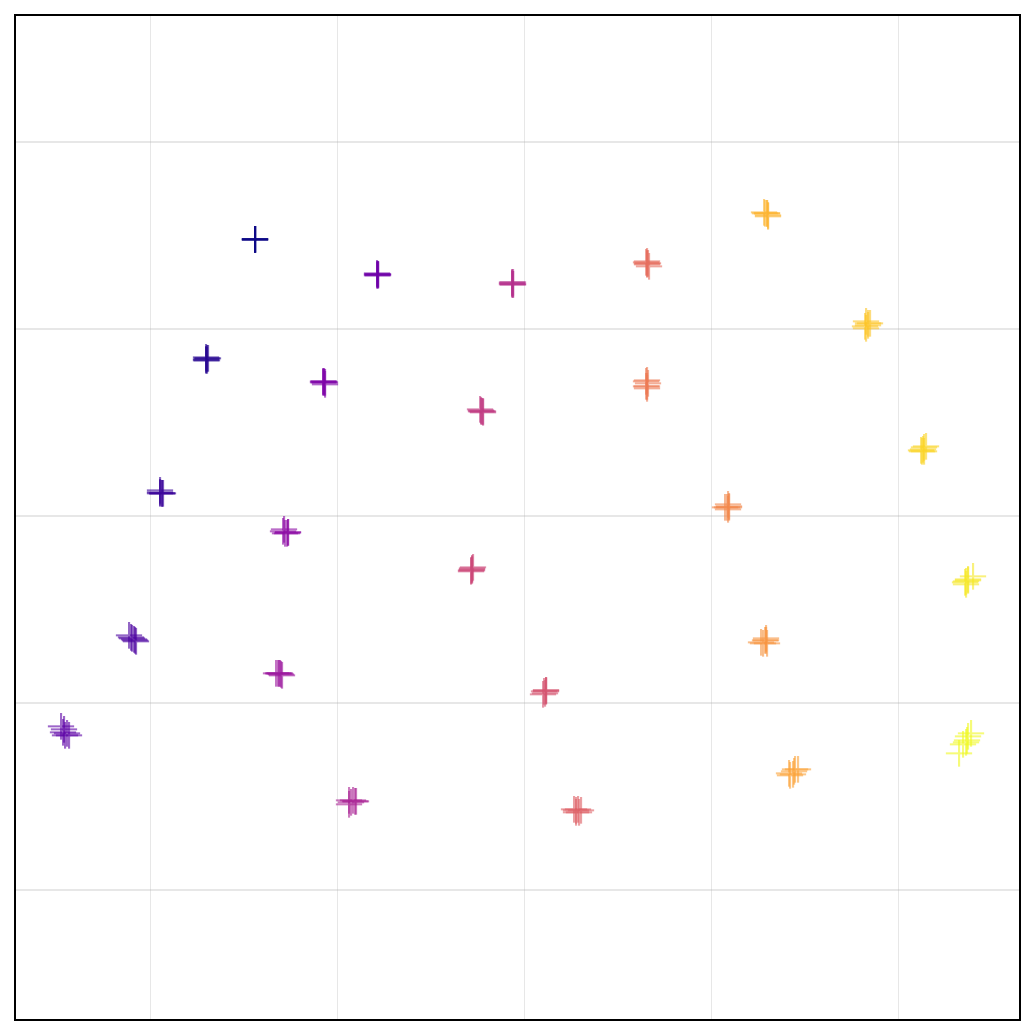}
        Adp. EMD ($k=25$)
    \end{minipage}
    \\
    \begin{minipage}{0.19\linewidth}
        \centering
        \includegraphics[width=\linewidth]{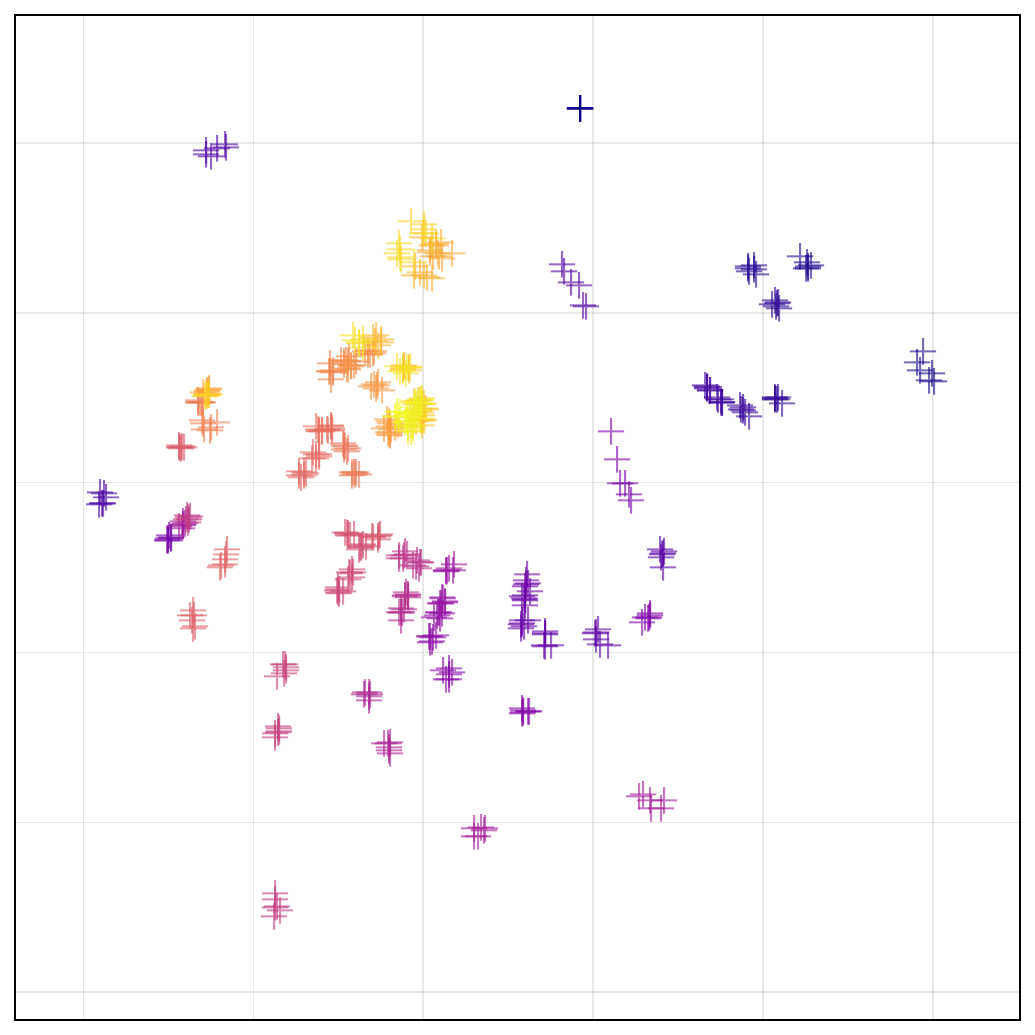}
        Chamfer ($k=81$)
    \end{minipage}
    \begin{minipage}{0.19\linewidth}
        \centering
        \includegraphics[width=\linewidth]{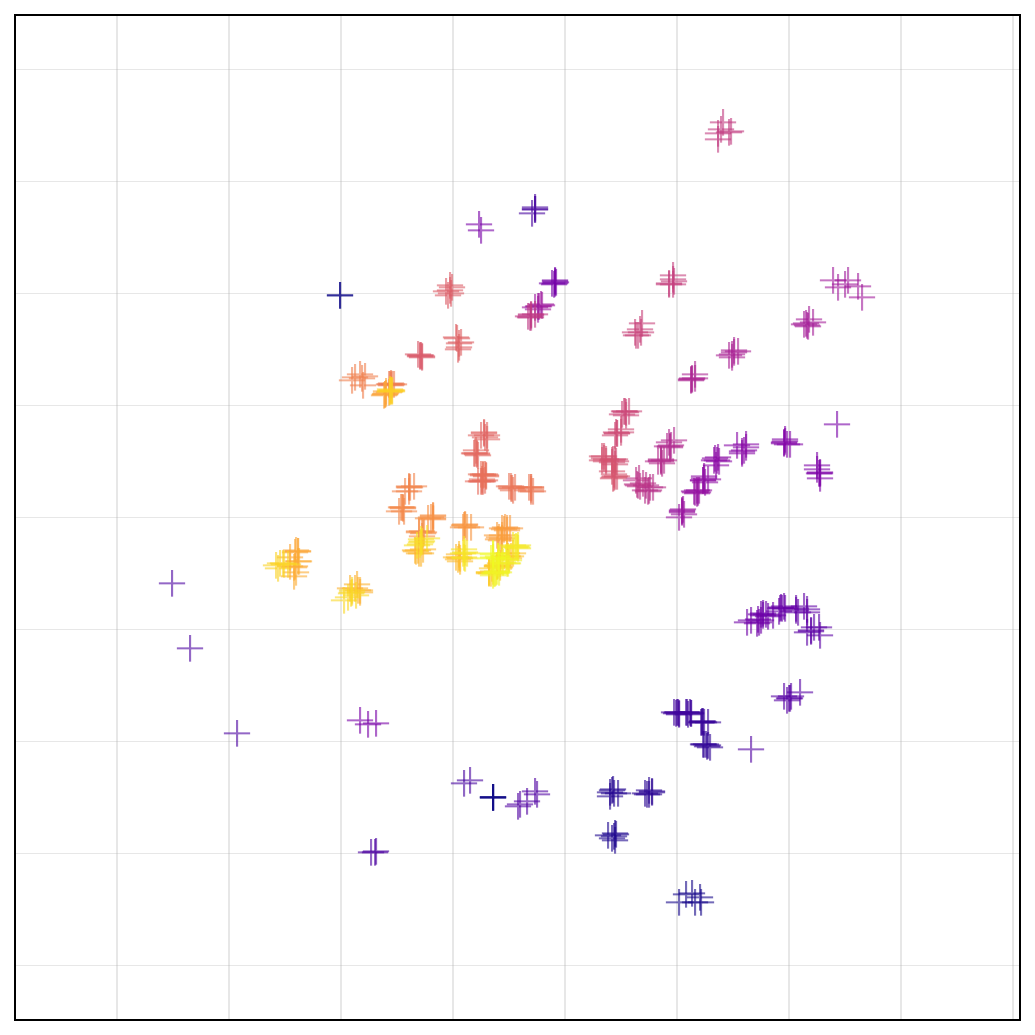}
        Hausdorff ($k=81$)
    \end{minipage}
    \begin{minipage}{0.19\linewidth}
        \centering
        \includegraphics[width=\linewidth]{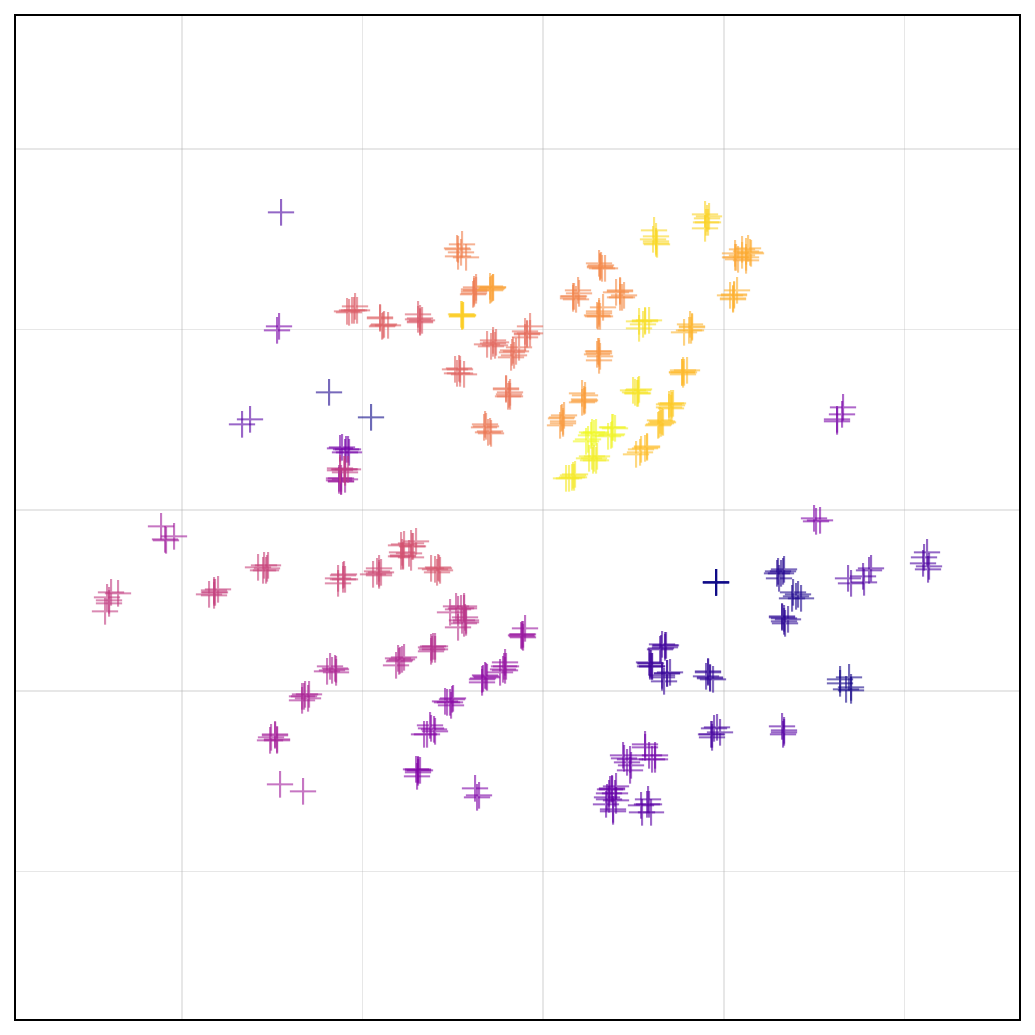}
        2-Wass. ($k=81$)
    \end{minipage}
    \begin{minipage}{0.19\linewidth}
        \centering
        \includegraphics[width=\linewidth]{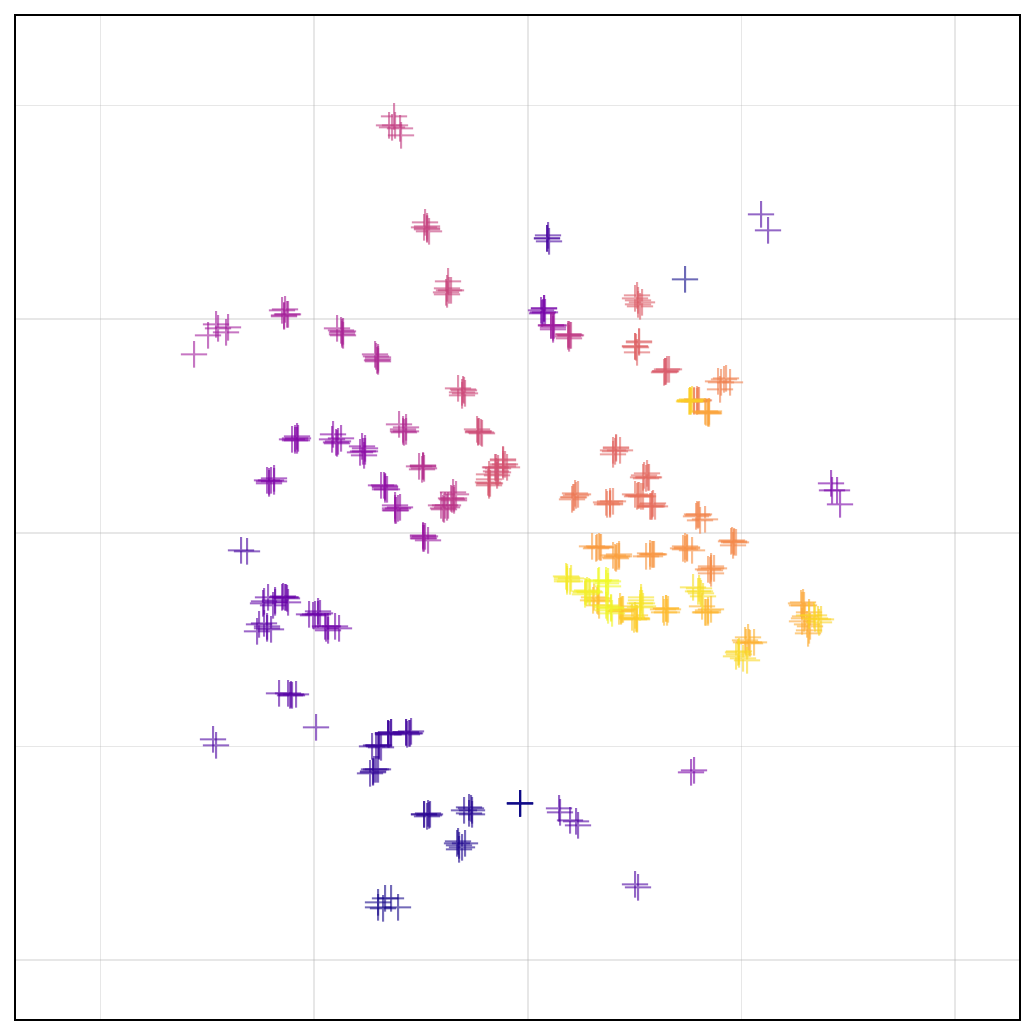}
        EMD ($k=81$)
    \end{minipage}
    \begin{minipage}{0.19\linewidth}
        \centering
        \includegraphics[width=\linewidth]{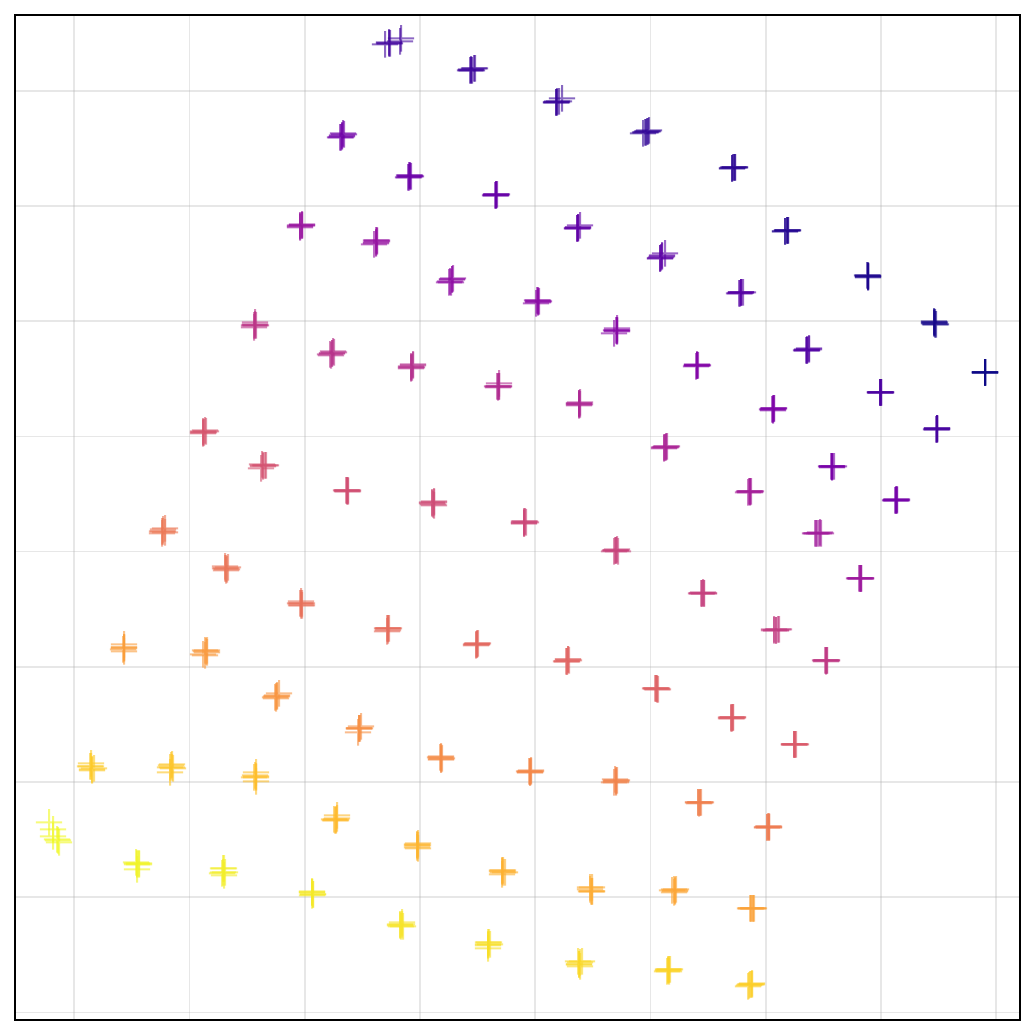}
        Adp. EMD ($k=81$)
    \end{minipage}
    \\
    \caption{Comparison of t-SNE visualization of clustering results with cluster-centric BK-tree under different distance metrics and true numbers of clusters, with inconsistent combat unit scales across clusters}
    \label{fig:classification_results_inconsistent}
\end{figure*}

Figure~\ref{fig:classification_results_inconsistent} shows the comparison for scenarios with inconsistent combat unit scale. The distant virtual points introduced by adapted EMD effectively maintain state separation performance under inconsistent unit scales, thereby avoiding incorrect clustering.

Note that the t-SNE in Figure~\ref{fig:classification_results} and Figure~\ref{fig:classification_results_inconsistent} is used solely for visualization, not for dimensionality reduction during clustering. 

\newpage
\subsection{Parameter Sensitivity Analysis of DenStream}

\begin{figure}[ht]
	\centering
		\centering
		\includegraphics[width=\linewidth]{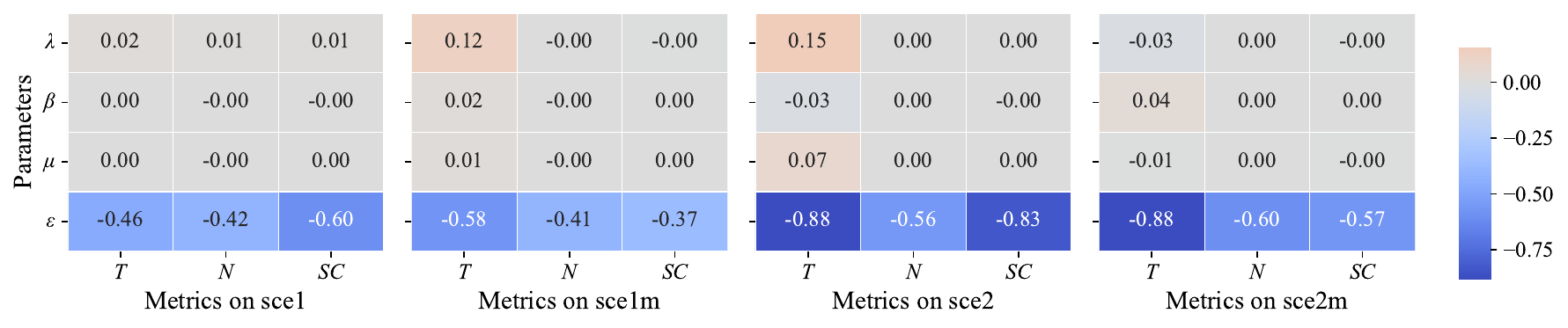}
    \caption{Correlation heatmap between parameters for DenStream and clustering performance metrics in different scenarios}
    \label{fig:denstream_correlation_heatmap}
\end{figure}

The correlation analysis in Figure~\ref{fig:denstream_correlation_heatmap} elucidates the distinct roles of DenStream parameters. While $\lambda$ controls temporal decay and $\beta, \mu$ govern micro-cluster dynamics, their minimal correlation with performance metrics suggests a relatively stable distribution in the StarCraft II data stream, where temporal evolution exerts negligible impact on the cluster architecture. In contrast, the neighborhood radius $\epsilon$ emerges as the dominant factor, underscoring that inter-point spatial distance is the primary determinant of clustering efficacy in this context. Consequently, $\epsilon$ is identified as the critical parameter for optimization. For the performance comparison in \text{Figure~7} in the main text, the secondary parameters were fixed at $\lambda=0.5$, $\beta=0.1$, and $\mu=1.0$ to ensure a focused evaluation of the $\epsilon$ impact.

\newpage
\subsection{Detailed Performance Data of State Stream Clustering}

Tables~\ref{tab:Comparison of clustering performance metrics of cluster-centric BK-tree} and \ref{tab:Comparison of clustering performance metrics of DenStream} present the detailed experimental results for the cluster-centric BK-tree and the DenStream algorithm, respectively. These tables record the execution time ($T$), the number of clusters ($N$), and the clustering quality ($Q$) across various scenarios under different clustering thresholds ($\epsilon$).

\begin{table*}[ht]
\centering
\renewcommand{\arraystretch}{0.9}
\caption{Comparison of clustering performance metrics of the cluster-centric BK-tree with various $\epsilon$ across different scenarios}
\label{tab:Comparison of clustering performance metrics of cluster-centric BK-tree}
\begin{tabular}{@{}ccccccccccccc@{}}
\toprule
Param. & \multicolumn{3}{c}{sce1} & \multicolumn{3}{c}{sce1m} & \multicolumn{3}{c}{sce2} & \multicolumn{3}{c}{sce2m} \\
\cmidrule(lr){1-1}\cmidrule(lr){2-4}\cmidrule(lr){5-7}\cmidrule(lr){8-10}\cmidrule(lr){11-13}
 $\epsilon$ & \multicolumn{1}{c}{$T$} & $N$ & $Q$ & \multicolumn{1}{c}{$T$} & $N$ & $Q$ & \multicolumn{1}{c}{$T$} & $N$ & $Q$ & \multicolumn{1}{c}{$T$} & $N$ & $Q$ \\
\midrule
 0.10 & 15.448 & 286 & 0.597 & 24.238 & 690 & 0.576 & - & - & - & - & - & -  \\
 0.15 & 10.350 & 106 & 0.644 & 15.725 & 236 & 0.613 & - & - & - & - & - & -  \\
 0.20 & 10.444 & 60 & 0.676 & 12.471 & 120 & 0.627 & - & - & - & - & - & -  \\
 0.25 & 10.307 & 40 & 0.749 & 10.851 & 74 & 0.650 & - & - & - & - & - & -  \\
 0.50 & 10.199 & 23 & 0.903 & 10.581 & 29 & 0.868 & 21.758 & 415 & 0.516 & 20.940 & 344 & 0.512  \\
 0.60 & 10.146 & 21 & 0.923 & 10.145 & 25 & 0.909 & 20.414 & 256 & 0.539 & 17.447 & 218 & 0.555  \\
 0.70 & 10.097 & 20 & 0.939 & 10.374 & 24 & 0.926 & 19.761 & 172 & 0.604 & 16.184 & 158 & 0.600  \\
 0.75 & 10.231 & 19 & 0.938 & 10.265 & 23 & 0.924 & 19.569 & 146 & 0.633 & 16.485 & 134 & 0.636  \\
 0.80 & 10.396 & 19 & 0.938 & 10.423 & 23 & 0.925 & 20.051 & 127 & 0.664 & 16.505 & 117 & 0.673 \\
 0.85 & 10.684 & 19 & 0.938 & 10.618 & 23 & 0.925 & 20.390 & 105 & 0.682 & 17.132 & 105 & 0.689  \\
 0.90 & 10.146 & 19 & 0.938 & 10.662 & 23 & 0.925 & 21.656 & 96 & 0.687 & 17.751 & 95 & 0.703  \\
 1.00 & 13.176 & 19 & 0.938 & 12.890 & 23 & 0.924 & 26.801 & 85 & 0.729 & 20.140 & 86 & 0.740  \\
 1.10 & 11.174 & 16 & 0.487 & 11.237 & 22 & 0.617 & 28.500 & 78 & 0.694 & 22.566 & 76 & 0.638  \\
 1.20 & 9.254 & 14 & 0.598 & 9.017 & 19 & 0.651 & 24.429 & 73 & 0.592 & 19.821 & 72 & 0.587  \\
 1.25 & 9.137 & 12 & 0.615 & 8.599 & 17 & 0.795 & 22.414 & 72 & 0.599 & 18.641 & 70 & 0.594  \\
 1.30 & 9.139 & 11 & 0.663 & 8.509 & 15 & 0.822 & 20.336 & 70 & 0.573 & 16.500 & 68 & 0.566  \\
 1.40 & 9.057 & 10 & 0.649 & 8.645 & 13 & 0.830 & 18.429 & 62 & 0.500 & 15.003 & 60 & 0.572  \\
 1.50 & 8.769 & 10 & 0.663 & 8.570 & 11 & 0.824 & 16.252 & 59 & 0.501 & 13.569 & 57 & 0.603  \\
 2.50 & 7.151& 4 & 0.349 & 7.073 & 5 & 0.754 & 13.342 & 19 & 0.572 & 11.292 & 20 & 0.595  \\
\bottomrule
\end{tabular}
\end{table*}

\begin{table*}[ht]
\vspace{-15pt}
\centering
\renewcommand{\arraystretch}{0.9}
\caption{Comparison of clustering performance metrics of DenStream with different $\epsilon$ across various scenarios}
\label{tab:Comparison of clustering performance metrics of DenStream}
\begin{tabular}{@{}ccccccccccccc@{}}
\toprule
Param. & \multicolumn{3}{c}{sce1} & \multicolumn{3}{c}{sce1m} & \multicolumn{3}{c}{sce2} & \multicolumn{3}{c}{sce2m} \\
\cmidrule(lr){1-1}\cmidrule(lr){2-4}\cmidrule(lr){5-7}\cmidrule(lr){8-10}\cmidrule(lr){11-13}
 $\epsilon$ & \multicolumn{1}{c}{$T$} & $N$ & $Q$ & \multicolumn{1}{c}{$T$} & $N$ & $Q$ & \multicolumn{1}{c}{$T$} & $N$ & $Q$ & \multicolumn{1}{c}{$T$} & $N$ & $Q$ \\
\midrule
 0.10 & 75.182 & 225 & 0.938 & 158.459 & 537 & 0.925 & - & - & - & - & - & - \\
 0.50 & 44.463 & 23 & 0.937 & 46.983 & 28 & 0.925 & 276.884 & 329 & 0.738 & 232.813 & 266 & 0.728 \\
 1.00 & 43.932 & 19 & 0.939 & 49.694 & 23 & 0.928 & 207.578 & 82 & 0.726 & 169.649 & 78 & 0.720 \\
 1.10 & 40.225 & 16 & 0.899 & 42.289 & 19 & 0.866 & 212.884 & 77 & 0.748 & 162.528 & 75 & 0.742 \\
 1.20 & 35.044 & 13 & 0.755 & 37.738 & 18 & 0.811 & 201.567 & 72 & 0.705 & 156.619 & 71 & 0.716 \\
 1.25 & 34.294 & 12 & 0.651 & 39.024 & 16 & 0.557 & 205.272 & 71 & 0.703 & 150.622 & 67 & 0.715 \\
 1.30 & 33.384 & 11 & 0.446 & 39.146 & 14 & 0.499 & 205.159 & 70 & 0.708 & 153.797 & 66 & 0.720 \\
 1.40 & 32.959 & 10 & 0.381 & 35.721 & 13 & 0.374 & 196.817 & 62 & 0.650 & 157.778 & 58 & 0.611 \\
 1.50 & 31.280 & 9 & 0.202 & 35.370 & 11 & 0.137 & 234.638 & 58 & 0.616 & 154.382 & 52 & 0.552 \\
 2.00 & 30.532 & 9 & 0.158 & 33.610 & 11 & 0.241 & 179.407 & 32 & 0.296 & 151.479 & 31 & 0.277 \\
 2.50 & 18.589 & 4 & 0.329 & 19.742 & 5 & 0.391 & 121.905 & 19 & 0.437 & 102.243 & 19 & 0.482 \\
\bottomrule
\end{tabular}
\end{table*}

Taking the results of the sce1 scenario in Table~\ref{tab:Comparison of clustering performance metrics of cluster-centric BK-tree} as a specific example, the internal efficiency logic of the cluster-centric BK-tree can be elucidated through the non-monotonic trend of $T$. When the clustering threshold is at its maximum ($\epsilon = 2.50$), the minimum value of $T$ occurs because the large threshold allows a large amount of data to be greedily assigned to the same cluster, which also accounts for the minimum value of $Q$ at this point. As $\epsilon$ decreases to $1.00$, $T$ reaches its first peak. 

The comparative data in Table~\ref{tab:Comparison of clustering performance metrics of DenStream} further underscores these findings. In contrast to the stable time consumption of the cluster-centric BK-tree, the DenStream algorithm exhibits a substantial increase in $T$ within complex scenarios (\textit{sce2} and \textit{sce2m}), proving that the proposed combination of the state distance metric and the cluster-centric BK-tree structure is uniquely suited for maintaining real-time performance in massive, high-dimensional RTS data streams.

\newpage
\subsection{Tactic Label Codebook}

The tactic label codebook is presented in Table~\ref{tab:tactic_code}, providing a structured taxonomy for the qualitative analysis of micromanagement behaviors. This codebook categorizes tactic labels into four hierarchical dimensions:

\begin{itemize}
    \item Basic Actions (IDs 0--6): Represent explicit primitive actions in unit engagements, including greedy or threat-focused targeting, exfiltration, and deceptive actions such as feints.
    \item Collaboration Granularity (IDs 7--9): Characterize the degree of local collaboration among units, ranging from independent operations to centralized collaboration.
    \item Collaboration Persistence (IDs 10--11): Describe the temporal stability of the collaboration mode, identifying whether units maintain a consistent mode or undergo tactical switching.
    \item Tactical Characteristics (IDs 12--17): Represent high-level tactical profiles, such as sustained offensive/defensive stances and hybrid actions.
\end{itemize}

\begin{table}[ht]
\centering
\renewcommand{\arraystretch}{1.2}
\caption{Tactic Label Codebook}
\label{tab:tactic_code}
\begin{tabular}{clp{11cm}}
\toprule
\textbf{ID} & \textbf{Tactic Label} & \textbf{Description of the Criteria} \\
\midrule
0 & a-greedy & Always greedily attack the nearest or weakest targets first. \\
1 & a-threat-focused & Always strike the currently most threatening targets first. \\
2 & a-exfiltration & Planned withdrawal from enemy contact. \\
3 & a-massing & Gather units to concentrate forces at a single point to achieve local superiority. \\
4 & a-feint & Deceptive attack to draw enemy reserves away from the real main effort. \\
5 & a-sacrificial feint & A feint with the sacrifice of individual units. \\
6 & a-inefficiency & Inefficiency actions generated by distractors. \\\midrule
7 & k-independent & Units operate without local collaboration. \\
8 & k-distributed & Units conduct local collaboration based on the distribution. \\
9 & k-centralized & Units perform in a centralized manner. \\\midrule
10 & c-sustained collaboration & Maintain the same collaborative model (IDs 7-9) continuously. \\
11 & c-switched collaboration & Switch collaboration modes (IDs 7-9). \\\midrule
12 & t-sustained attack & Sustained attack (IDs 0-1). \\
13 & t-sustained defense & Sustained defense (ID 2). \\
14 & t-sustained hybrid & Sustained hybrid actions (IDs 3-5). \\
15 & t-switched tactic & Shift between distinct tactical modes (IDs 12-14). \\
16 & t-inefficient tactic & Inefficient tactic with conducting inefficiency actions (ID 6). \\
17 & t-unknown tactic & Behavior does not match any recognized pattern in the current codebook. \\
\bottomrule
\end{tabular}
\end{table}

\newpage
\subsection{Tactic Codebook}

Selected examples of tactical codes and their semantic interpretations are detailed in Table~\ref{tab:tactic example}. These codes represent the primary tactical profiles analyzed in \text{Sec.~4.6}, encompassing various combinations of basic actions, collaboration granularities, and collaboration transitions. Specifically, the table illustrates a range of tactics including fundamental greedy or threat-focused attacks and complex coordination modes such as distributed or centralized collaboration. Furthermore, it identifies whether these patterns remain sustained or undergo tactical switching.

\begin{table}[ht]
\centering
\renewcommand{\arraystretch}{1.2}
\caption{Some of the tactic codes and their descriptions}
\label{tab:tactic example}
\begin{tabularx}{\textwidth}{@{}>{\centering\arraybackslash}lcX@{}}
\hline
\textbf{ID} & \textbf{Tactic Code} & \textbf{Description} \\
\hline
0 & \makecell{0000000 000 00 000000} & No obvious tactic\\
1 & \makecell{1000000 010 10 100000} & Greedy attack, sustained collaboration based on the distribution\\
2 & \makecell{1000000 111 01 100000} & Greedy attack, switched collaboration modes among independent, distributed and centralized mode\\
3 & \makecell{1000000 011 01 100000} & Greedy attack, switched collaboration modes among distributed and centralized mode\\
4 & \makecell{1000100 111 01 000100} & Greedy attack and feint, switched collaboration modes among independent, distributed and centralized mode\\
5 & \makecell{1000010 011 01 000100} & Greedy attack and sacrificial feint, switched collaboration modes among distributed and centralized mode\\
6 & \makecell{1000000 110 01 100000} & Greedy attack, switched collaboration modes among independent and distributed mode\\
7 & \makecell{1000000 100 10 100000} & Greedy attack, without local collaboration\\
8 & \makecell{1000000 001 10 100000} & Greedy attack, sustained centralized collaboration mode\\
9 & \makecell{1001000 011 01 000100} & Greedy attack and massing, switched collaboration modes among distributed and centralized mode\\
10 & \makecell{1001000 001 10 000100} & Greedy attack and massing, sustained centralized collaboration mode\\
11 & \makecell{1100000 111 01 100000} & Greedy and threat-focused attack, switched collaboration modes among independent, distributed and centralized mode\\
12 & \makecell{1000100 011 01 000100} & Greedy attack and feint, switched collaboration modes among distributed and centralized mode\\
\hline
\end{tabularx}
\end{table}

\subsection{Action Sequence and Tactic Pattern Analysis Based on Sankey Diagrams}

\begin{figure*}[!htp]
    \centering
    \begin{minipage}{0.333\linewidth}
        \centering
        \includegraphics[width=\linewidth]{fitness_landscape_agglomerative_k3.pdf}
    \end{minipage}
    \begin{minipage}{0.19\linewidth}
        \centering
        \includegraphics[width=\linewidth]{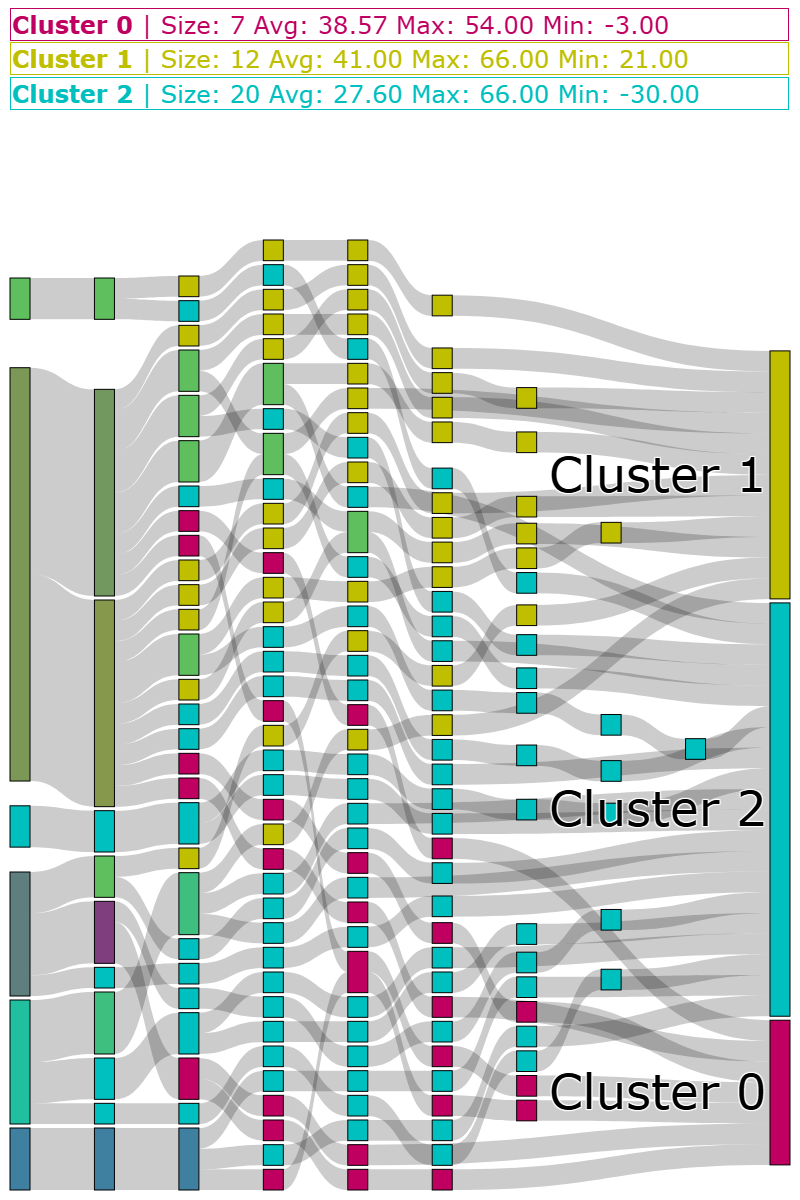}
    \end{minipage}
    \begin{minipage}{0.19\linewidth}
        \centering
        \includegraphics[width=\linewidth]{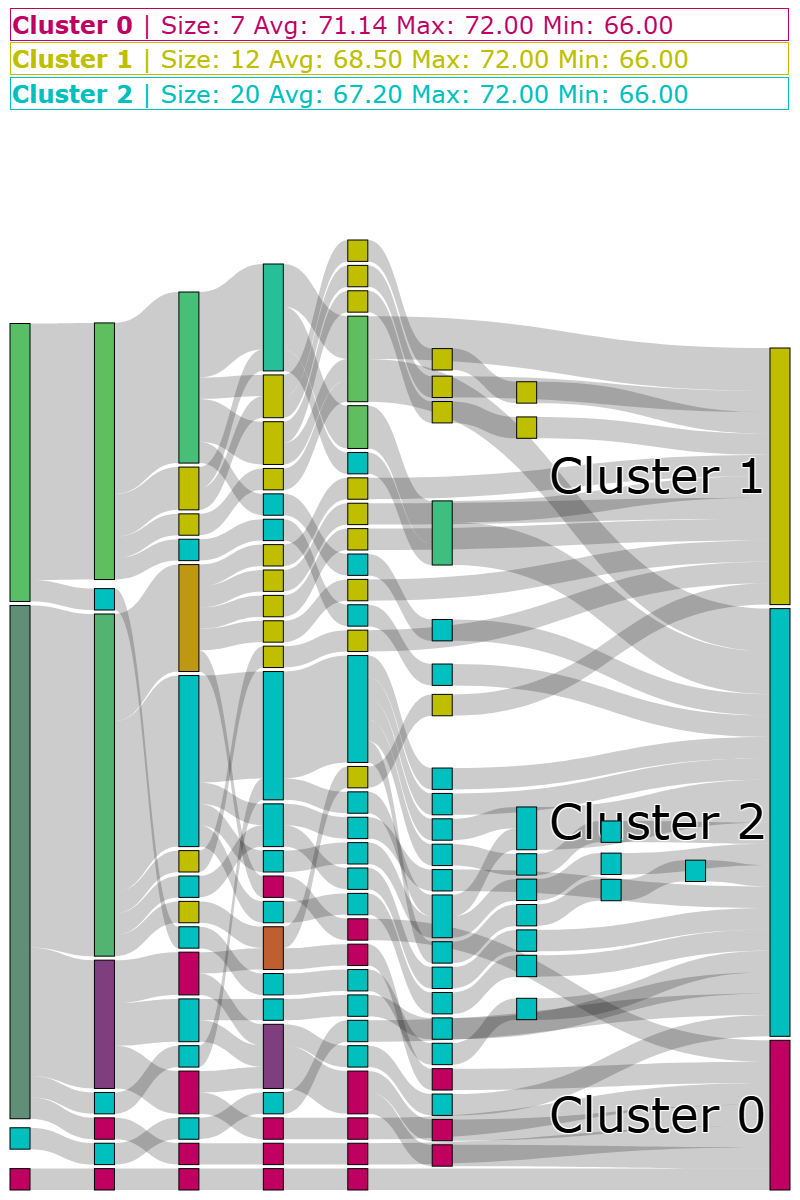}
    \end{minipage}
    \begin{minipage}{0.19\linewidth}
        \centering
        \includegraphics[width=\linewidth]{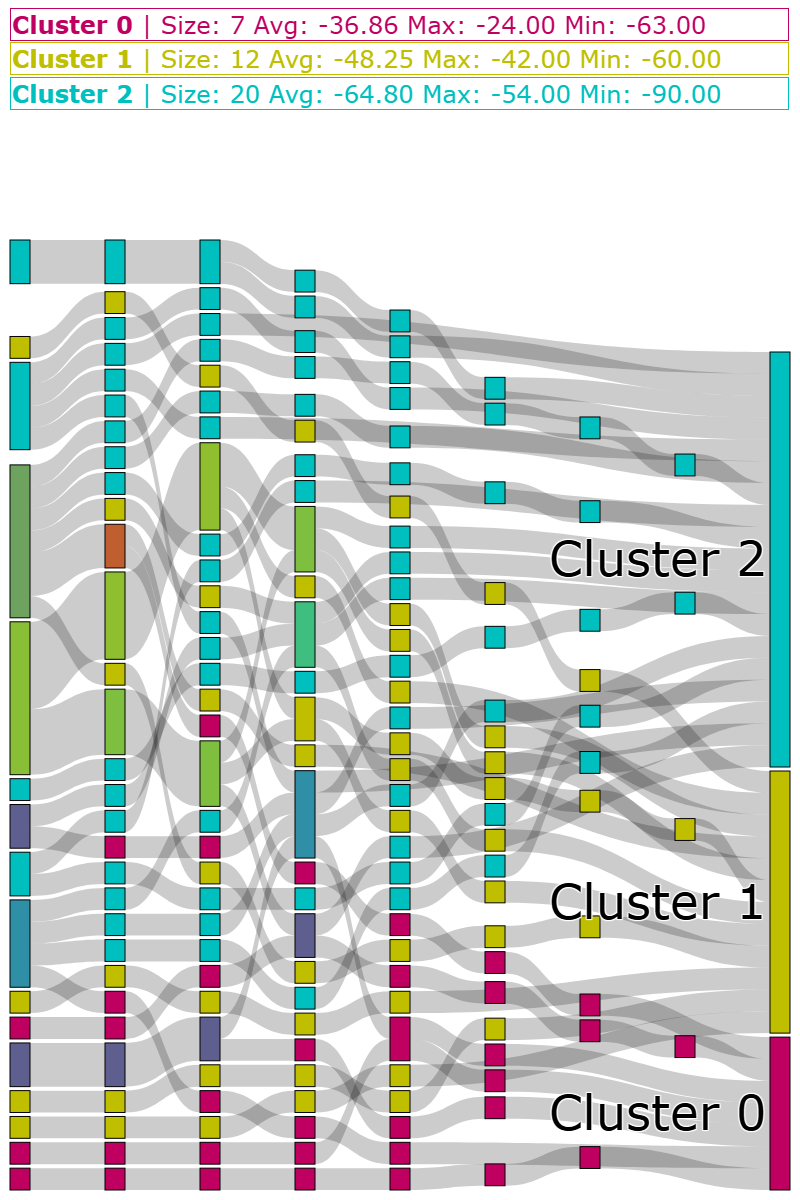}
    \end{minipage}
    \\
    \begin{minipage}{0.333\linewidth}
        \centering
        \includegraphics[width=\linewidth]{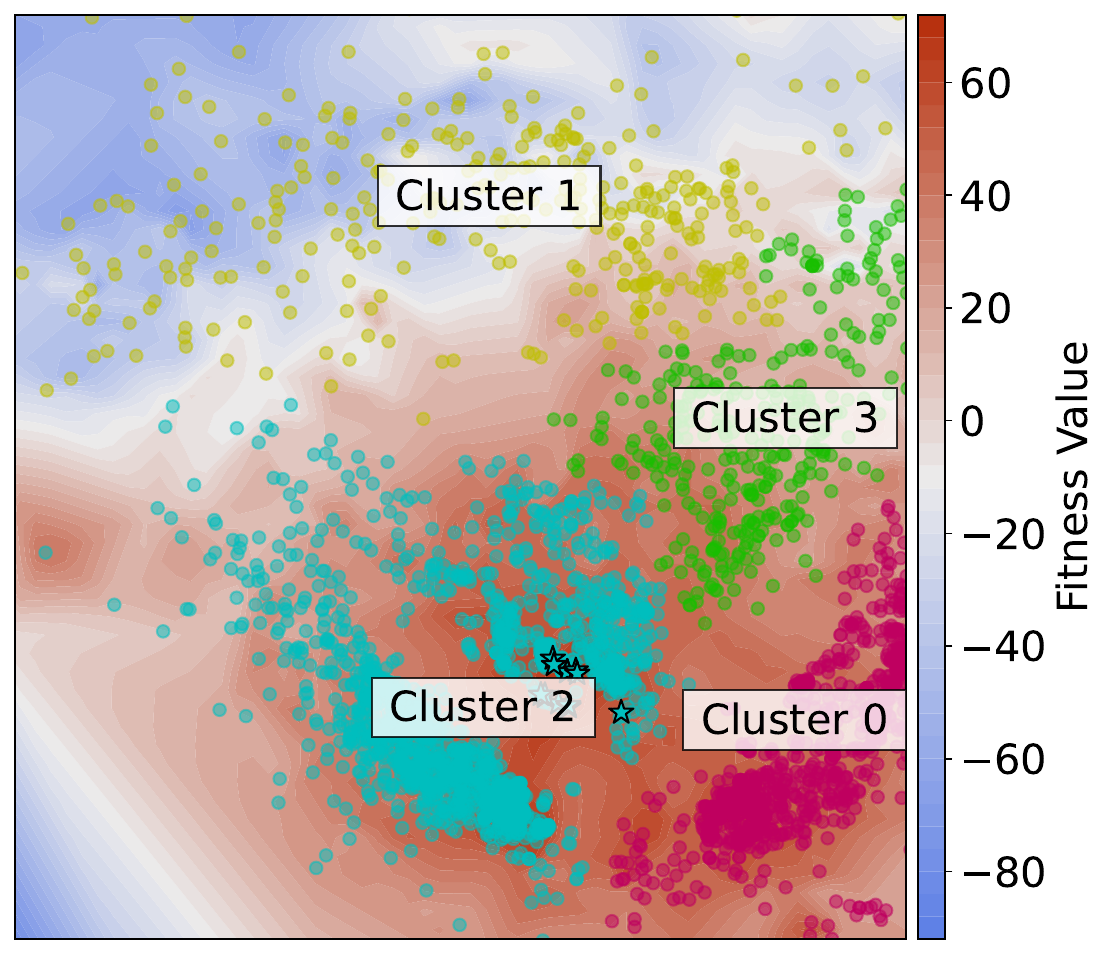}
    \end{minipage}
    \begin{minipage}{0.19\linewidth}
        \centering
        \includegraphics[width=\linewidth]{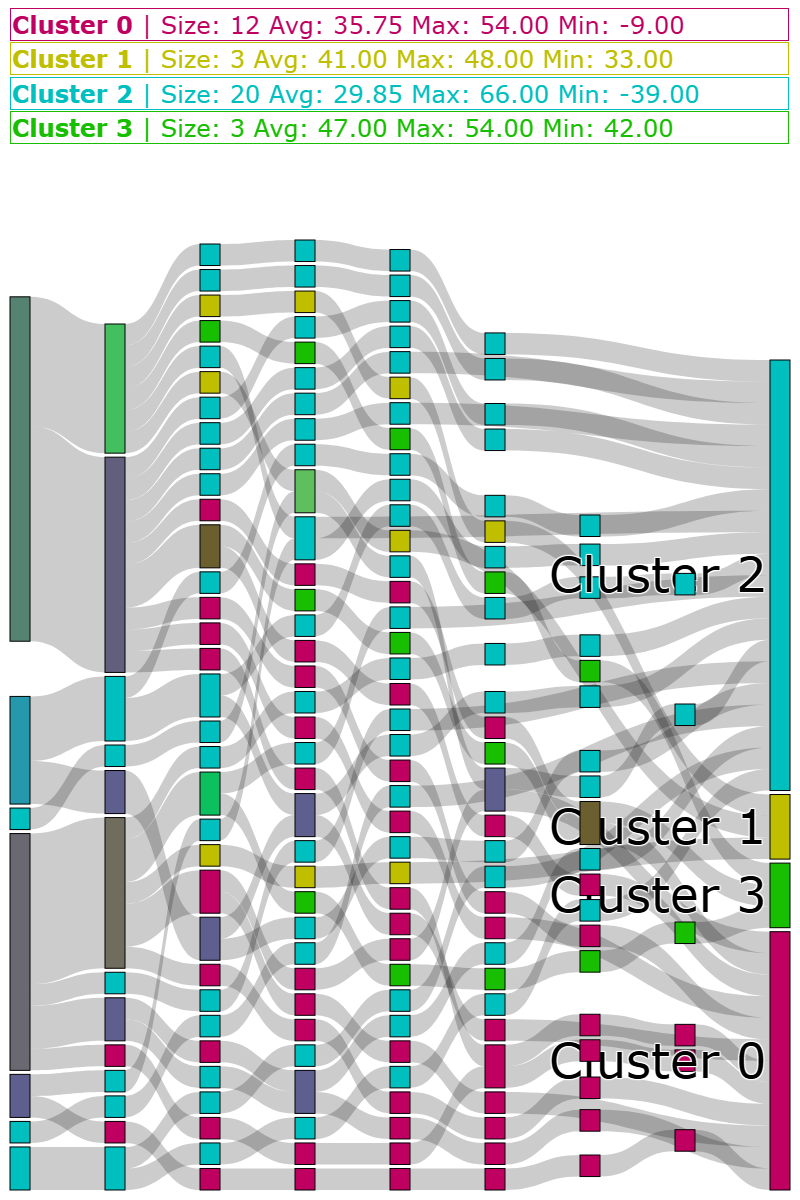}
    \end{minipage}
    \begin{minipage}{0.19\linewidth}
        \centering
        \includegraphics[width=\linewidth]{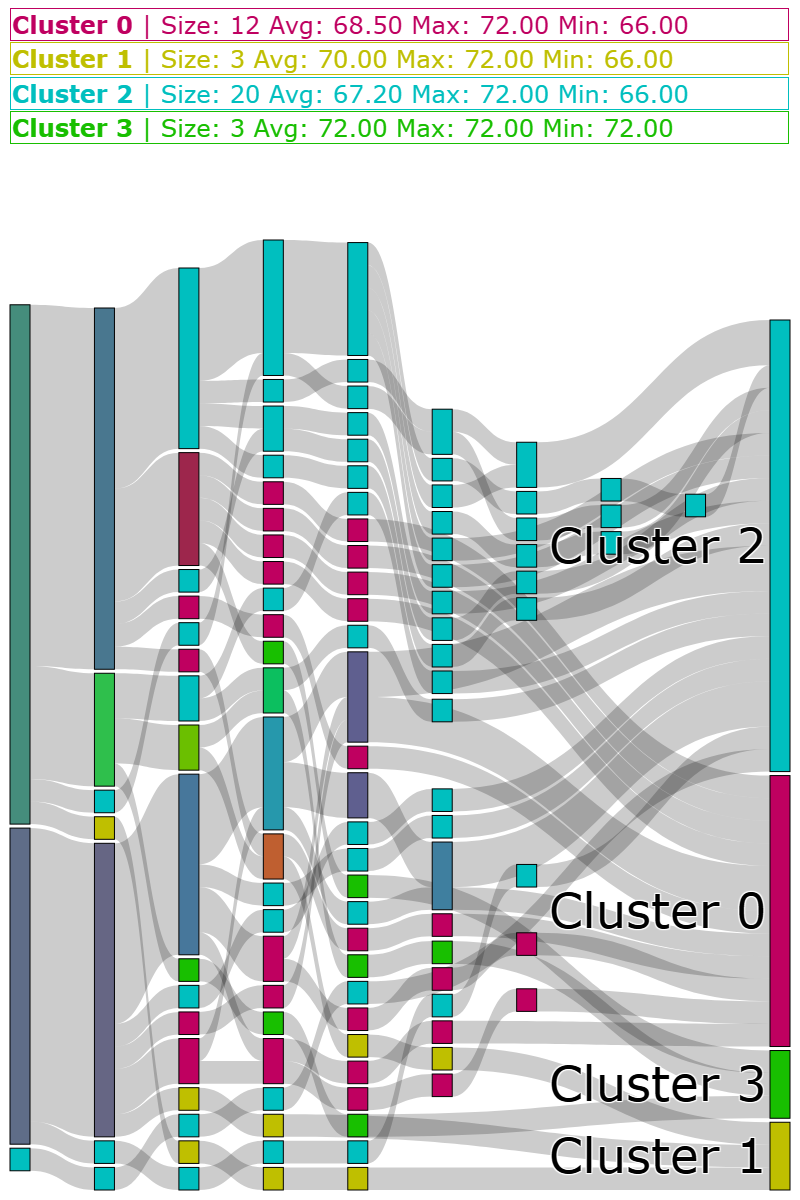}
    \end{minipage}
    \begin{minipage}{0.19\linewidth}
        \centering
        \includegraphics[width=\linewidth]{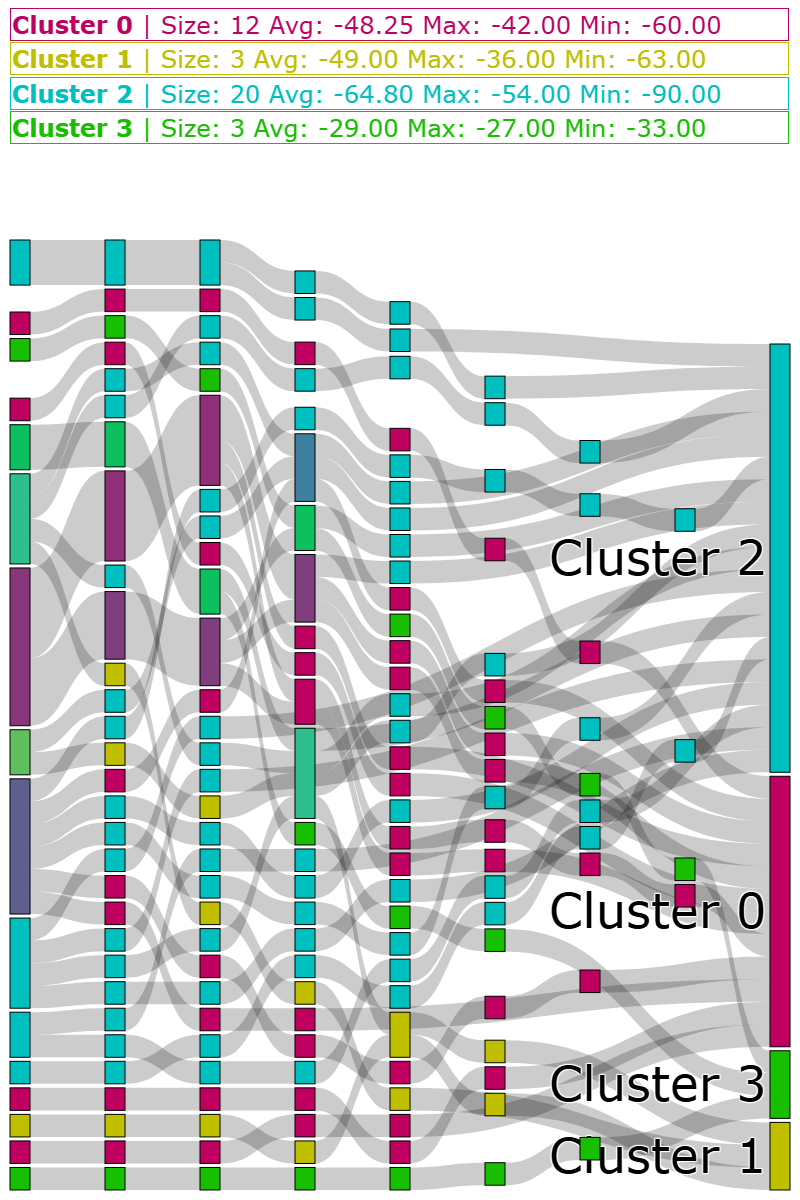}
    \end{minipage}
    \\
    \begin{minipage}{0.333\linewidth}
        \centering
        \includegraphics[width=\linewidth]{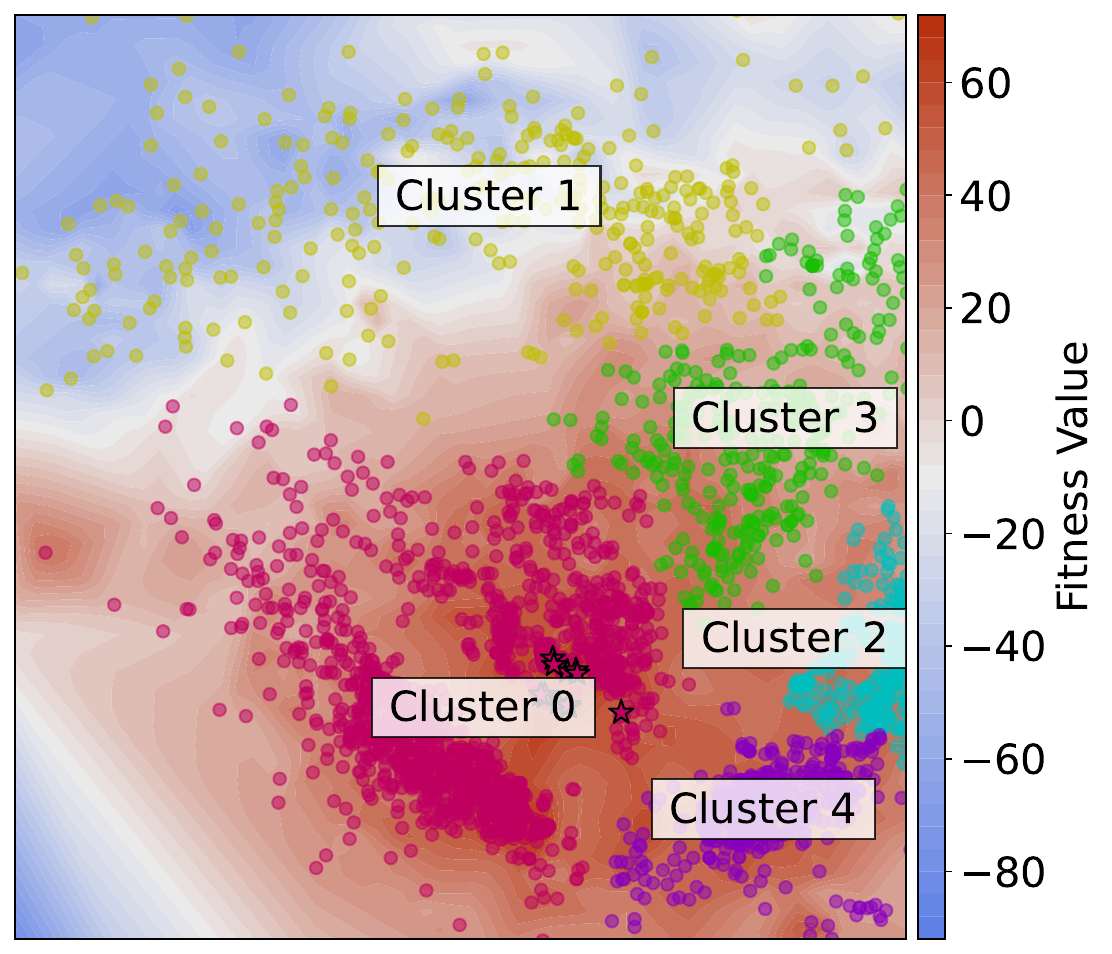}
    \end{minipage}
    \begin{minipage}{0.19\linewidth}
        \centering
        \includegraphics[width=\linewidth]{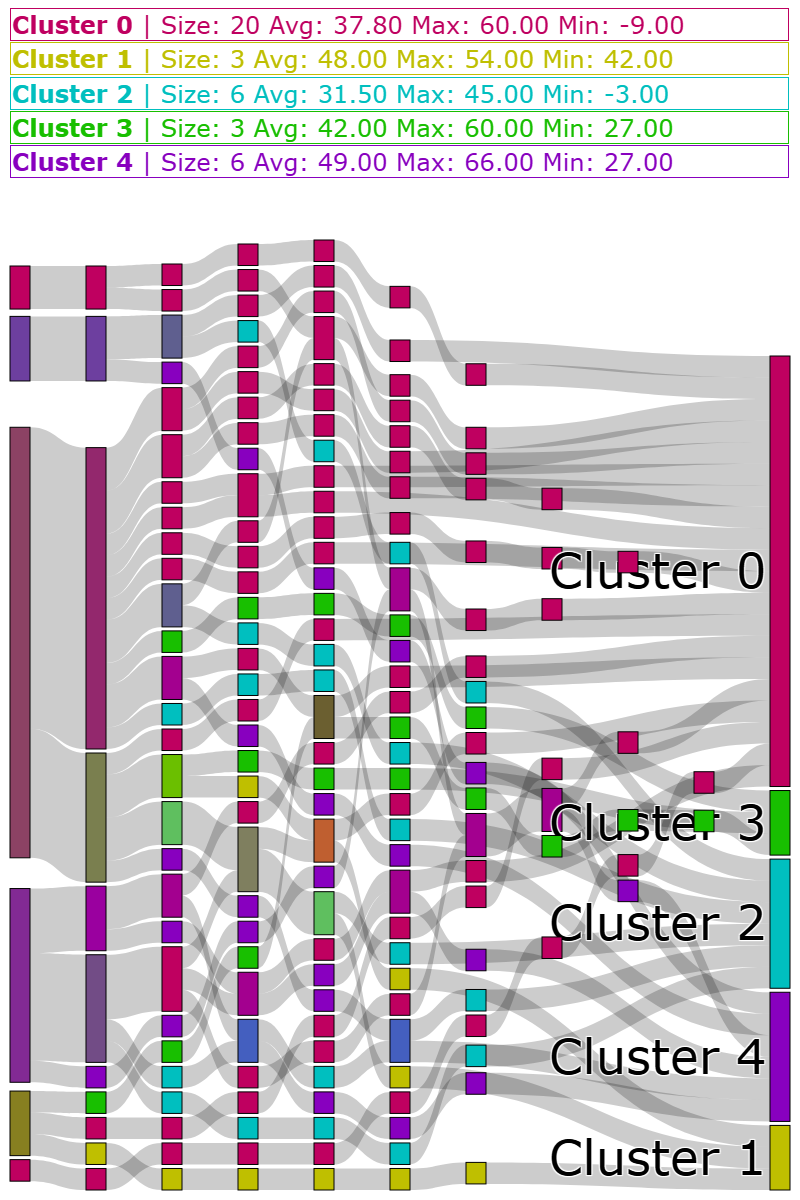}
    \end{minipage}
    \begin{minipage}{0.19\linewidth}
        \centering
        \includegraphics[width=\linewidth]{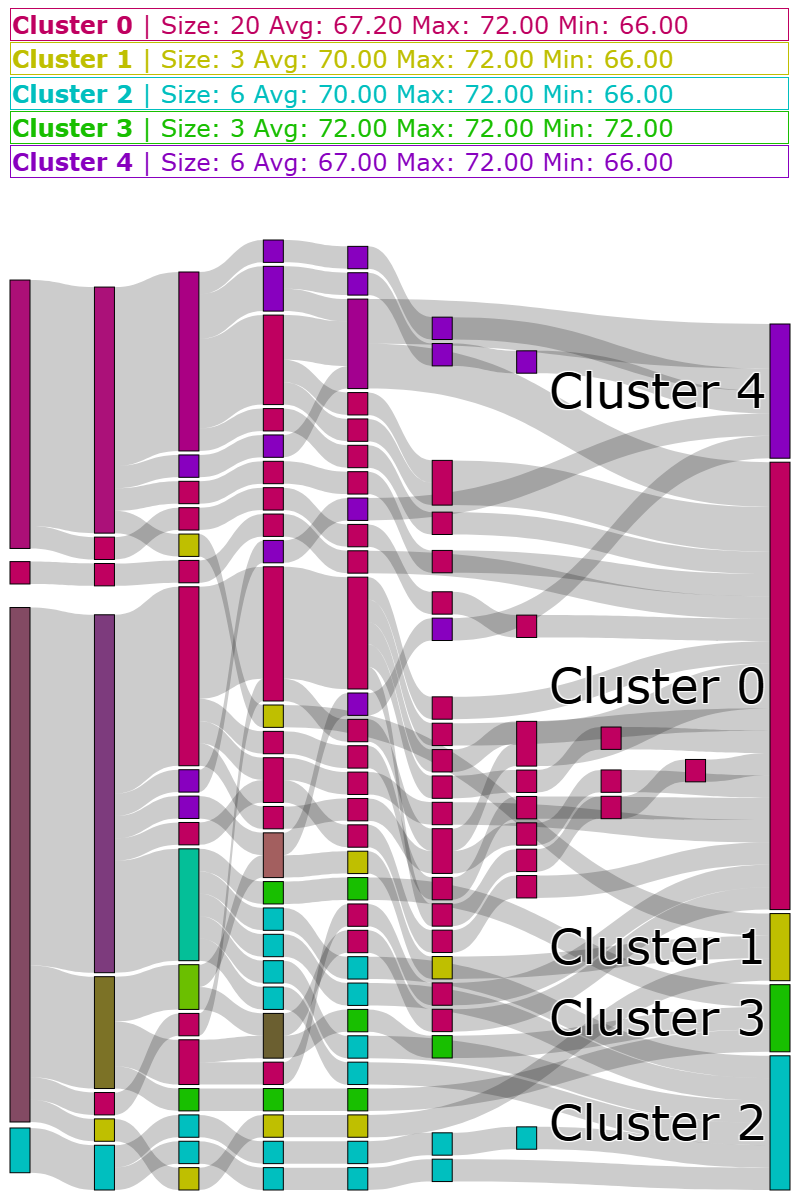}
    \end{minipage}
    \begin{minipage}{0.19\linewidth}
        \centering
        \includegraphics[width=\linewidth]{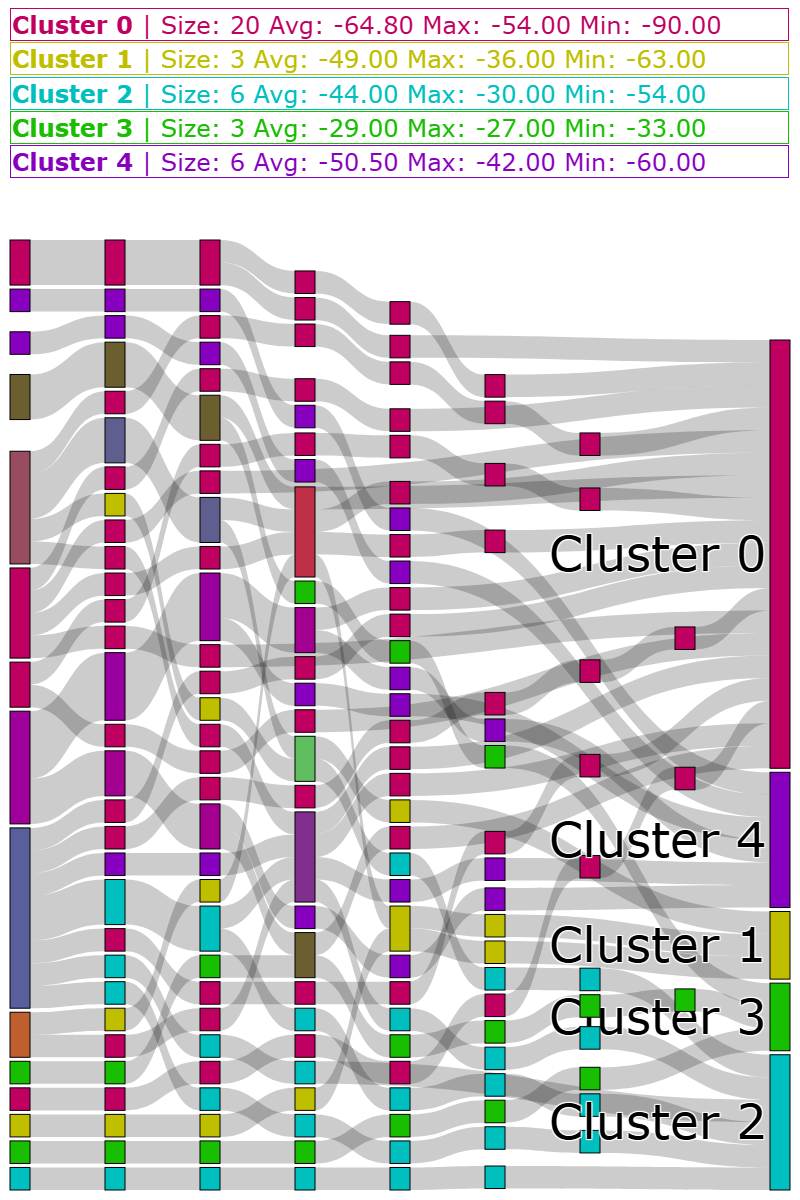}
    \end{minipage}
    \\
    \begin{minipage}{0.333\linewidth}
        \centering
        \includegraphics[width=\linewidth]{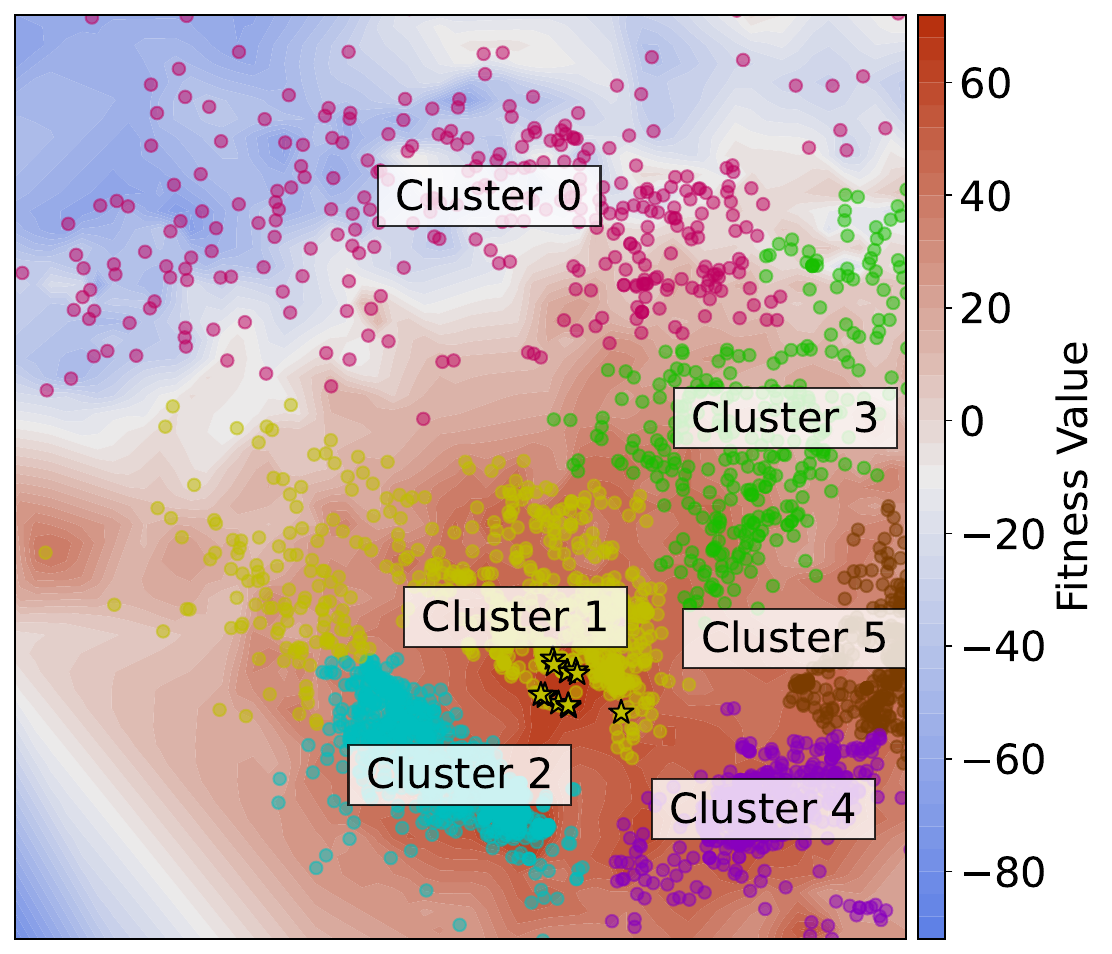}
        fitness landscape
    \end{minipage}
    \begin{minipage}{0.19\linewidth}
        \centering
        \includegraphics[width=\linewidth]{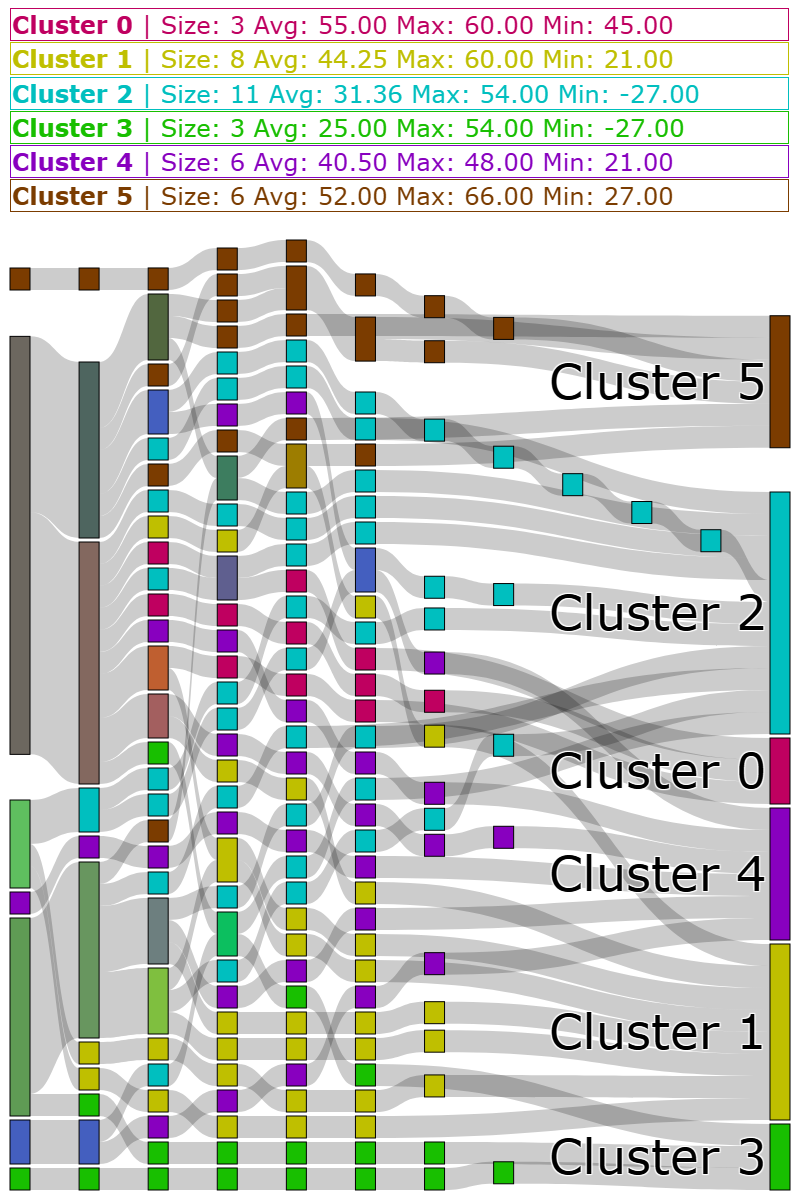}
        unsorted
    \end{minipage}
    \begin{minipage}{0.19\linewidth}
        \centering
        \includegraphics[width=\linewidth]{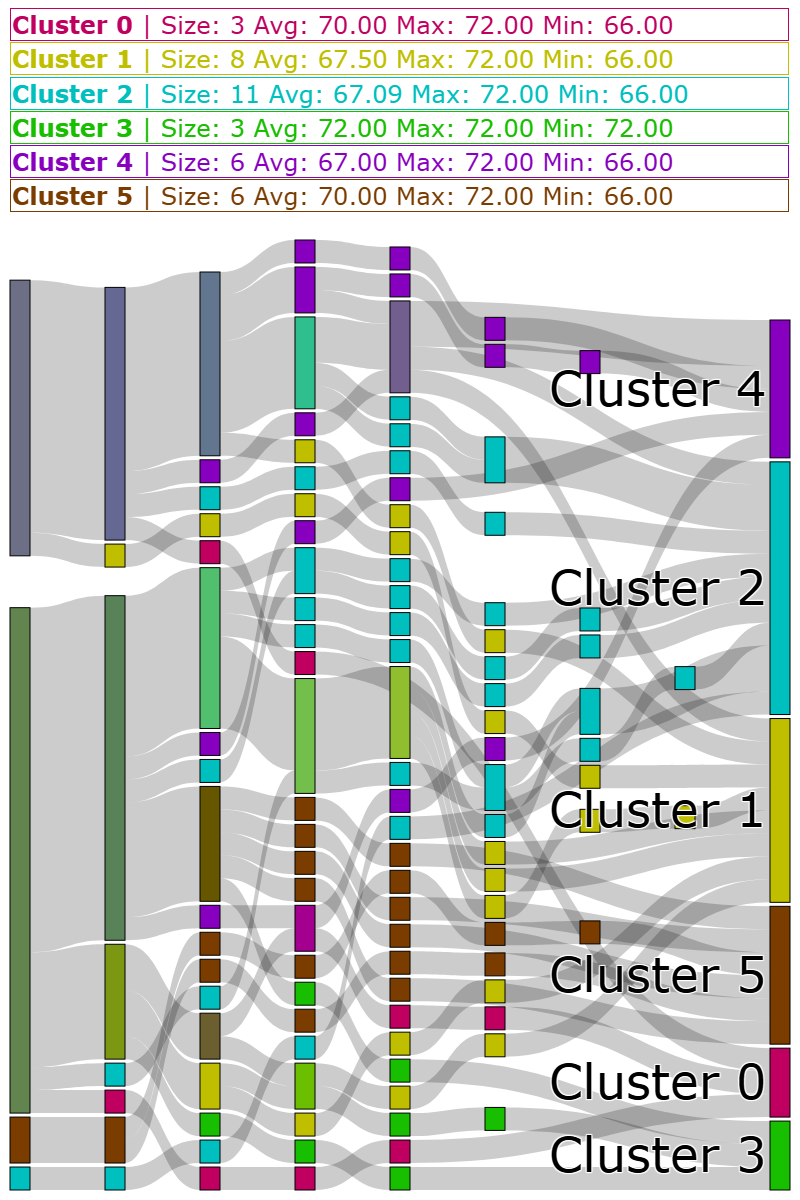}
        win
    \end{minipage}
    \begin{minipage}{0.19\linewidth}
        \centering
        \includegraphics[width=\linewidth]{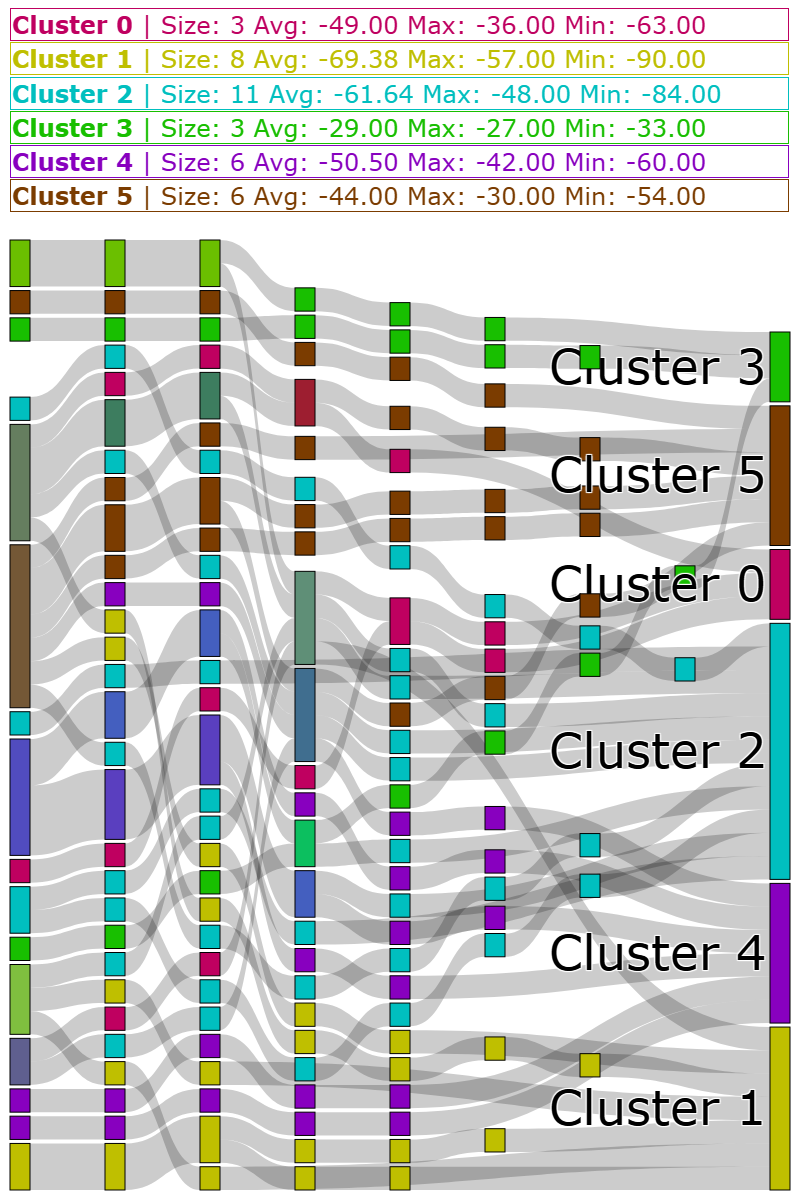}
        loss
    \end{minipage}
    \\
    \caption{Comprehensive sankey diagram of action sequences patterns with Ward Agglomerative Clustering algorithm for clustering}
    \label{fig:appendix_comprehensive_sankey_diagram_Agglomerative}
\end{figure*}

\begin{figure*}[!htp]
    \centering
    \begin{minipage}{0.333\linewidth}
        \centering
        \includegraphics[width=\linewidth]{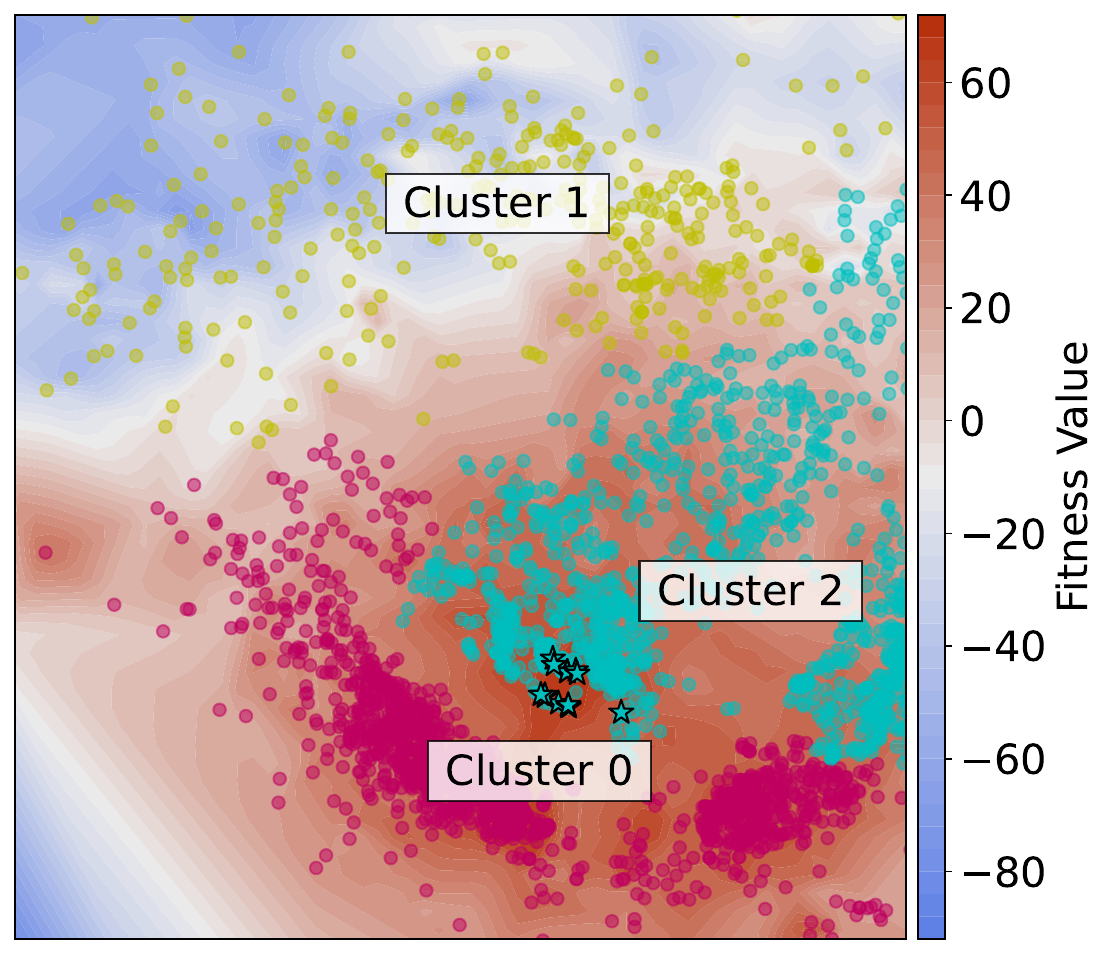}
    \end{minipage}
    \begin{minipage}{0.19\linewidth}
        \centering
        \includegraphics[width=\linewidth]{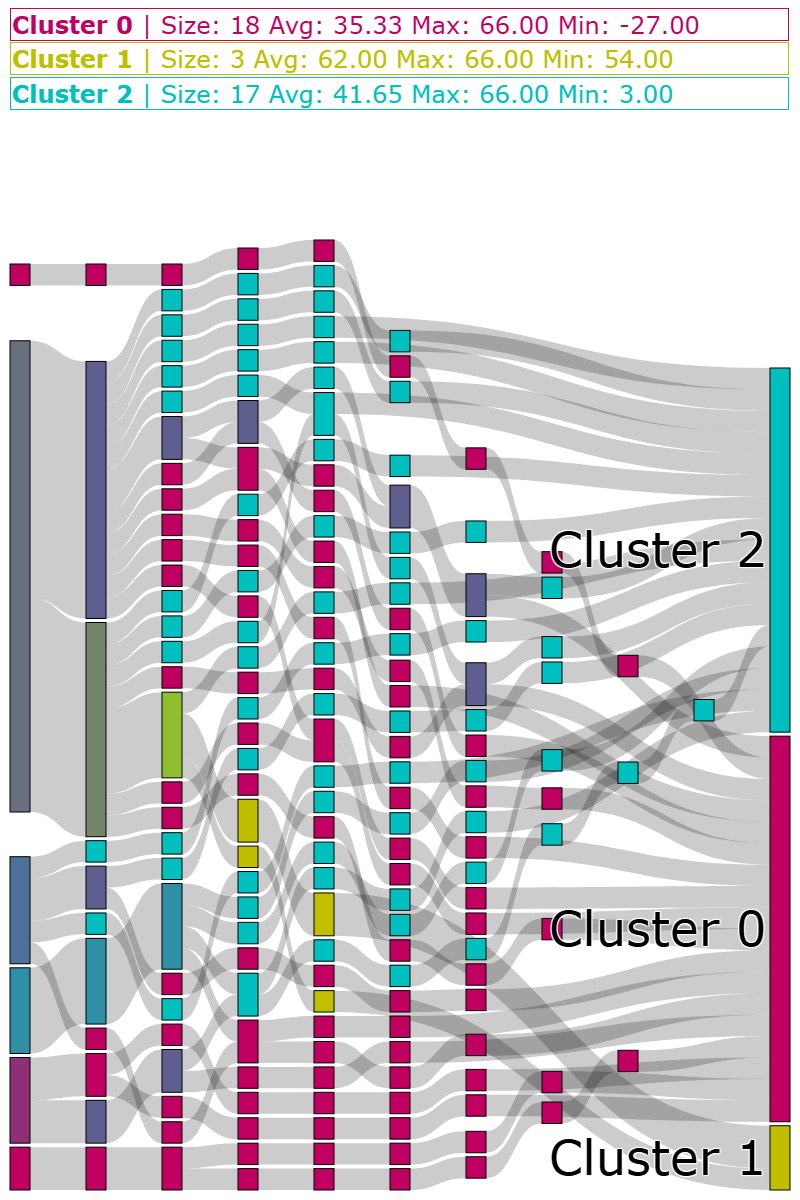}
    \end{minipage}
    \begin{minipage}{0.19\linewidth}
        \centering
        \includegraphics[width=\linewidth]{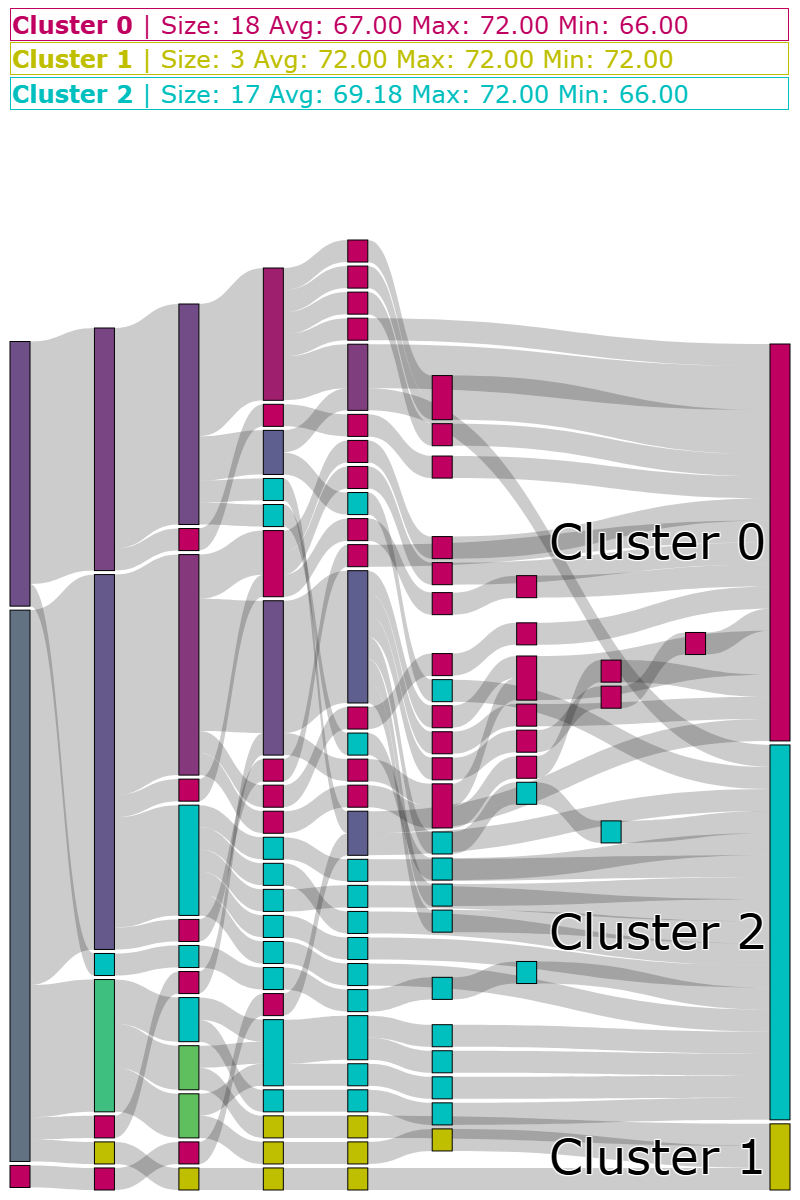}
    \end{minipage}
    \begin{minipage}{0.19\linewidth}
        \centering
        \includegraphics[width=\linewidth]{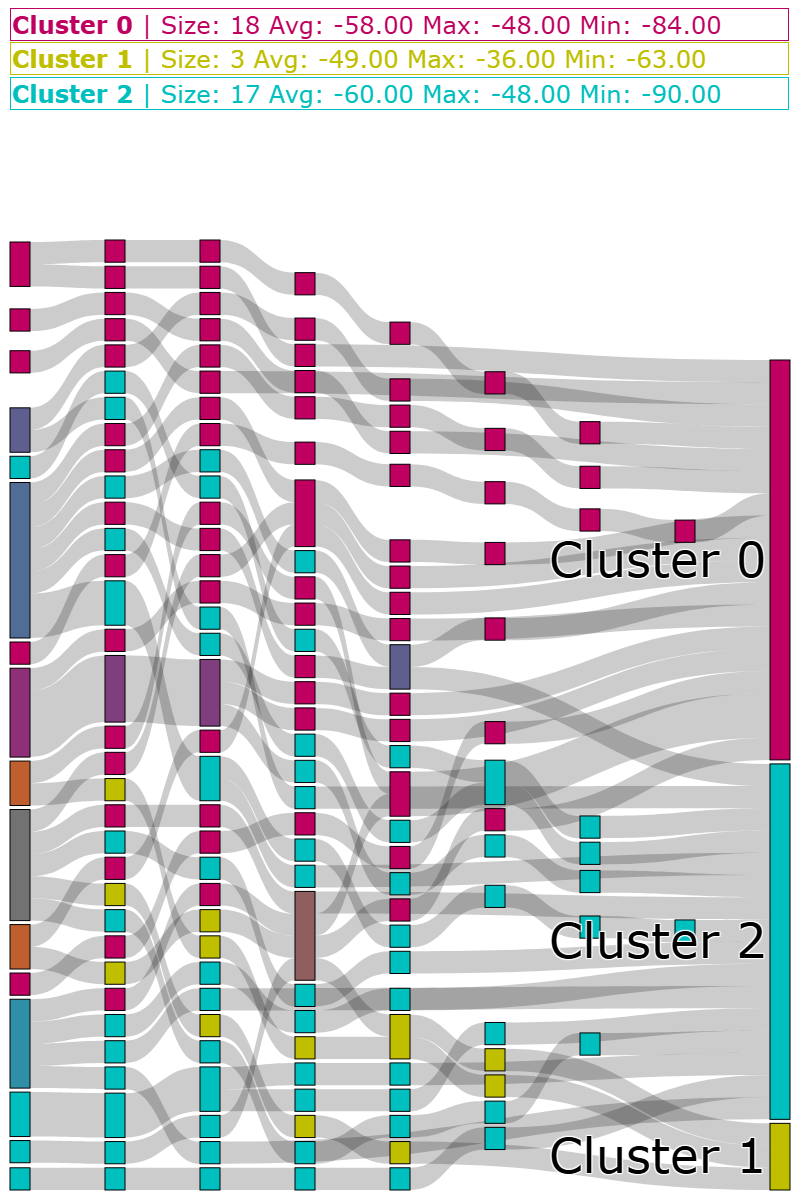}
    \end{minipage}
    \\
    \begin{minipage}{0.333\linewidth}
        \centering
        \includegraphics[width=\linewidth]{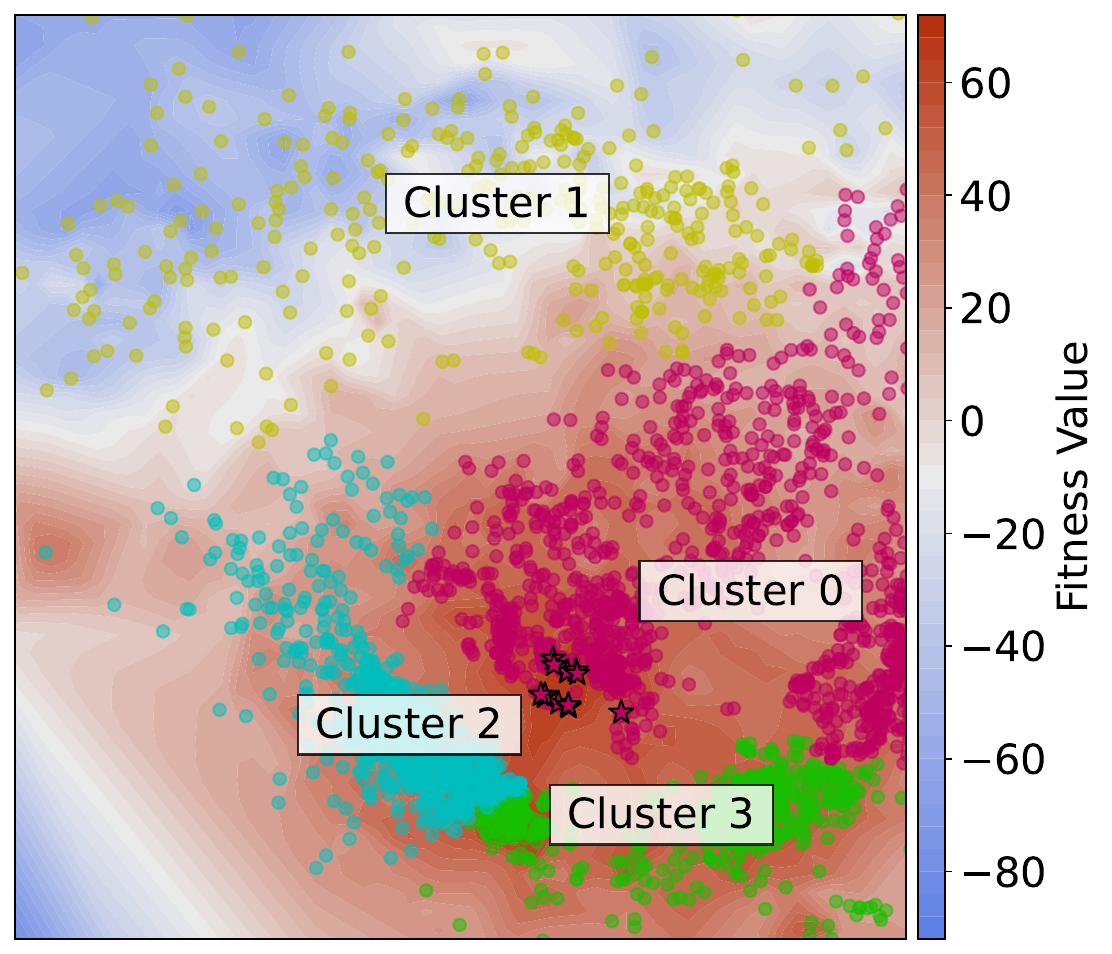}
    \end{minipage}
    \begin{minipage}{0.19\linewidth}
        \centering
        \includegraphics[width=\linewidth]{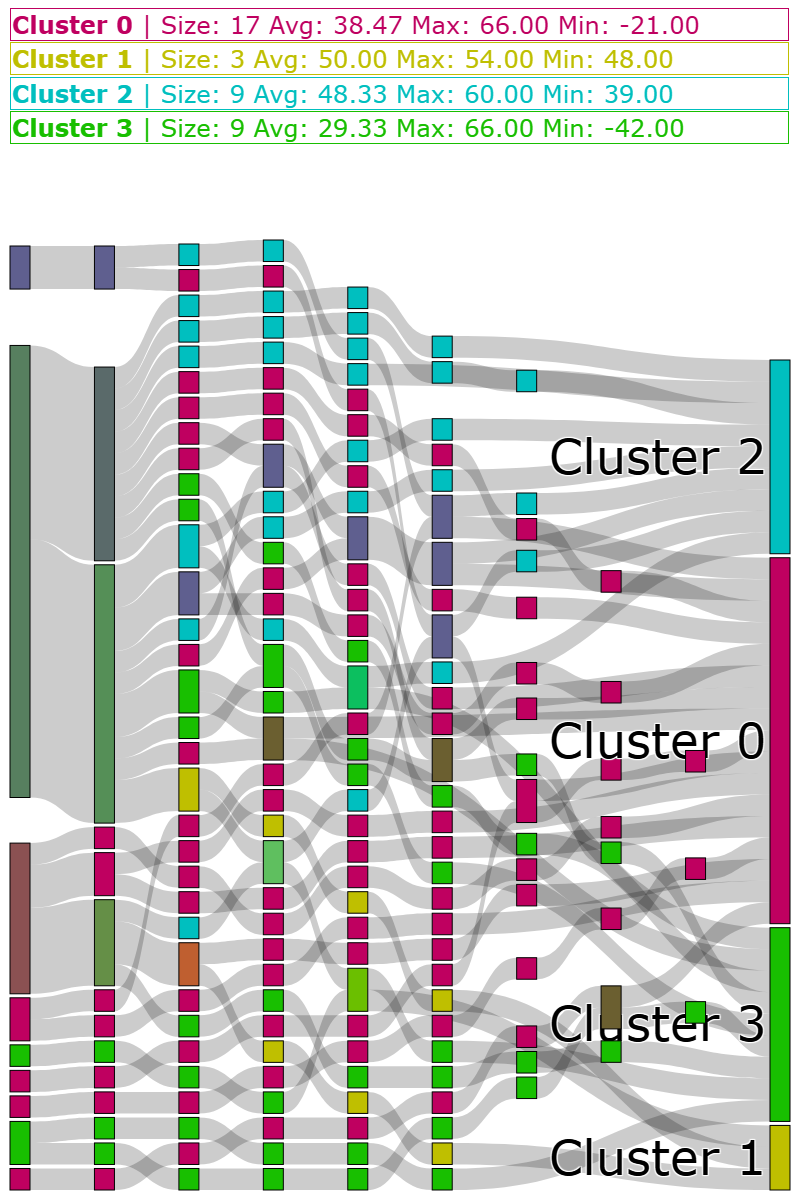}
    \end{minipage}
    \begin{minipage}{0.19\linewidth}
        \centering
        \includegraphics[width=\linewidth]{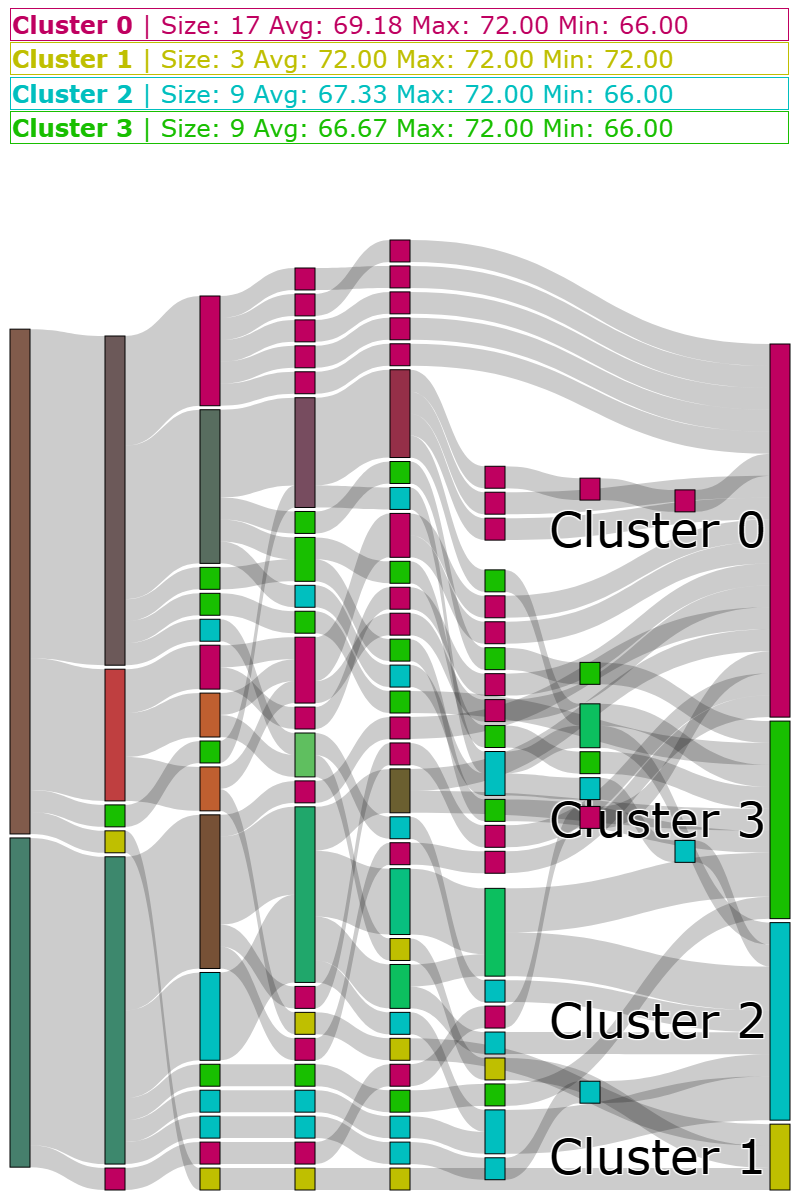}
    \end{minipage}
    \begin{minipage}{0.19\linewidth}
        \centering
        \includegraphics[width=\linewidth]{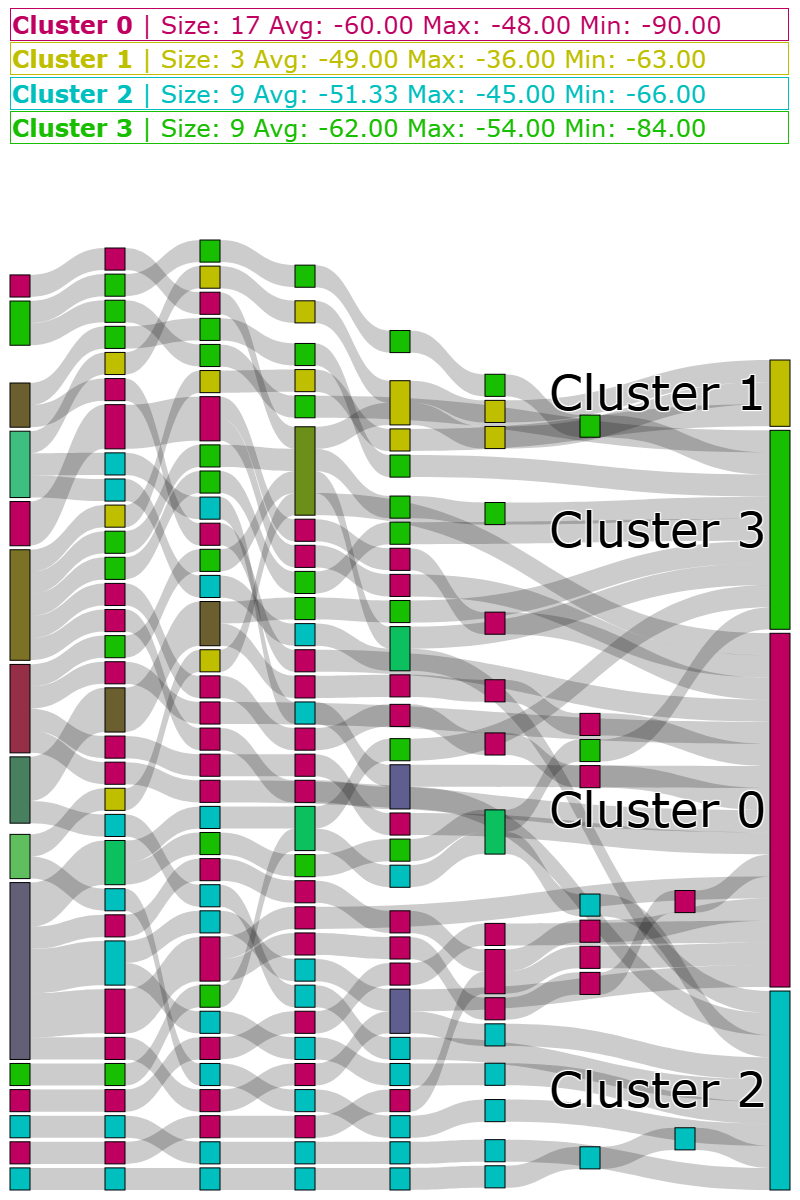}
    \end{minipage}
    \\
    \begin{minipage}{0.333\linewidth}
        \centering
        \includegraphics[width=\linewidth]{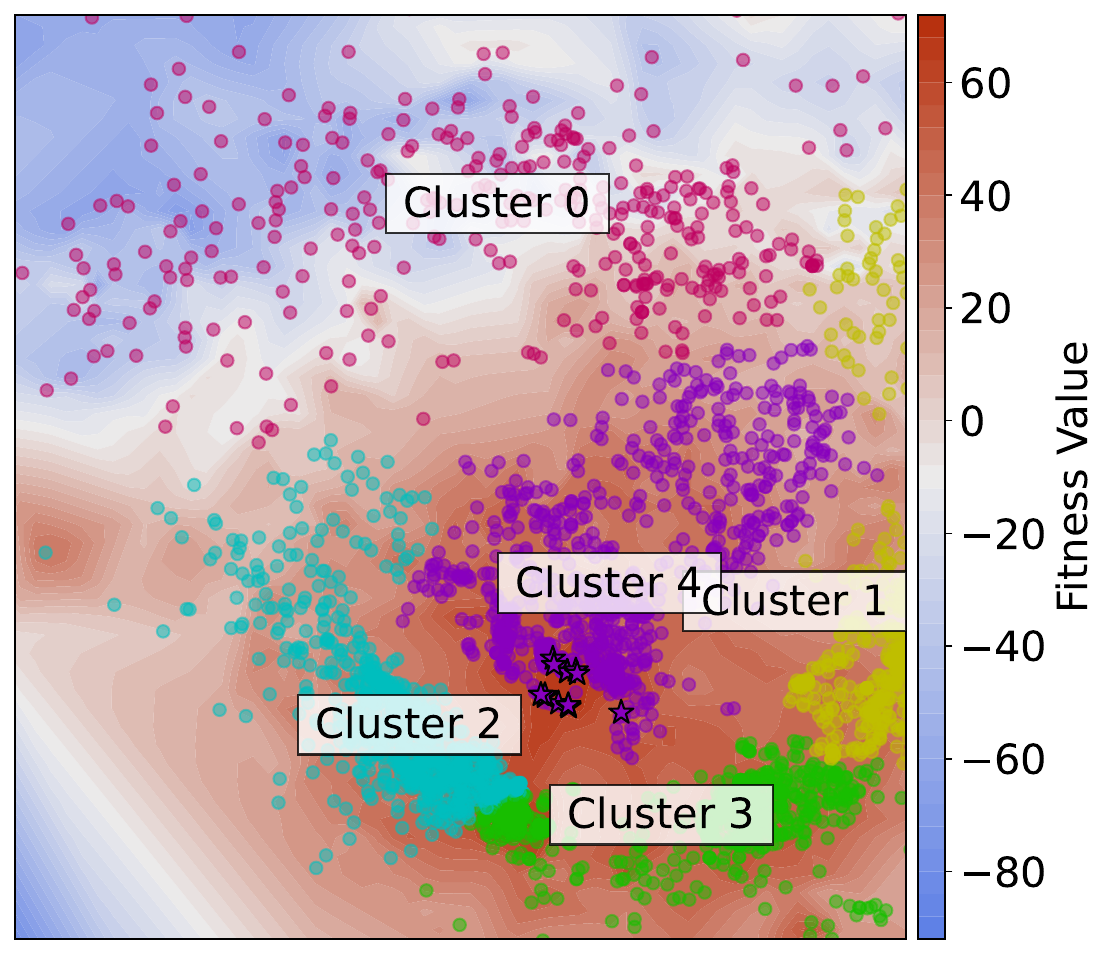}
    \end{minipage}
    \begin{minipage}{0.19\linewidth}
        \centering
        \includegraphics[width=\linewidth]{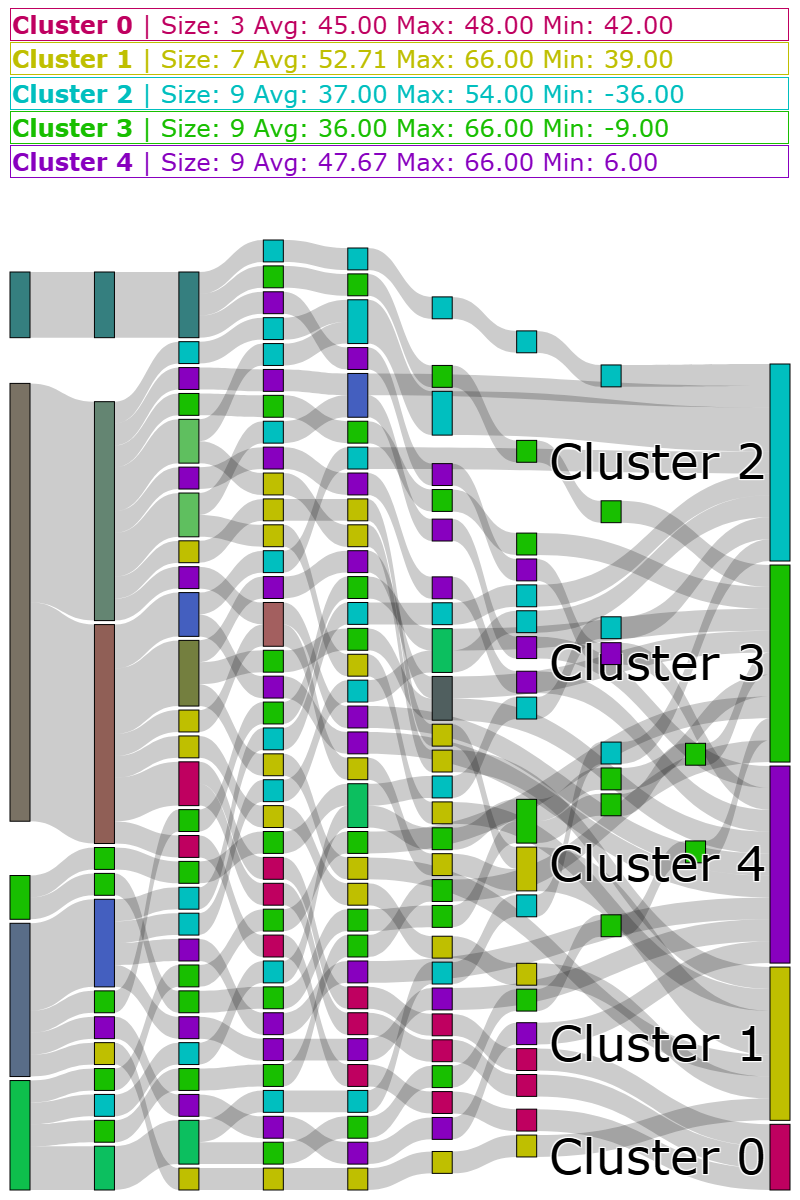}
    \end{minipage}
    \begin{minipage}{0.19\linewidth}
        \centering
        \includegraphics[width=\linewidth]{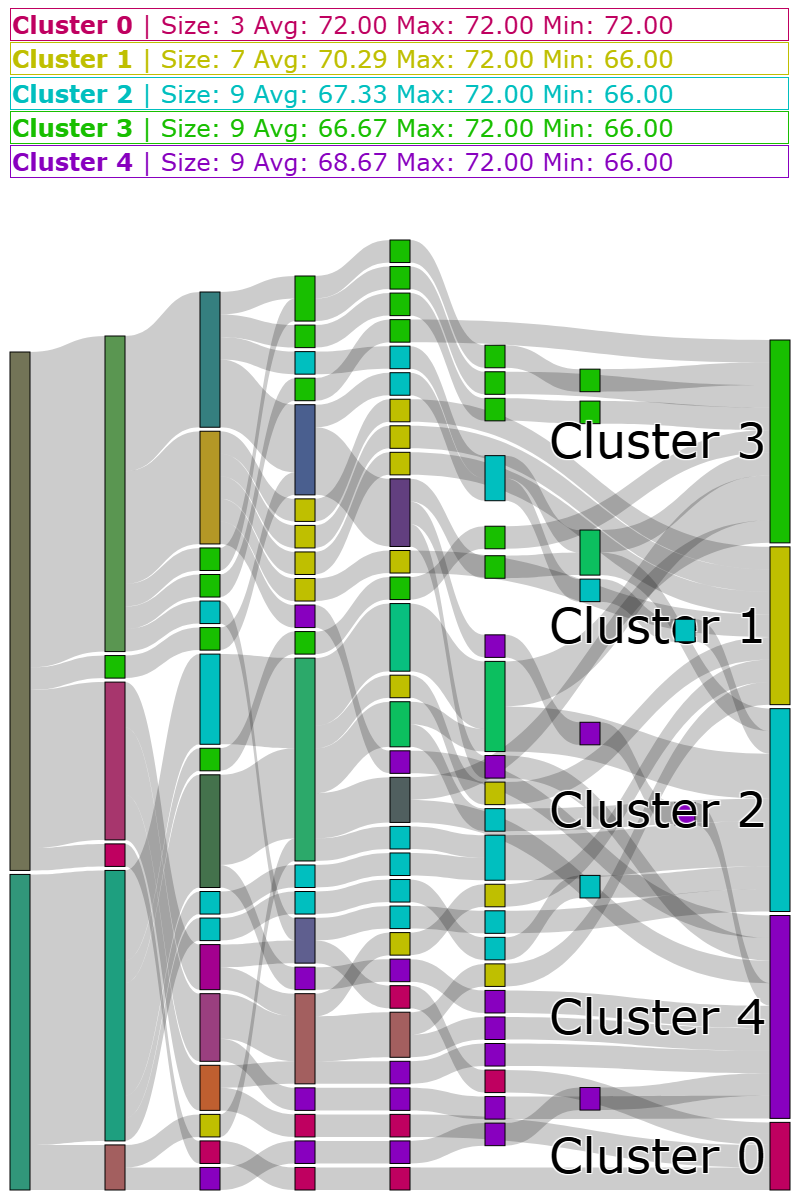}
    \end{minipage}
    \begin{minipage}{0.19\linewidth}
        \centering
        \includegraphics[width=\linewidth]{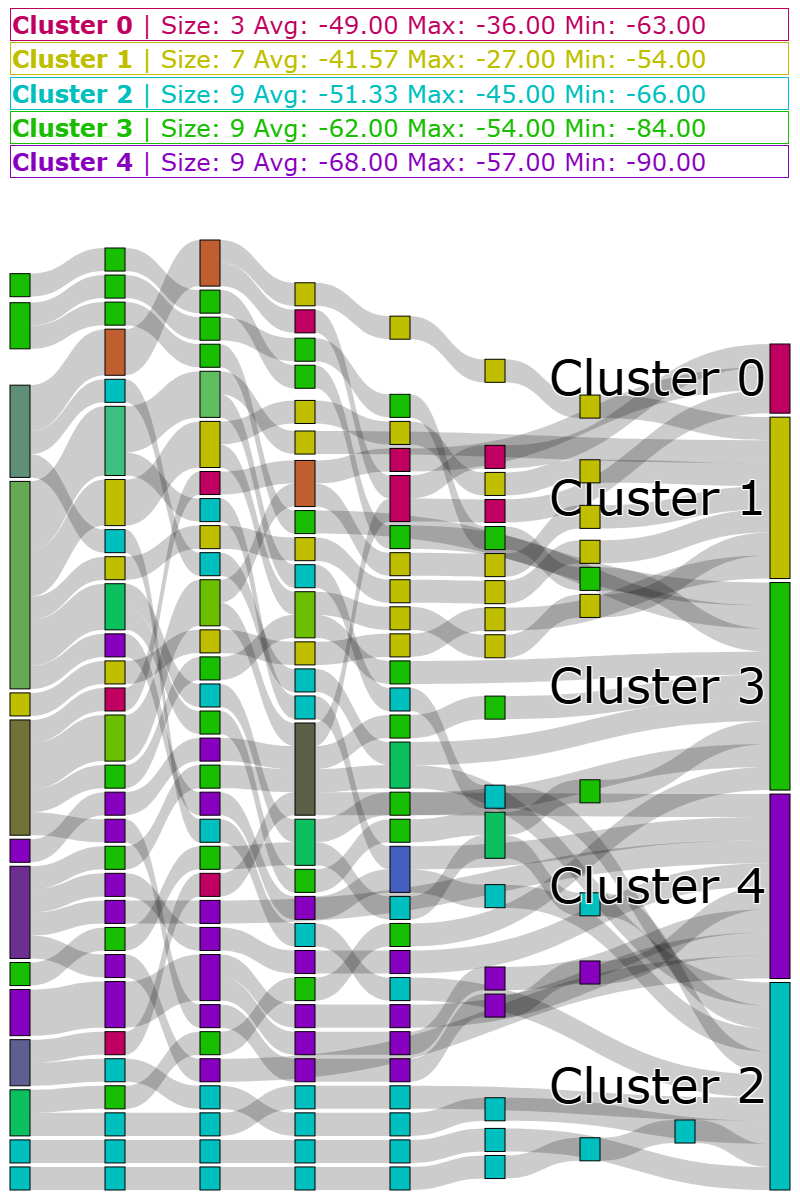}
    \end{minipage}
    \\
    \begin{minipage}{0.333\linewidth}
        \centering
        \includegraphics[width=\linewidth]{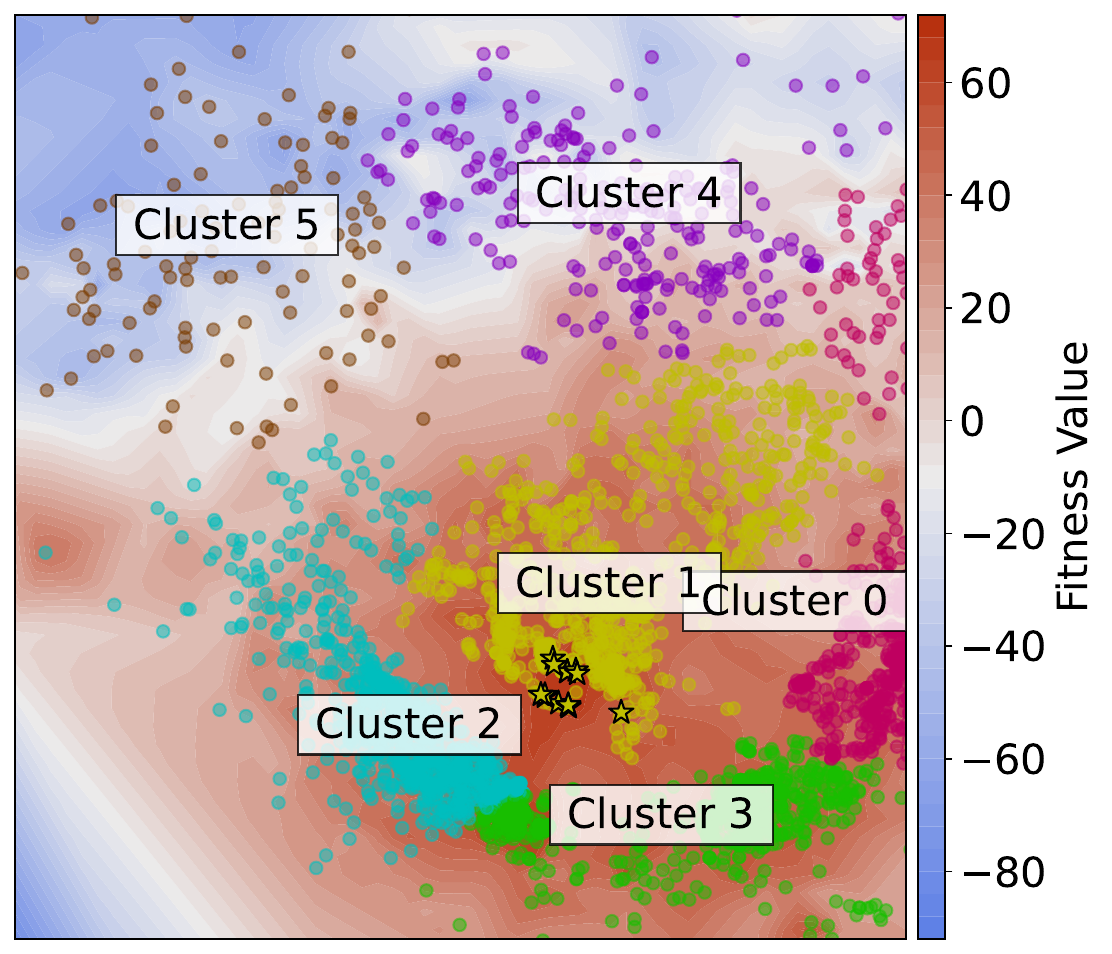}
        fitness landscape
    \end{minipage}
    \begin{minipage}{0.19\linewidth}
        \centering
        \includegraphics[width=\linewidth]{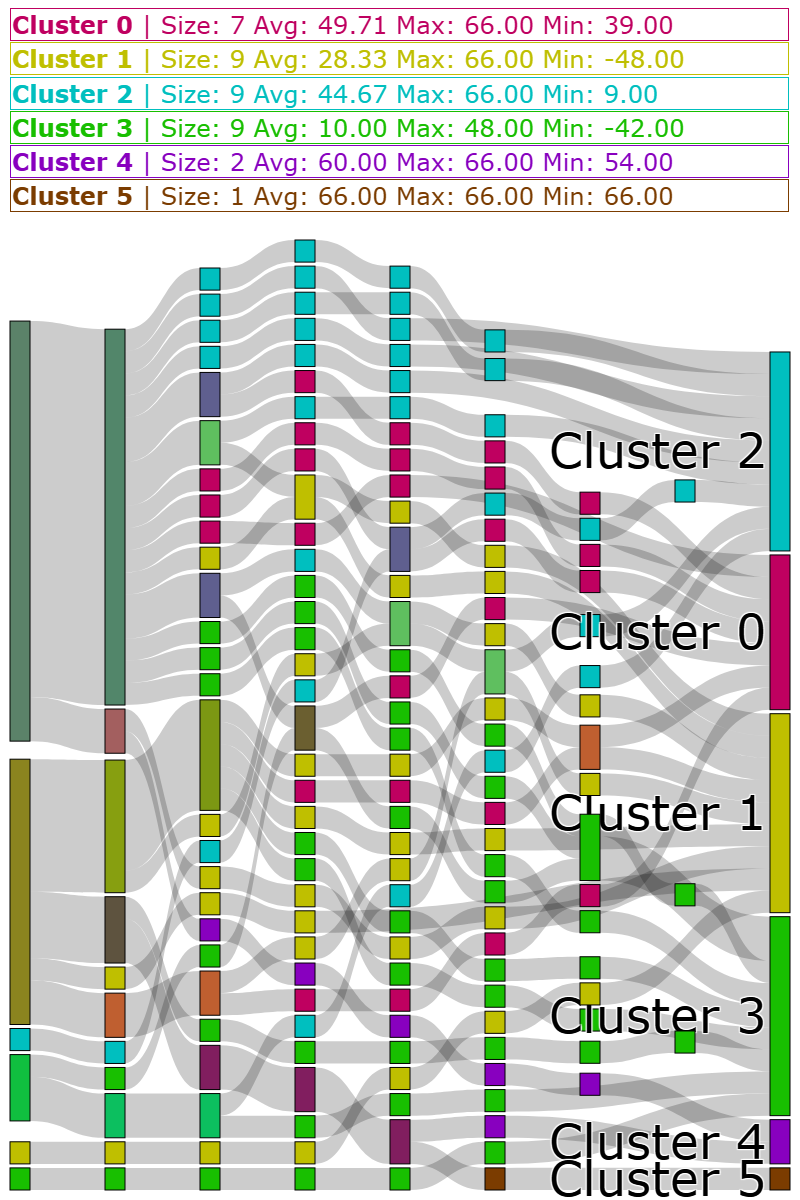}
        unsorted
    \end{minipage}
    \begin{minipage}{0.19\linewidth}
        \centering
        \includegraphics[width=\linewidth]{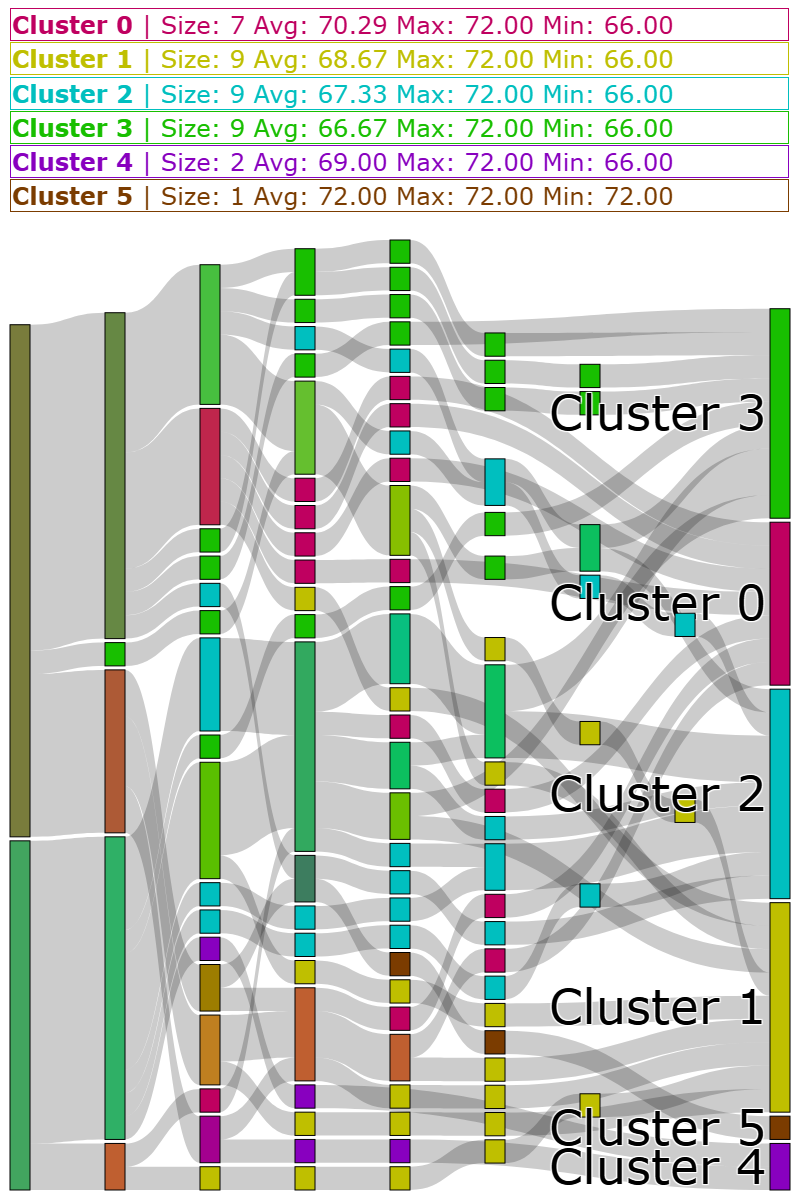}
        win
    \end{minipage}
    \begin{minipage}{0.19\linewidth}
        \centering
        \includegraphics[width=\linewidth]{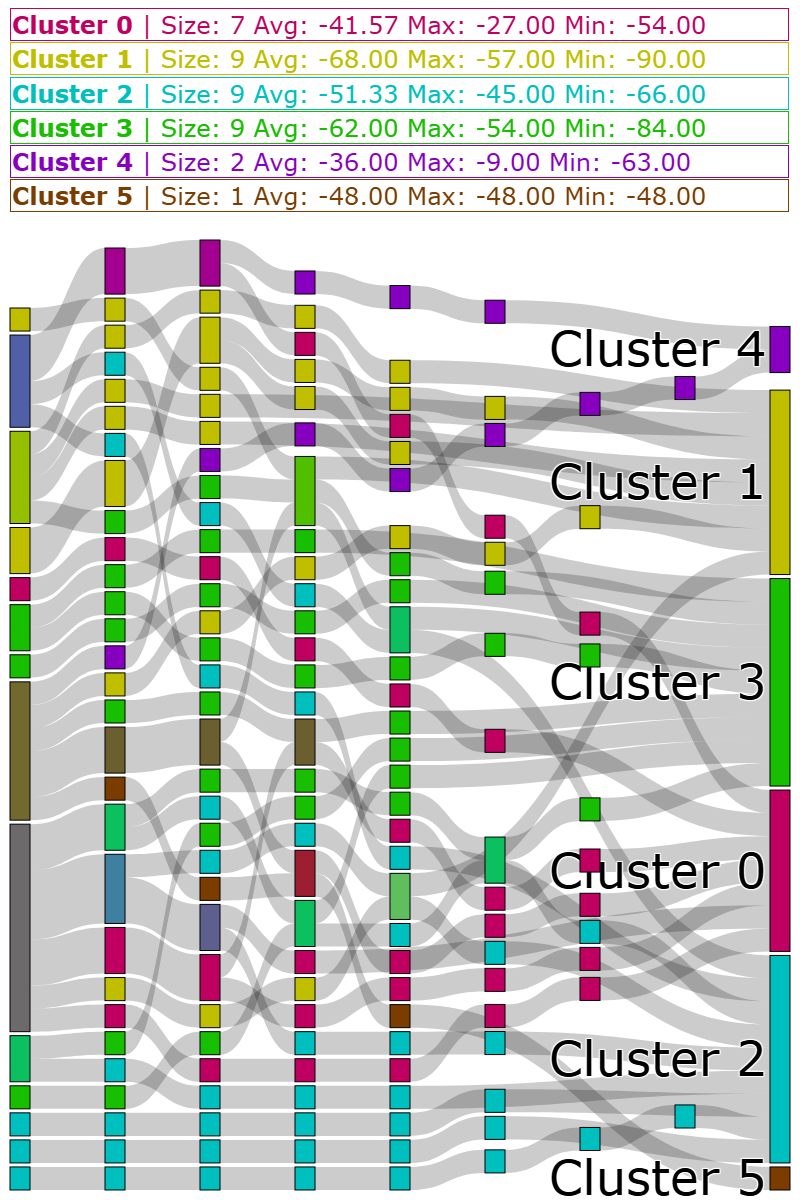}
        loss
    \end{minipage}
    \\
    \caption{Comprehensive sankey diagram of action sequences patterns with birch algorithm for clustering}
    \label{fig:comprehensive_sankey_diagram_birch}
\end{figure*}

\begin{figure*}[!htp]
    \centering
    \begin{minipage}{0.333\linewidth}
        \centering
        \includegraphics[width=\linewidth]{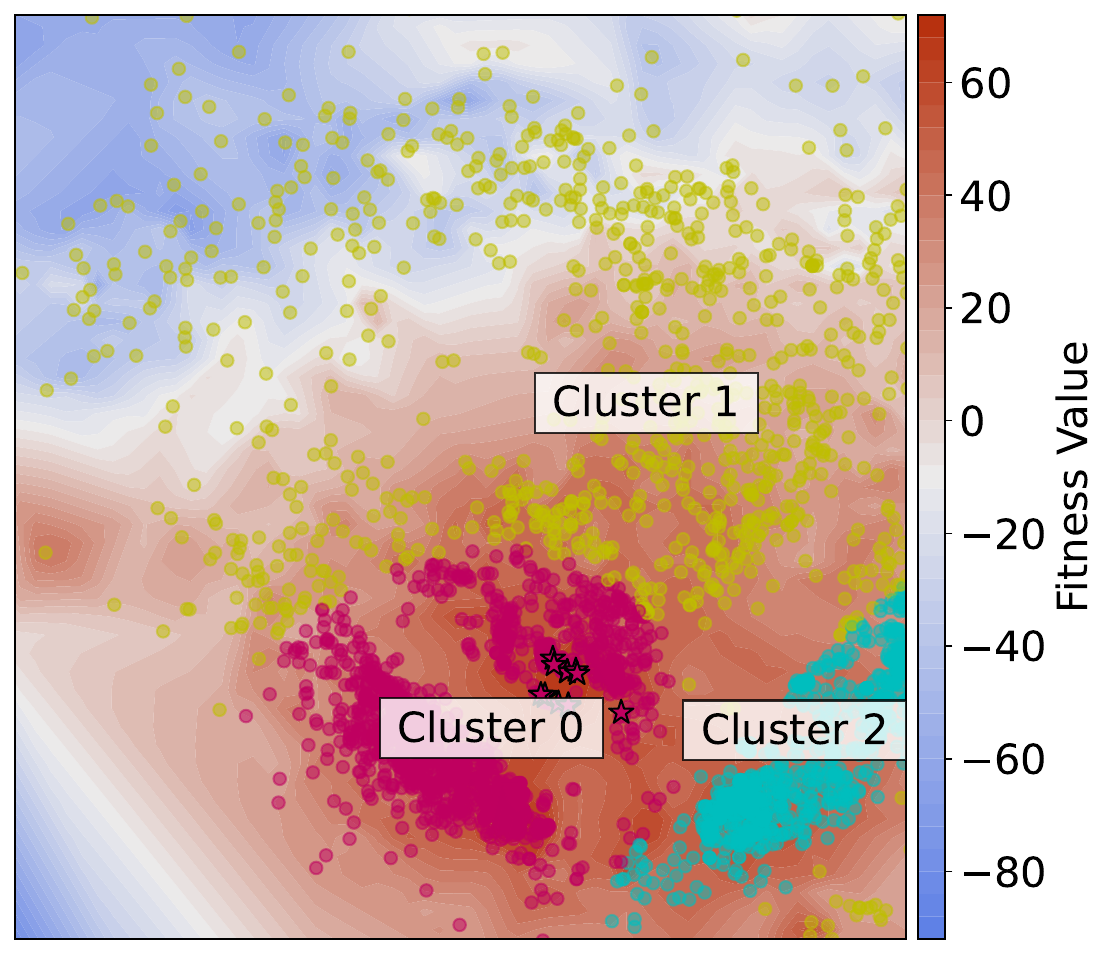}
    \end{minipage}
    \begin{minipage}{0.19\linewidth}
        \centering
        \includegraphics[width=\linewidth]{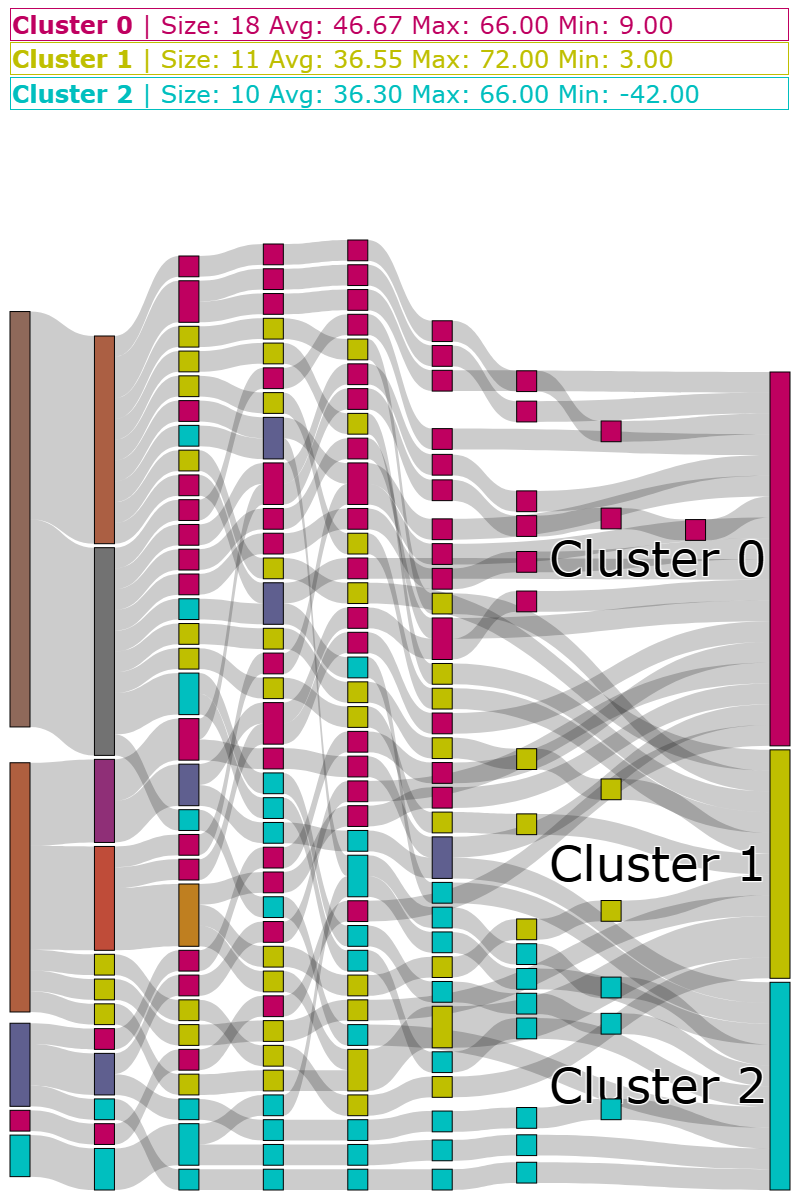}
    \end{minipage}
    \begin{minipage}{0.19\linewidth}
        \centering
        \includegraphics[width=\linewidth]{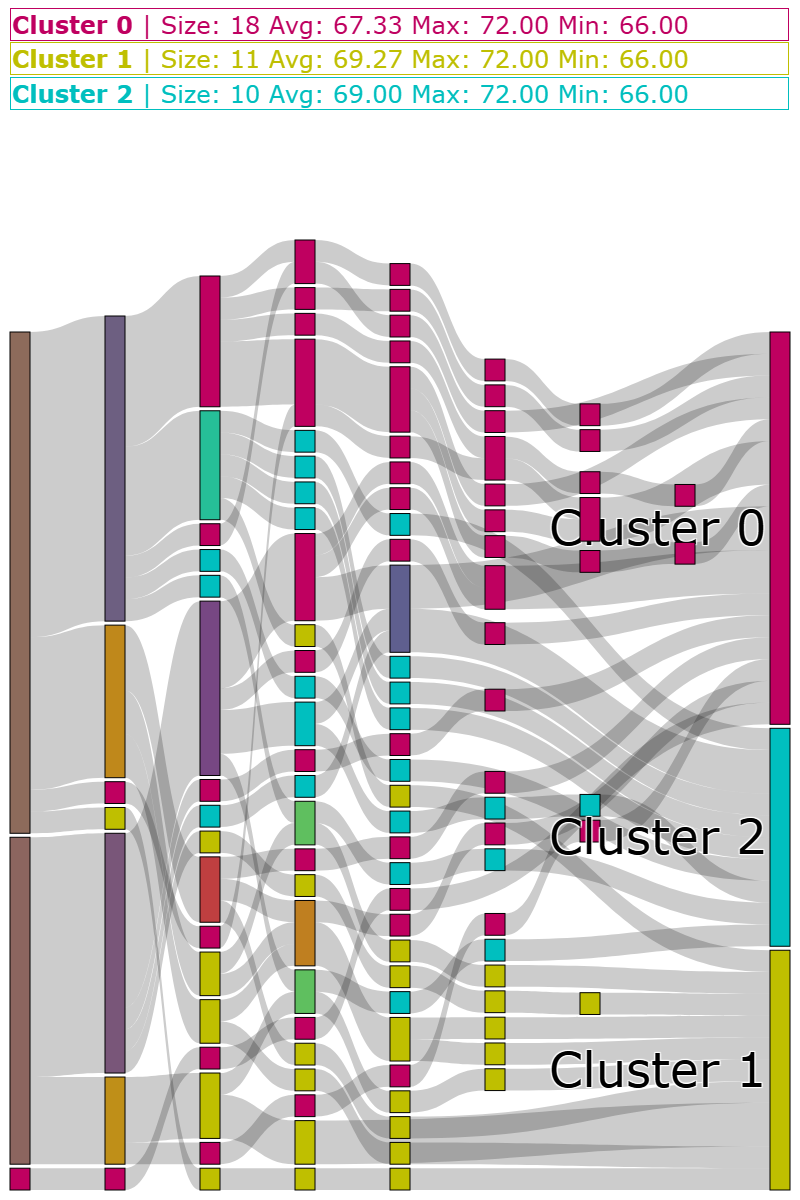}
    \end{minipage}
    \begin{minipage}{0.19\linewidth}
        \centering
        \includegraphics[width=\linewidth]{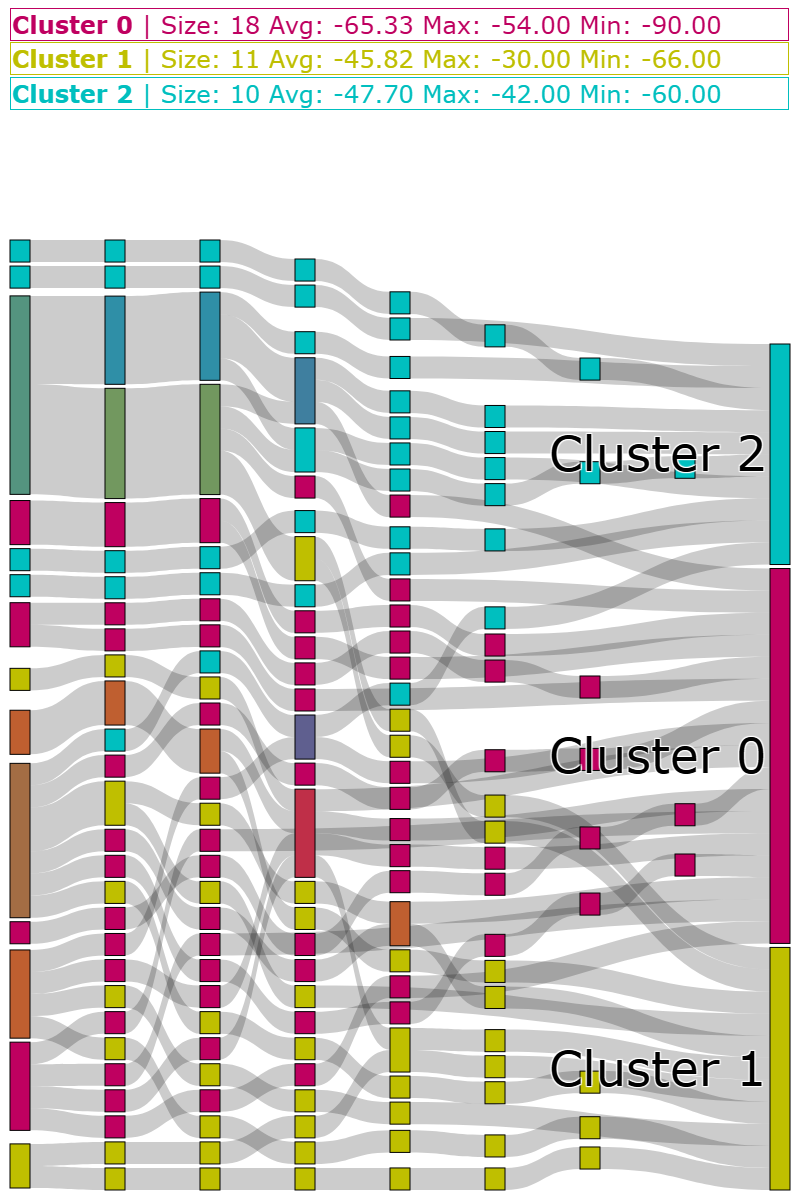}
    \end{minipage}
    \\
    \begin{minipage}{0.333\linewidth}
        \centering
        \includegraphics[width=\linewidth]{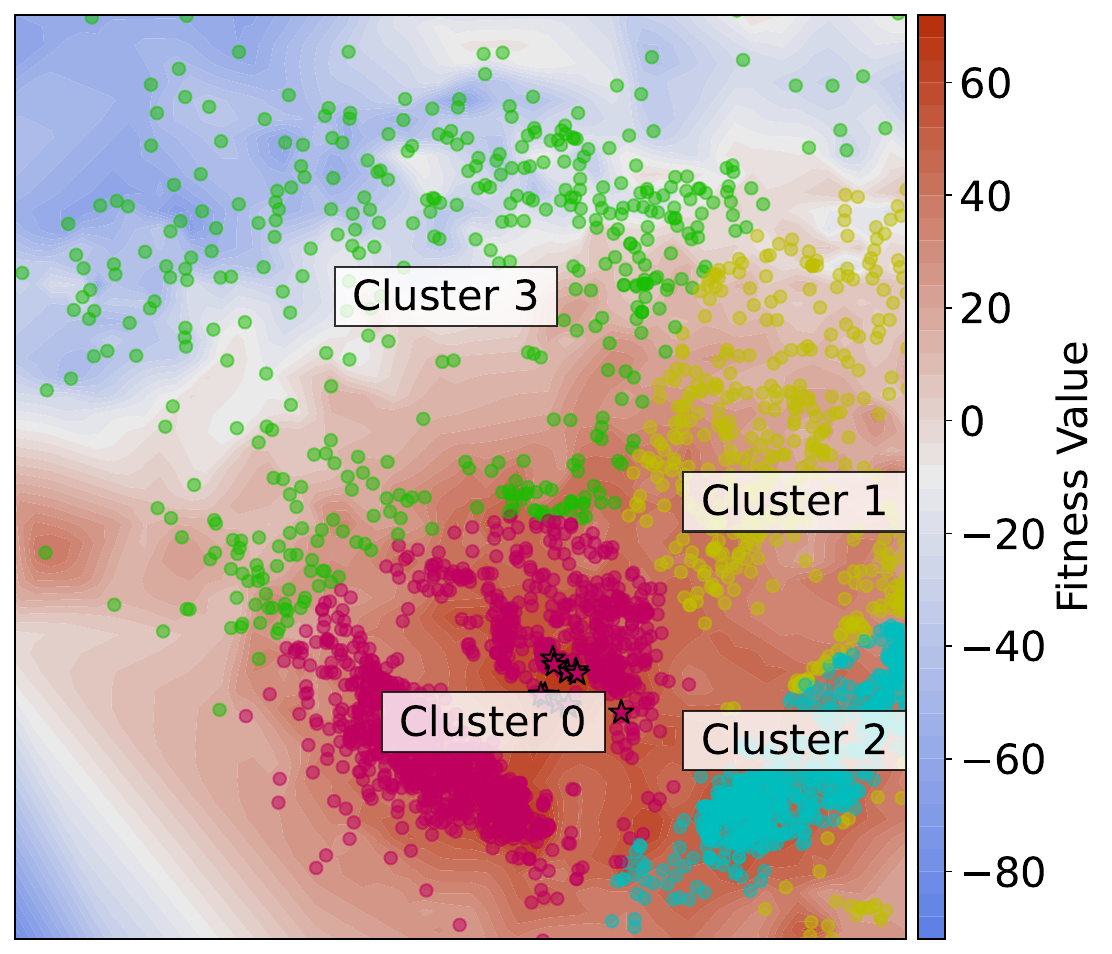}
    \end{minipage}
    \begin{minipage}{0.19\linewidth}
        \centering
        \includegraphics[width=\linewidth]{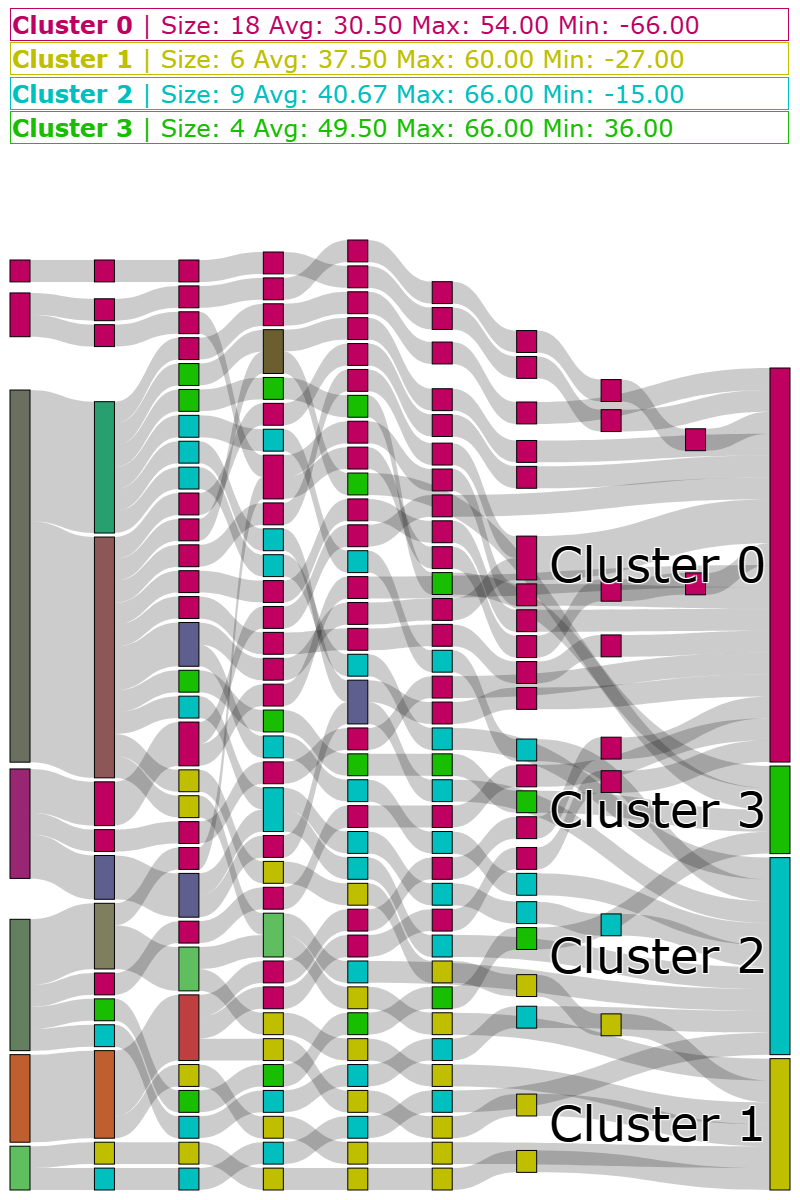}
    \end{minipage}
    \begin{minipage}{0.19\linewidth}
        \centering
        \includegraphics[width=\linewidth]{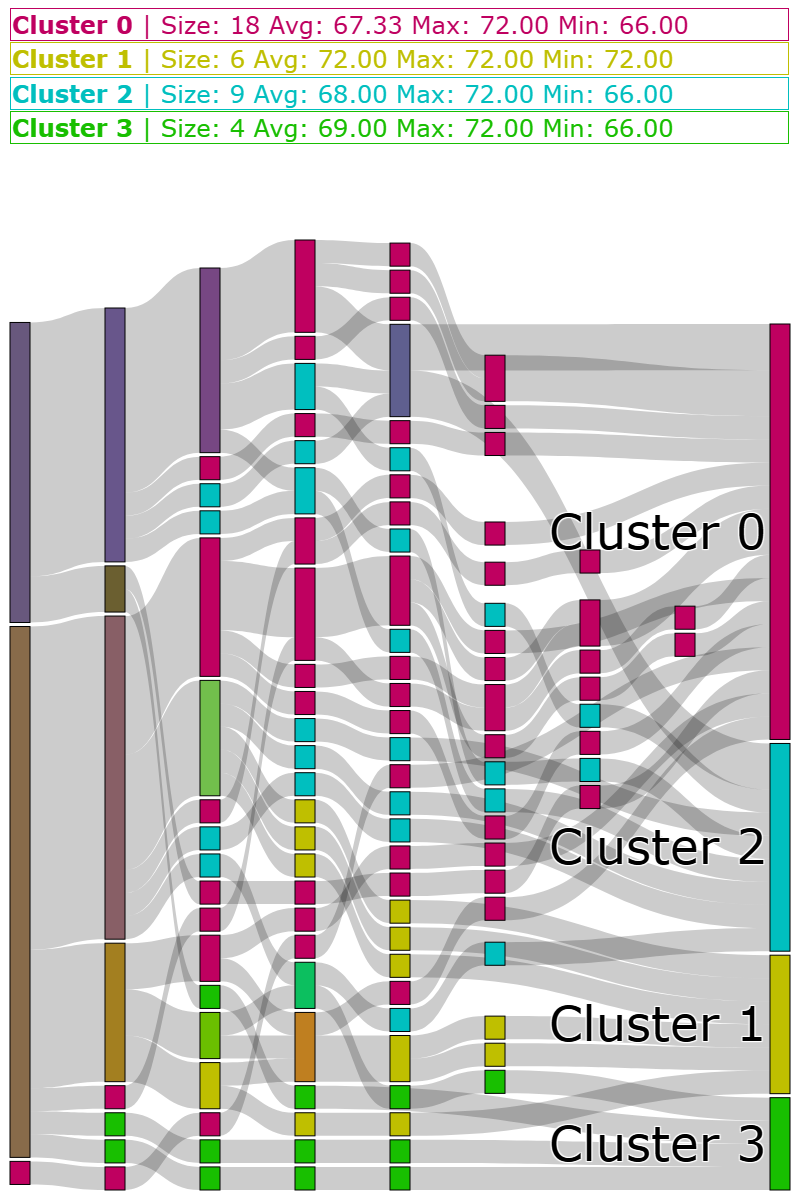}
    \end{minipage}
    \begin{minipage}{0.19\linewidth}
        \centering
        \includegraphics[width=\linewidth]{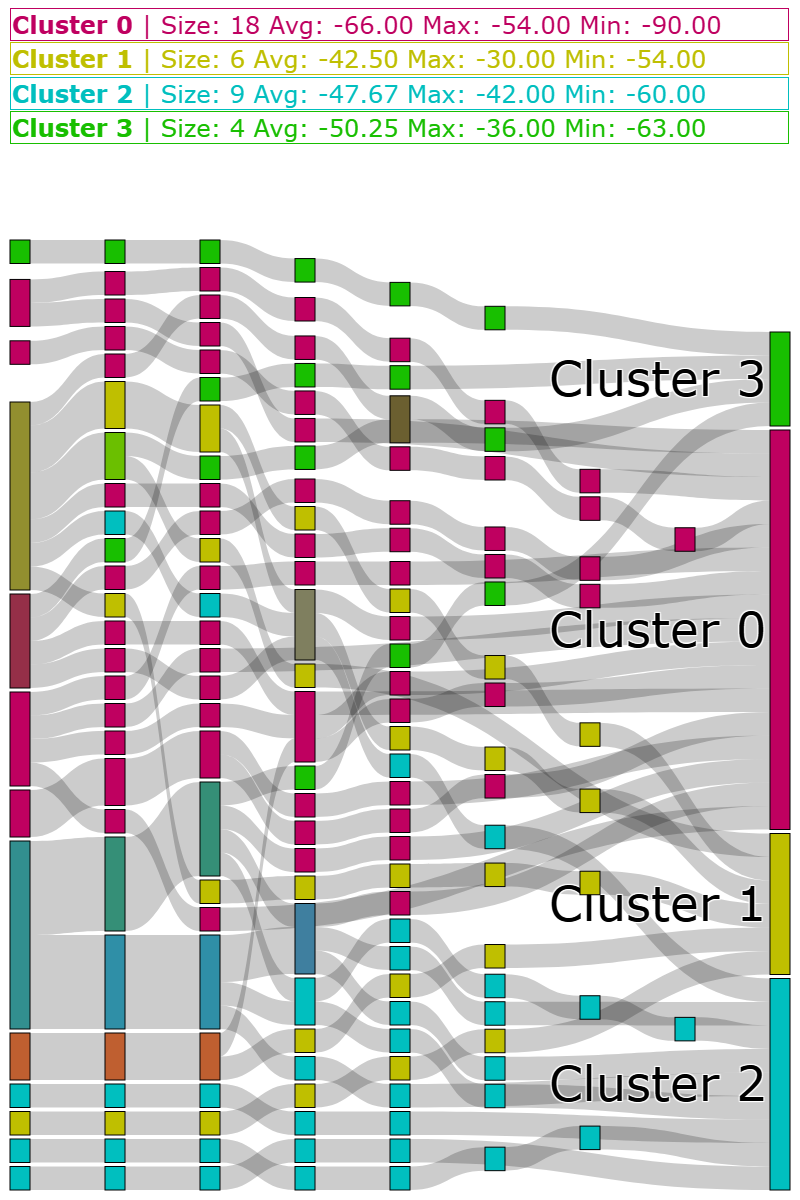}
    \end{minipage}
    \\
    \begin{minipage}{0.333\linewidth}
        \centering
        \includegraphics[width=\linewidth]{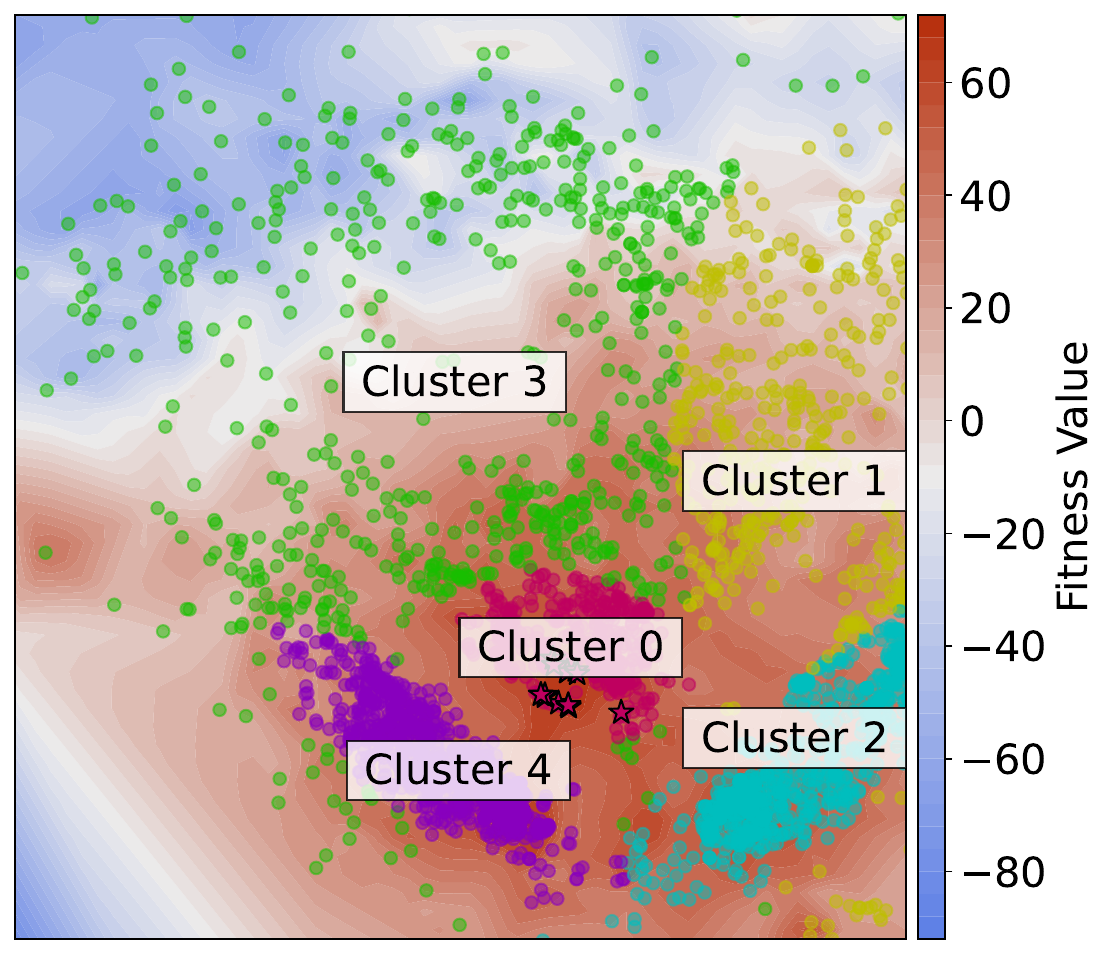}
    \end{minipage}
    \begin{minipage}{0.19\linewidth}
        \centering
        \includegraphics[width=\linewidth]{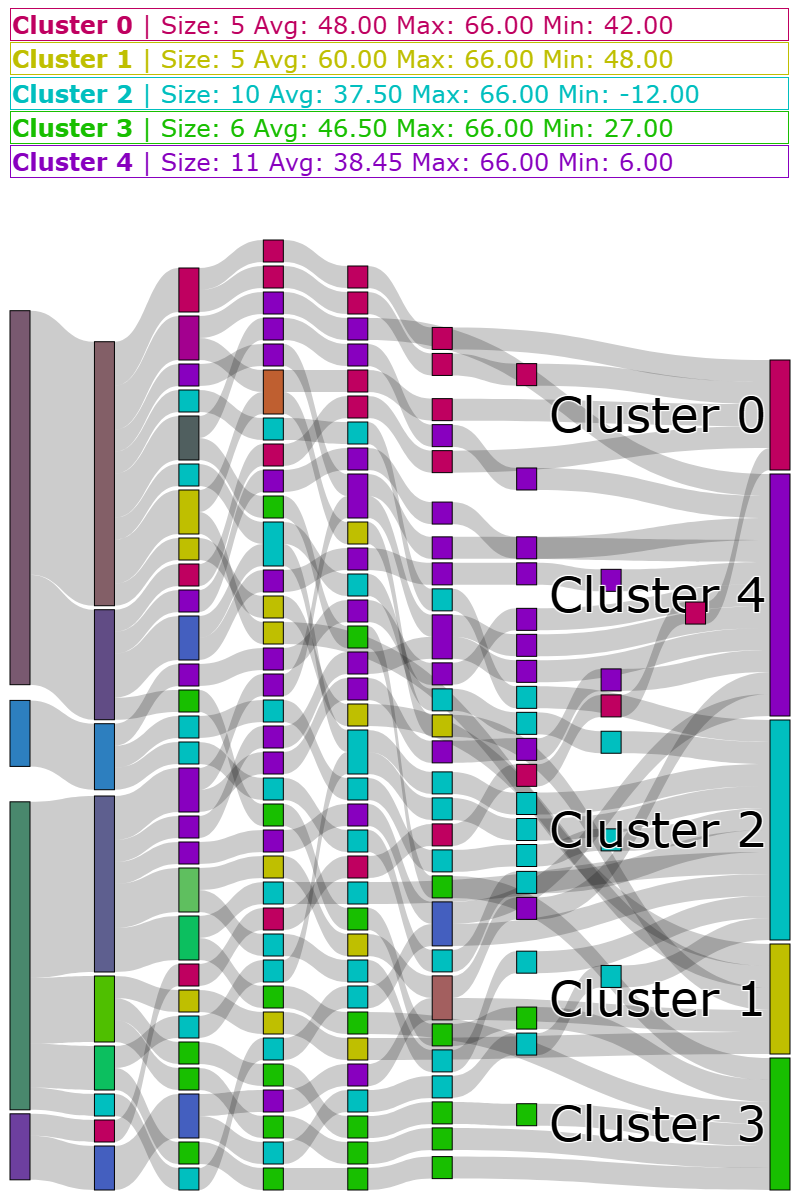}
    \end{minipage}
    \begin{minipage}{0.19\linewidth}
        \centering
        \includegraphics[width=\linewidth]{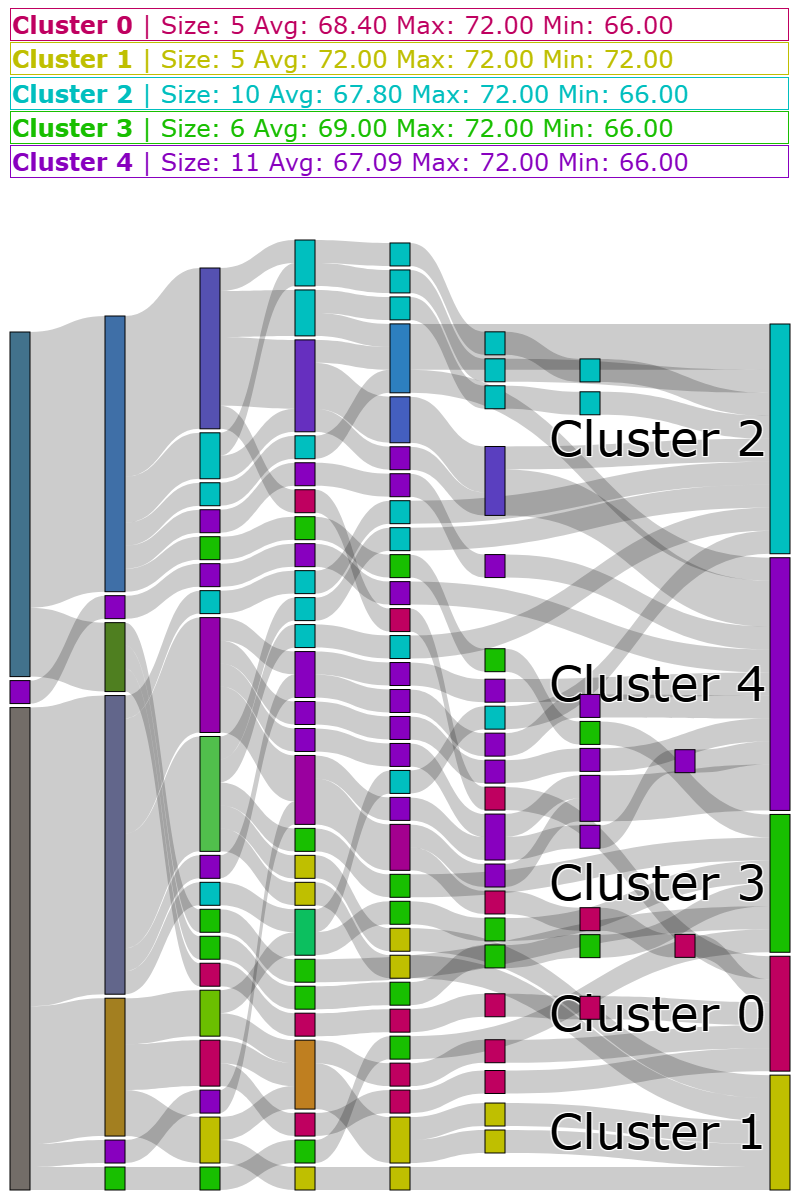}
    \end{minipage}
    \begin{minipage}{0.19\linewidth}
        \centering
        \includegraphics[width=\linewidth]{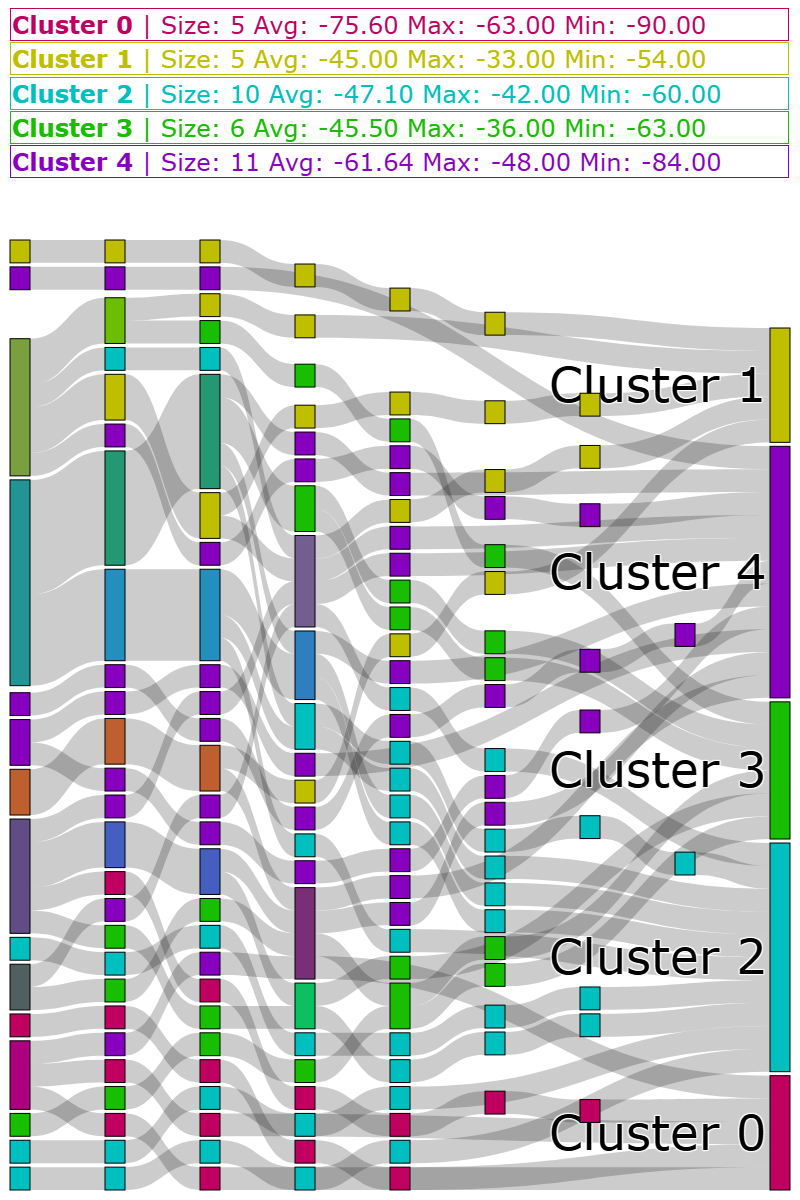}
    \end{minipage}
    \\
    \begin{minipage}{0.333\linewidth}
        \centering
        \includegraphics[width=\linewidth]{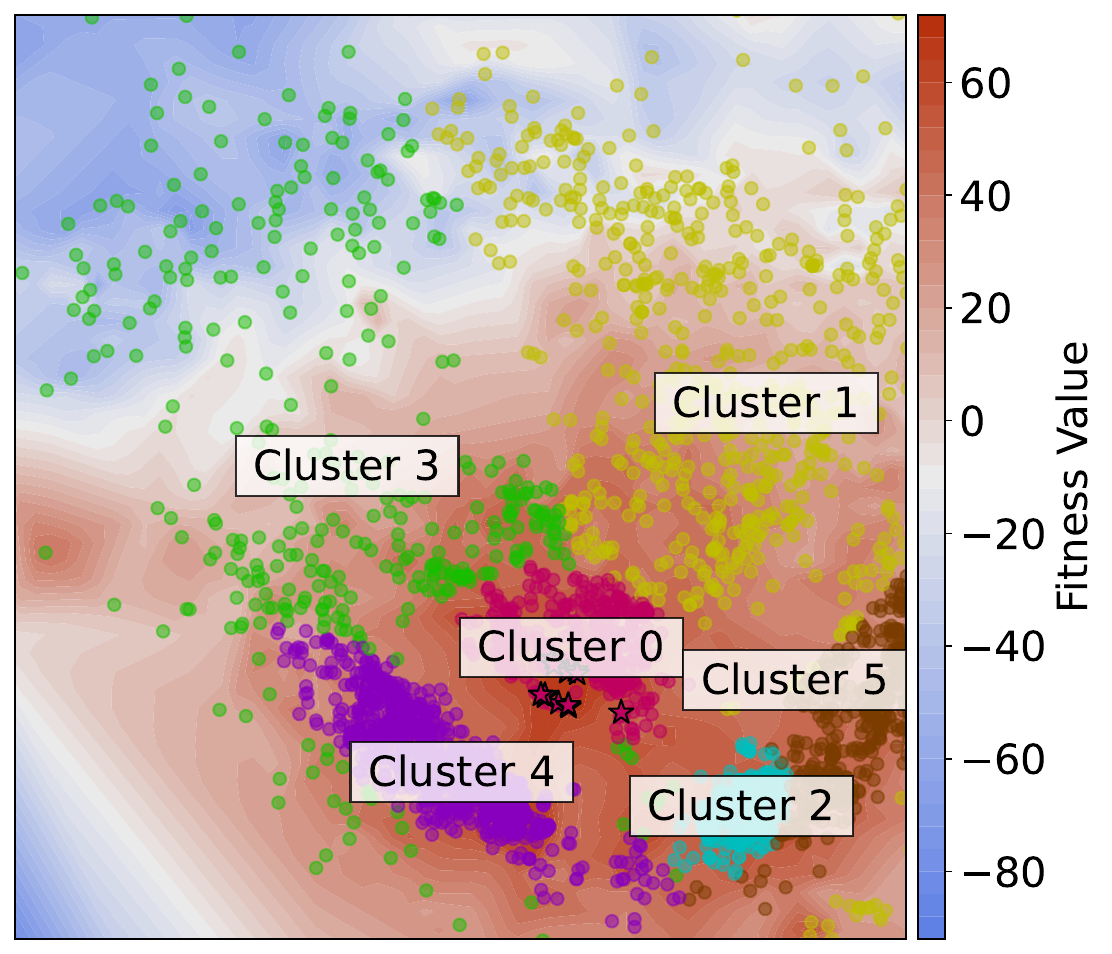}
        fitness landscape
    \end{minipage}
    \begin{minipage}{0.19\linewidth}
        \centering
        \includegraphics[width=\linewidth]{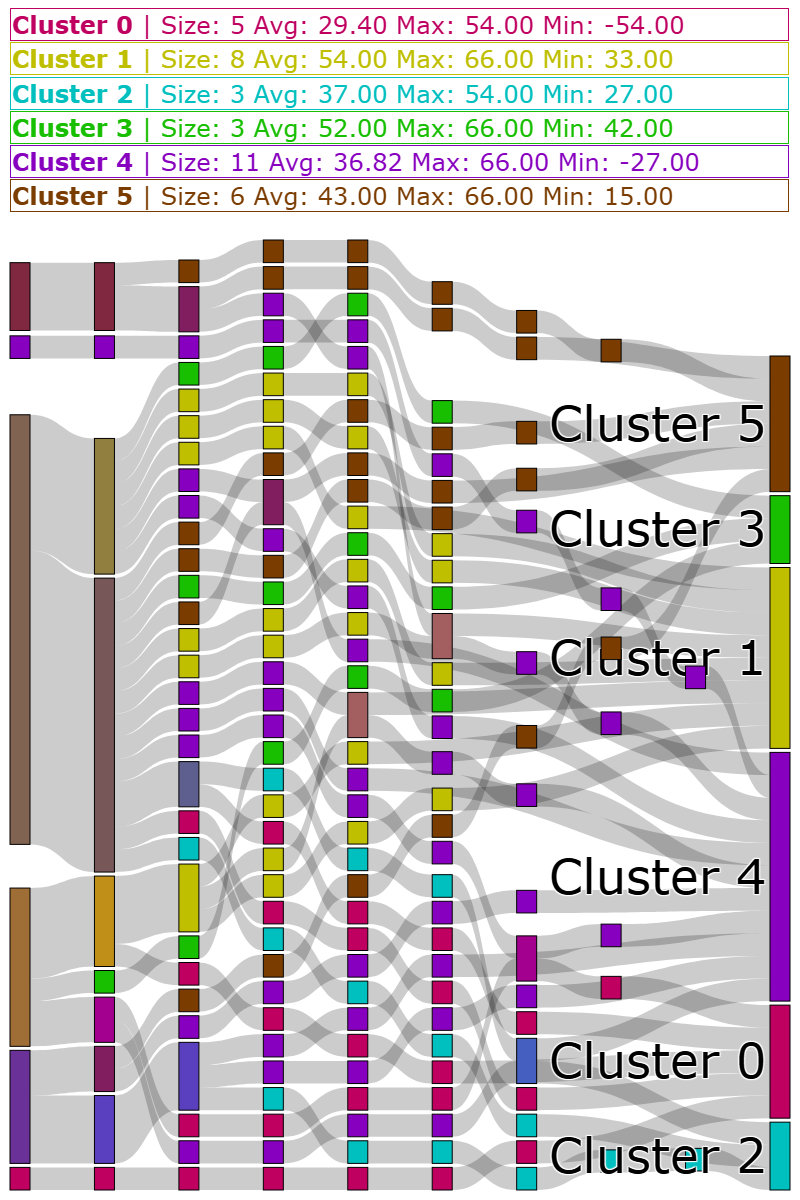}
        unsorted
    \end{minipage}
    \begin{minipage}{0.19\linewidth}
        \centering
        \includegraphics[width=\linewidth]{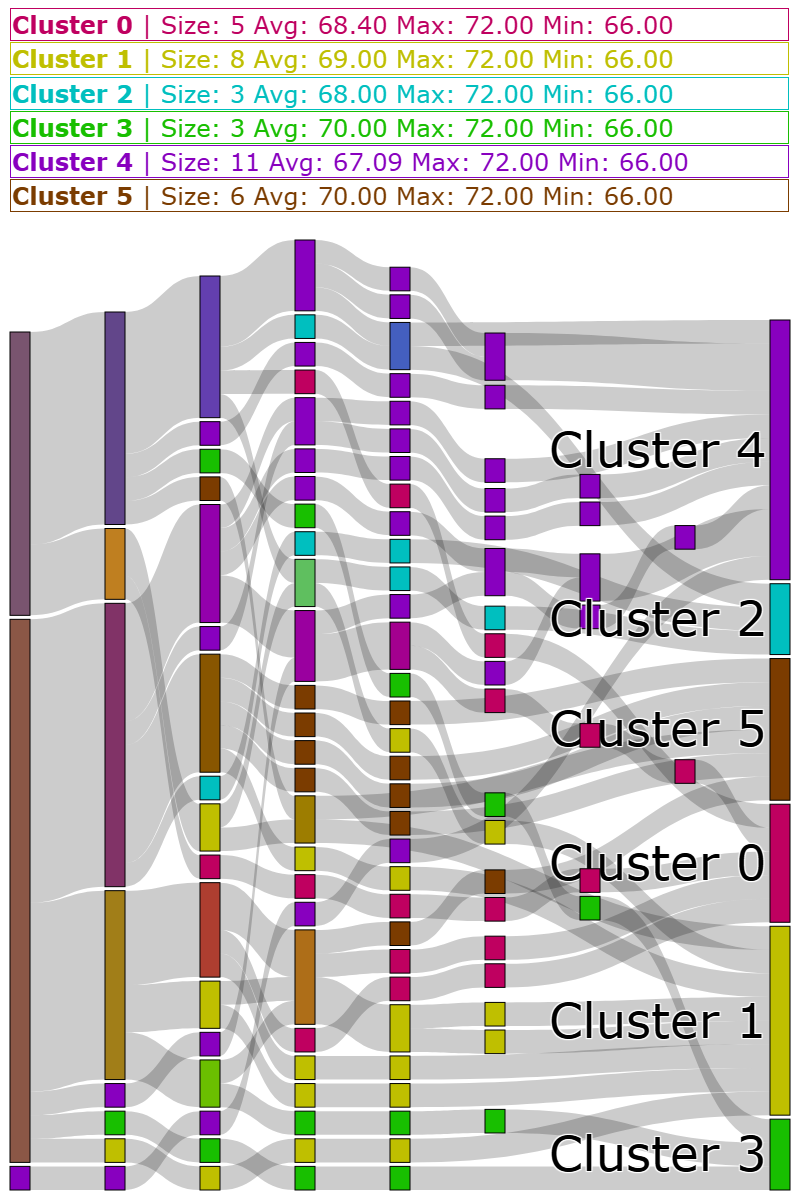}
        win
    \end{minipage}
    \begin{minipage}{0.19\linewidth}
        \centering
        \includegraphics[width=\linewidth]{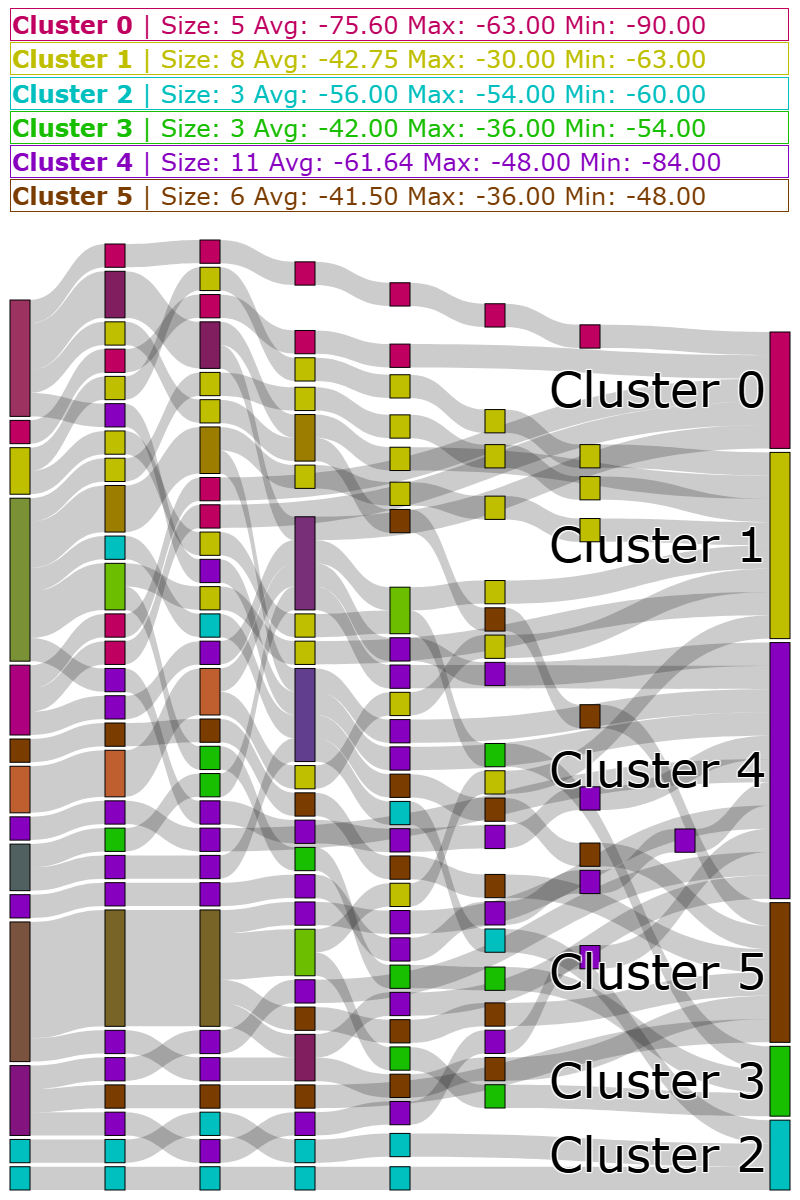}
        loss
    \end{minipage}
    \\
    \caption{Comprehensive sankey diagram of action sequences patterns with gmm algorithm for clustering}
    \label{fig:comprehensive_sankey_diagram_gmm}
\end{figure*}

\begin{figure*}[!htp]
    \centering
    \begin{minipage}{0.333\linewidth}
        \centering
        \includegraphics[width=\linewidth]{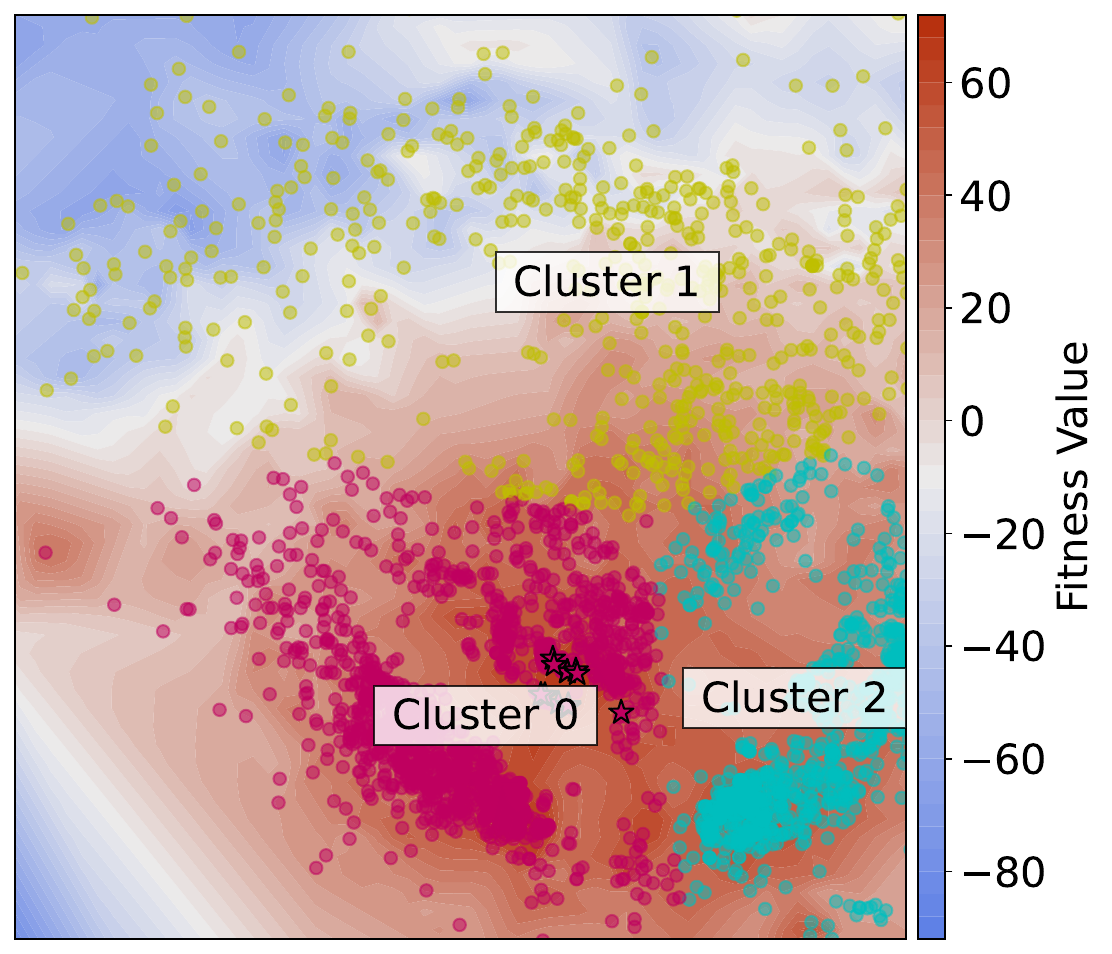}
    \end{minipage}
    \begin{minipage}{0.19\linewidth}
        \centering
        \includegraphics[width=\linewidth]{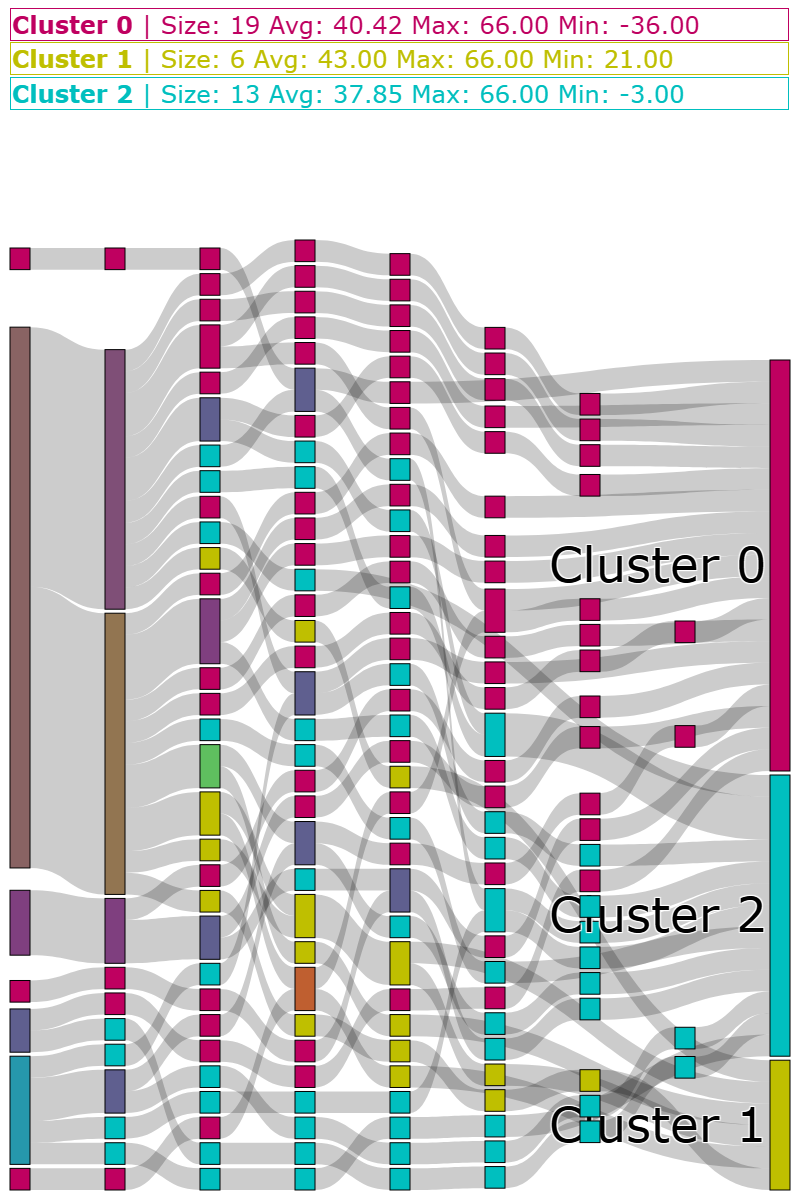}
    \end{minipage}
    \begin{minipage}{0.19\linewidth}
        \centering
        \includegraphics[width=\linewidth]{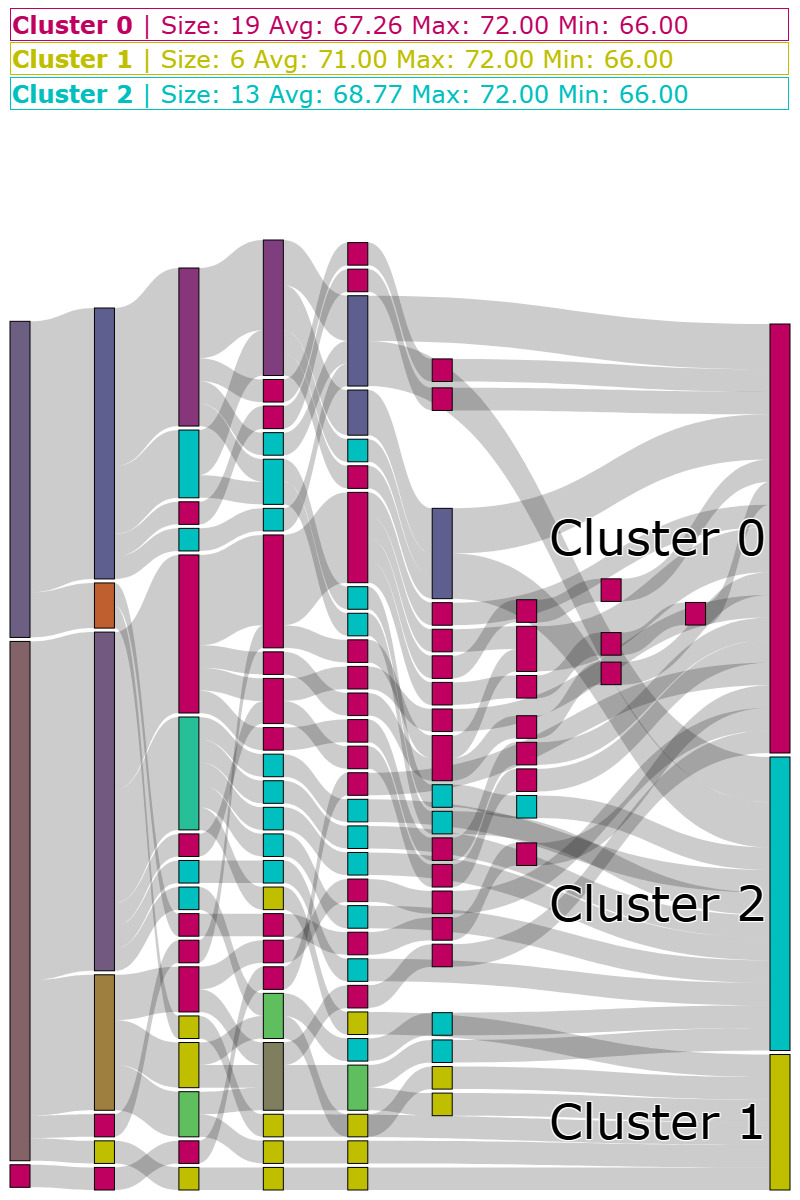}
    \end{minipage}
    \begin{minipage}{0.19\linewidth}
        \centering
        \includegraphics[width=\linewidth]{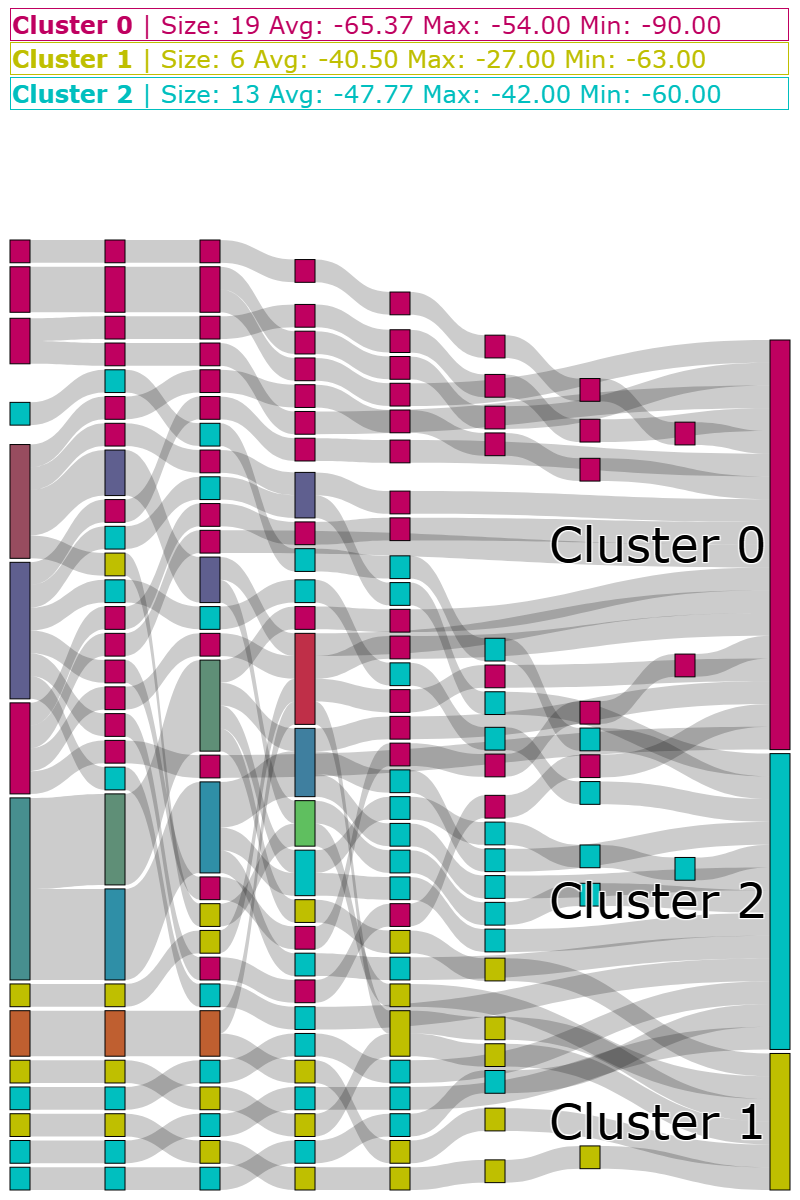}
    \end{minipage}
    \\
    \begin{minipage}{0.333\linewidth}
        \centering
        \includegraphics[width=\linewidth]{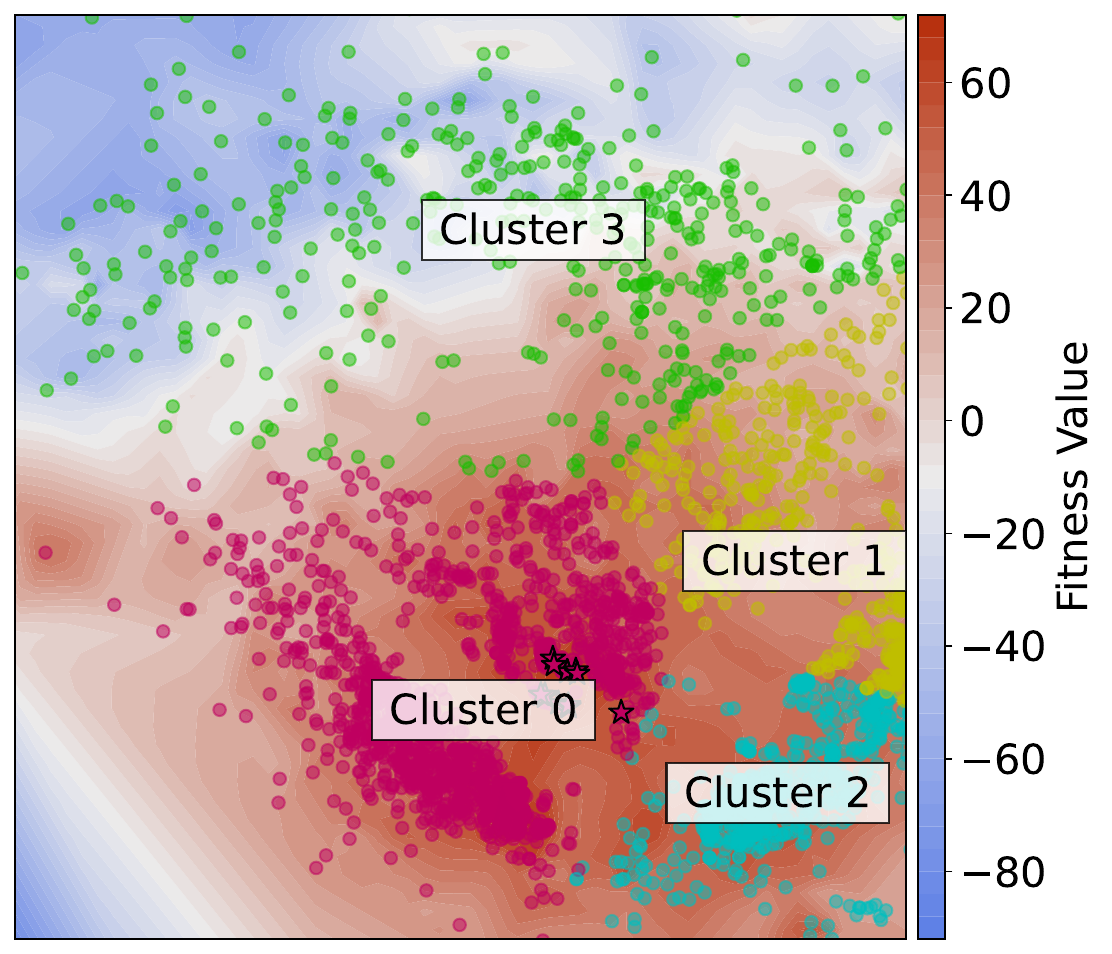}
    \end{minipage}
    \begin{minipage}{0.19\linewidth}
        \centering
        \includegraphics[width=\linewidth]{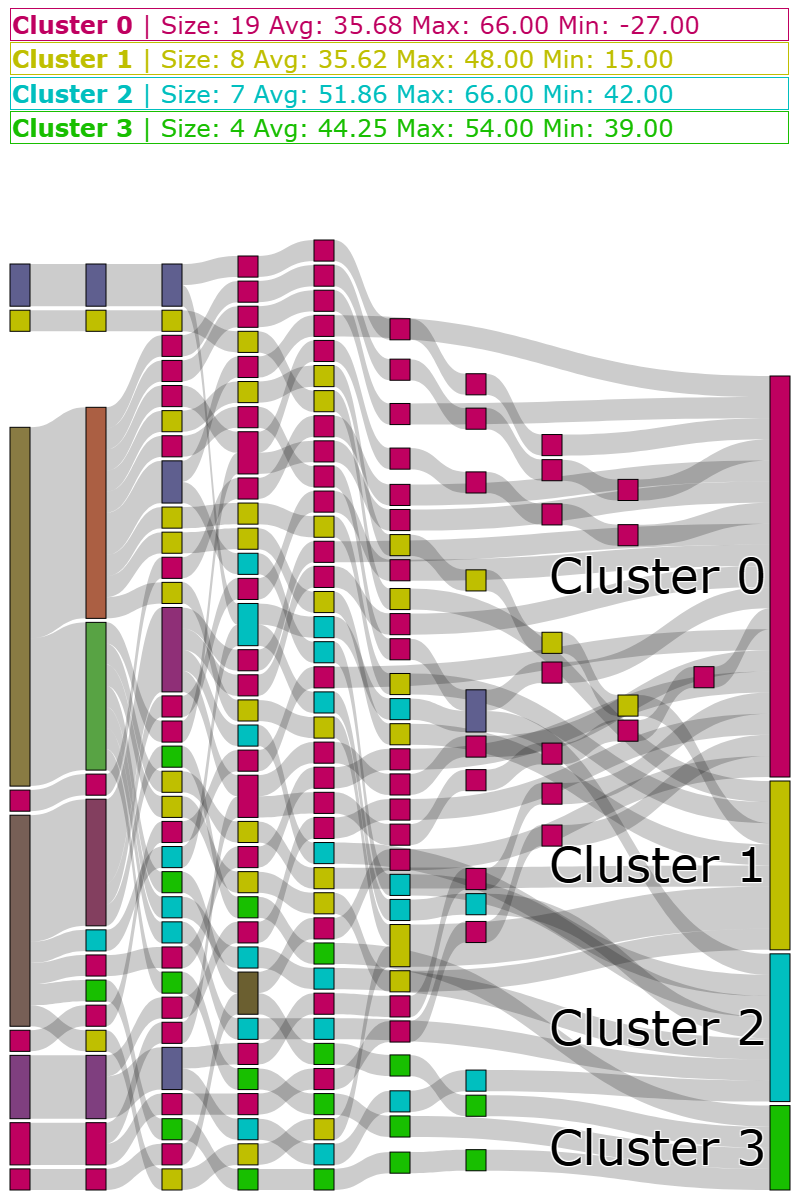}
    \end{minipage}
    \begin{minipage}{0.19\linewidth}
        \centering
        \includegraphics[width=\linewidth]{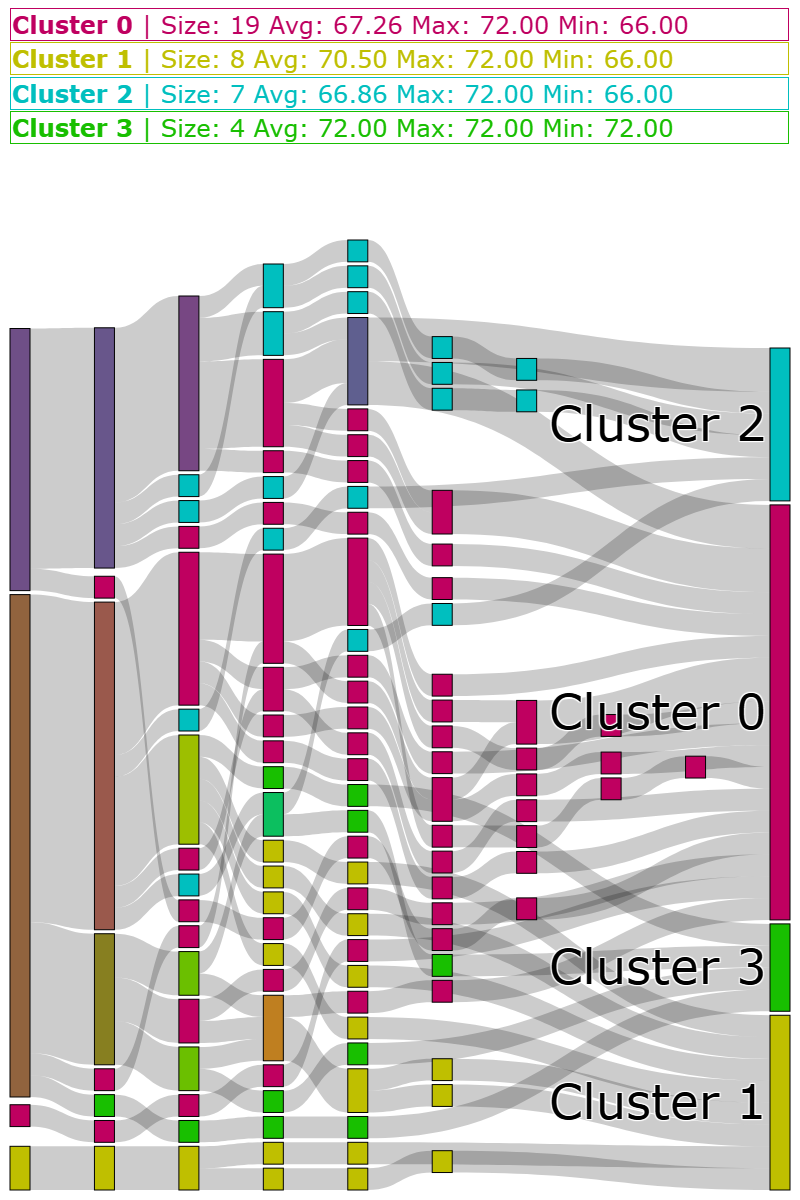}
    \end{minipage}
    \begin{minipage}{0.19\linewidth}
        \centering
        \includegraphics[width=\linewidth]{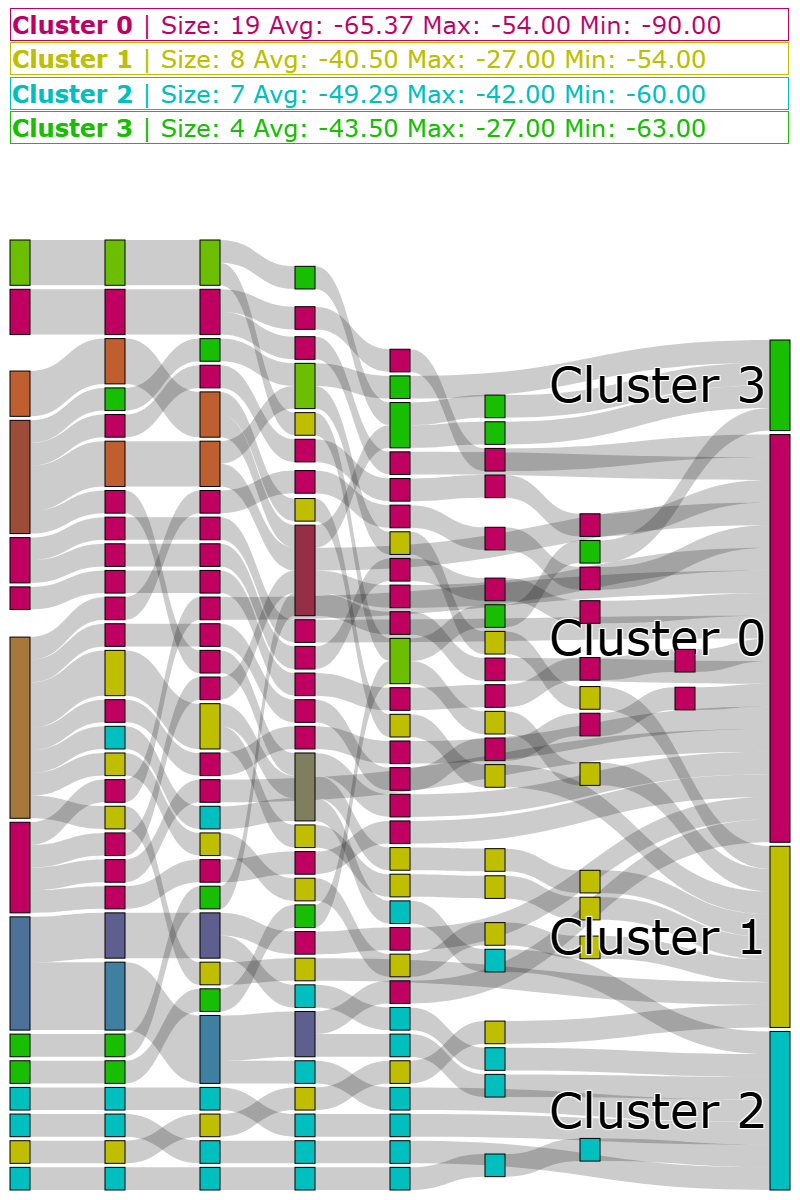}
    \end{minipage}
    \\
    \begin{minipage}{0.333\linewidth}
        \centering
        \includegraphics[width=\linewidth]{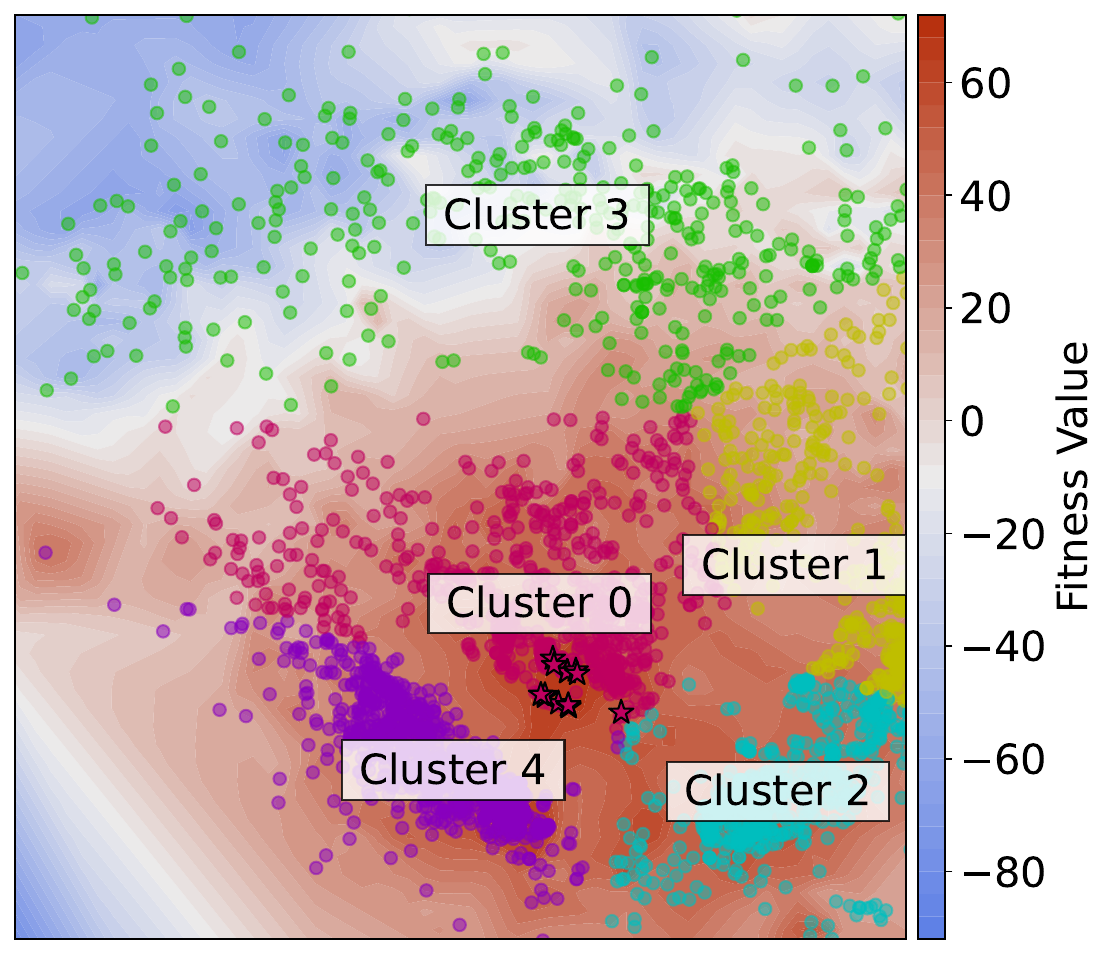}
    \end{minipage}
    \begin{minipage}{0.19\linewidth}
        \centering
        \includegraphics[width=\linewidth]{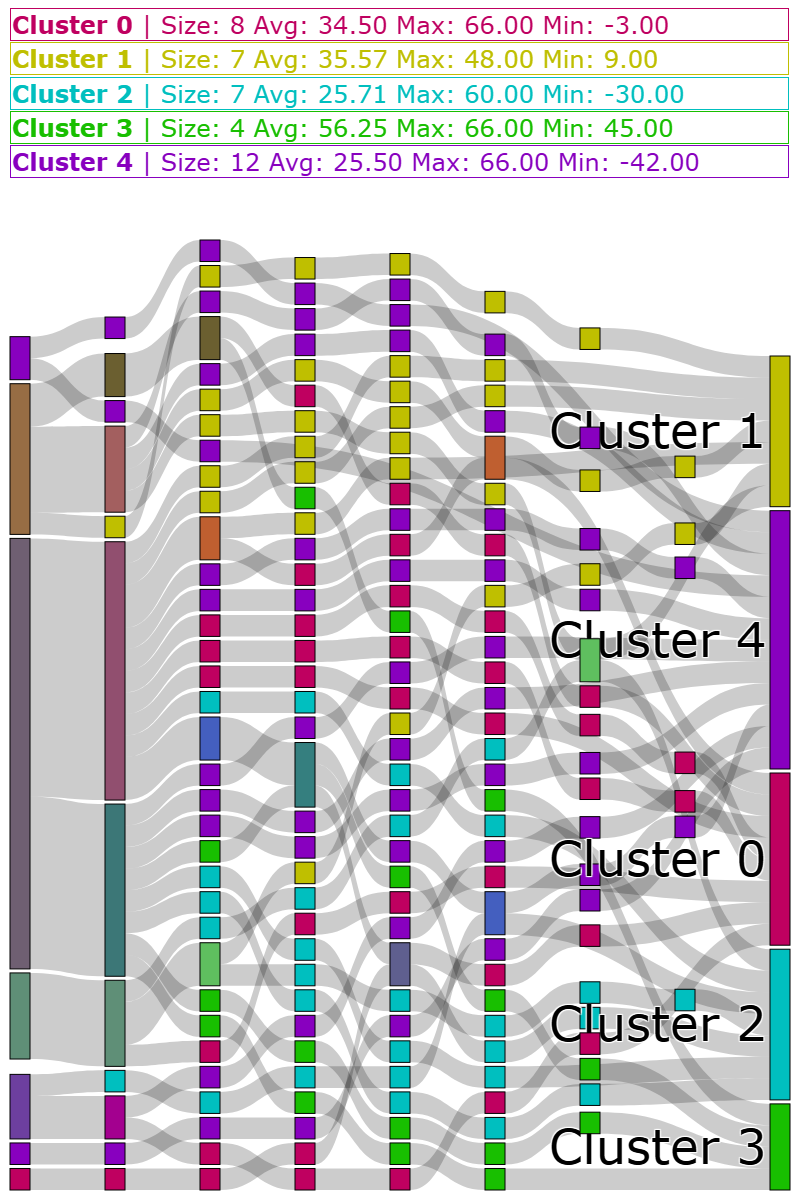}
    \end{minipage}
    \begin{minipage}{0.19\linewidth}
        \centering
        \includegraphics[width=\linewidth]{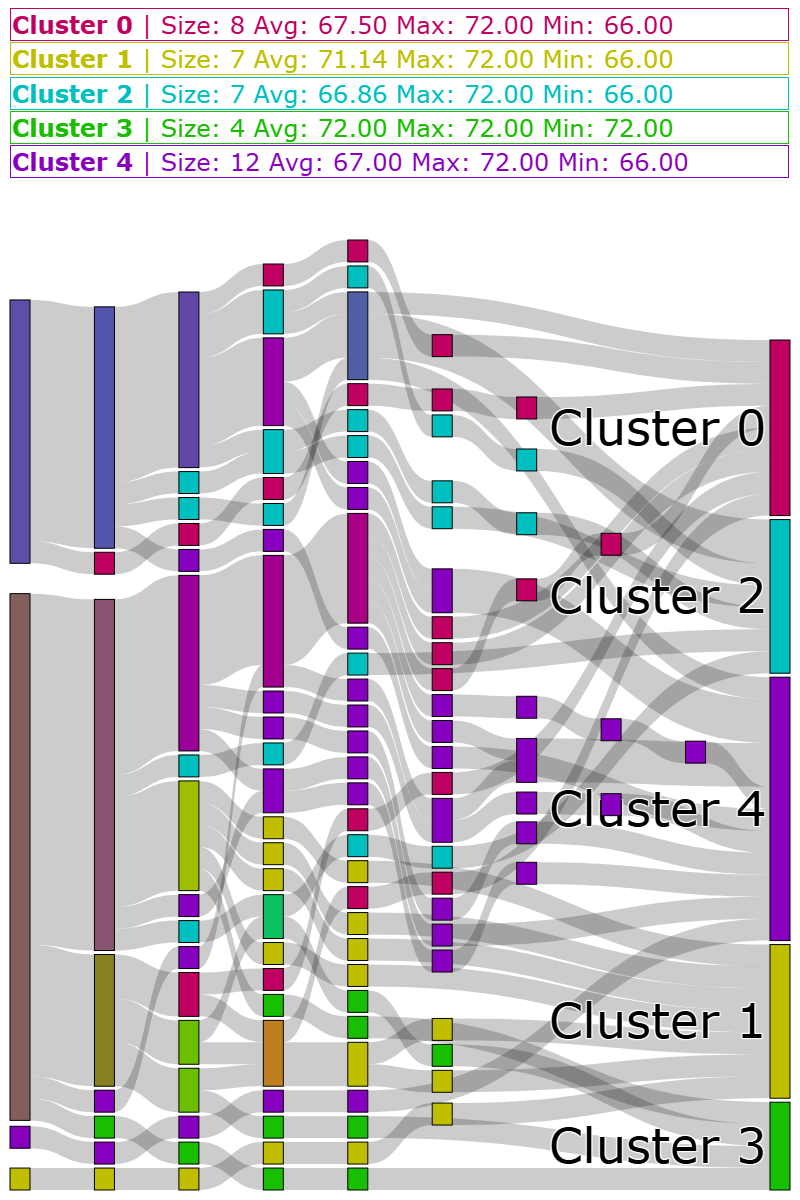}
    \end{minipage}
    \begin{minipage}{0.19\linewidth}
        \centering
        \includegraphics[width=\linewidth]{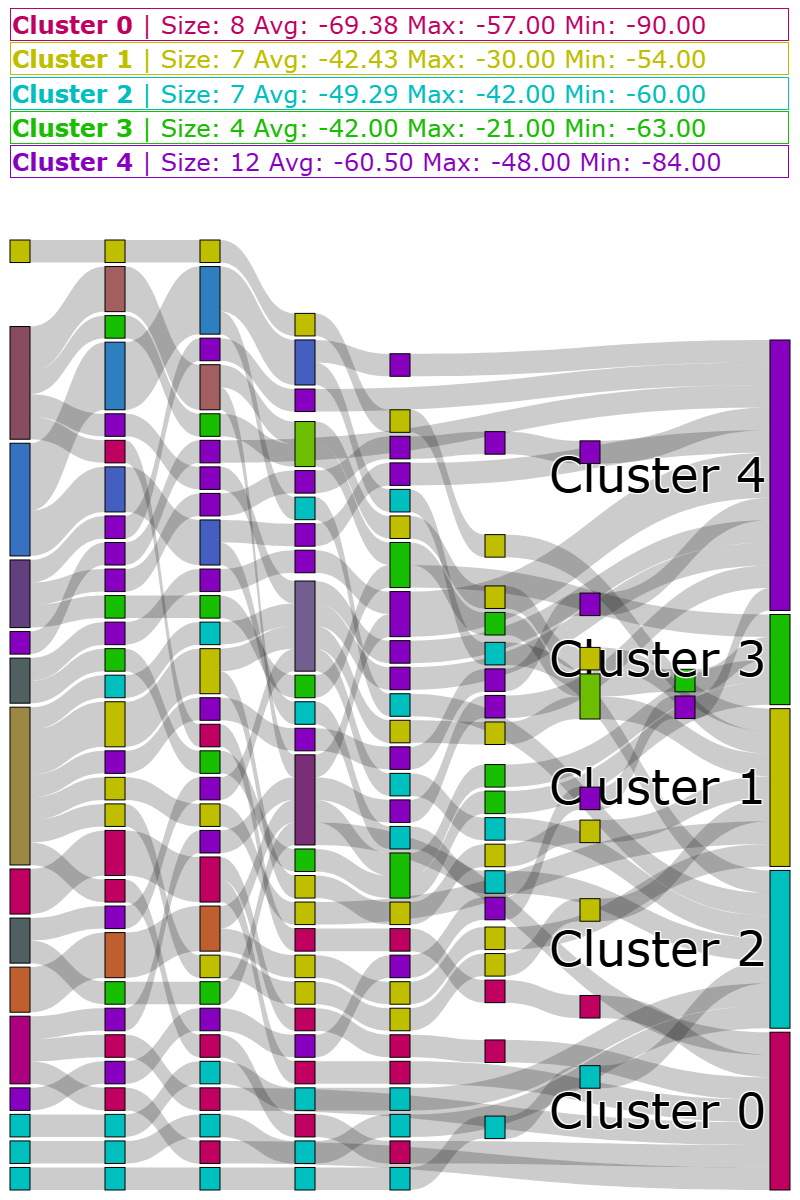}
    \end{minipage}
    \\
    \begin{minipage}{0.333\linewidth}
        \centering
        \includegraphics[width=\linewidth]{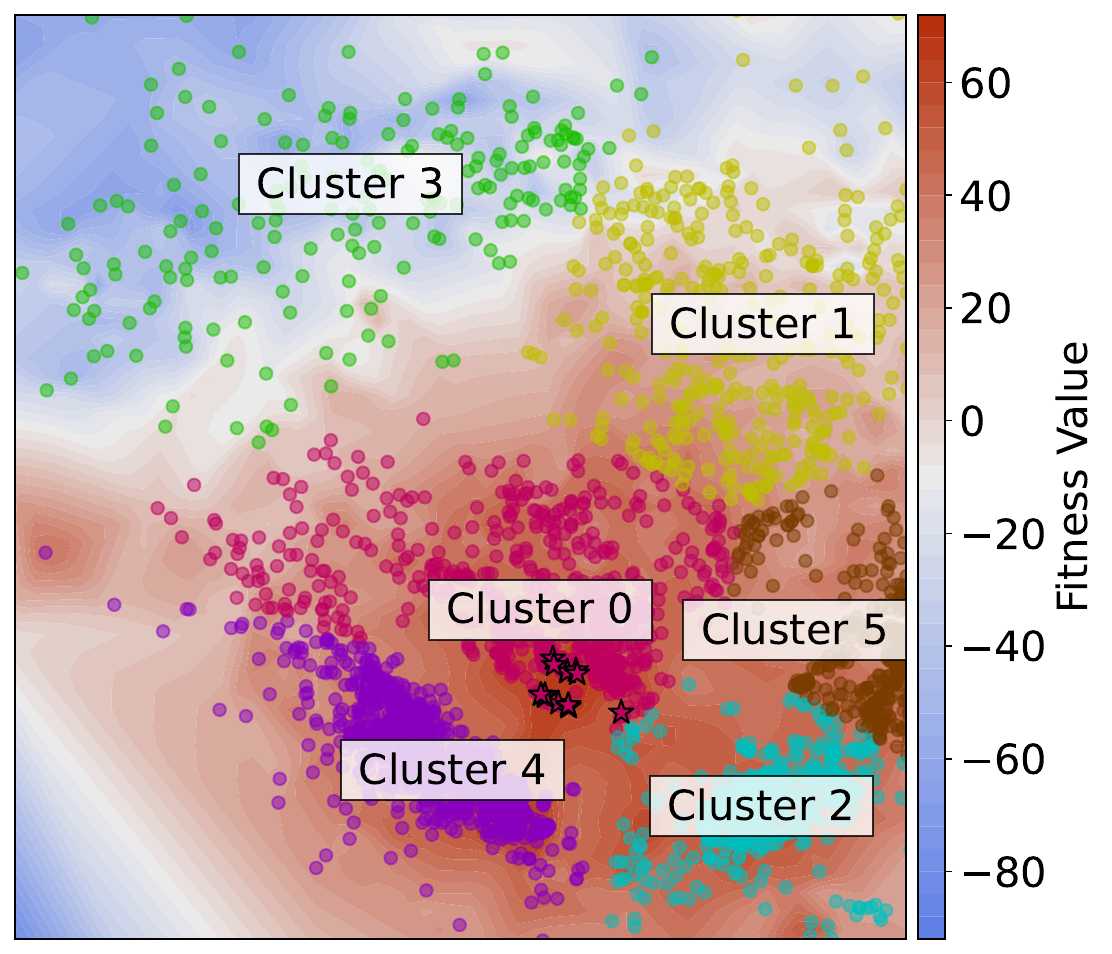}
        fitness landscape
    \end{minipage}
    \begin{minipage}{0.19\linewidth}
        \centering
        \includegraphics[width=\linewidth]{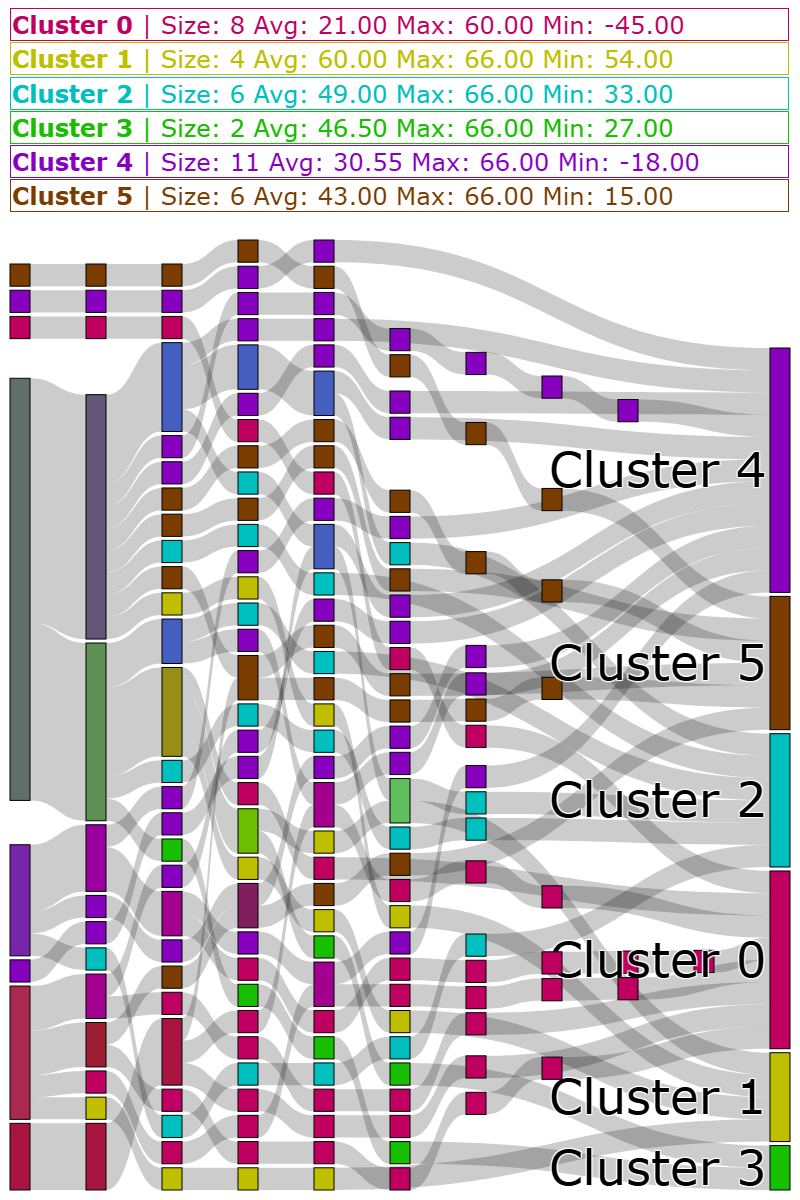}
        unsorted
    \end{minipage}
    \begin{minipage}{0.19\linewidth}
        \centering
        \includegraphics[width=\linewidth]{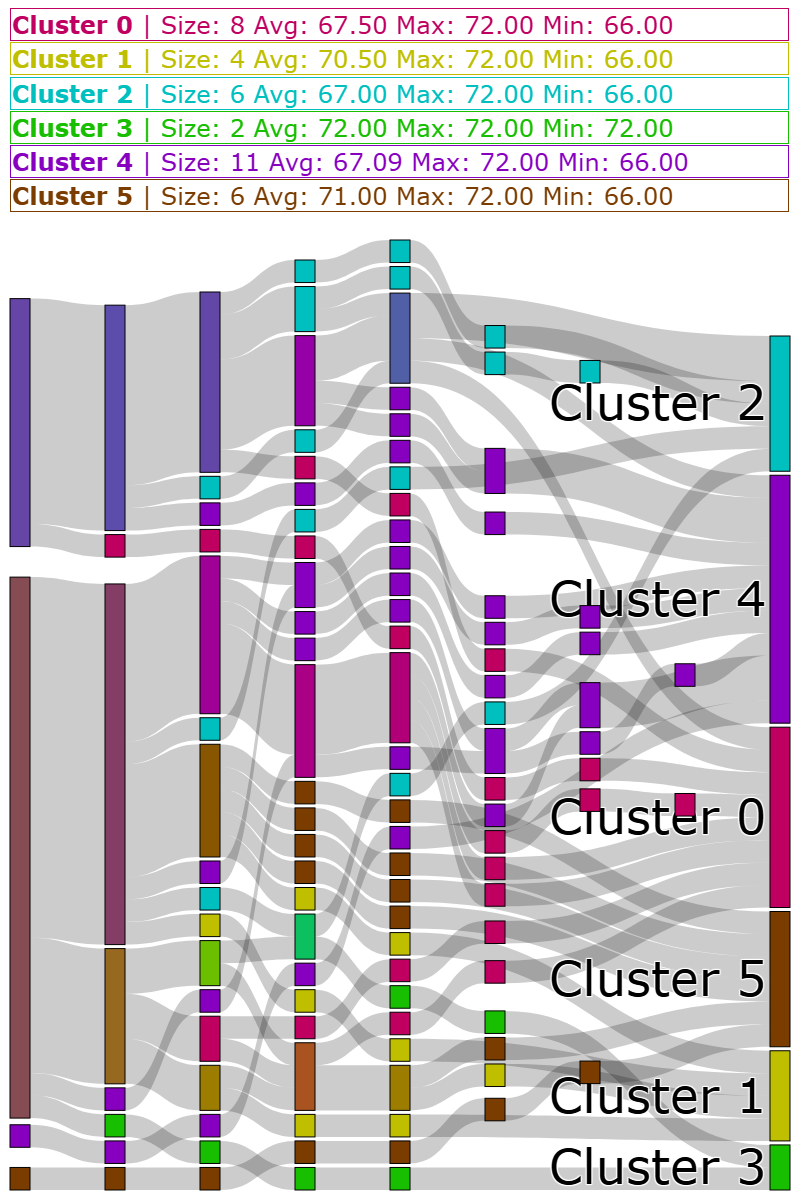}
        win
    \end{minipage}
    \begin{minipage}{0.19\linewidth}
        \centering
        \includegraphics[width=\linewidth]{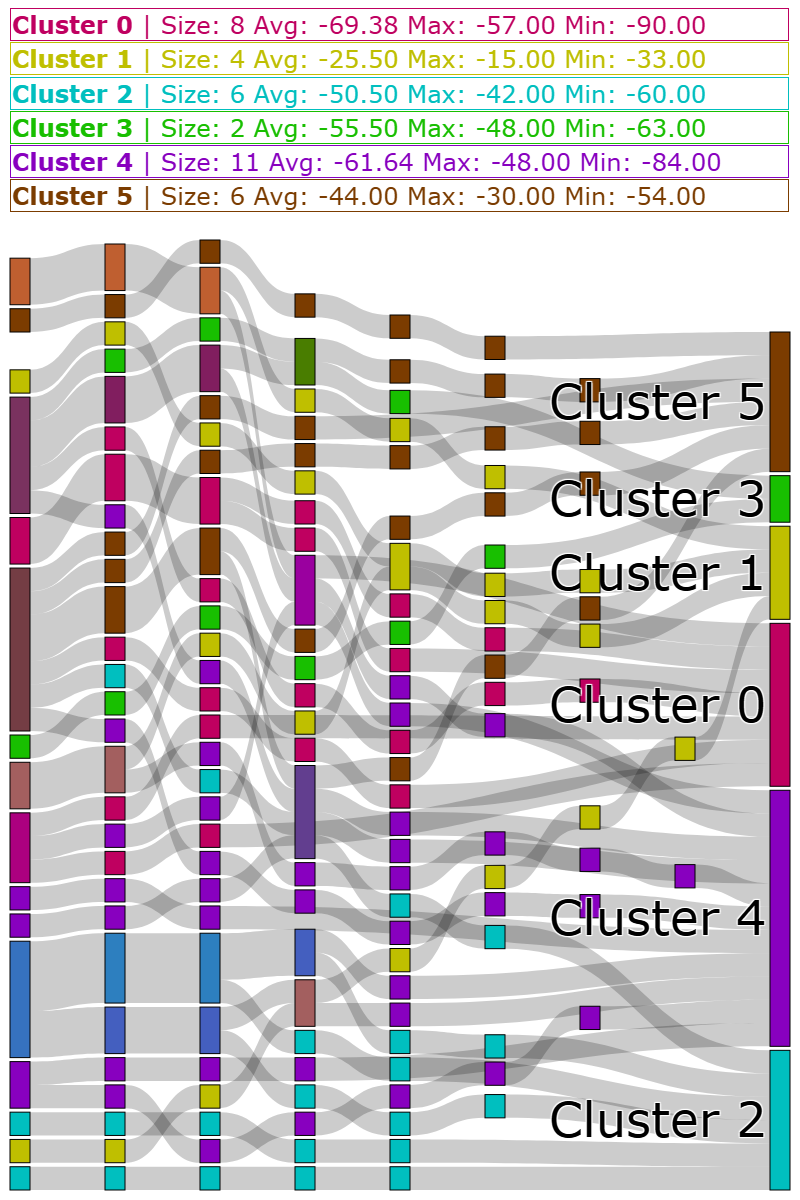}
        loss
    \end{minipage}
    \\
    \caption{Comprehensive sankey diagram of action sequences patterns with $k$-means algorithm for clustering}
    \label{fig:comprehensive_sankey_diagram_kmeans}
\end{figure*}

\begin{figure*}[!htp]
    \centering
    \begin{minipage}{0.333\linewidth}
        \centering
        \includegraphics[width=\linewidth]{fitness_landscape_agglomerative_k3.pdf}
    \end{minipage}
    \begin{minipage}{0.19\linewidth}
        \centering
        \includegraphics[width=\linewidth]{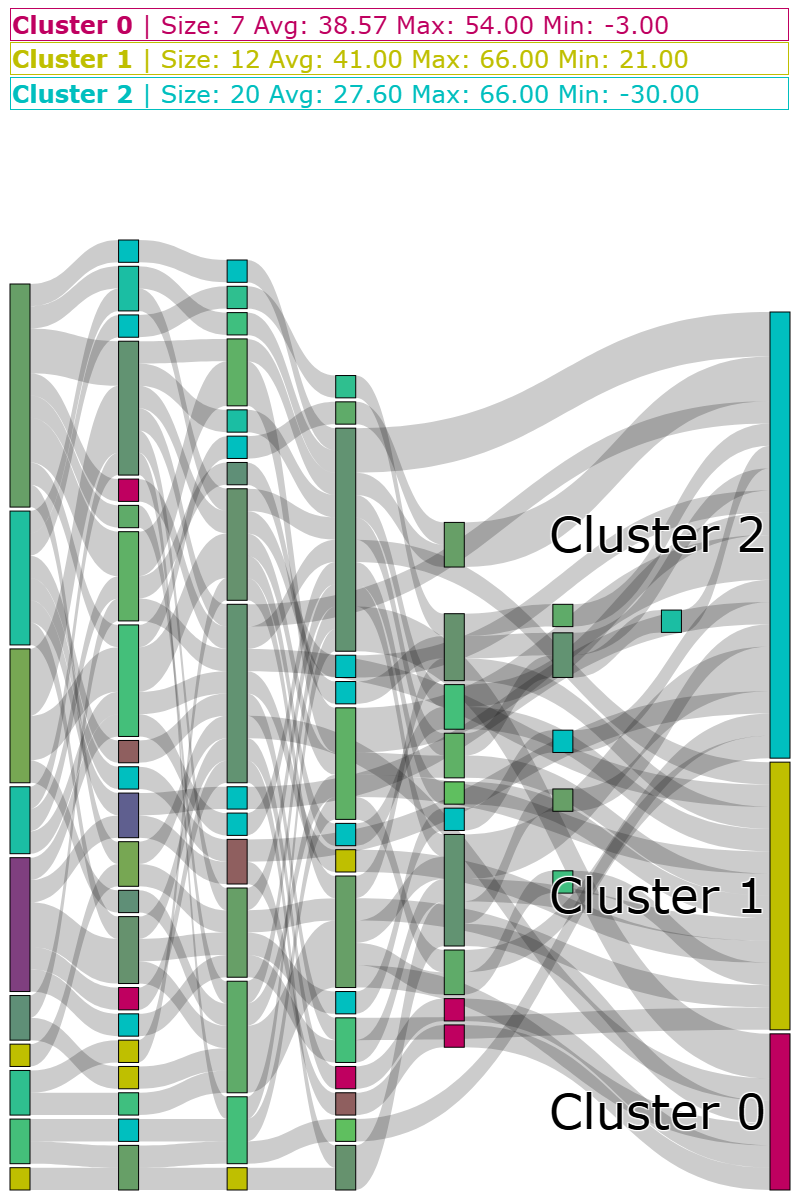}
    \end{minipage}
    \begin{minipage}{0.19\linewidth}
        \centering
        \includegraphics[width=\linewidth]{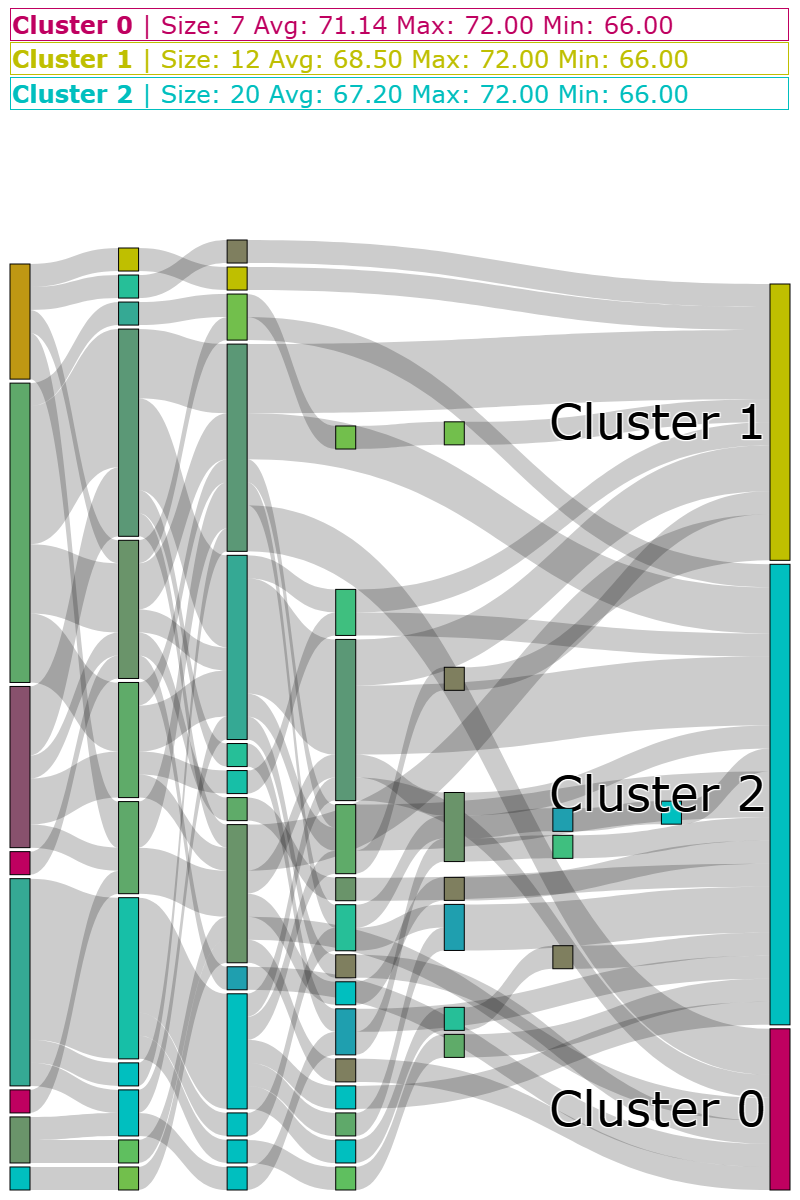}
    \end{minipage}
    \begin{minipage}{0.19\linewidth}
        \centering
        \includegraphics[width=\linewidth]{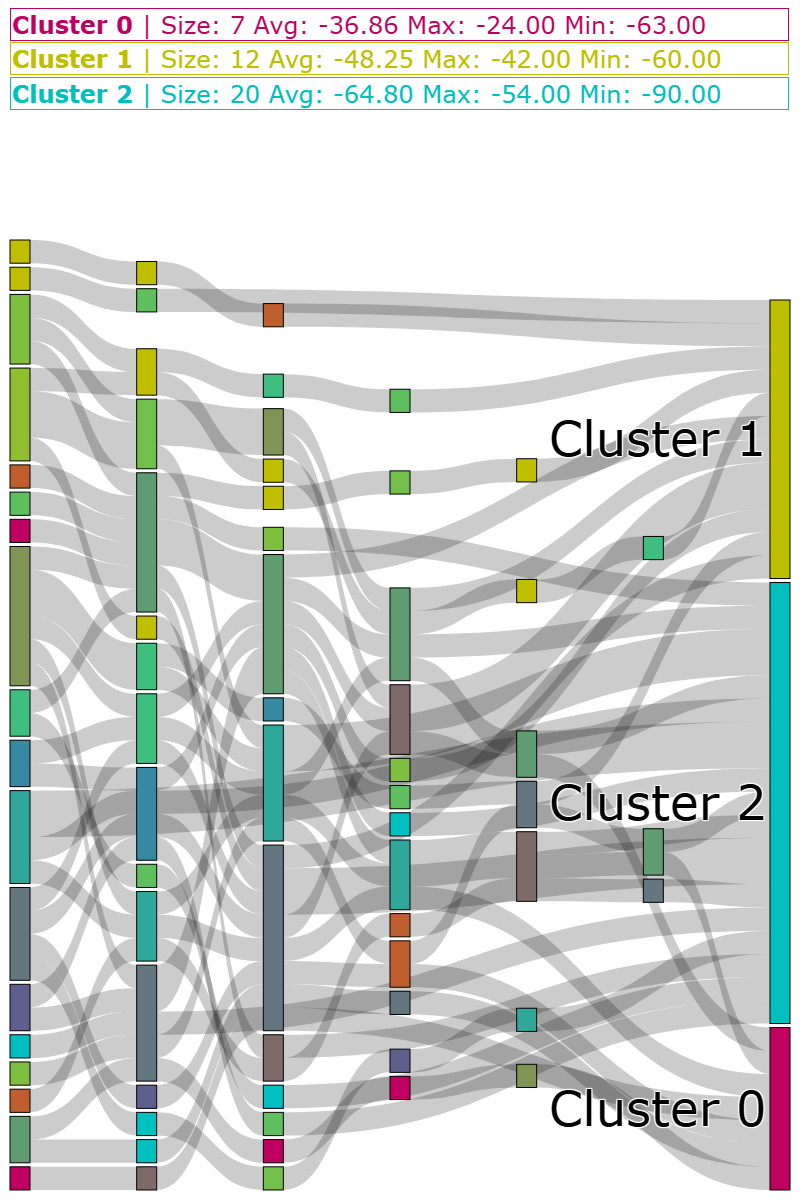}
    \end{minipage}
    \\
    \begin{minipage}{0.333\linewidth}
        \centering
        \includegraphics[width=\linewidth]{fitness_landscape_agglomerative_k4.pdf}
    \end{minipage}
    \begin{minipage}{0.19\linewidth}
        \centering
        \includegraphics[width=\linewidth]{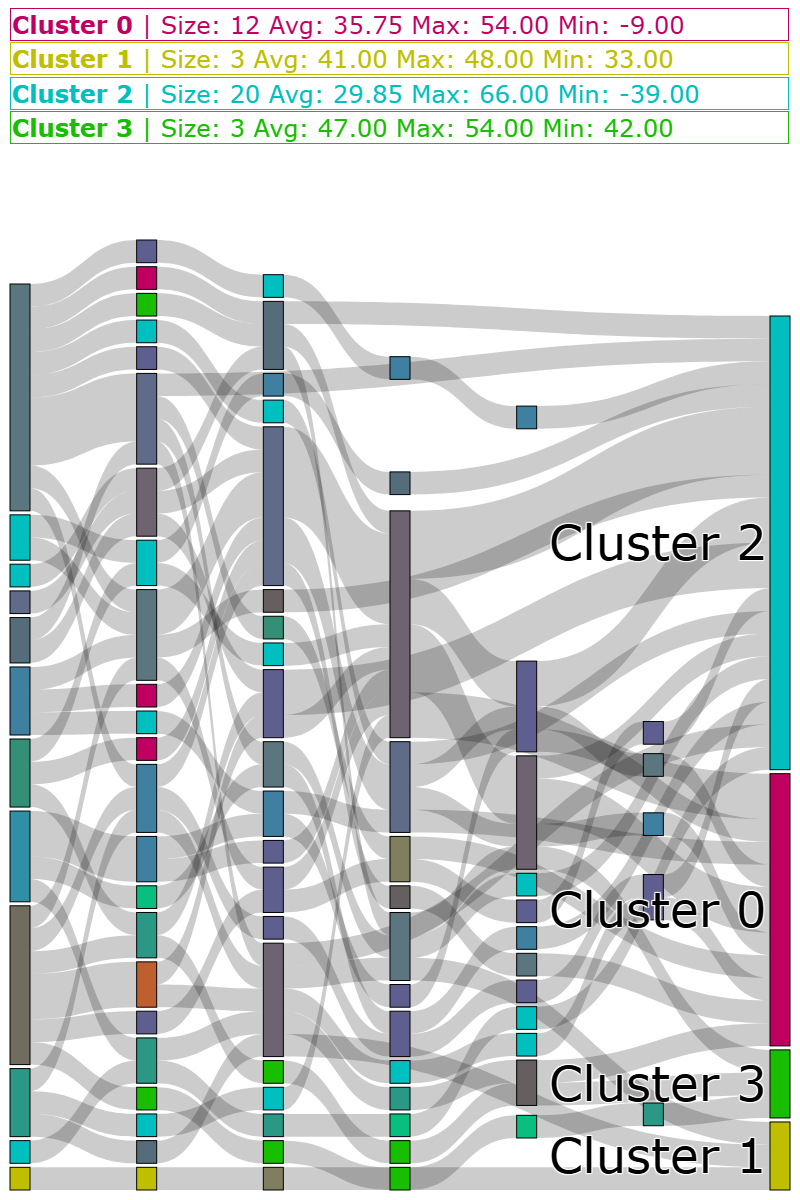}
    \end{minipage}
    \begin{minipage}{0.19\linewidth}
        \centering
        \includegraphics[width=\linewidth]{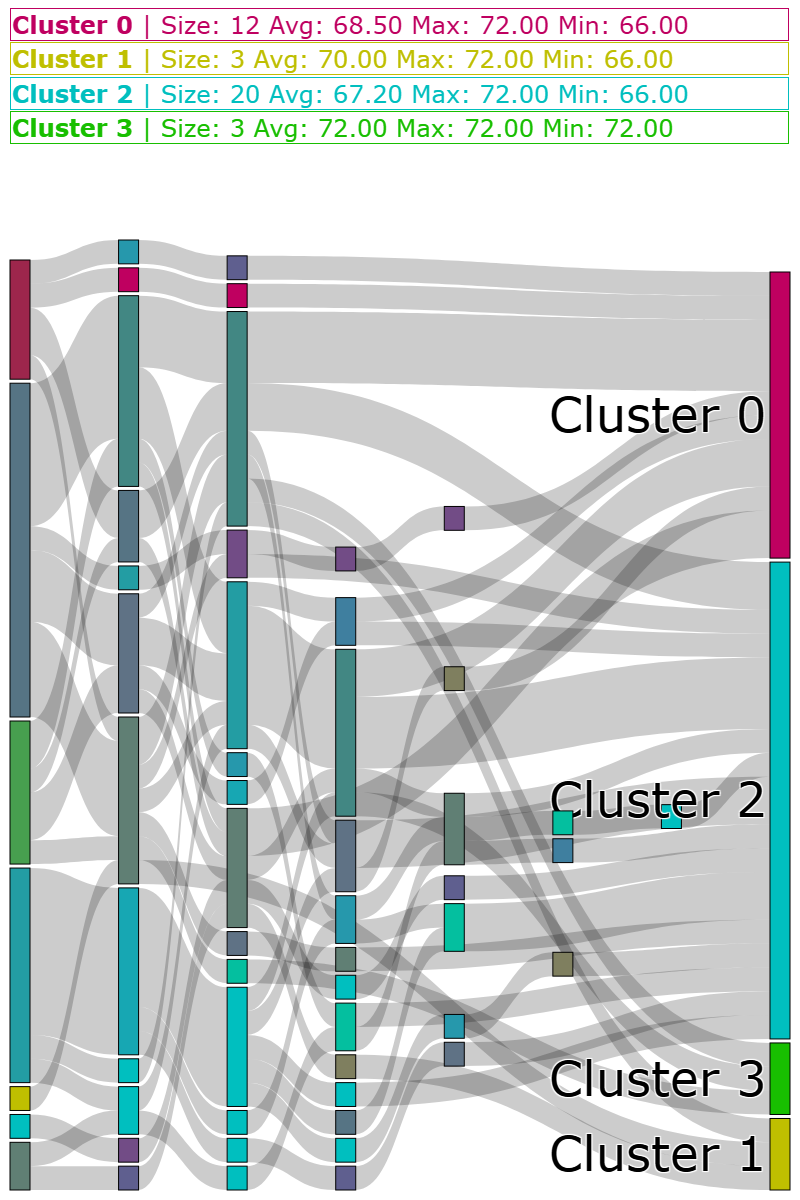}
    \end{minipage}
    \begin{minipage}{0.19\linewidth}
        \centering
        \includegraphics[width=\linewidth]{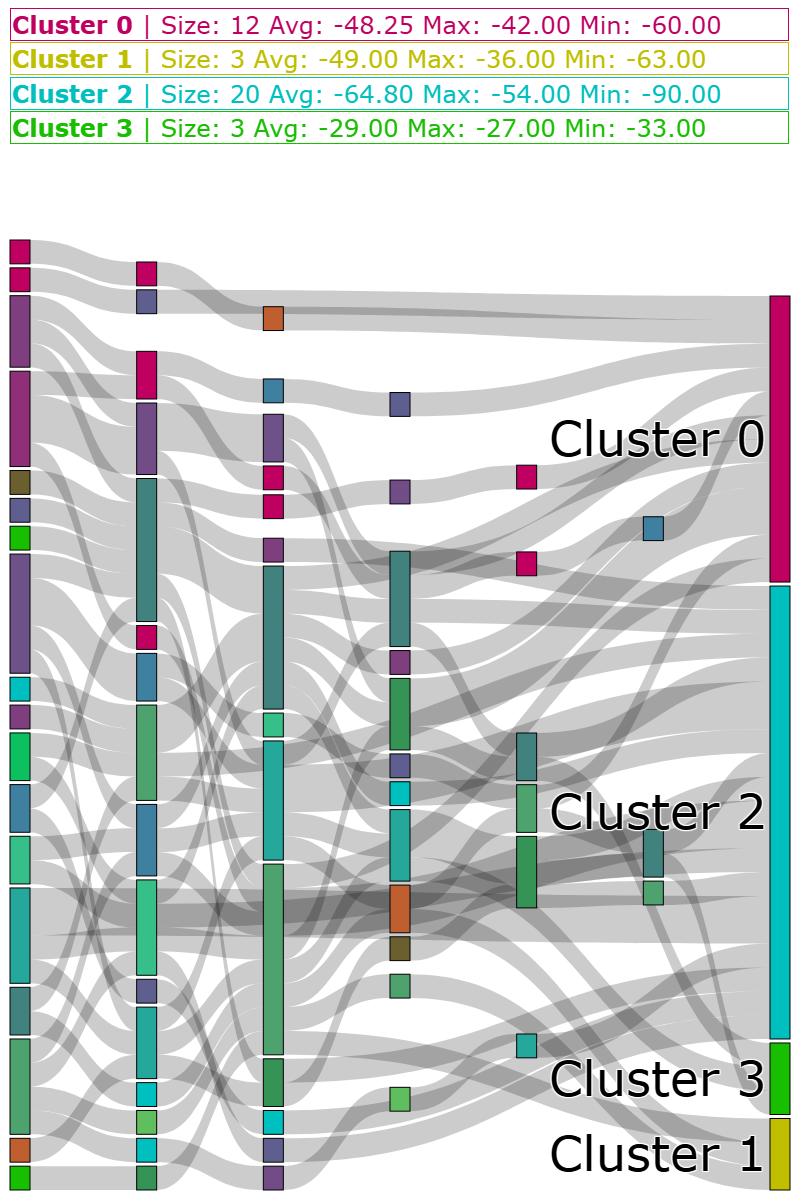}
    \end{minipage}
    \\
    \begin{minipage}{0.333\linewidth}
        \centering
        \includegraphics[width=\linewidth]{fitness_landscape_agglomerative_k5.pdf}
    \end{minipage}
    \begin{minipage}{0.19\linewidth}
        \centering
        \includegraphics[width=\linewidth]{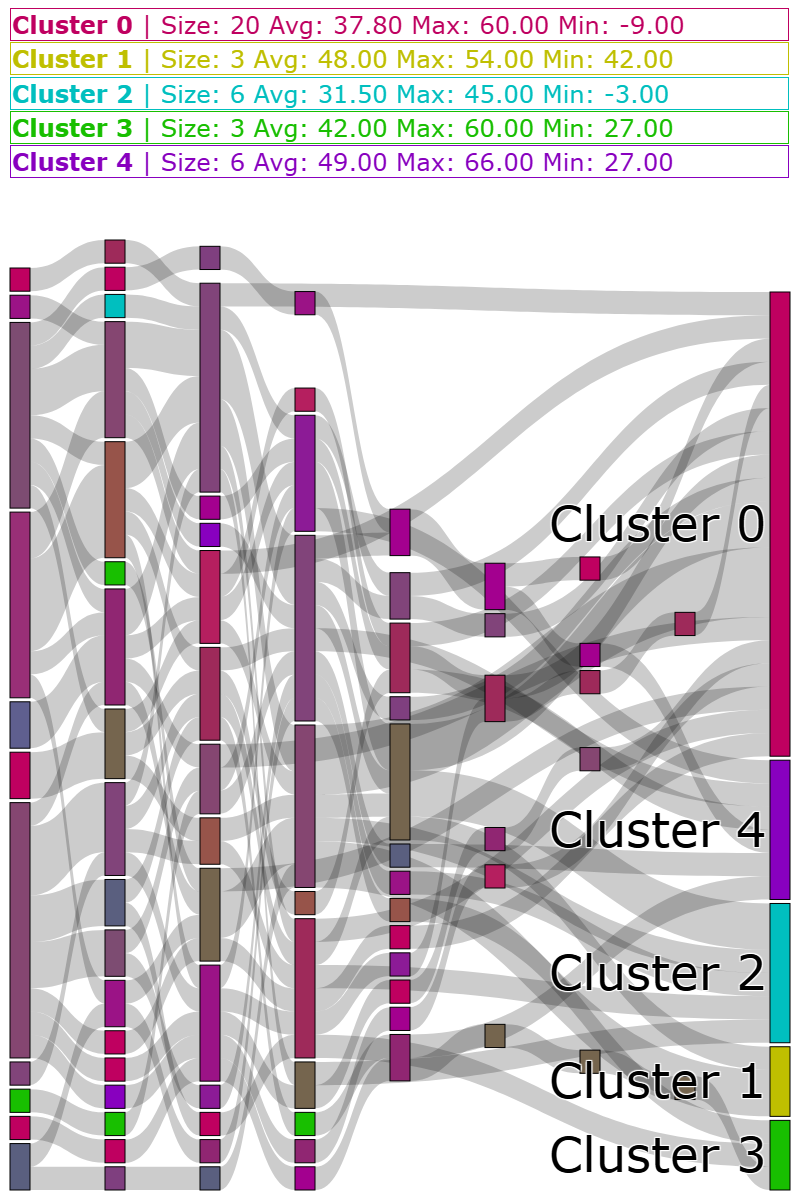}
    \end{minipage}
    \begin{minipage}{0.19\linewidth}
        \centering
        \includegraphics[width=\linewidth]{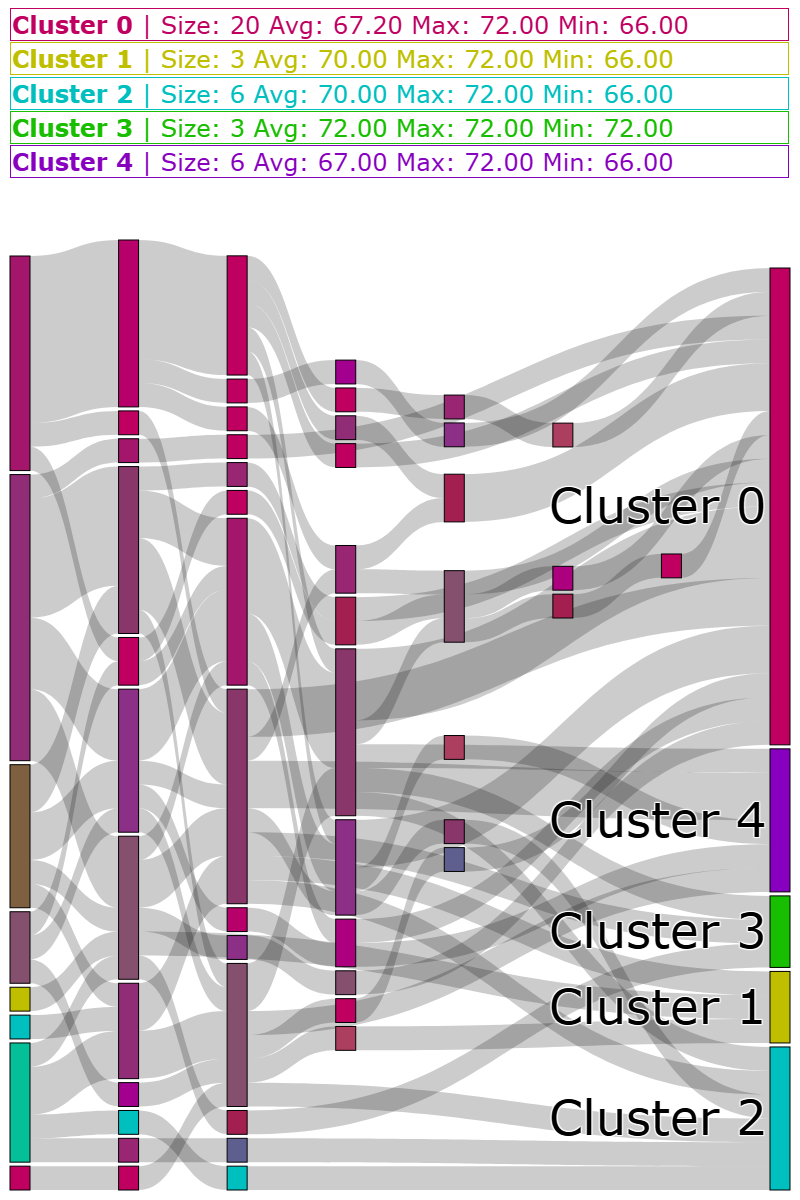}
    \end{minipage}
    \begin{minipage}{0.19\linewidth}
        \centering
        \includegraphics[width=\linewidth]{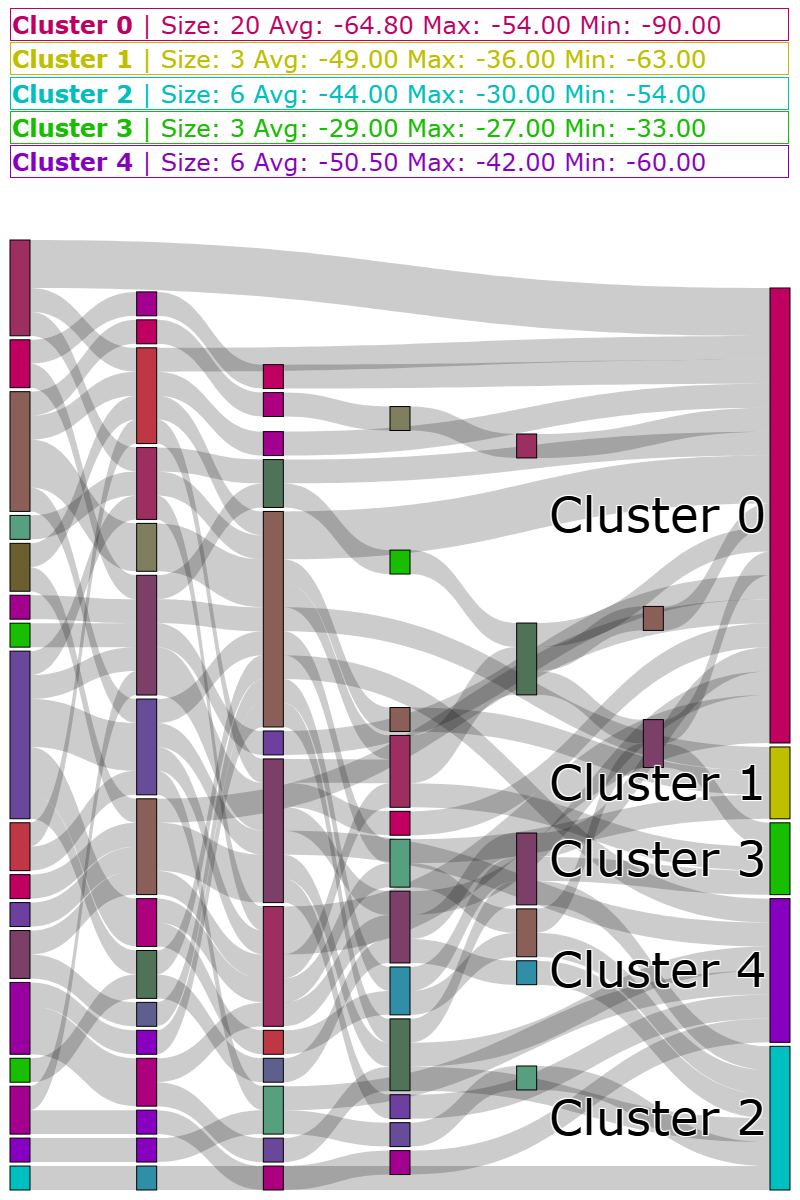}
    \end{minipage}
    \\
    \begin{minipage}{0.333\linewidth}
        \centering
        \includegraphics[width=\linewidth]{fitness_landscape_agglomerative_k6.pdf}
        fitness landscape
    \end{minipage}
    \begin{minipage}{0.19\linewidth}
        \centering
        \includegraphics[width=\linewidth]{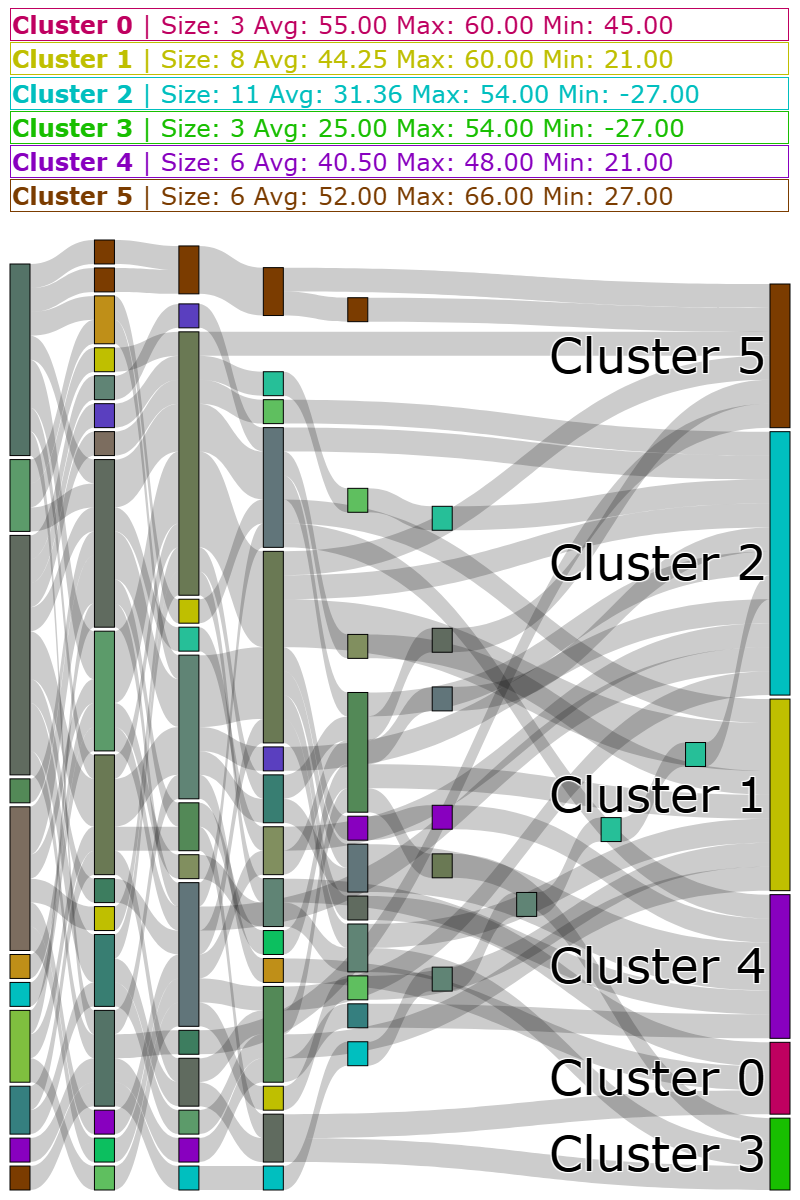}
        unsorted
    \end{minipage}
    \begin{minipage}{0.19\linewidth}
        \centering
        \includegraphics[width=\linewidth]{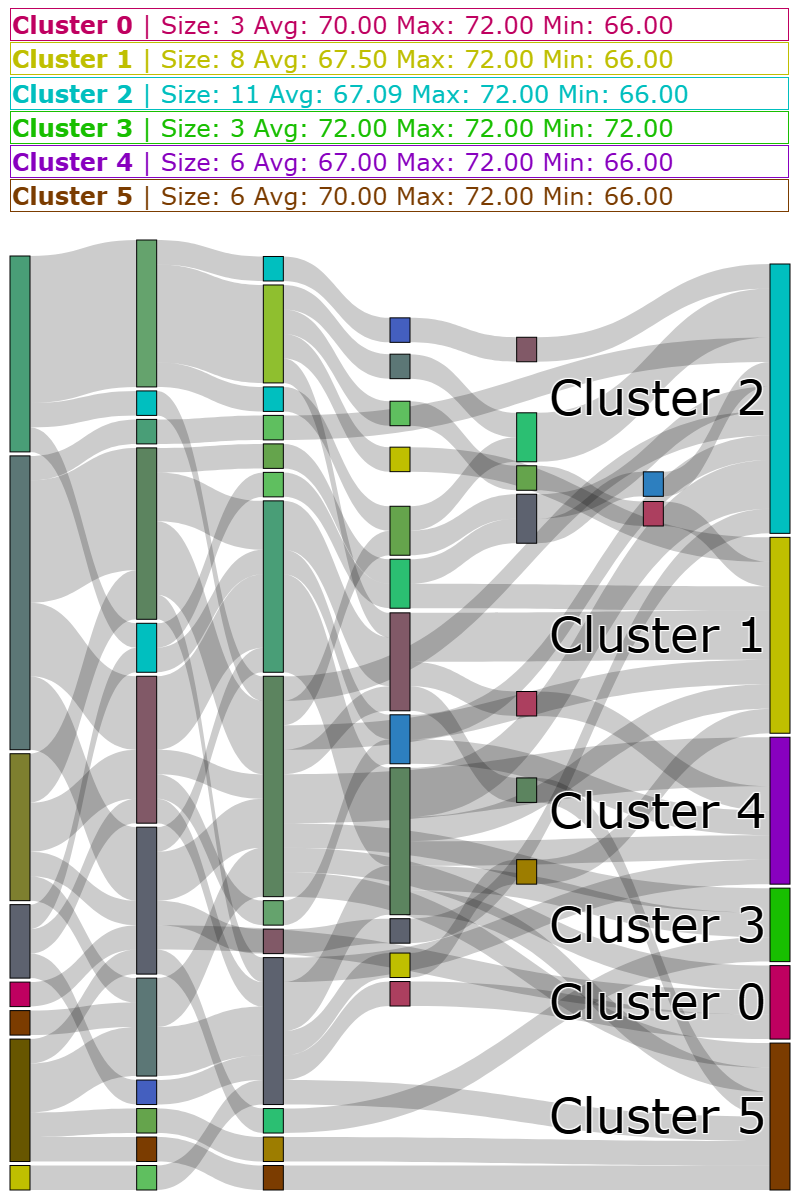}
        win
    \end{minipage}
    \begin{minipage}{0.19\linewidth}
        \centering
        \includegraphics[width=\linewidth]{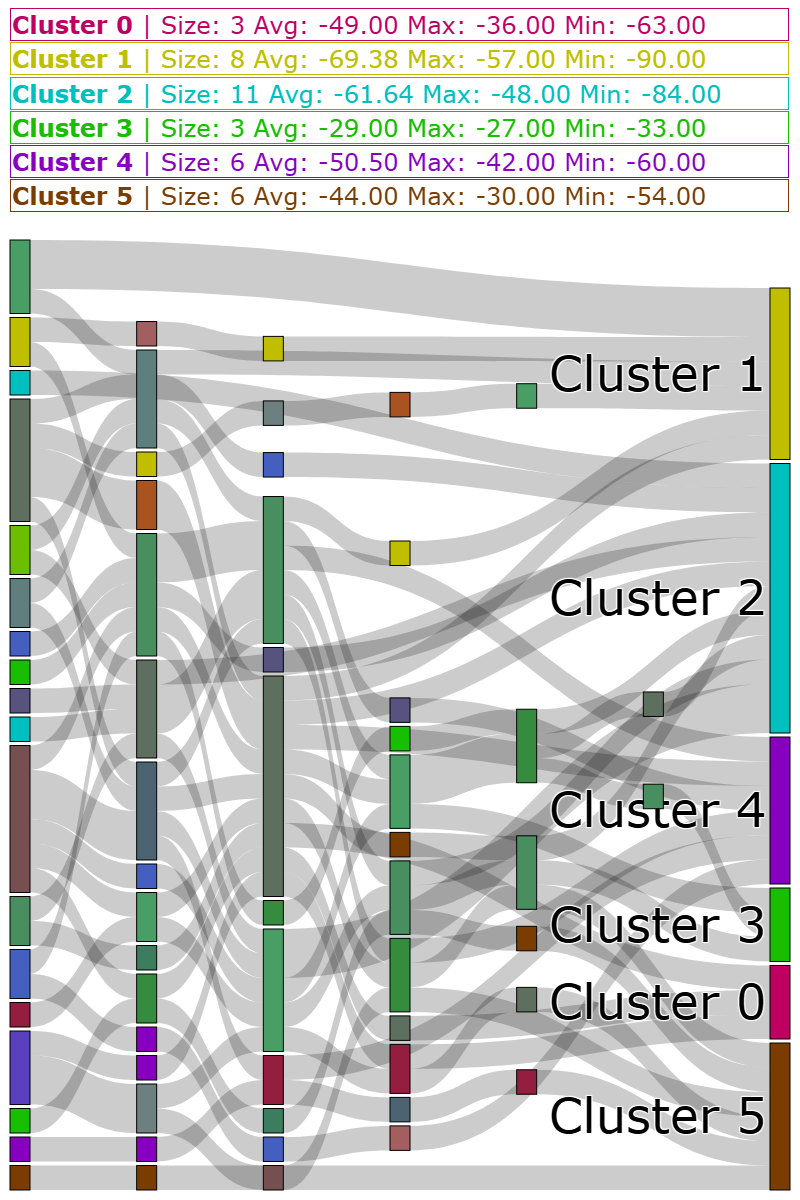}
        loss
    \end{minipage}
    \\
    \caption{Comprehensive sankey diagram of tactic patterns with Ward Agglomerative Clustering algorithm for clustering}
    \label{fig:comprehensive_sankey_tactic_diagram_Agglomerative}
\end{figure*}

\begin{figure*}[!htp]
    \centering
    \begin{minipage}{0.333\linewidth}
        \centering
        \includegraphics[width=\linewidth]{fitness_landscape_birch_k3.pdf}
    \end{minipage}
    \begin{minipage}{0.19\linewidth}
        \centering
        \includegraphics[width=\linewidth]{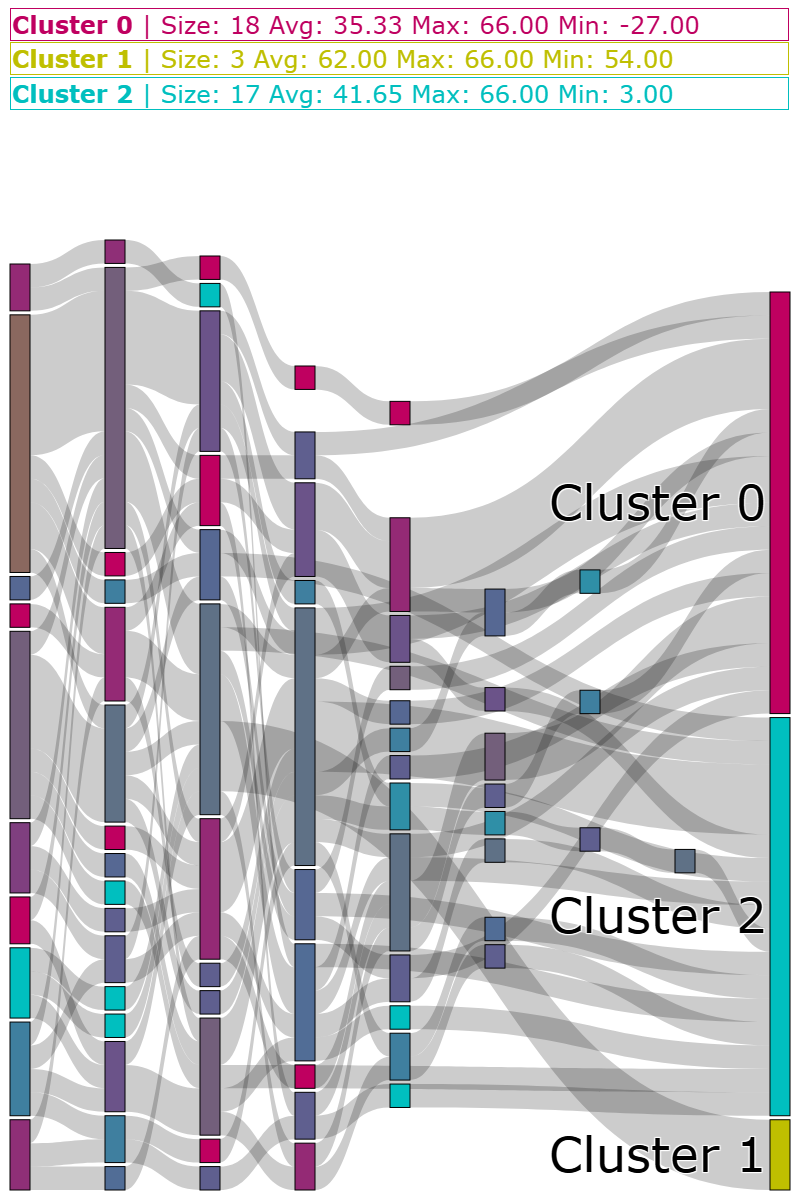}
    \end{minipage}
    \begin{minipage}{0.19\linewidth}
        \centering
        \includegraphics[width=\linewidth]{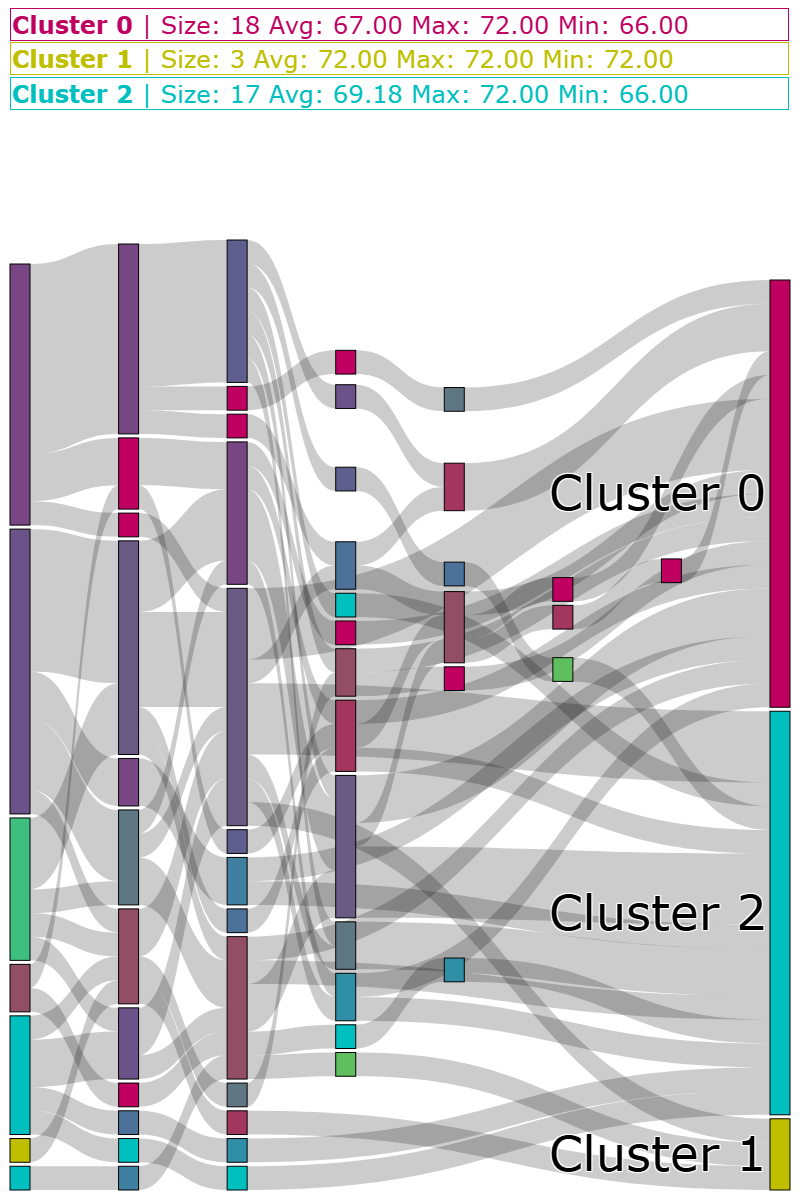}
    \end{minipage}
    \begin{minipage}{0.19\linewidth}
        \centering
        \includegraphics[width=\linewidth]{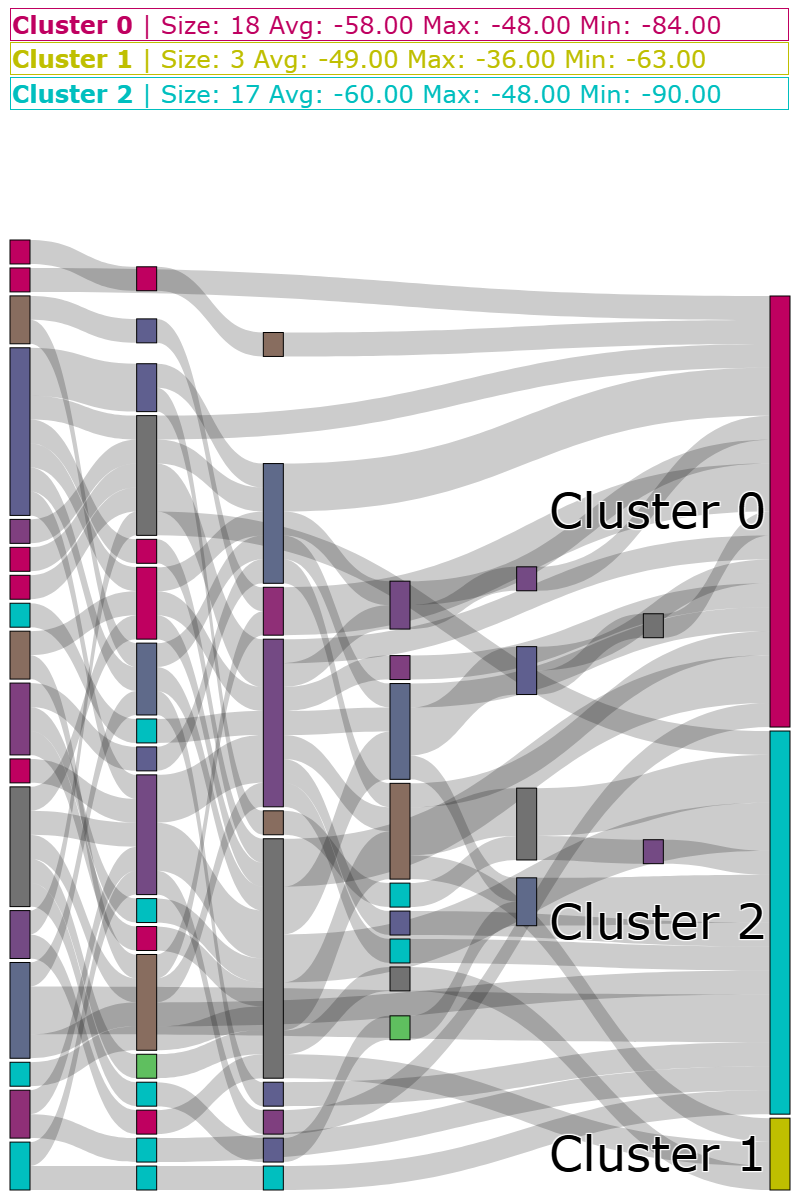}
    \end{minipage}
    \\
    \begin{minipage}{0.333\linewidth}
        \centering
        \includegraphics[width=\linewidth]{fitness_landscape_birch_k4.pdf}
    \end{minipage}
    \begin{minipage}{0.19\linewidth}
        \centering
        \includegraphics[width=\linewidth]{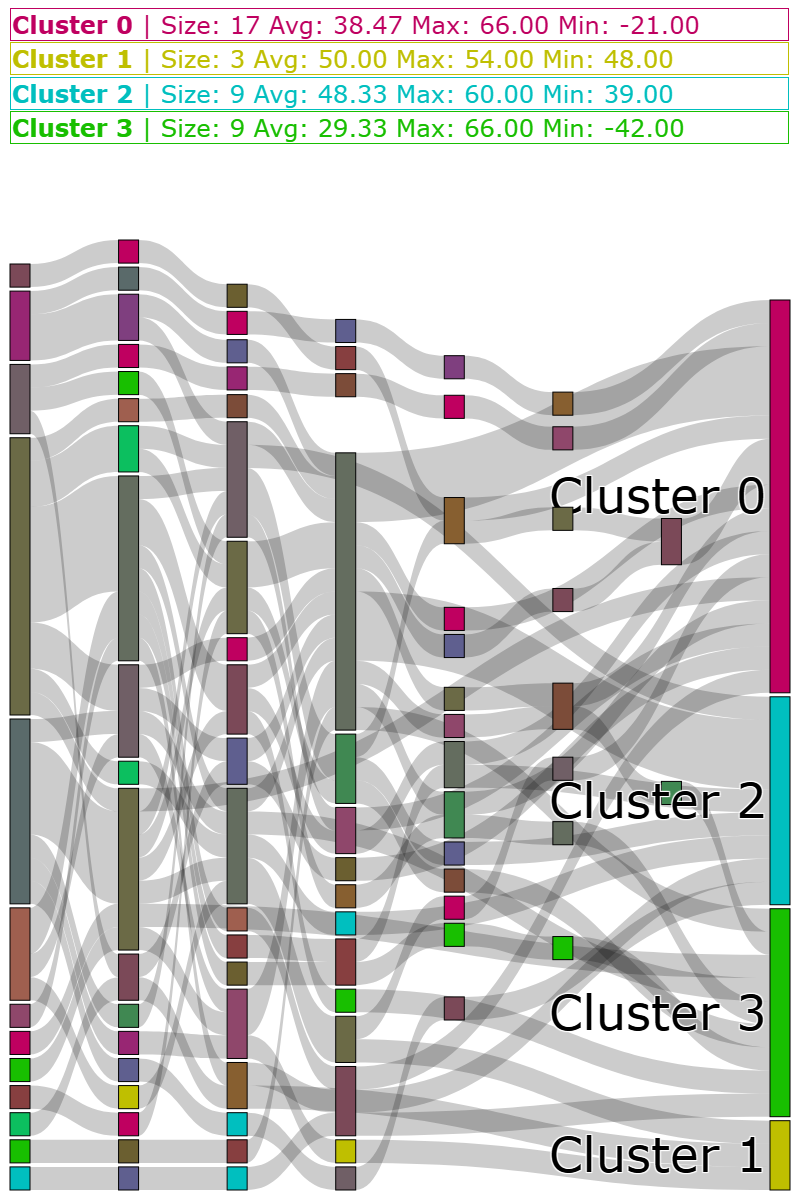}
    \end{minipage}
    \begin{minipage}{0.19\linewidth}
        \centering
        \includegraphics[width=\linewidth]{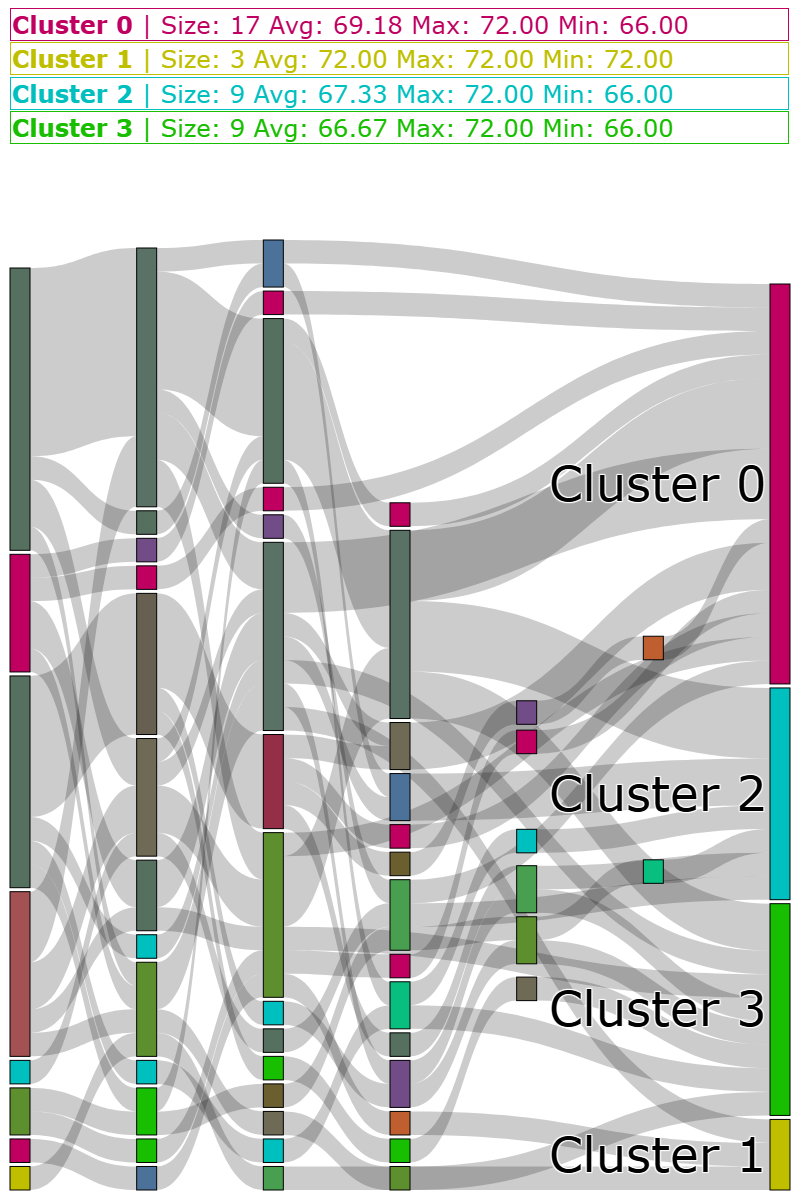}
    \end{minipage}
    \begin{minipage}{0.19\linewidth}
        \centering
        \includegraphics[width=\linewidth]{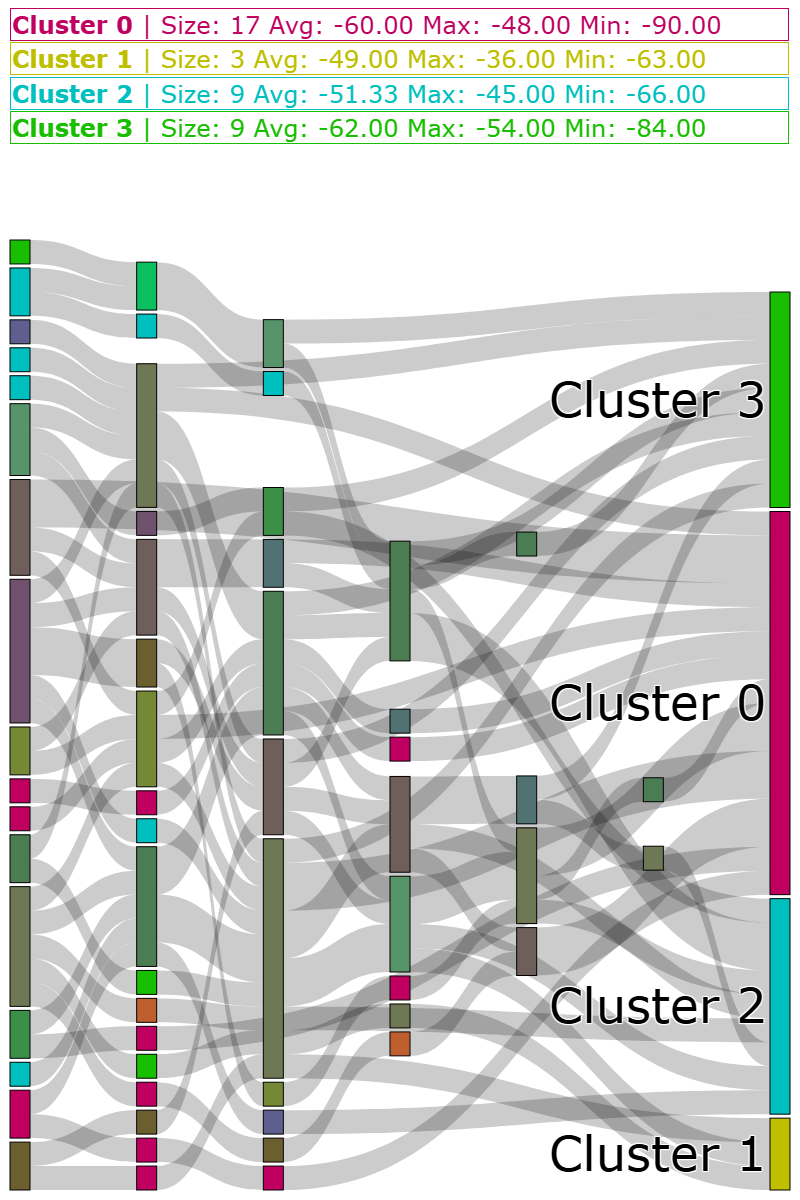}
    \end{minipage}
    \\
    \begin{minipage}{0.333\linewidth}
        \centering
        \includegraphics[width=\linewidth]{fitness_landscape_birch_k5.pdf}
    \end{minipage}
    \begin{minipage}{0.19\linewidth}
        \centering
        \includegraphics[width=\linewidth]{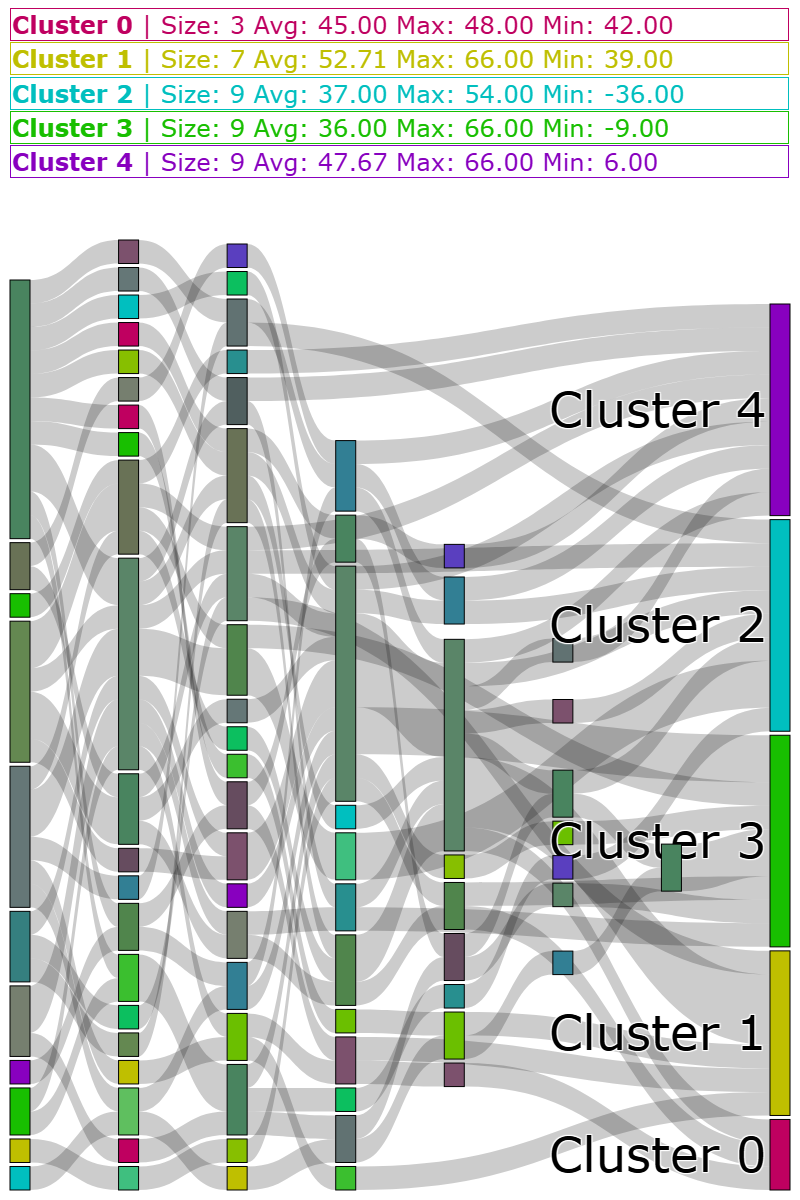}
    \end{minipage}
    \begin{minipage}{0.19\linewidth}
        \centering
        \includegraphics[width=\linewidth]{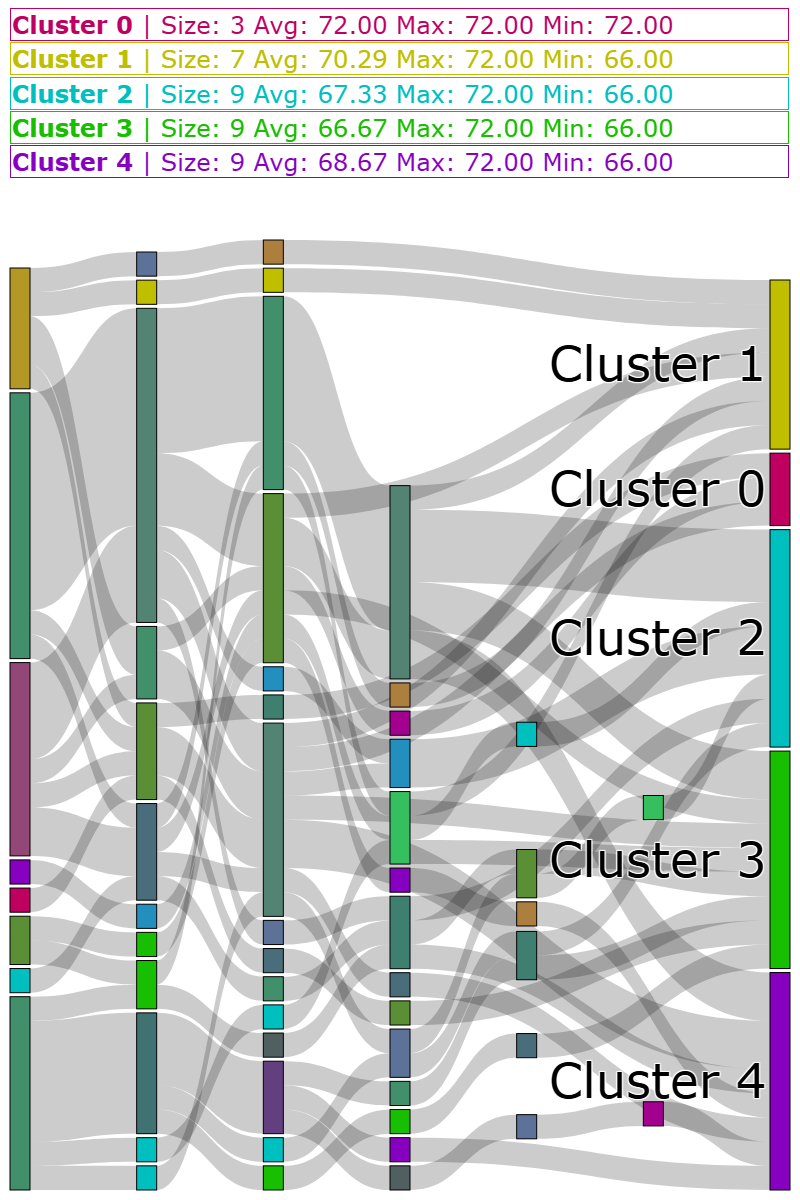}
    \end{minipage}
    \begin{minipage}{0.19\linewidth}
        \centering
        \includegraphics[width=\linewidth]{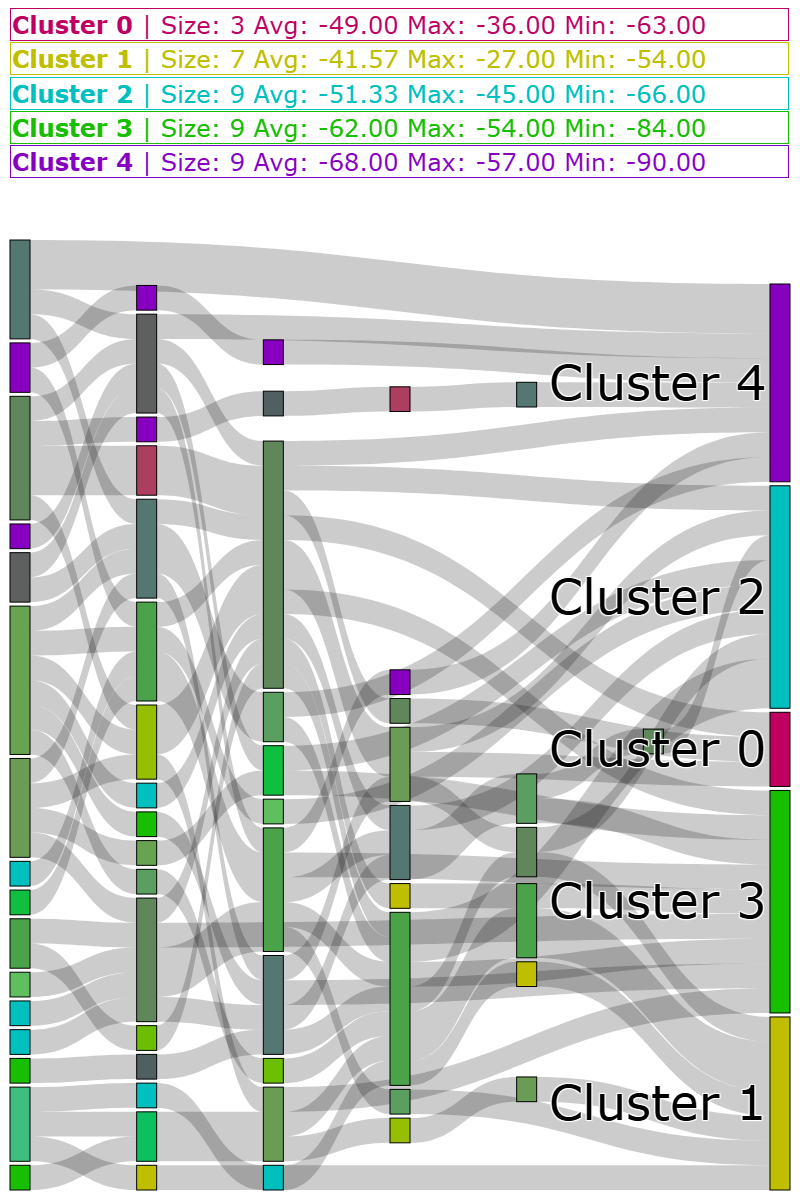}
    \end{minipage}
    \\
    \begin{minipage}{0.333\linewidth}
        \centering
        \includegraphics[width=\linewidth]{fitness_landscape_birch_k6.pdf}
        fitness landscape
    \end{minipage}
    \begin{minipage}{0.19\linewidth}
        \centering
        \includegraphics[width=\linewidth]{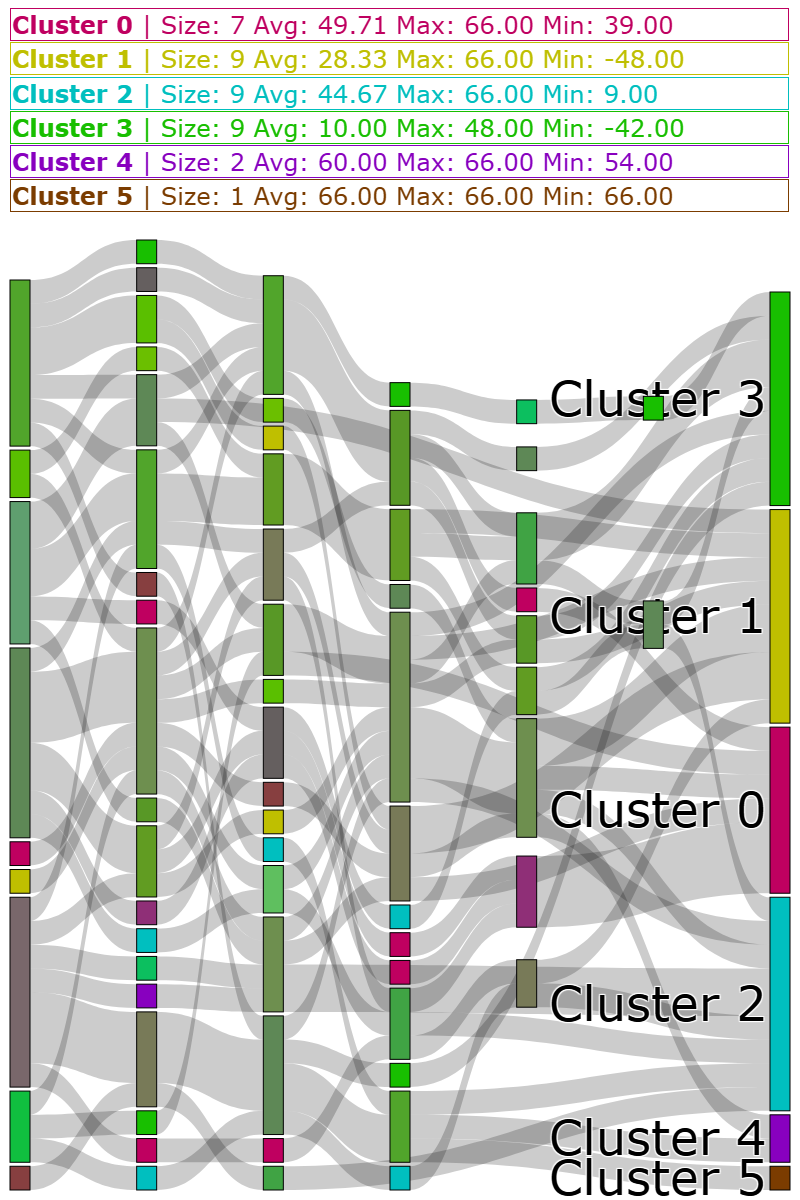}
        unsorted
    \end{minipage}
    \begin{minipage}{0.19\linewidth}
        \centering
        \includegraphics[width=\linewidth]{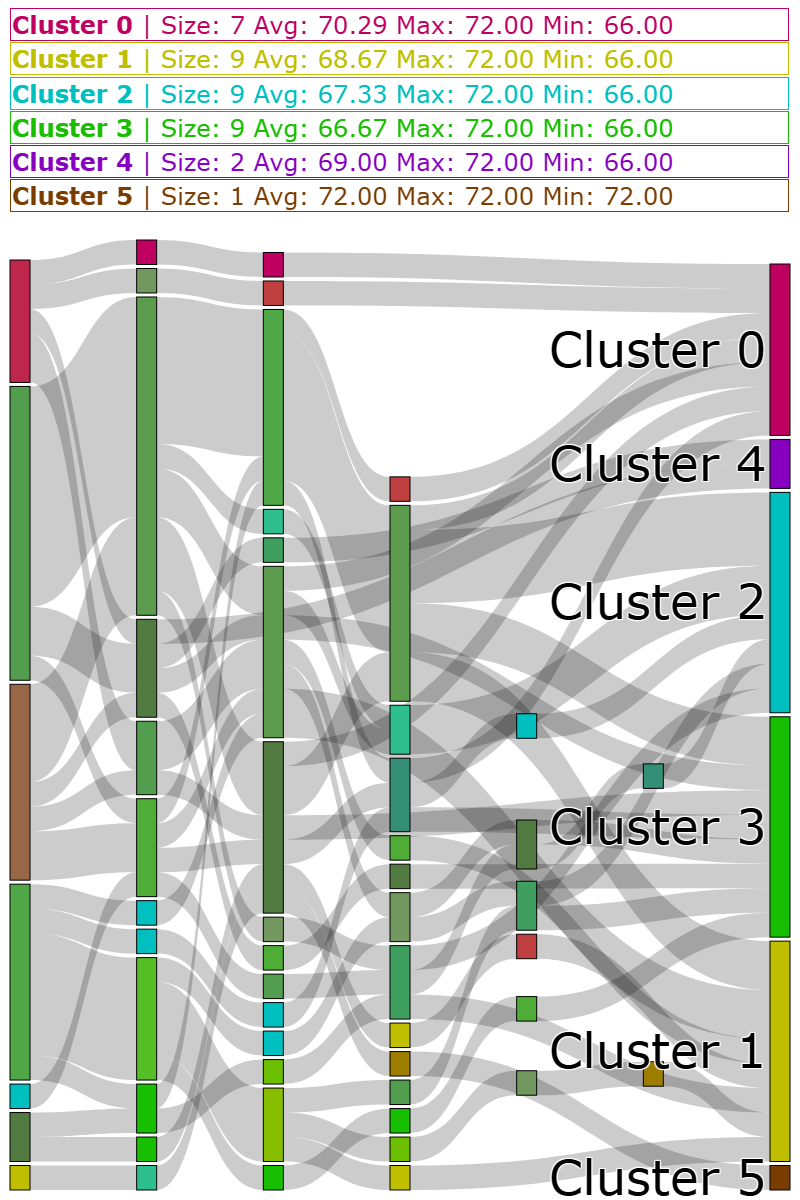}
        win
    \end{minipage}
    \begin{minipage}{0.19\linewidth}
        \centering
        \includegraphics[width=\linewidth]{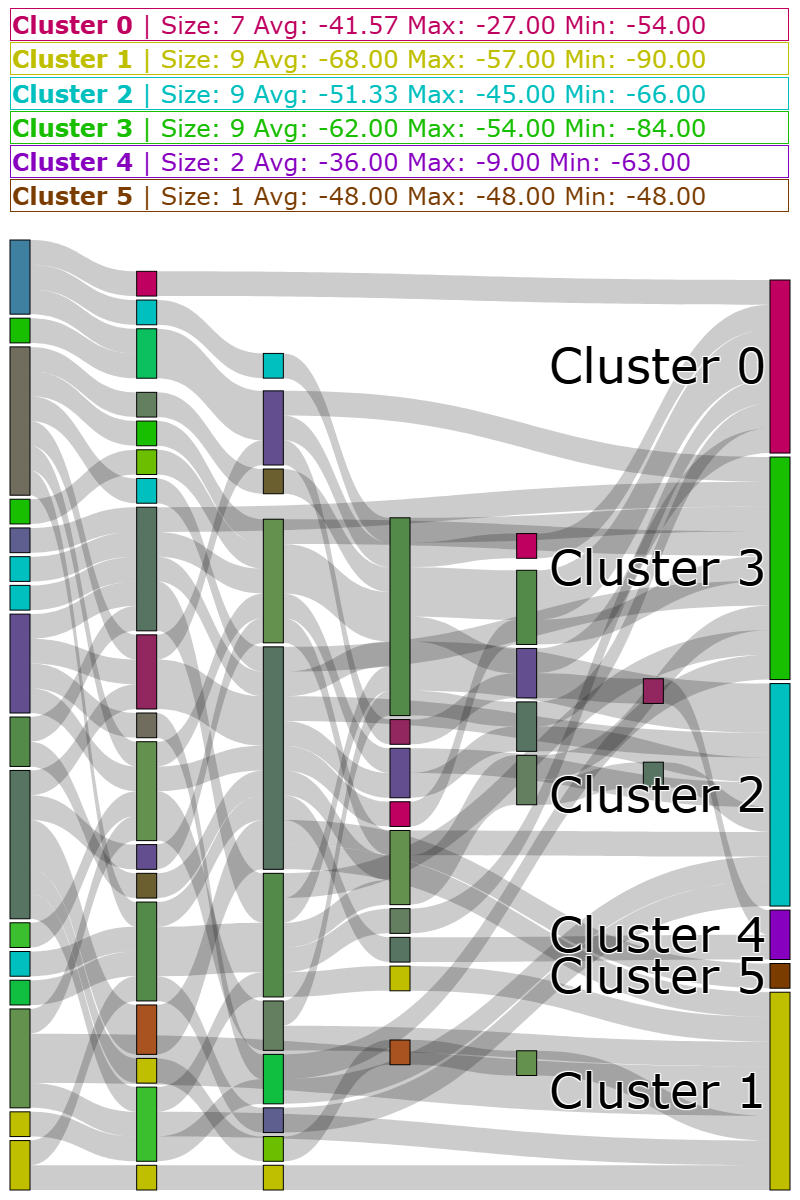}
        loss
    \end{minipage}
    \\
    \caption{Comprehensive sankey diagram of tactic patterns with birch algorithm for clustering}
    \label{fig:comprehensive_sankey_tactic_diagram_birch}
\end{figure*}

\begin{figure*}[!htp]
    \centering
    \begin{minipage}{0.333\linewidth}
        \centering
        \includegraphics[width=\linewidth]{fitness_landscape_gmm_k3.pdf}
    \end{minipage}
    \begin{minipage}{0.19\linewidth}
        \centering
        \includegraphics[width=\linewidth]{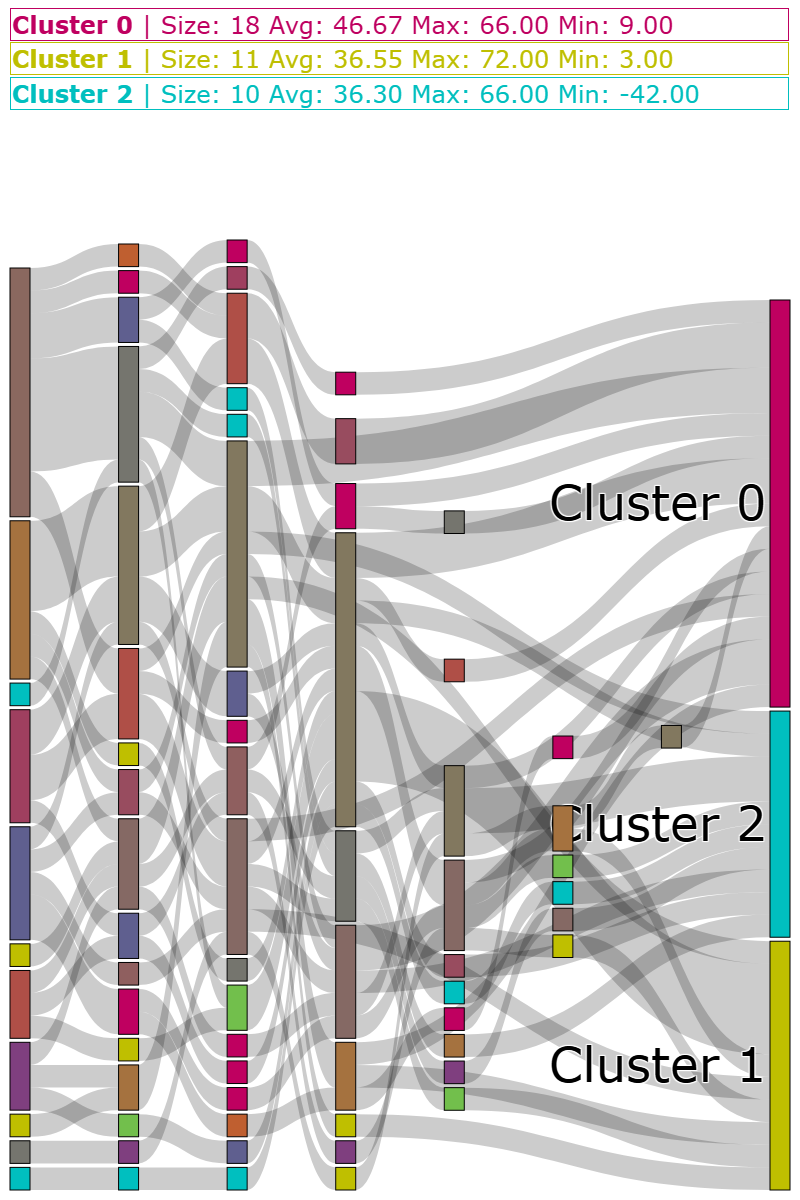}
    \end{minipage}
    \begin{minipage}{0.19\linewidth}
        \centering
        \includegraphics[width=\linewidth]{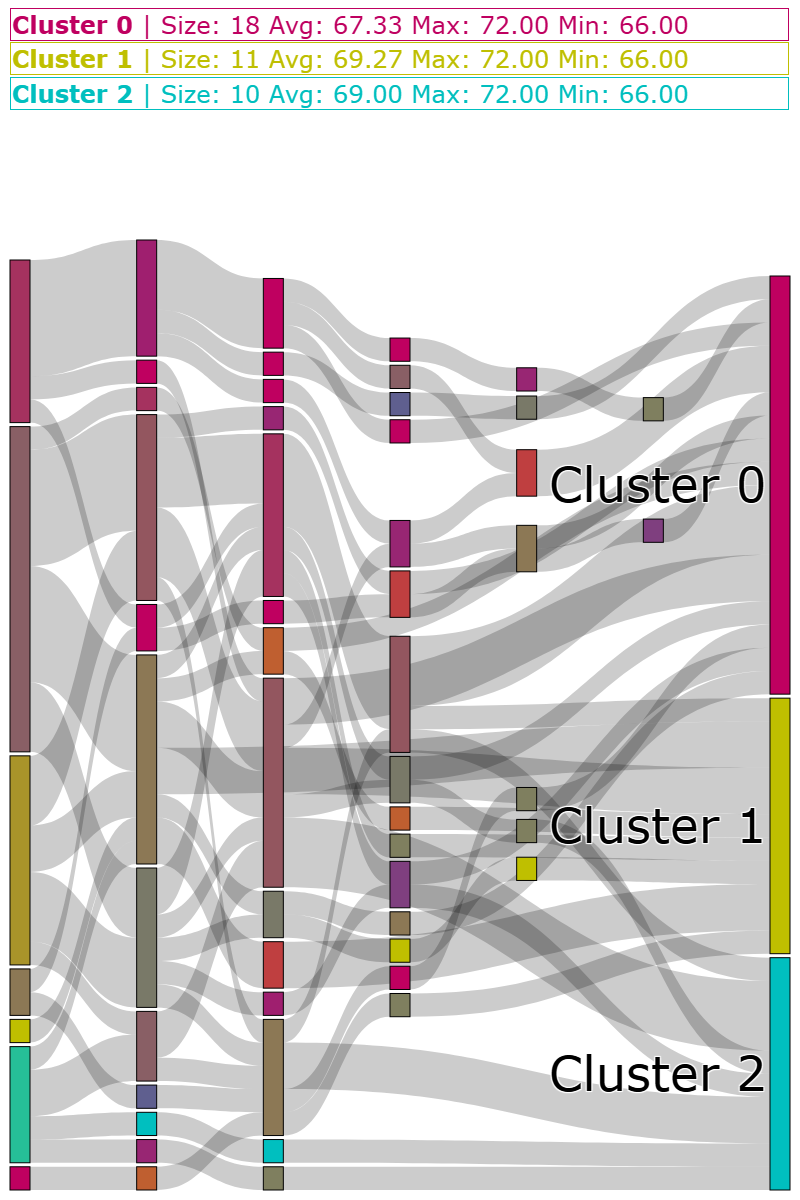}
    \end{minipage}
    \begin{minipage}{0.19\linewidth}
        \centering
        \includegraphics[width=\linewidth]{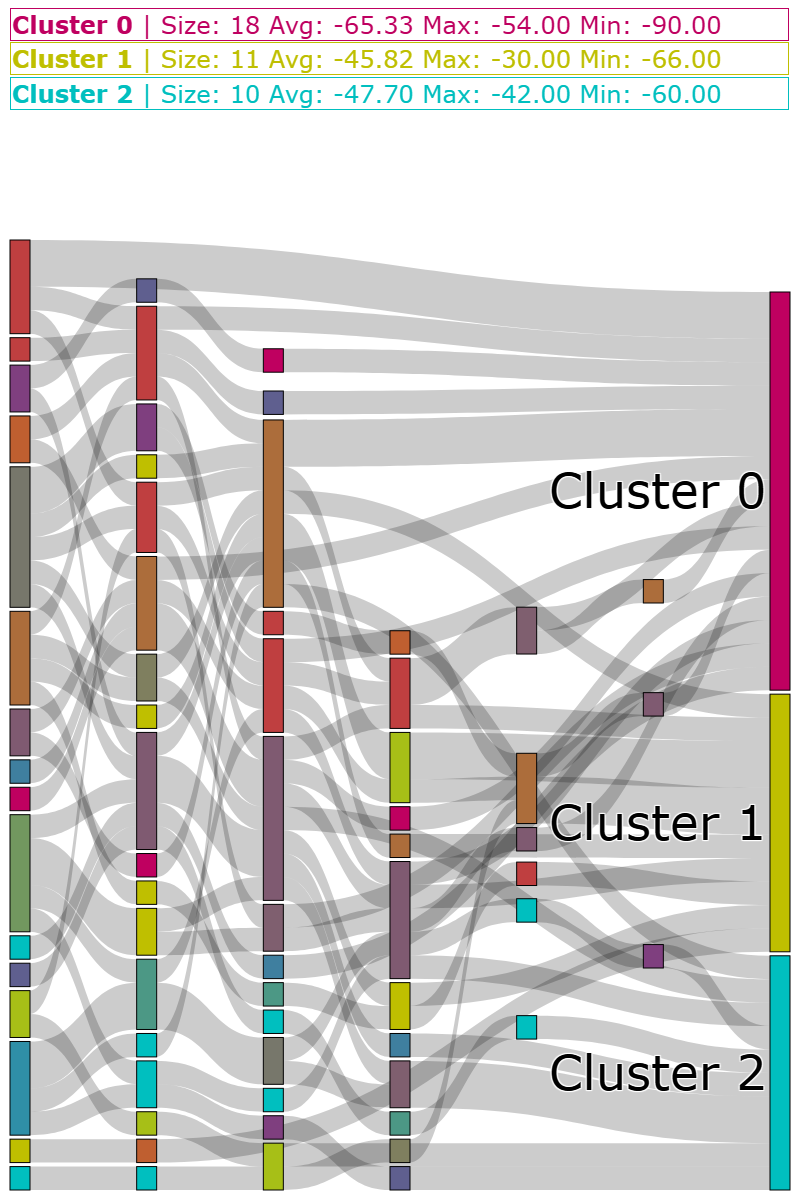}
    \end{minipage}
    \\
    \begin{minipage}{0.333\linewidth}
        \centering
        \includegraphics[width=\linewidth]{fitness_landscape_gmm_k4.pdf}
    \end{minipage}
    \begin{minipage}{0.19\linewidth}
        \centering
        \includegraphics[width=\linewidth]{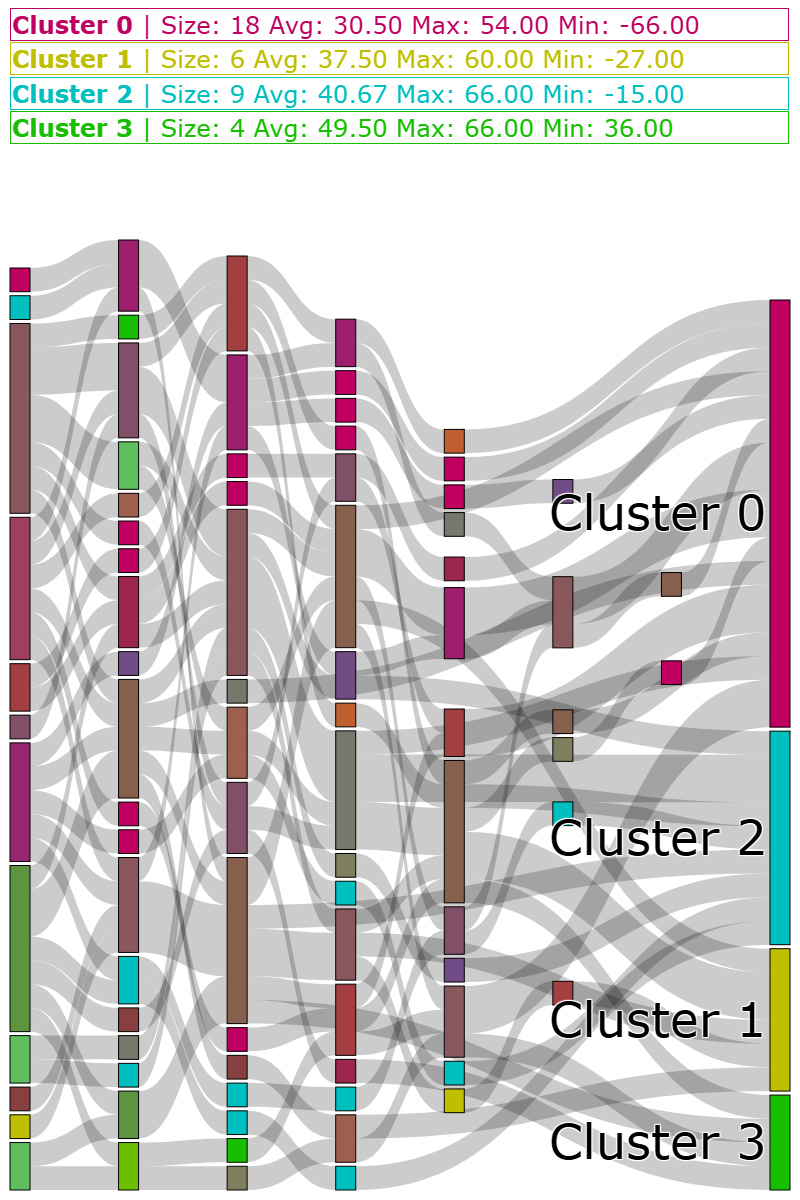}
    \end{minipage}
    \begin{minipage}{0.19\linewidth}
        \centering
        \includegraphics[width=\linewidth]{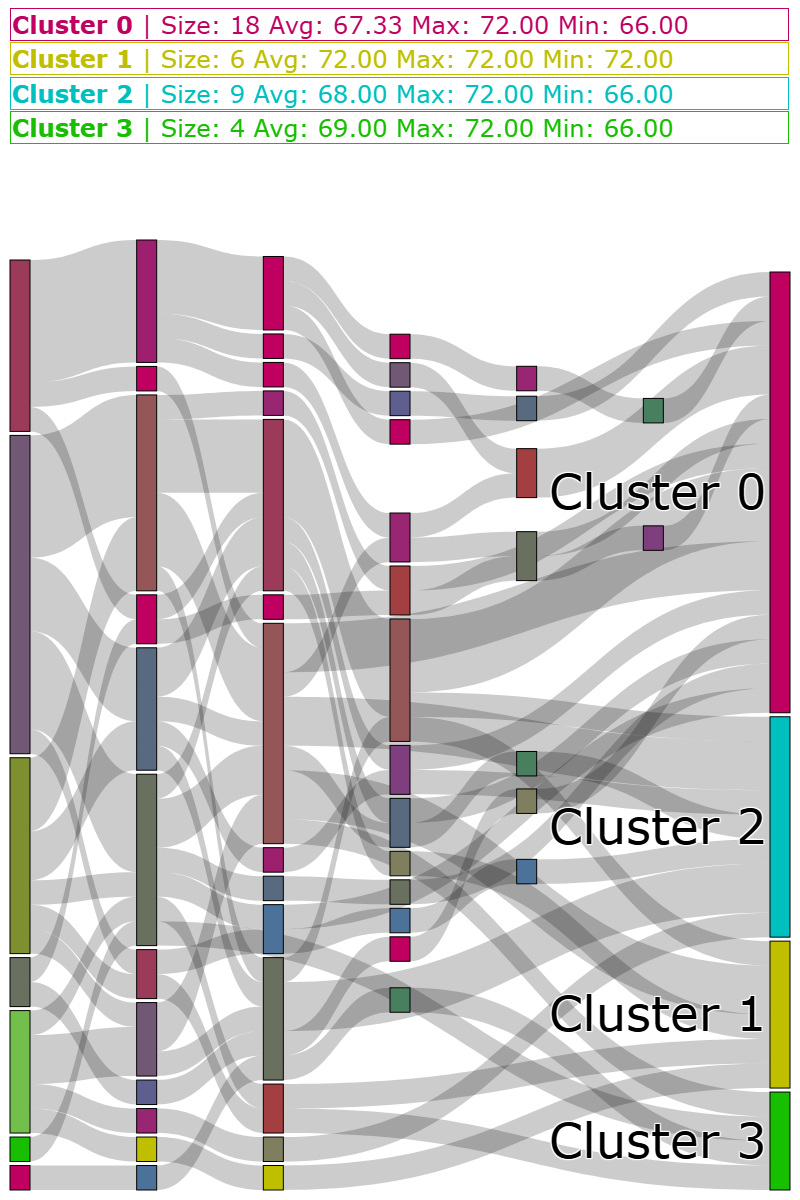}
    \end{minipage}
    \begin{minipage}{0.19\linewidth}
        \centering
        \includegraphics[width=\linewidth]{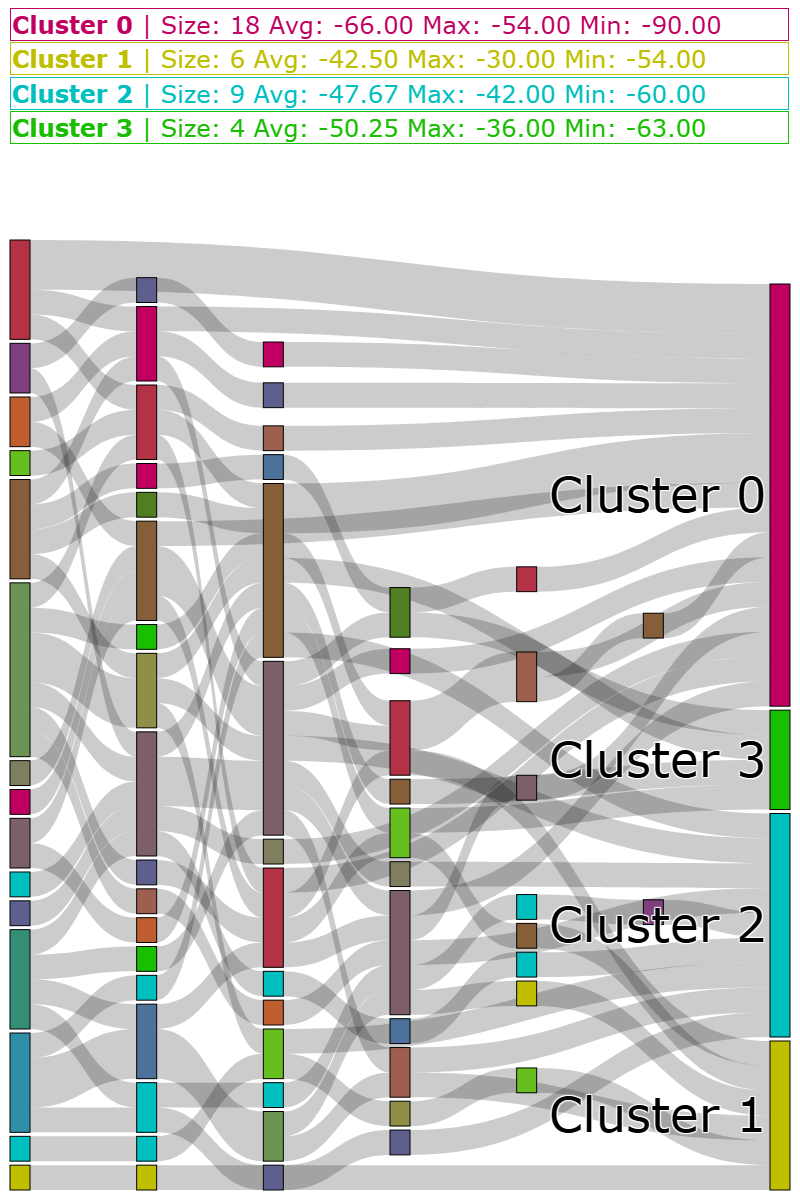}
    \end{minipage}
    \\
    \begin{minipage}{0.333\linewidth}
        \centering
        \includegraphics[width=\linewidth]{fitness_landscape_gmm_k5.pdf}
    \end{minipage}
    \begin{minipage}{0.19\linewidth}
        \centering
        \includegraphics[width=\linewidth]{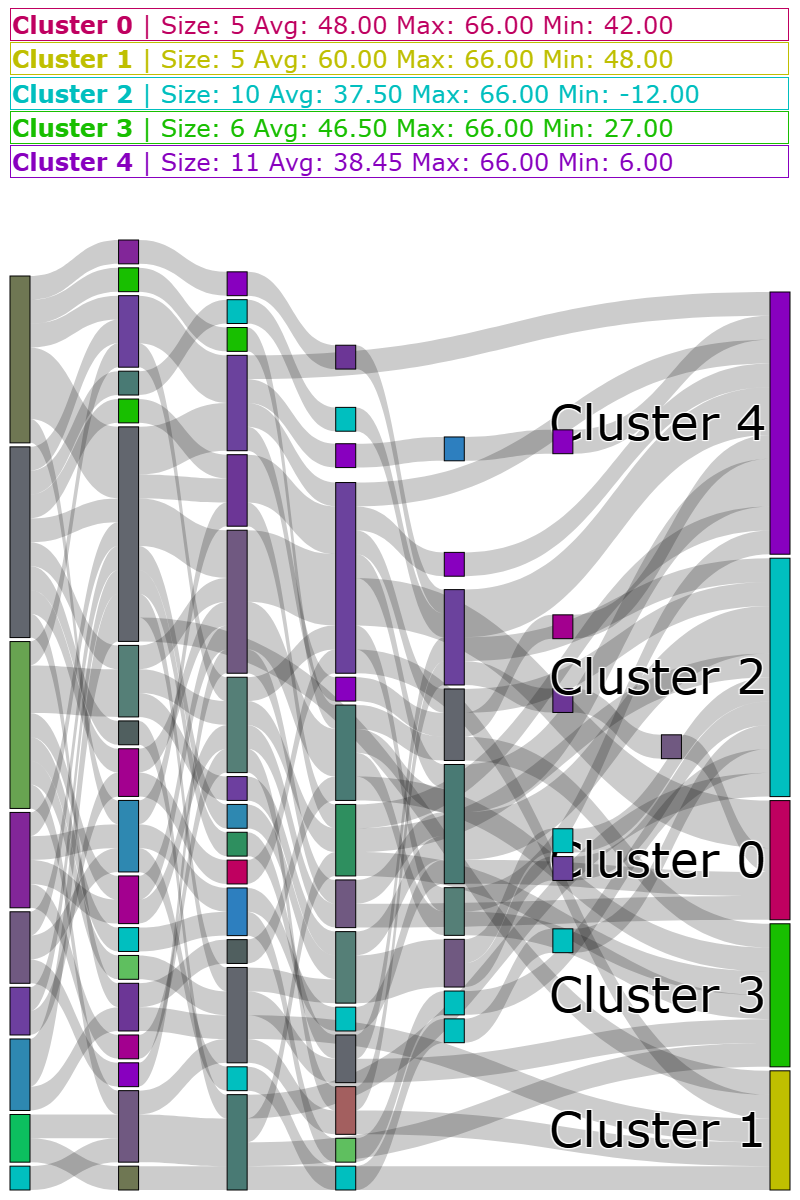}
    \end{minipage}
    \begin{minipage}{0.19\linewidth}
        \centering
        \includegraphics[width=\linewidth]{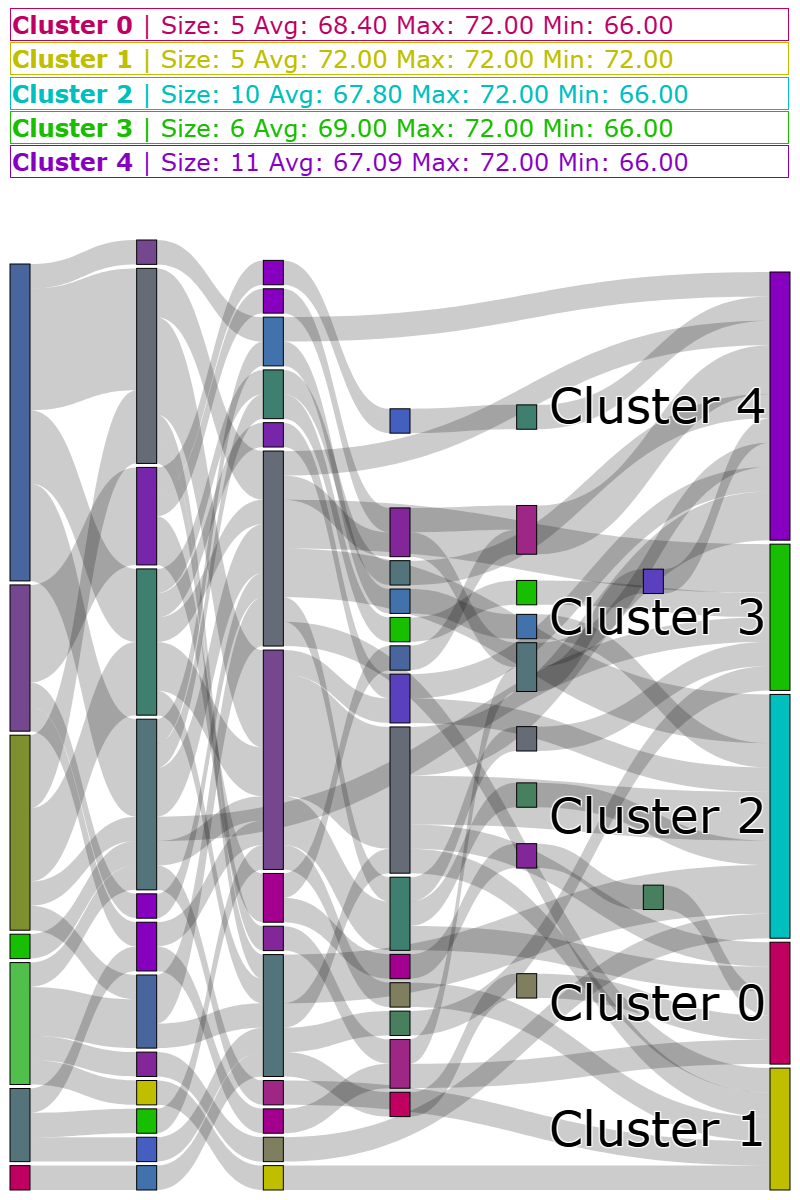}
    \end{minipage}
    \begin{minipage}{0.19\linewidth}
        \centering
        \includegraphics[width=\linewidth]{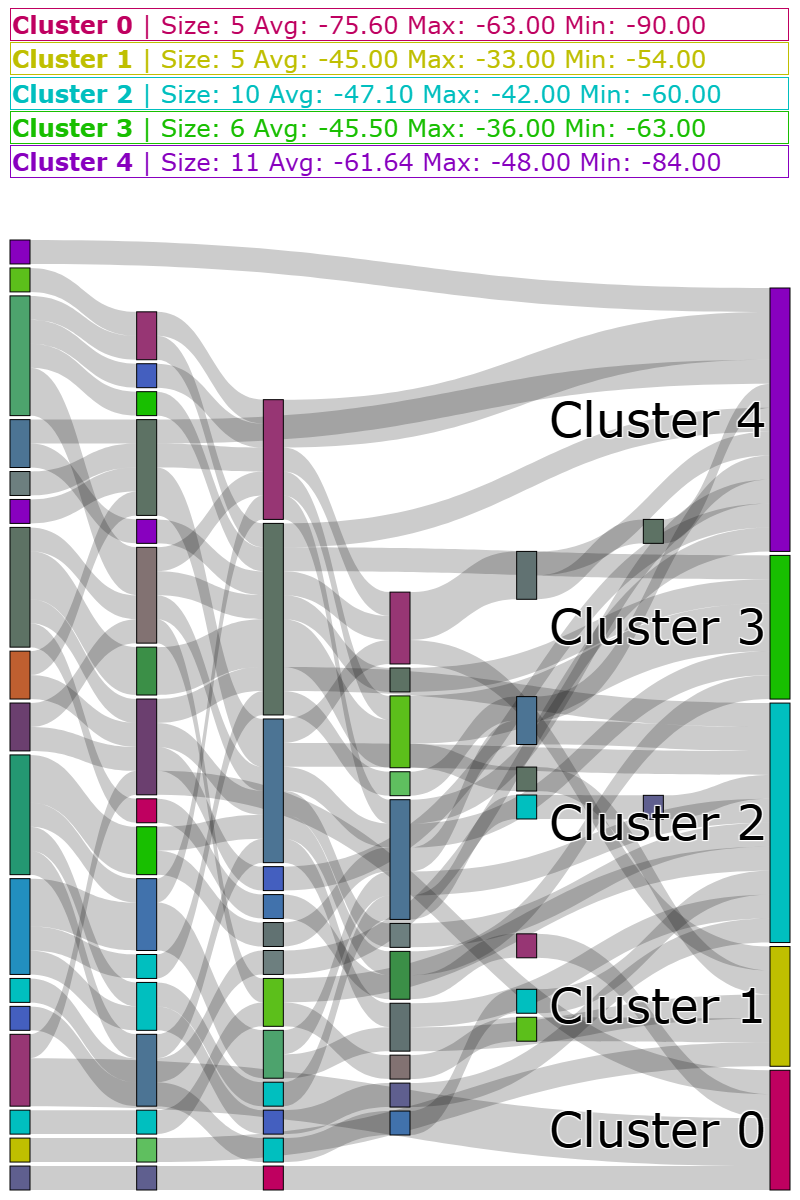}
    \end{minipage}
    \\
    \begin{minipage}{0.333\linewidth}
        \centering
        \includegraphics[width=\linewidth]{fitness_landscape_gmm_k6.pdf}
        fitness landscape
    \end{minipage}
    \begin{minipage}{0.19\linewidth}
        \centering
        \includegraphics[width=\linewidth]{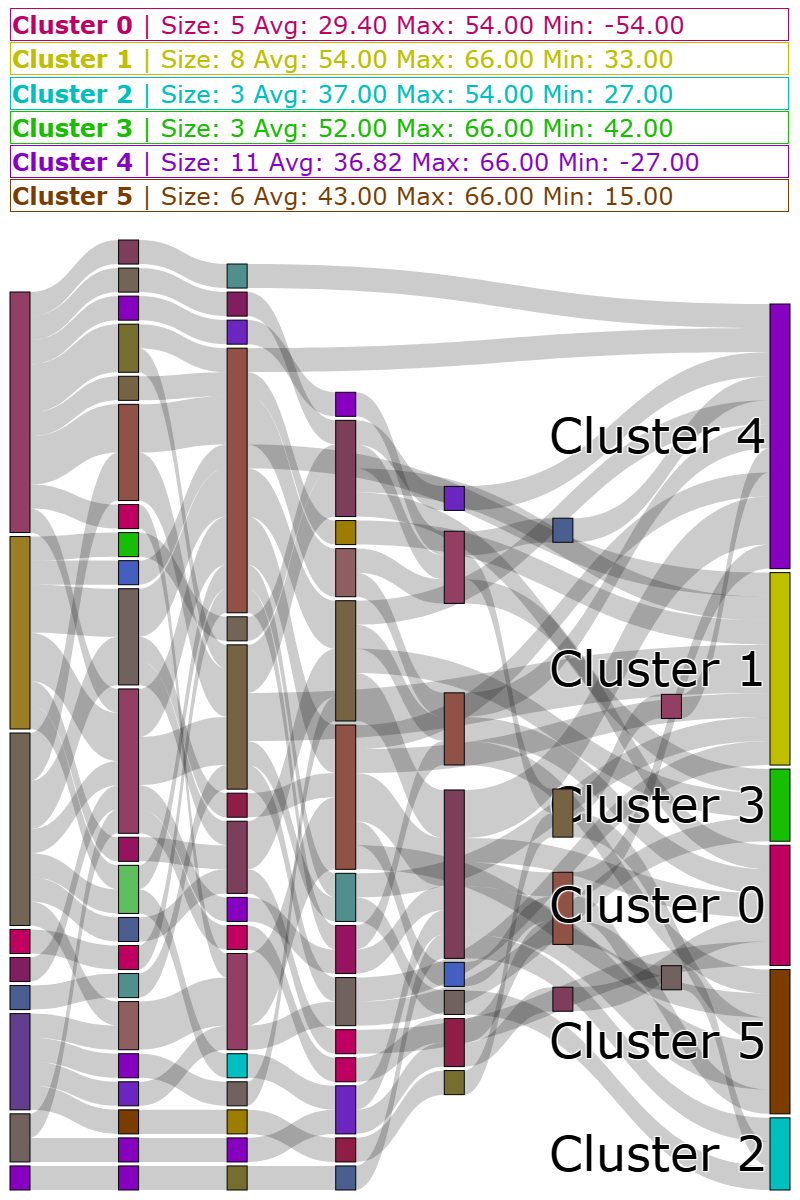}
        unsorted
    \end{minipage}
    \begin{minipage}{0.19\linewidth}
        \centering
        \includegraphics[width=\linewidth]{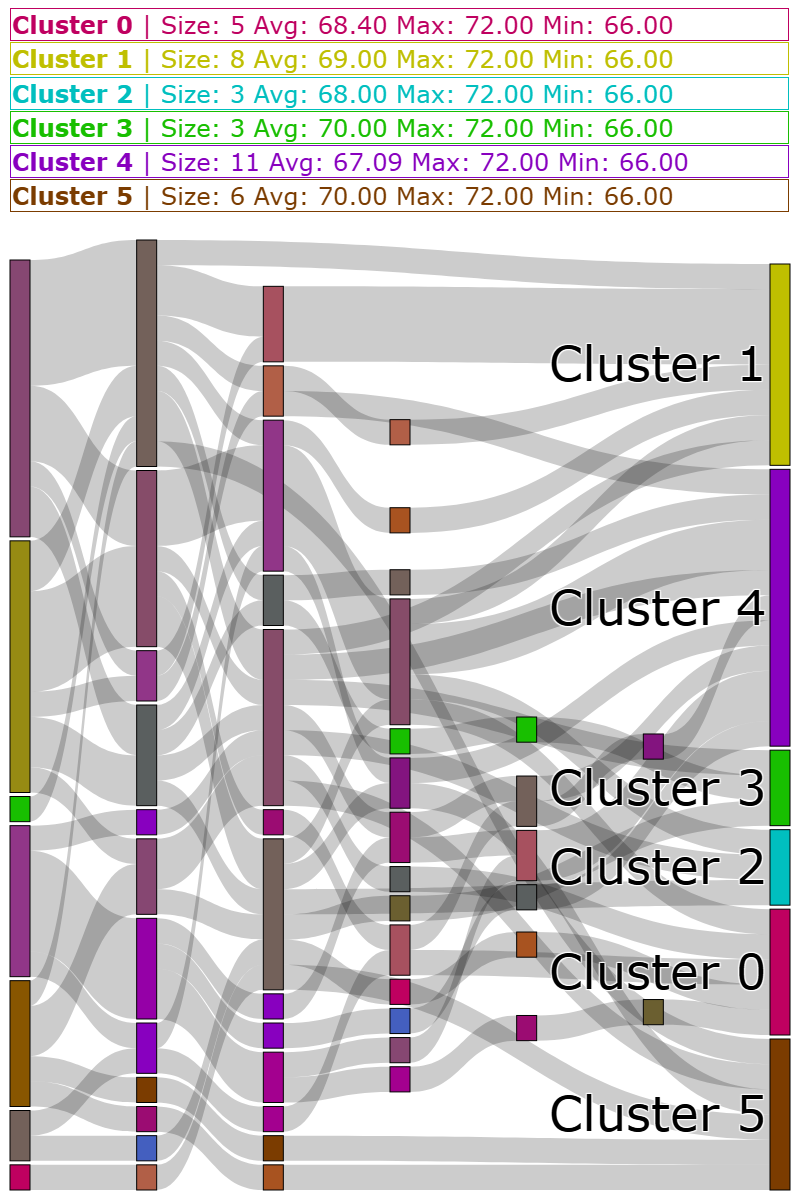}
        win
    \end{minipage}
    \begin{minipage}{0.19\linewidth}
        \centering
        \includegraphics[width=\linewidth]{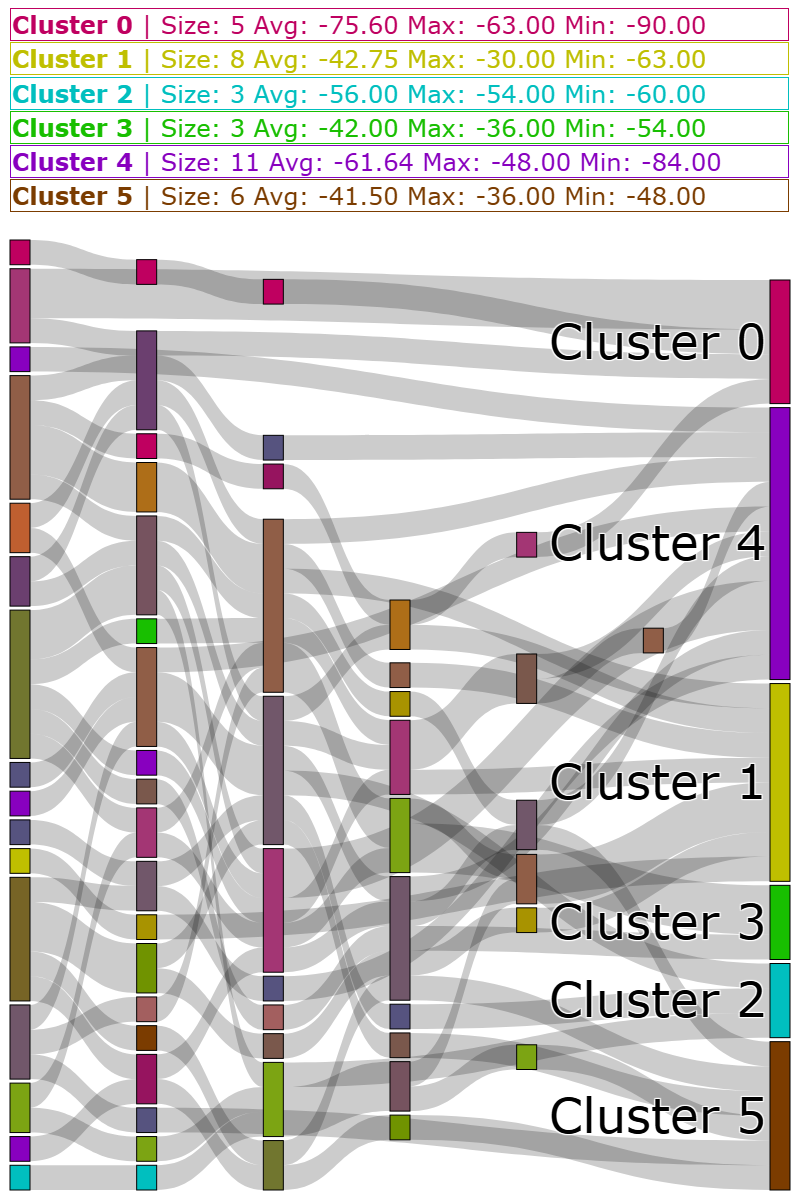}
        loss
    \end{minipage}
    \\
    \caption{Comprehensive sankey diagram of tactic patterns with gmm algorithm for clustering}
    \label{fig:comprehensive_sankey_tactic_diagram_gmm}
\end{figure*}

\begin{figure*}[!htp]
    \centering
    \begin{minipage}{0.333\linewidth}
        \centering
        \includegraphics[width=\linewidth]{fitness_landscape_kmeans_k3.pdf}
    \end{minipage}
    \begin{minipage}{0.19\linewidth}
        \centering
        \includegraphics[width=\linewidth]{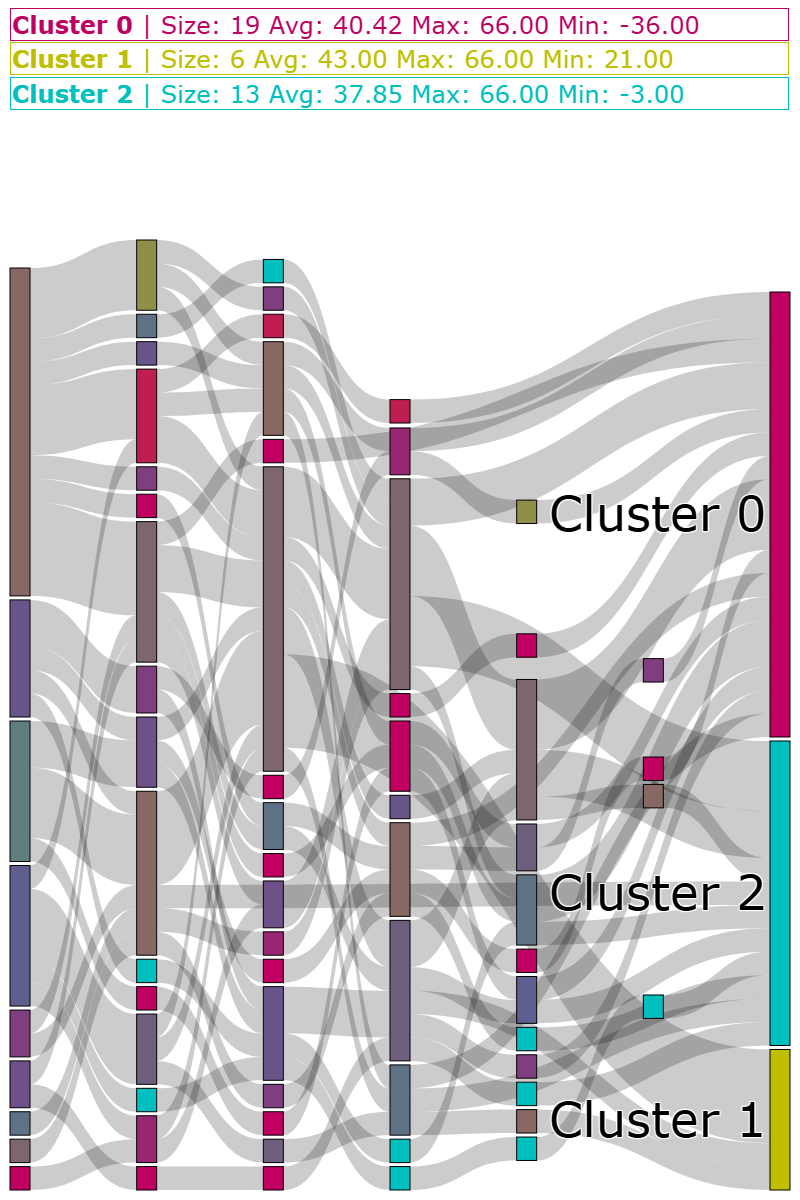}
    \end{minipage}
    \begin{minipage}{0.19\linewidth}
        \centering
        \includegraphics[width=\linewidth]{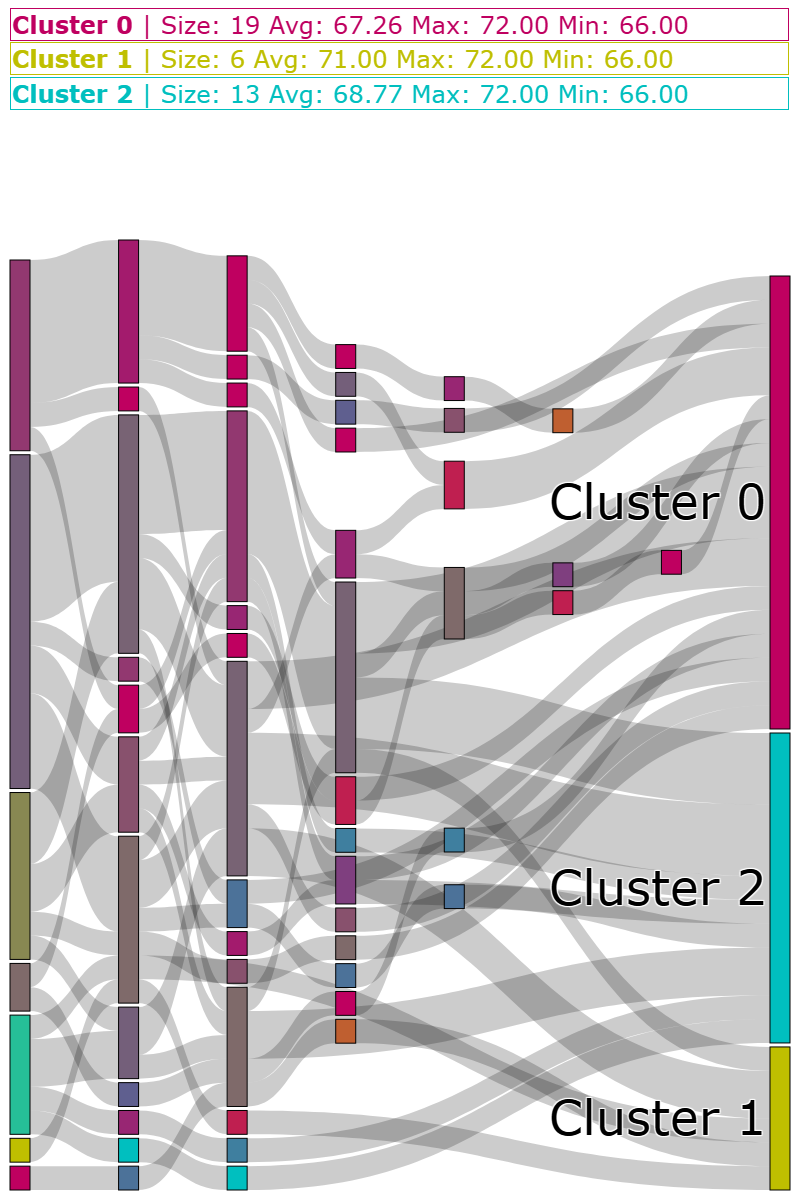}
    \end{minipage}
    \begin{minipage}{0.19\linewidth}
        \centering
        \includegraphics[width=\linewidth]{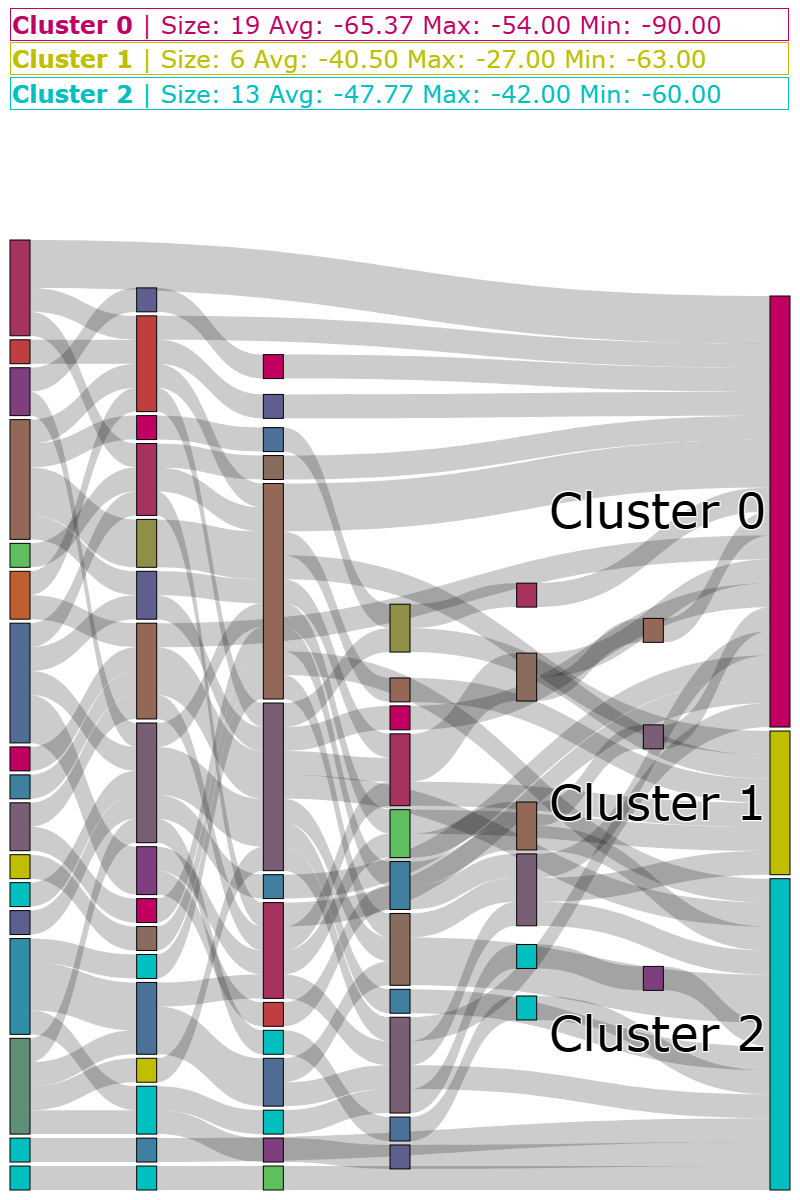}
    \end{minipage}
    \\
    \begin{minipage}{0.333\linewidth}
        \centering
        \includegraphics[width=\linewidth]{fitness_landscape_kmeans_k4.pdf}
    \end{minipage}
    \begin{minipage}{0.19\linewidth}
        \centering
        \includegraphics[width=\linewidth]{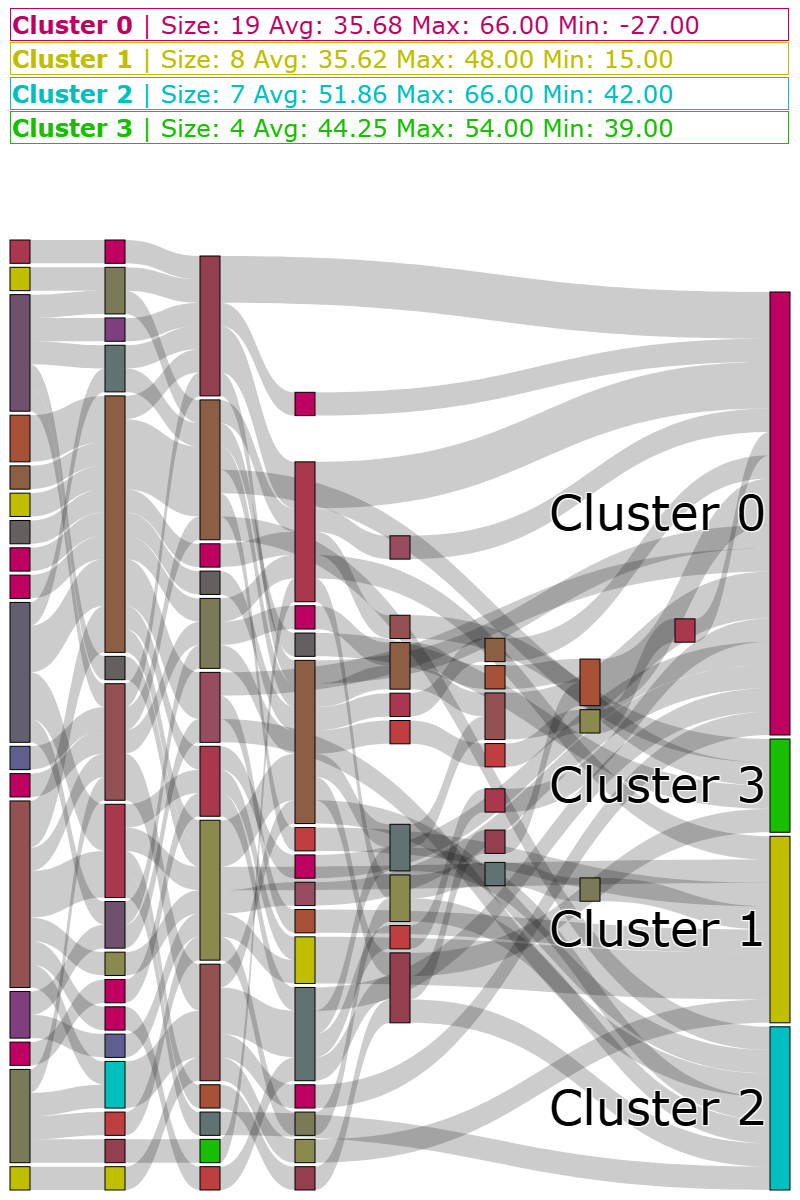}
    \end{minipage}
    \begin{minipage}{0.19\linewidth}
        \centering
        \includegraphics[width=\linewidth]{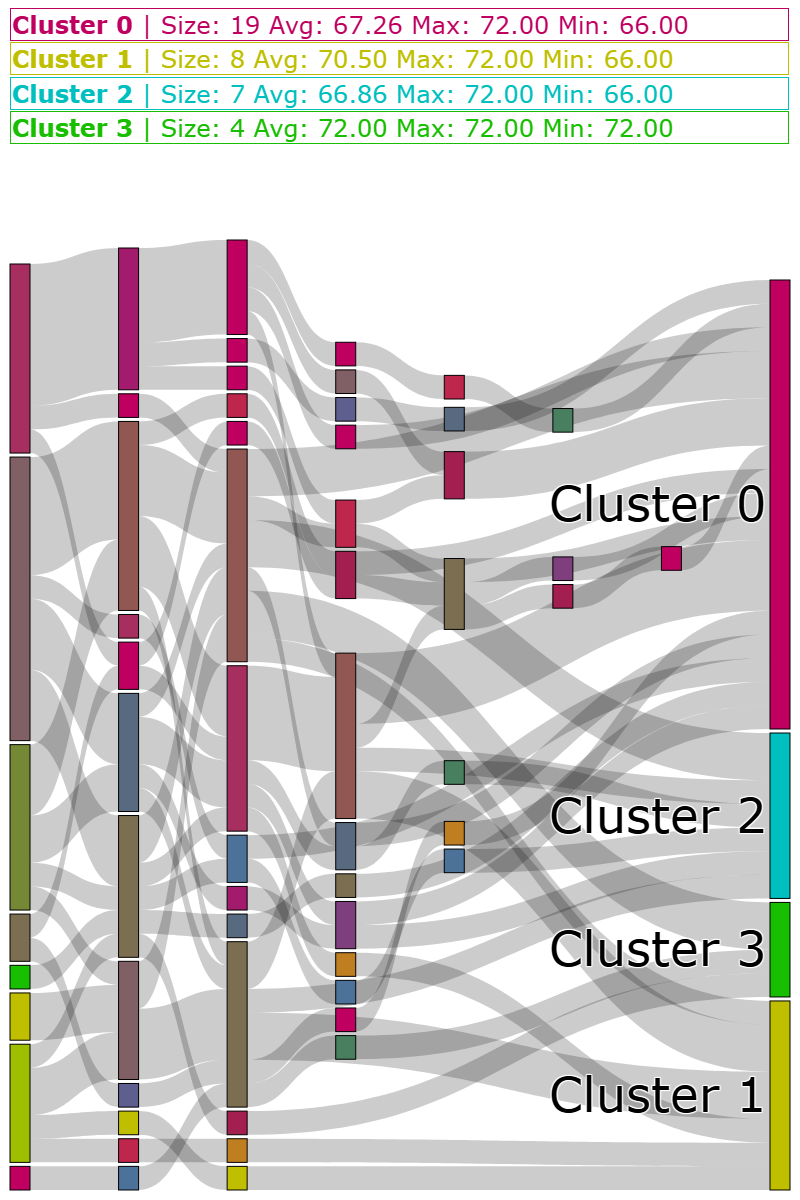}
    \end{minipage}
    \begin{minipage}{0.19\linewidth}
        \centering
        \includegraphics[width=\linewidth]{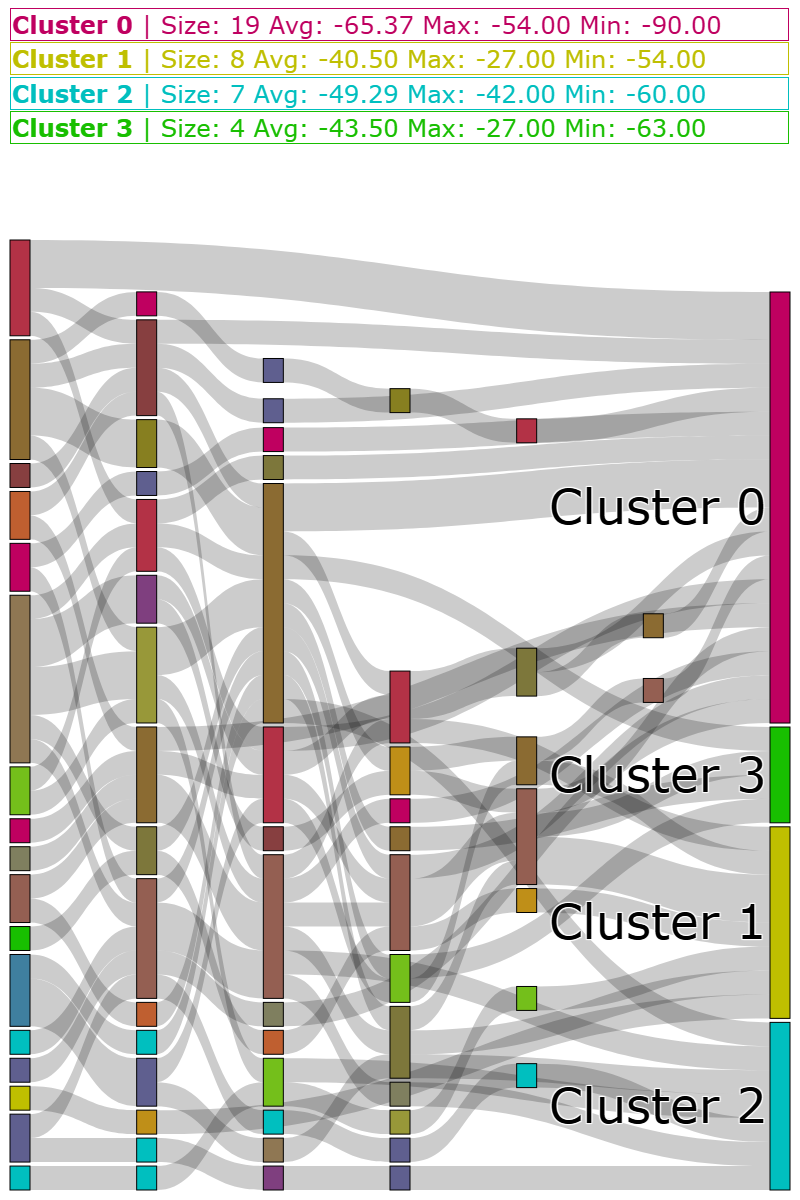}
    \end{minipage}
    \\
    \begin{minipage}{0.333\linewidth}
        \centering
        \includegraphics[width=\linewidth]{fitness_landscape_kmeans_k5.pdf}
    \end{minipage}
    \begin{minipage}{0.19\linewidth}
        \centering
        \includegraphics[width=\linewidth]{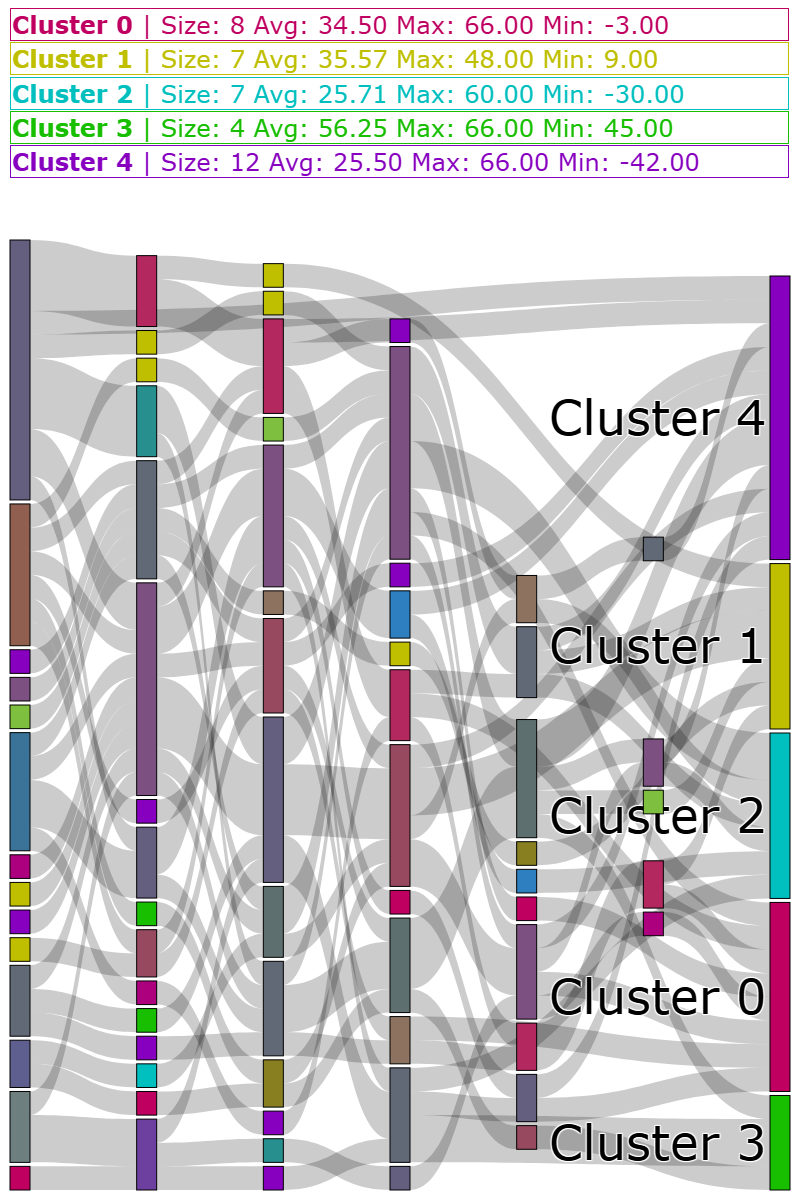}
    \end{minipage}
    \begin{minipage}{0.19\linewidth}
        \centering
        \includegraphics[width=\linewidth]{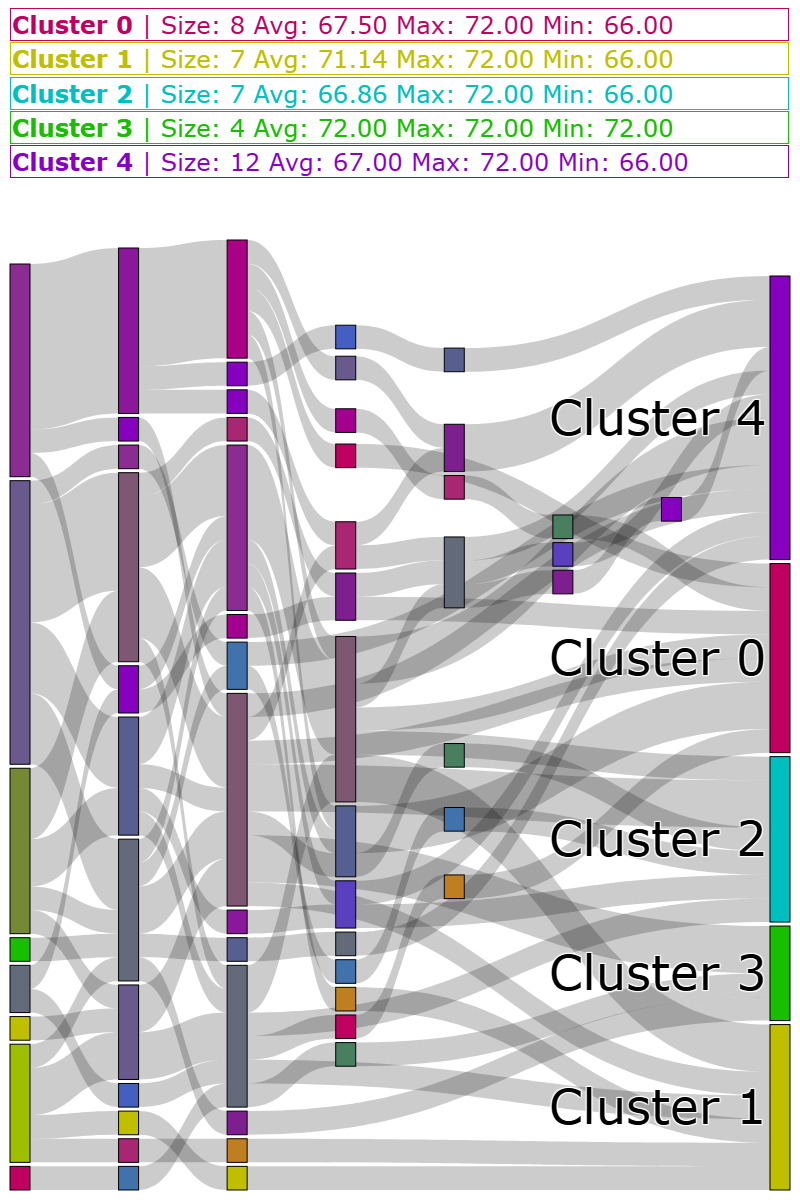}
    \end{minipage}
    \begin{minipage}{0.19\linewidth}
        \centering
        \includegraphics[width=\linewidth]{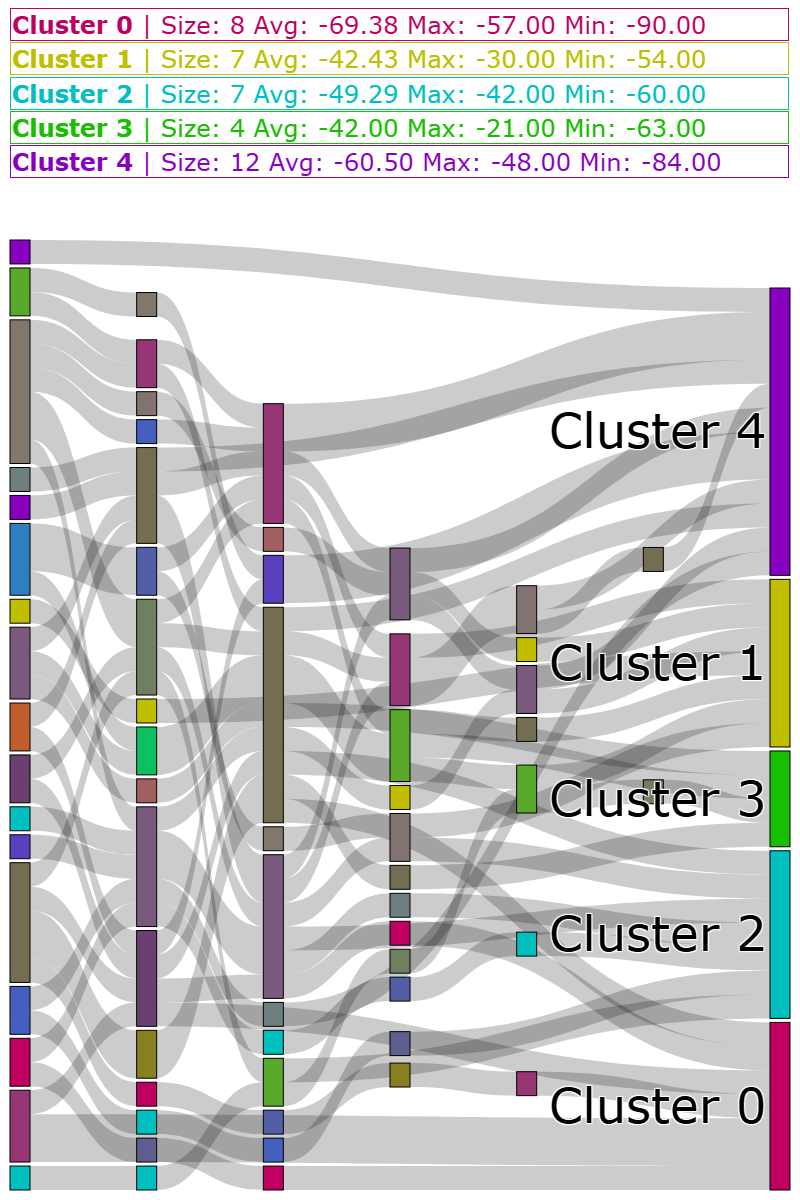}
    \end{minipage}
    \\
    \begin{minipage}{0.333\linewidth}
        \centering
        \includegraphics[width=\linewidth]{fitness_landscape_kmeans_k6.pdf}
        fitness landscape
    \end{minipage}
    \begin{minipage}{0.19\linewidth}
        \centering
        \includegraphics[width=\linewidth]{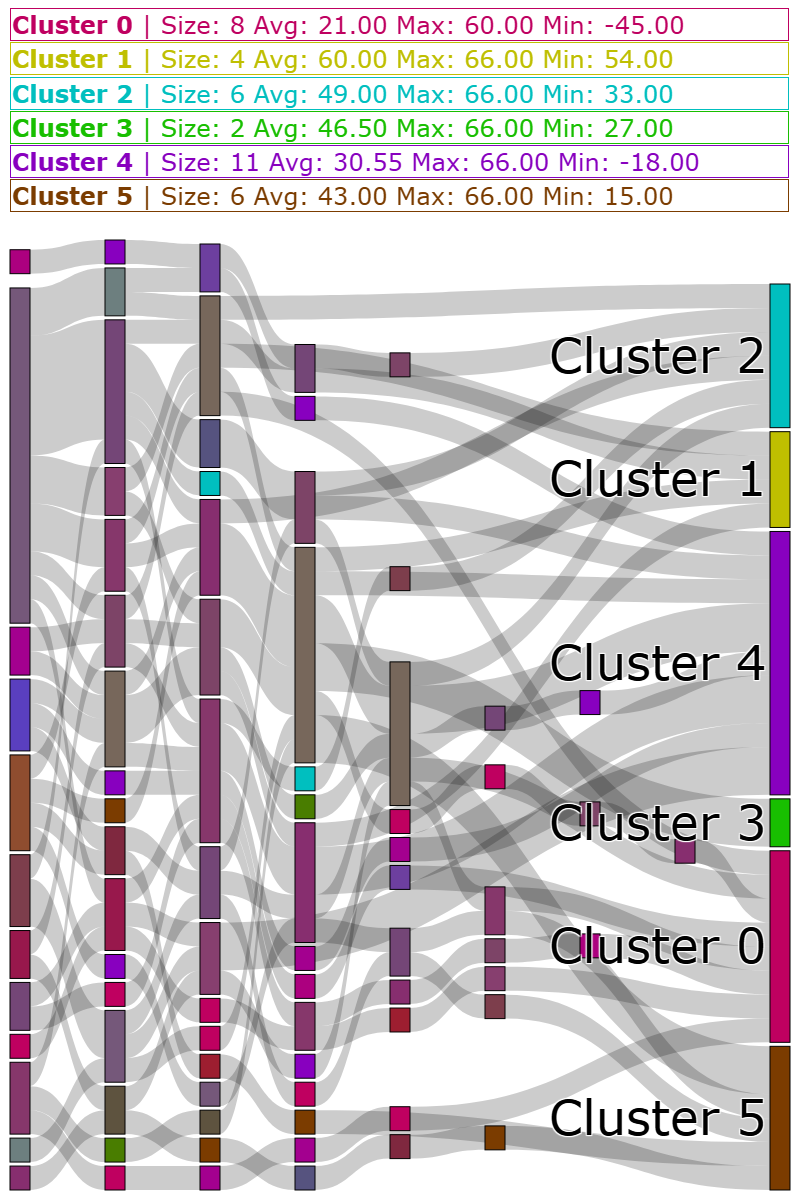}
        unsorted
    \end{minipage}
    \begin{minipage}{0.19\linewidth}
        \centering
        \includegraphics[width=\linewidth]{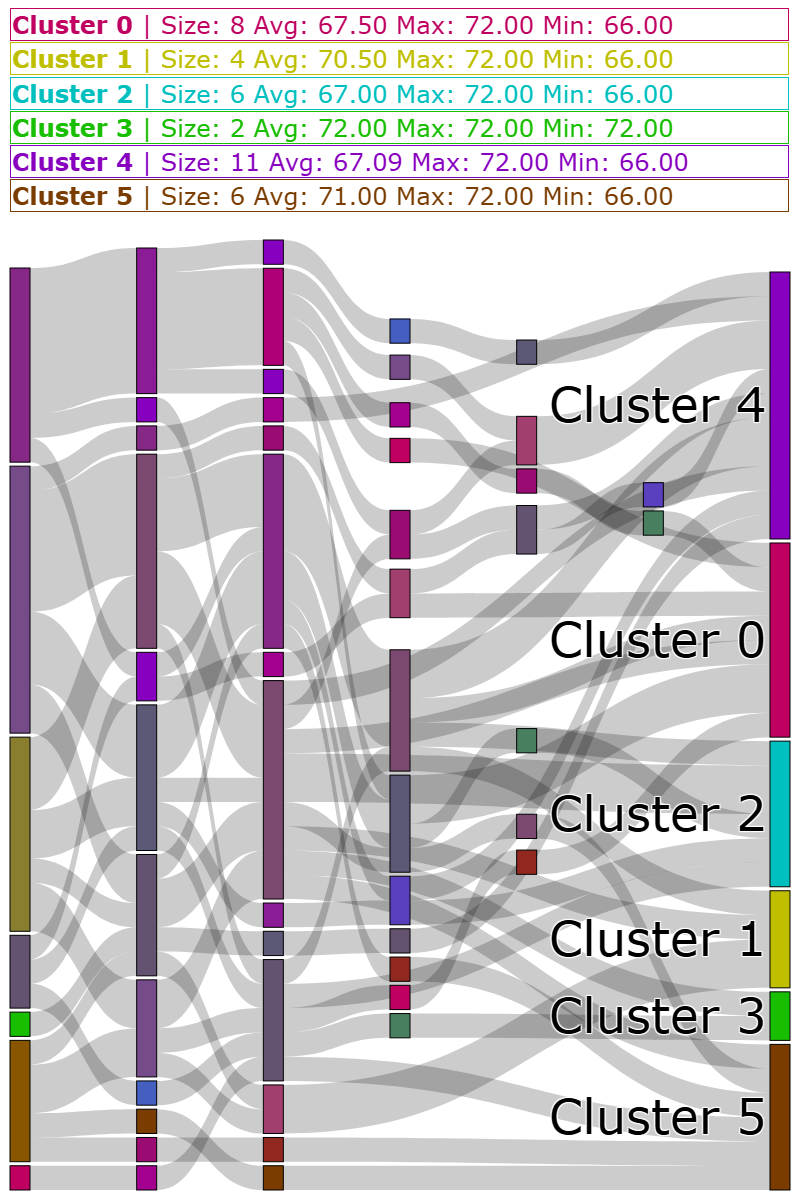}
        win
    \end{minipage}
    \begin{minipage}{0.19\linewidth}
        \centering
        \includegraphics[width=\linewidth]{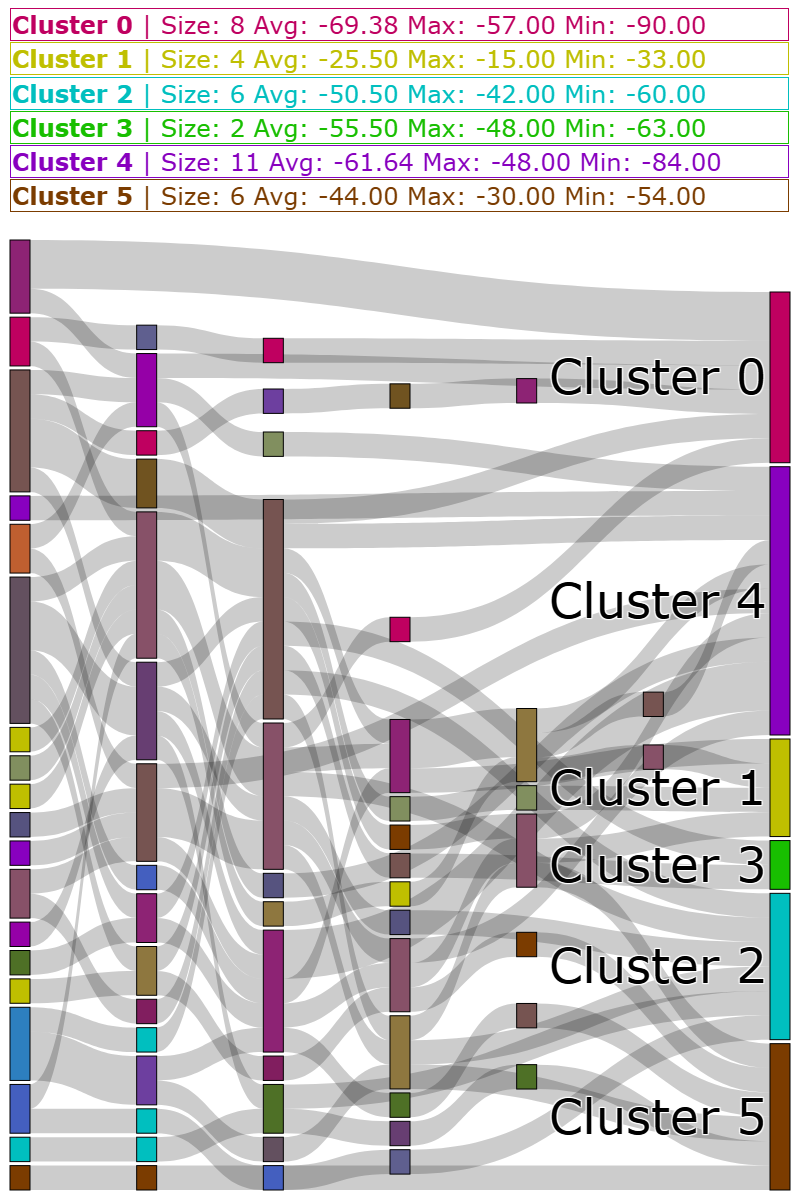}
        loss
    \end{minipage}
    \\
    \caption{Comprehensive sankey diagram of tactic patterns with $k$-means algorithm for clustering}
    \label{fig:comprehensive_sankey_tactic_diagram_kmeans}
\end{figure*}

\Cref{fig:appendix_comprehensive_sankey_diagram_Agglomerative,fig:comprehensive_sankey_diagram_birch,fig:comprehensive_sankey_diagram_gmm,fig:comprehensive_sankey_diagram_kmeans} and \Cref{fig:comprehensive_sankey_tactic_diagram_Agglomerative,fig:comprehensive_sankey_tactic_diagram_birch,fig:comprehensive_sankey_tactic_diagram_gmm,fig:comprehensive_sankey_tactic_diagram_kmeans} present comprehensive Sankey diagrams for action sequence and tactic patterns, respectively, across multiple clustering algorithms (Ward Agglomerative Clustering, BIRCH, GMM, $k$-means) and various cluster numbers, providing extensive validation of the main findings presented in the main text. Tactical extraction is performed through rule-based multi-label annotation, which generates a large scale of tactics due to the combinatorial constitution of numerous labels.

It should be noted that multiple clustering algorithms with different numbers of clusters $k$ were applied to avoid coincidental conclusions resulting from specific algorithms and configurations. In the Sankey diagrams, circular flows caused by merged counting of the same patterns at different temporal occurrences have been handled by adding temporal subscripts for better visualization. Therefore, the rectangular blocks in the Sankey diagrams contain temporal slices of the same pattern at different time points. Additional statistical information for clustered samples, including sample size, maximum, minimum and average fitness values, is presented in the upper right corner of each diagram.

\newpage

\newpage
\subsection{Tactic Analysis}

\begin{figure*}[!htp]
    \centering
    \begin{minipage}{0.48\linewidth}
        \centering
        \includegraphics[width=\linewidth]{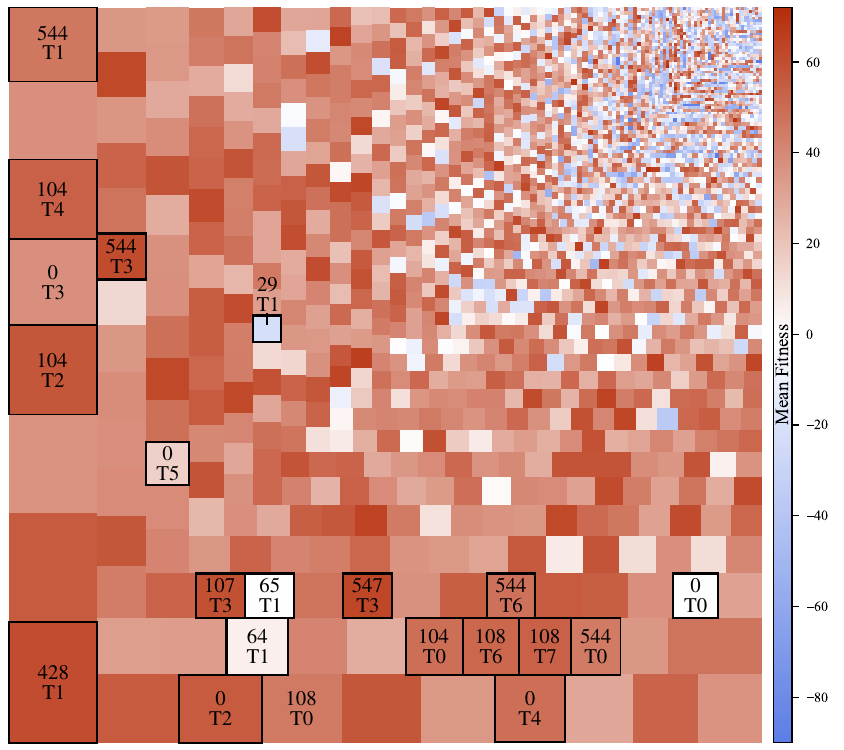}
        a) sce1
    \end{minipage}
    \begin{minipage}{0.48\linewidth}
        \centering
        \includegraphics[width=\linewidth]{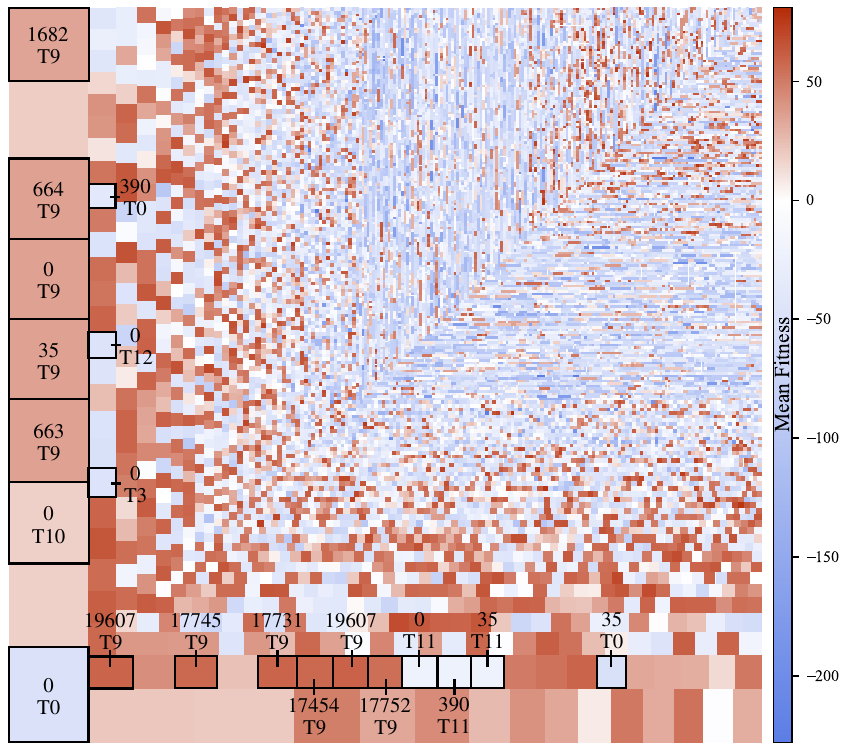}
        b) sce2
    \end{minipage}
    \caption{Treemap of state-tactic payoff correlation}
    \label{fig:appendix_state_tactic_correlation}
\end{figure*}

The detailed tactical analysis presented in Tables \ref{tab:tactical_analysis_sce1} and \ref{tab:tactical_analysis_sce2} provides comprehensive insights into state-tactic relationships across different scenarios. The analysis can be summarized through four key dimensions: 

\begin{enumerate}[leftmargin=18pt,label=(\arabic*),itemindent=0pt] 
    \item High win rate state-tactic pairs: This dimension examines pairs demonstrating superior average outcomes. It analyzes how specific tactical approaches achieve optimal performance in particular state configurations, such as independent greedy attacks versus collaborative tactics adapting to distinct tactical requirements and battlefield conditions.
    \item State-dependent tactical effectiveness: This dimension analyzes how different tactics yield significantly varying results within the same state. This phenomenon is particularly evident in opening states, where tactical selection critically determines final outcomes.
    \item State characteristics overriding tactical importance: This dimension investigates circumstances where different states achieve similar results regardless of tactical selection. Such cases typically occur in states with pronounced force superiority, where even basic tactics perform effectively due to inherent advantages.
    \item Disadvantaged state analysis: This dimension evaluates specific tactics that achieve low returns in certain states. It covers situations where ineffective tactics executed in balanced and disadvantaged situations, leading to negative outcomes.
\end{enumerate}
These detailed case studies and comparative analyses substantiate the macro-level patterns observed in the treemap visualization, providing empirical evidence for the state-dependent nature of tactical effectiveness and the critical role of opening tactics in determining final outcomes.

\begin{table*}[htp]
\centering
\caption{Detailed tactical analysis of representative states in sce1}
\label{tab:tactical_analysis_sce1}
\begin{tabular}{@{}>{\centering\arraybackslash}m{0.18\textwidth}>{\centering\arraybackslash}m{0.24\textwidth}m{0.54\textwidth}@{}}
\toprule
\textbf{state--tactic pair} & \textbf{state distribution} & \textbf{analysis} \\
\midrule
\vspace{-10pt}\makecell{104-T2\\107-T3\\428-T1\\544-T1\\547-T1} &
\vspace{8pt}\makecell{\vspace{-6pt}\includegraphics[width=\linewidth]{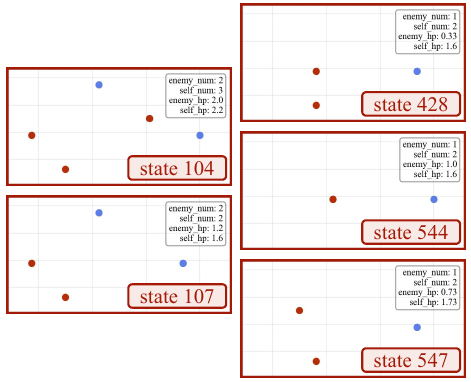}} &
\parbox[t]{0.54\textwidth}{\vspace{-58pt}
High win rate state-tactic pairs represent superior average outcomes when executing specific tactics in corresponding states. Tactic T1 represents independent greedy attack without collaboration consideration, which is applicable to states 428, 544 and 547 where the red side has clear advantages and the blue side has only one unit. In such scenarios, greedy attack is optimal and collaboration consideration has minimal impact on final actions. In contrast, tactics T2 and T3 introduce collaboration, corresponding to states 104 and 107 where targets are not unique and distributions are scattered, requiring coordinated efforts to concentrate firepower.} \\
\midrule
\vspace{16pt}\makecell{0-T0\\0-T2\\0-T3\\0-T4\\0-T5} &
\vspace{28pt}\makecell{\vspace{-6pt}\includegraphics[width=\linewidth]{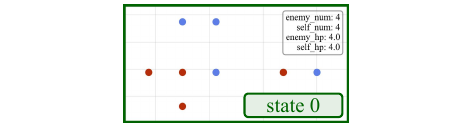}} &
\parbox[t]{0.54\textwidth}{\vspace{-36pt}
In specific states, different tactics yield significantly different results, with the most notable being state 0 (game opening). Under initial formation conditions where both sides are evenly matched, tactic T0 representing no obvious tactic achieves zero returns. Tactic T2, representing greedy attack with local collaboration, achieves maximum return. Building on this, tactic T4 introducing feint attack also achieves relatively high returns, but tactic T5 considering unit sacrifice for feint execution in unit-interleaved distributions shows insignificant returns.} \\
\midrule
\vspace{-36pt}\makecell{104-T0\\104-T2\\104-T4\\108-T0\\108-T6\\108-T7\\544-T0\\544-T1\\544-T3\\544-T6} &
\vspace{10pt}\makecell{\vspace{-6pt}\includegraphics[width=\linewidth]{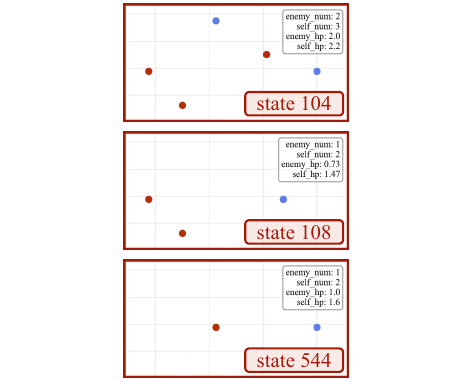}} &
\vspace{36pt}\parbox[t]{0.54\textwidth}{\vspace{-74pt}
In specific states, different tactics may produce similar results due more to state characteristics than tactic selection. States 104, 108, and 544 all feature clear red side superiority, where even tactic T0 without obvious tactical consideration can achieve high returns. For state 104, local collaboration needs to be considered to adapt to multiple targets, thus tactics T2 and T4 with switched collaboration modes are frequently utilized and demonstrate high returns. In contrast, states 108 and 544 have simpler distributions and more obvious advantages for the red, enabling simple or even non-coordinated tactics like T6 and T7 to perform well.} \\
\midrule
\vspace{-14pt}\makecell{0-T0\\29-T1\\64-T1\\65-T1} &
\vspace{-4pt}\makecell{\vspace{-6pt}\includegraphics[width=\linewidth]{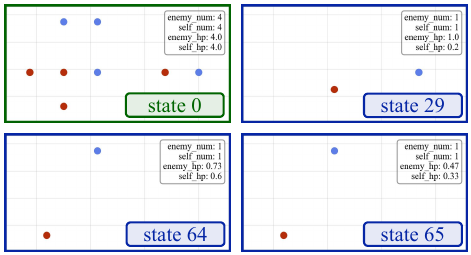}} &
\vspace{12pt}\parbox[t]{0.54\textwidth}{\vspace{-40pt}
Specific tactics in certain states produce low returns. In state 0, ineffective tactics prevent return accumulation, confirming the necessity of opening tactics for final outcomes. However, in some situations where the red side has fallen into clear disadvantage, continuing engagement with greedy attack T1 will undoubtedly result in negative returns.} \\
\bottomrule
\end{tabular}
\end{table*}

\begin{table*}[htp]
\centering
\caption{Detailed tactical analysis of representative states in sce2}
\label{tab:tactical_analysis_sce2}
\begin{tabular}{@{}>{\centering\arraybackslash}m{0.18\textwidth}>{\centering\arraybackslash}m{0.3\textwidth}m{0.5\textwidth}@{}}
\toprule
\textbf{state--tactic pair} & \textbf{state distribution} & \textbf{analysis} \\
\midrule
\vspace{-18pt}\makecell{17454-T9\\17731-T9\\17732-T9\\17745-T9\\17752-T9\\19607-T9} &
\vspace{6pt}\makecell{\vspace{-6pt}\includegraphics[width=\linewidth]{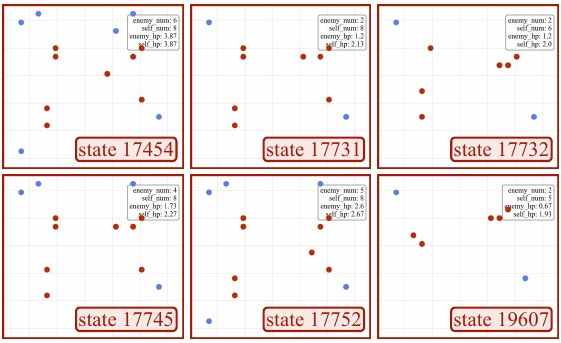}} &
\parbox[t]{0.5\textwidth}{\vspace{-52pt}
Specific tactics demonstrate high win rates across different states. These states are primarily in the early combat phases. In these states, the red side has already eliminated two units at the center of the blue side through local concentration of forces and coordinated firepower tactic T9, thereby establishing positional superiority. It should be noted that although states 17454 and 17752 have comparable total health values between both sides, these states still yield high overall returns due to numerical and firepower advantages of the red.} \\
\midrule
\vspace{-4pt}\makecell{0-T0\\0-T3\\0-T9\\0-T10\\0-T11\\0-T12} &
\vspace{22pt}\makecell{\vspace{-6pt}\includegraphics[width=\linewidth]{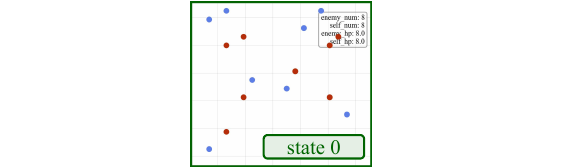}} &
\parbox[t]{0.5\textwidth}{\vspace{-40pt}
Different opening tactics for state 0 produce significantly different results, with this effect becoming more pronounced in the more challenging sce2 scenario. Tactic T0 with no obvious tactic achieves significant negative returns, while tactics T3, T11 and T12 are also unsuitable for this state. Tactics T9 and T10, which incorporate unit concentration, achieve positive returns, with T9 featuring switchable collaboration modes demonstrating higher returns compared to T10 with fixed collaboration mode.} \\
\midrule
\vspace{-6pt}\makecell{0-T9\\35-T9\\663-T9\\664-T9\\1682-T9} &
\vspace{12pt}\makecell{\vspace{-6pt}\includegraphics[width=\linewidth]{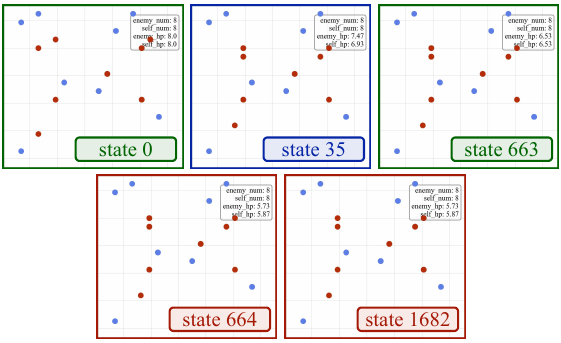}} &
\vspace{-6pt}\parbox[t]{0.5\textwidth}{\vspace{-56pt}
Certain states feature similar unit distributions but different health value distributions, where specific tactics demonstrate high efficiency. States 0 and 663 represent balanced situations between both sides, while state 35 shows clear disadvantage for the red side, and states 664 and 1682 indicate slight red side advantage. Despite different or even disadvantageous situations in these states, these state-tactic pairs occupy substantial area and achieve satisfactory returns. This demonstrates that the local concentration of forces and coordinated firepower tactic T9 exhibits broad applicability in such scenarios.} \\
\midrule
\vspace{-18pt}\makecell{0-T0\\0-T11\\35-T0\\35-T11\\390-T0\\390-T11} &
\vspace{4pt}\makecell{\vspace{-6pt}\includegraphics[width=\linewidth]{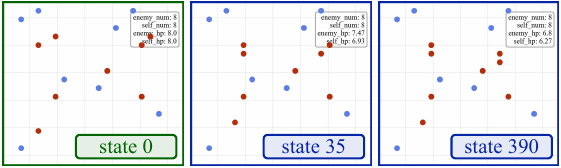}} &
\vspace{-6pt}\parbox[t]{0.5\textwidth}{\vspace{-30pt}
Besides no obvious tactics achieving low or negative returns in several opening states, tactic T11, which considers priority attacks on the most threatening targets, also demonstrates significant inefficiency. This tactic prioritizes enemy firepower without considering unit distribution factors, making it difficult to apply in opening states.} \\
\bottomrule
\end{tabular}
\end{table*}

\end{document}